%% file: iccv.tex
\documentclass[10pt,twocolumn,letterpaper]{article} 

\usepackage{cvpr}
\usepackage{times}
\usepackage{graphicx}
\usepackage{amsmath}
\usepackage{amssymb}
\usepackage{verbatim}
\usepackage{epstopdf}
\usepackage{epsfig}

\usepackage[pagebackref=true,breaklinks=true,letterpaper=true,colorlinks,bookmarks=false]{hyperref}

 \cvprfinalcopy 

\def\cvprPaperID{  Yaser Yacoob \\ UMIACS\\University of Maryland, College Park, MD 20742}

\begin{document}

\twocolumn

\title{Modeling Colors of Single Attribute Variations with Application to Food Appearance}

\author{
Yaser Yacoob\\
Computer Vision Laboratory-UMIACS\\
University of Maryland, College Park, MD 20742\\
{\tt\small yaser@umiacs.umd.edu}
}

\maketitle
\renewcommand{\thefootnote}{}


\begin{abstract}
This paper considers the intra-image  color-space of an object or a scene when these
are subject to a dominant single-source of variation.
The source of variation can be intrinsic or extrinsic (i.e., imaging conditions) to the object. We observe that the quantized colors 
for such objects typically lie on a planar subspace of RGB, and in some cases  linear or polynomial curves on
this plane are effective in capturing these color variations. We also observe that the inter-image color sub-spaces are robust as long
as drastic illumination change is not involved.

We illustrate the use of this analysis for: discriminating between shading-change and reflectance-change for patches, and 
object detection, segmentation and recognition based on a single exemplar. We focus on images of food items to illustrate the effectiveness of the proposed approach.

\end{abstract}


\input{iccv_all}



\bibliographystyle{plain}

\newpage
\rule{0pt}{1pt}\newpage
\rule{0pt}{1pt}\newpage
\rule{0pt}{1pt}\newpage
\rule{0pt}{1pt}\newpage
\rule{0pt}{1pt}\newpage
\rule{0pt}{1pt}\newpage
\rule{0pt}{1pt}\newpage
\rule{0pt}{1pt}\newpage

\end{document}

%% file: iccv_all.tex
\vspace{-0.1in}
\section{Background}

This paper studies the {\em intra-image} color appearance of objects that are subject to a {\em single} process that alters their 
color properties.
Specifically, we consider the existence and nature of the
subspace of the color appearance of an object for which an {\em unknown} but a single intrinsic or extrinsic attribute varies across
its surface.
Intrinsic attributes include changes that cause variations in the spectrum of light wavelengths absorbed/reflected under typical illumination.
For example, baking changes the reflected wavelengths with respect to the uniform reflectance of dough.
Extrinsic attributes are related to image formation including surface normal change, light absorption  coincident with
translucent objects, etc. Although these properties  apply {\em primarily} to intra-image appearance  we 
demonstrate some generalization to inter-image within-class objects  as long as illumination does not significantly 
alter the color space.

We assume that object  appearance takes on gradual-change in reflectance properties when exposed
to a single unknown process that is possibly challenging to quantify (e.g., baking dough).
 We focus on qualitative and empirical attributes.

\begin{figure*}
\centerline{
\psfig{figure=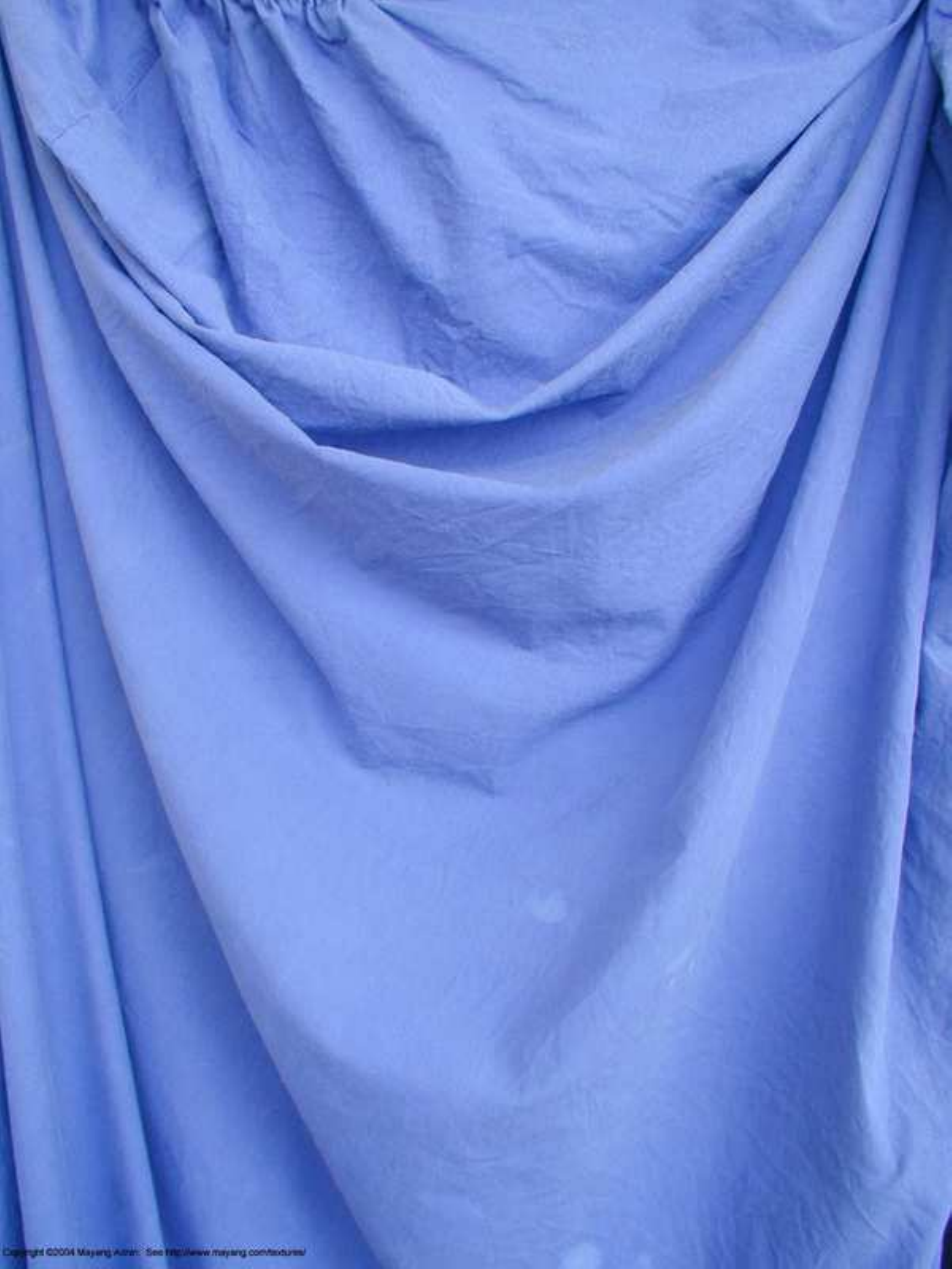,width=0.5in}
\psfig{figure=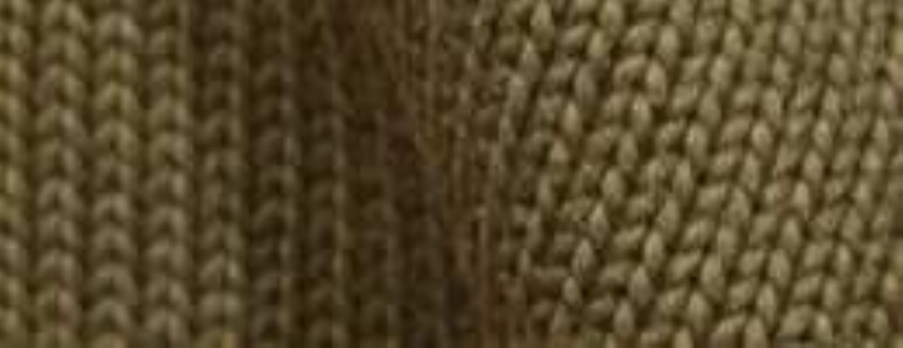,width=0.7in}
\psfig{figure=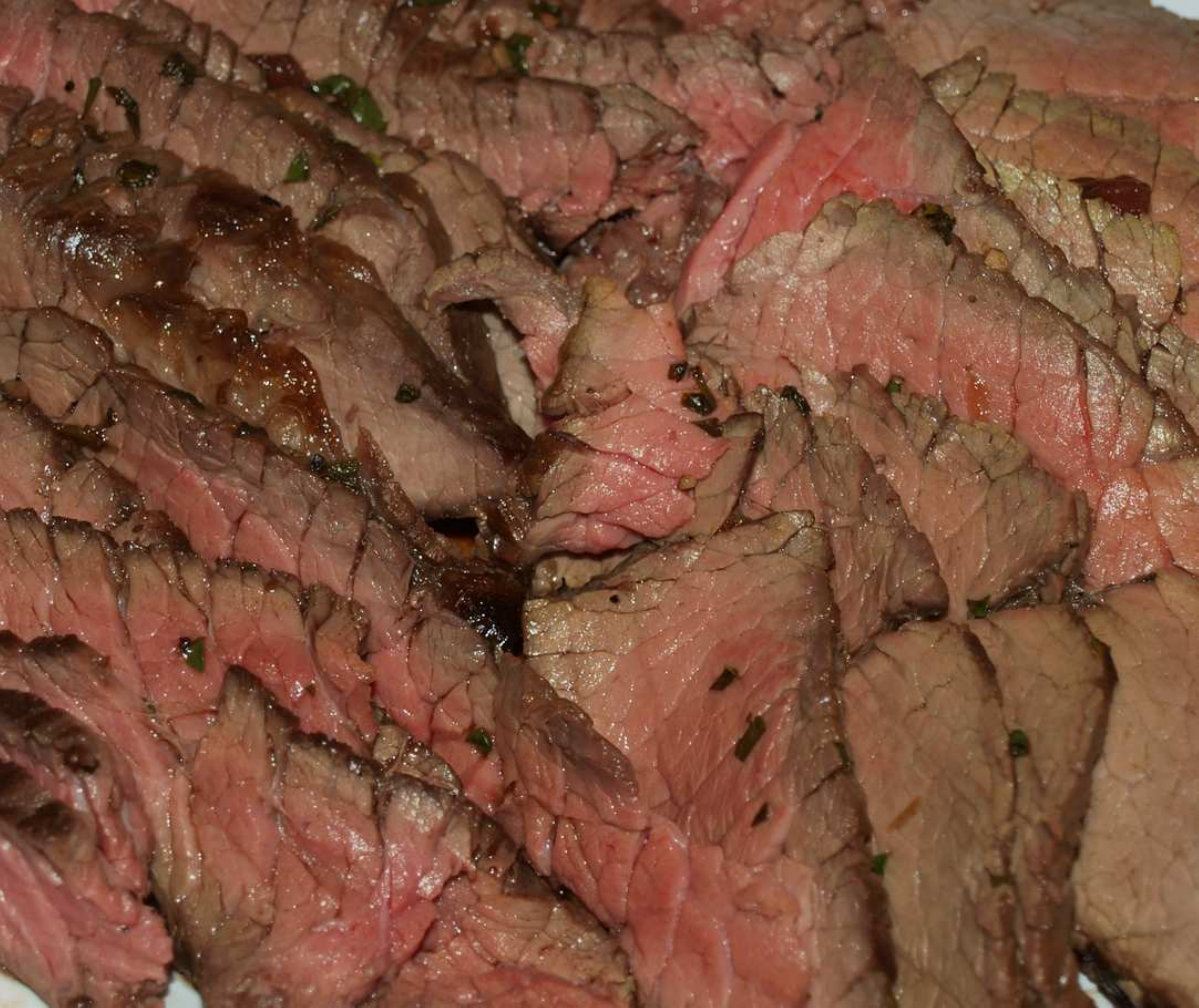,width=0.6in}
\psfig{figure=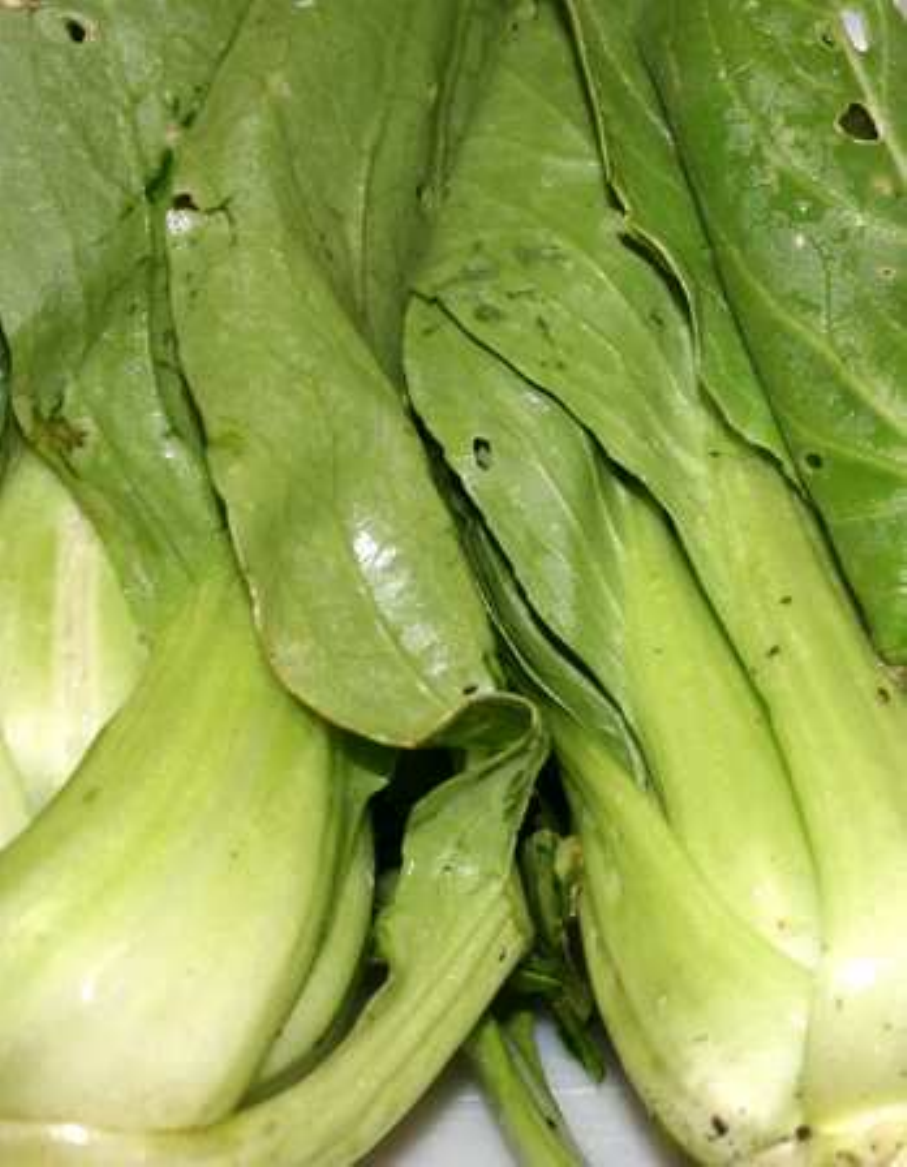,width=0.5in}
\psfig{figure=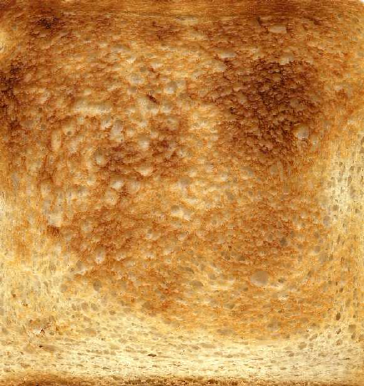,width=0.5in}
\psfig{figure=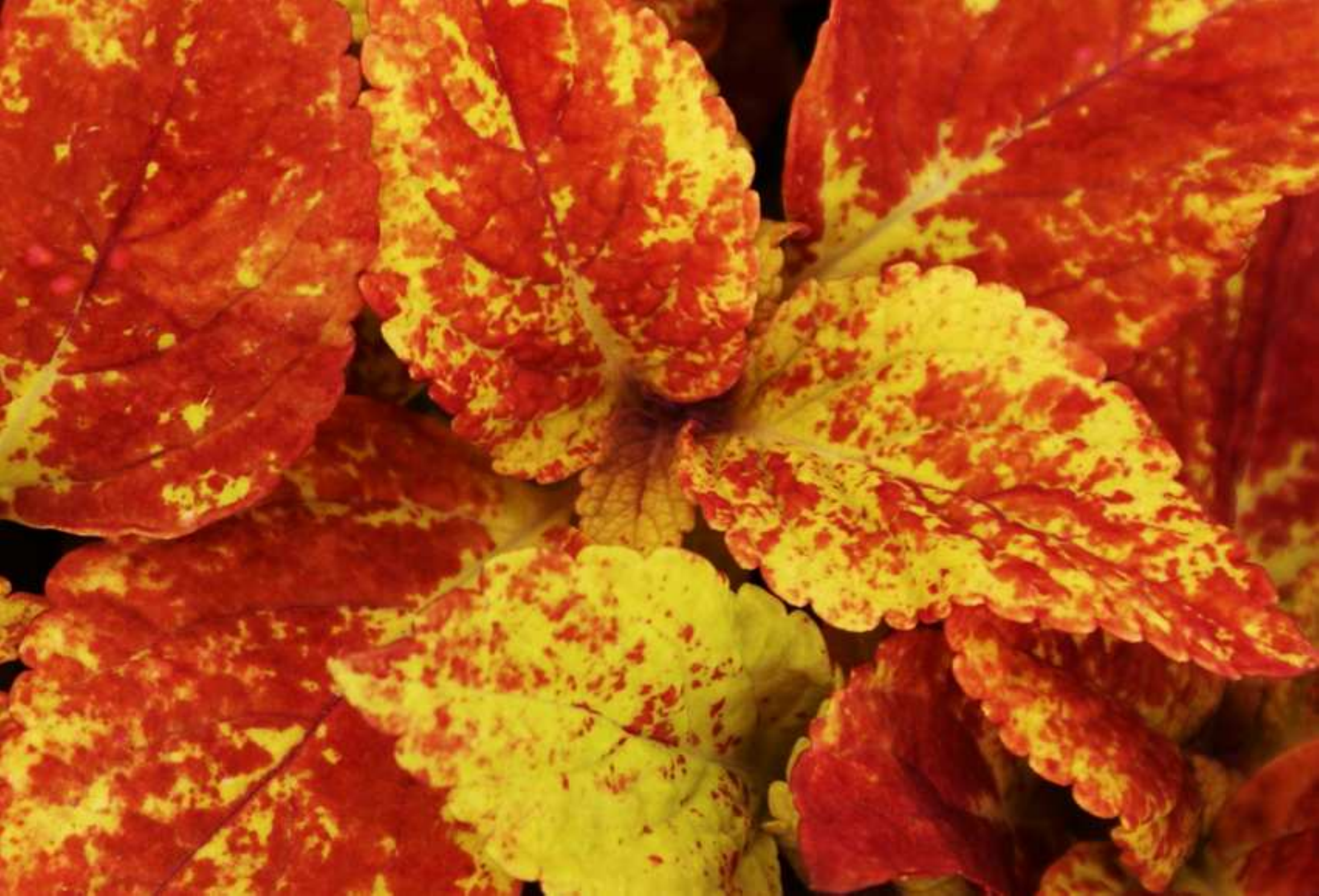,width=0.6in}
\psfig{figure=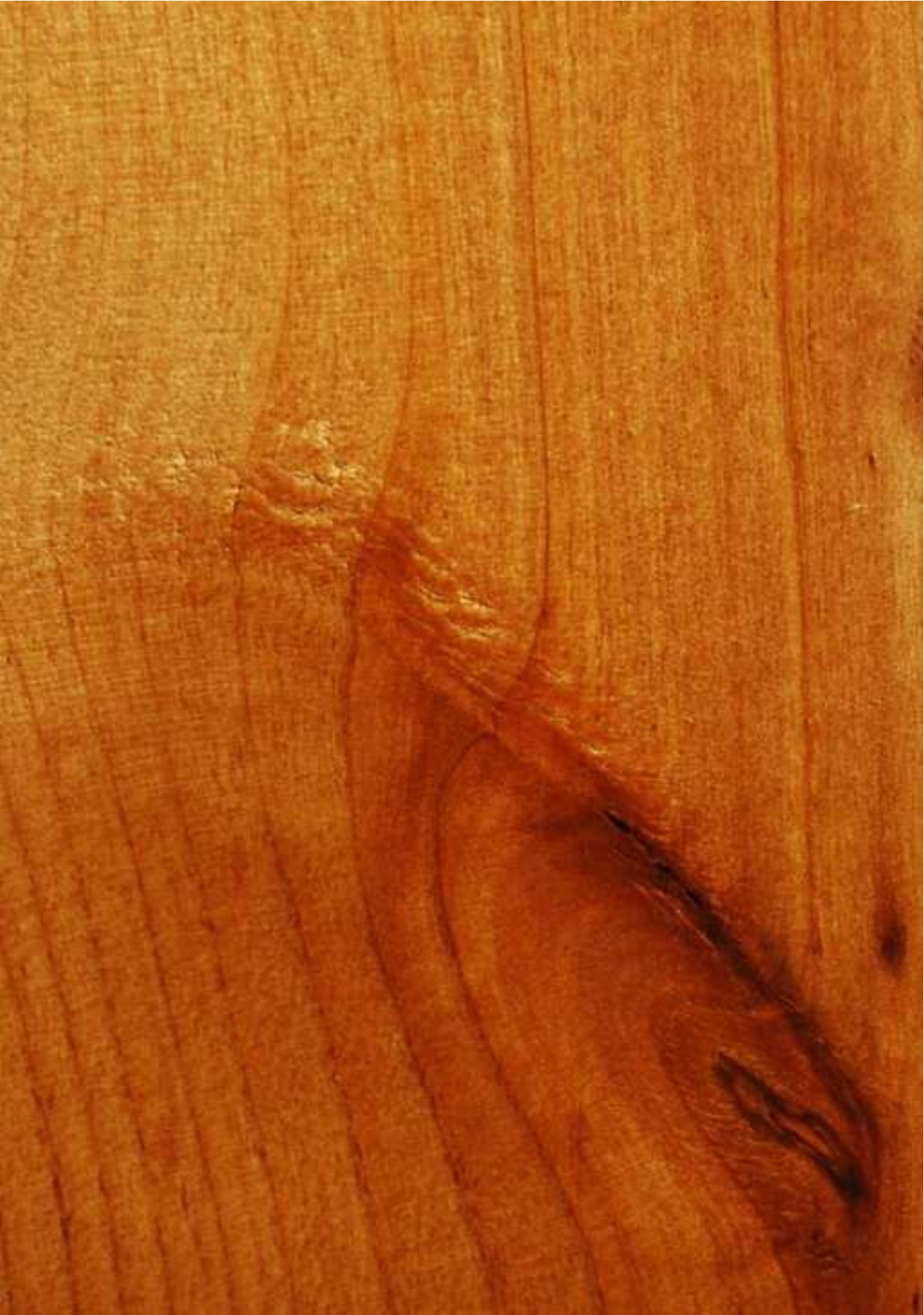,width=0.4in}
\psfig{figure=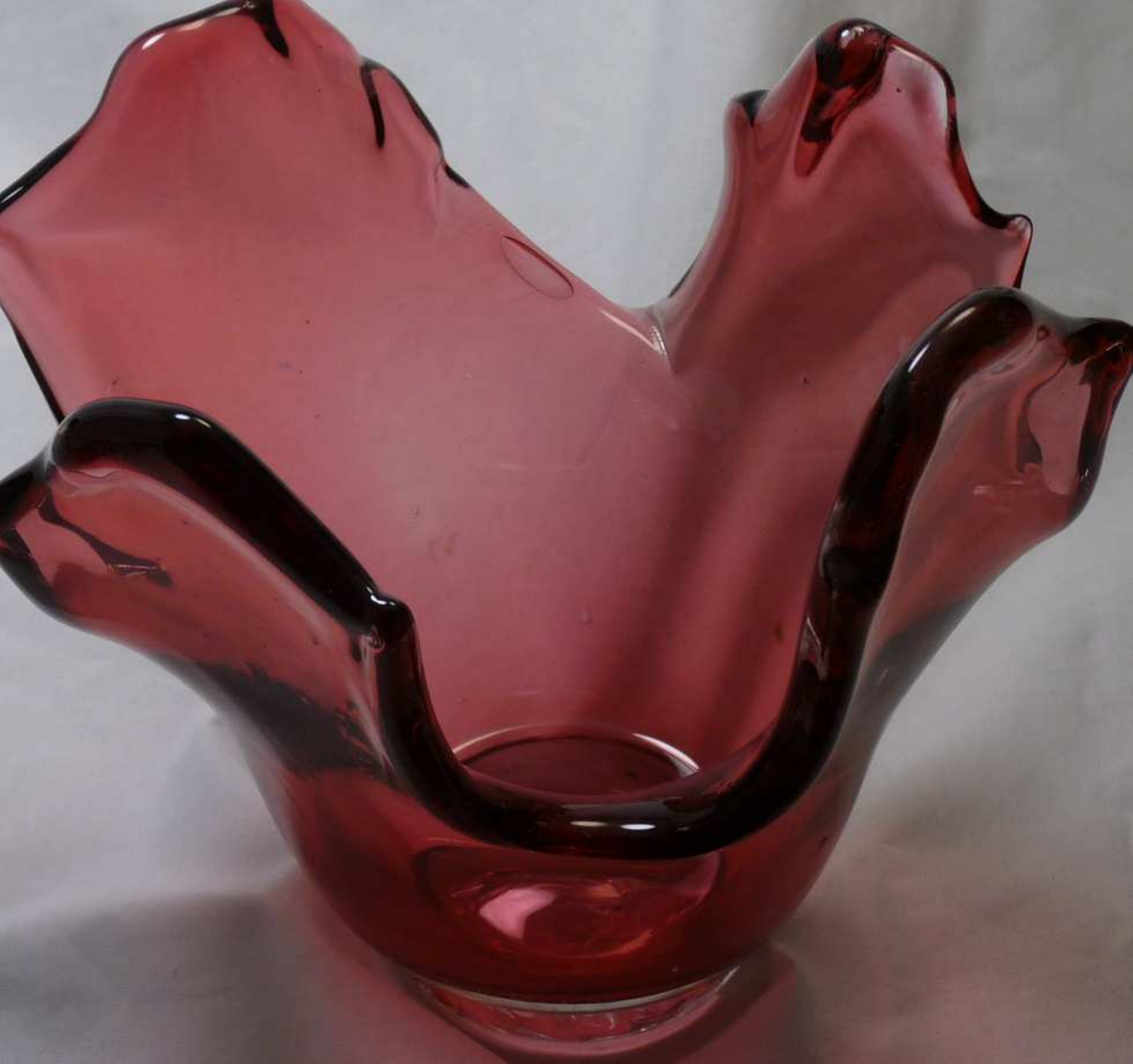,width=0.6in}
\psfig{figure=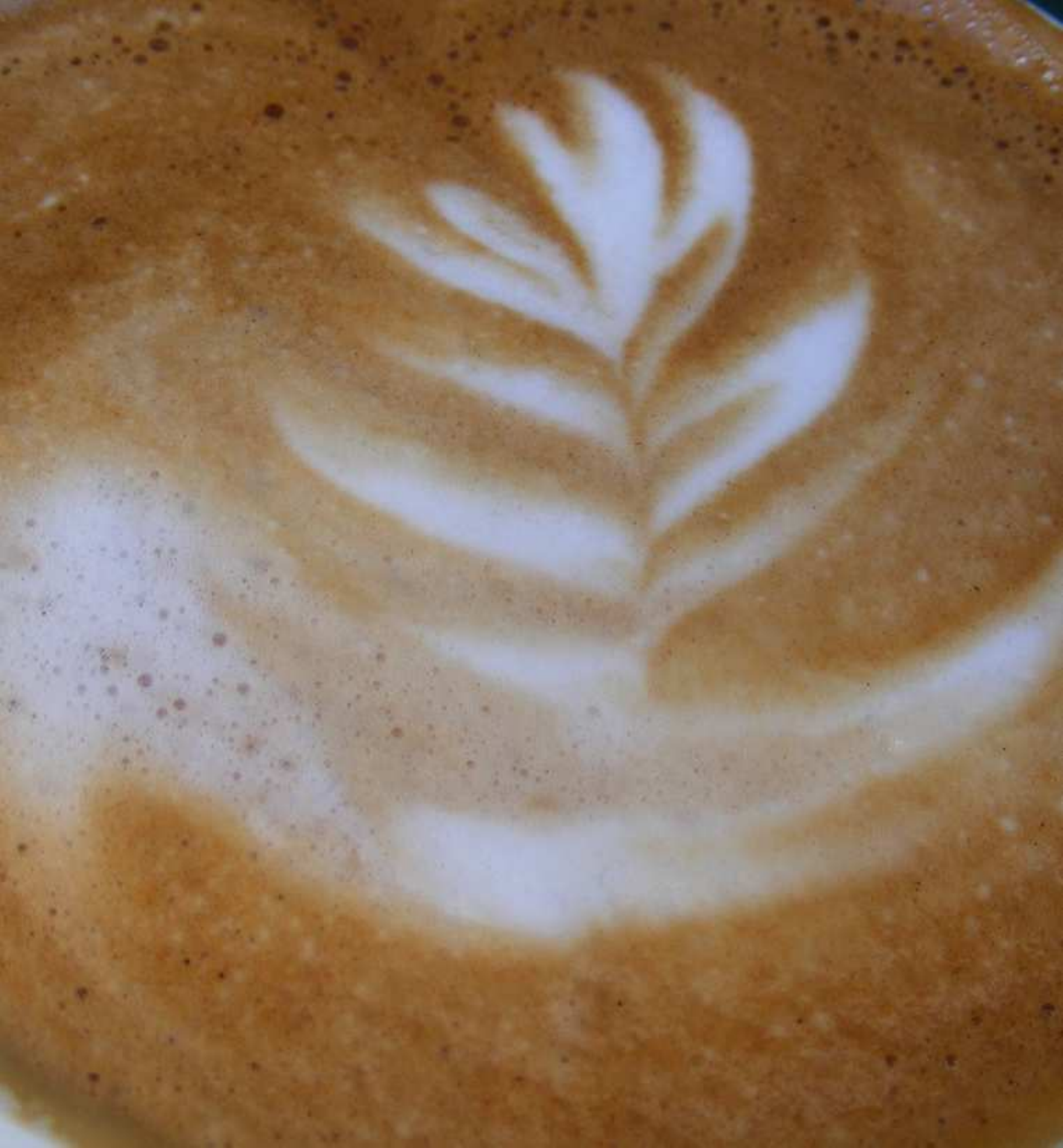,width=0.5in}
\psfig{figure=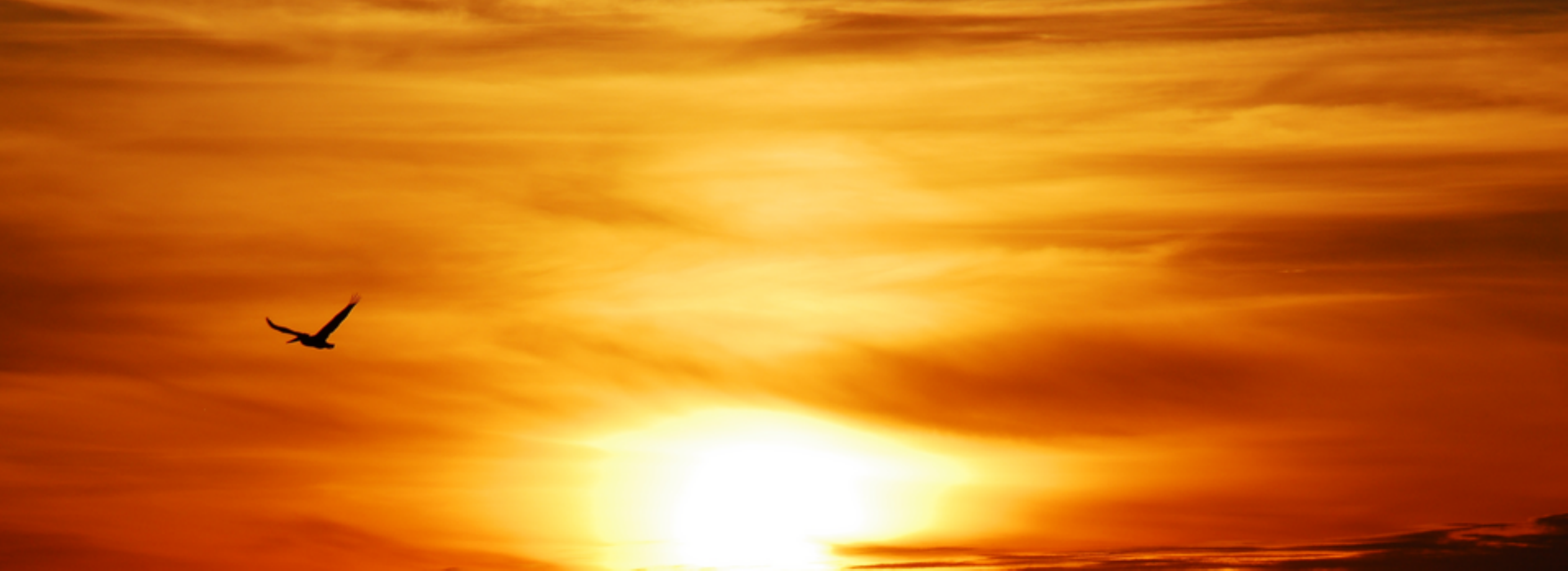,width=0.7in}
\psfig{figure=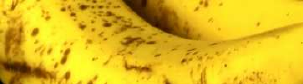,width=0.7in}
\psfig{figure=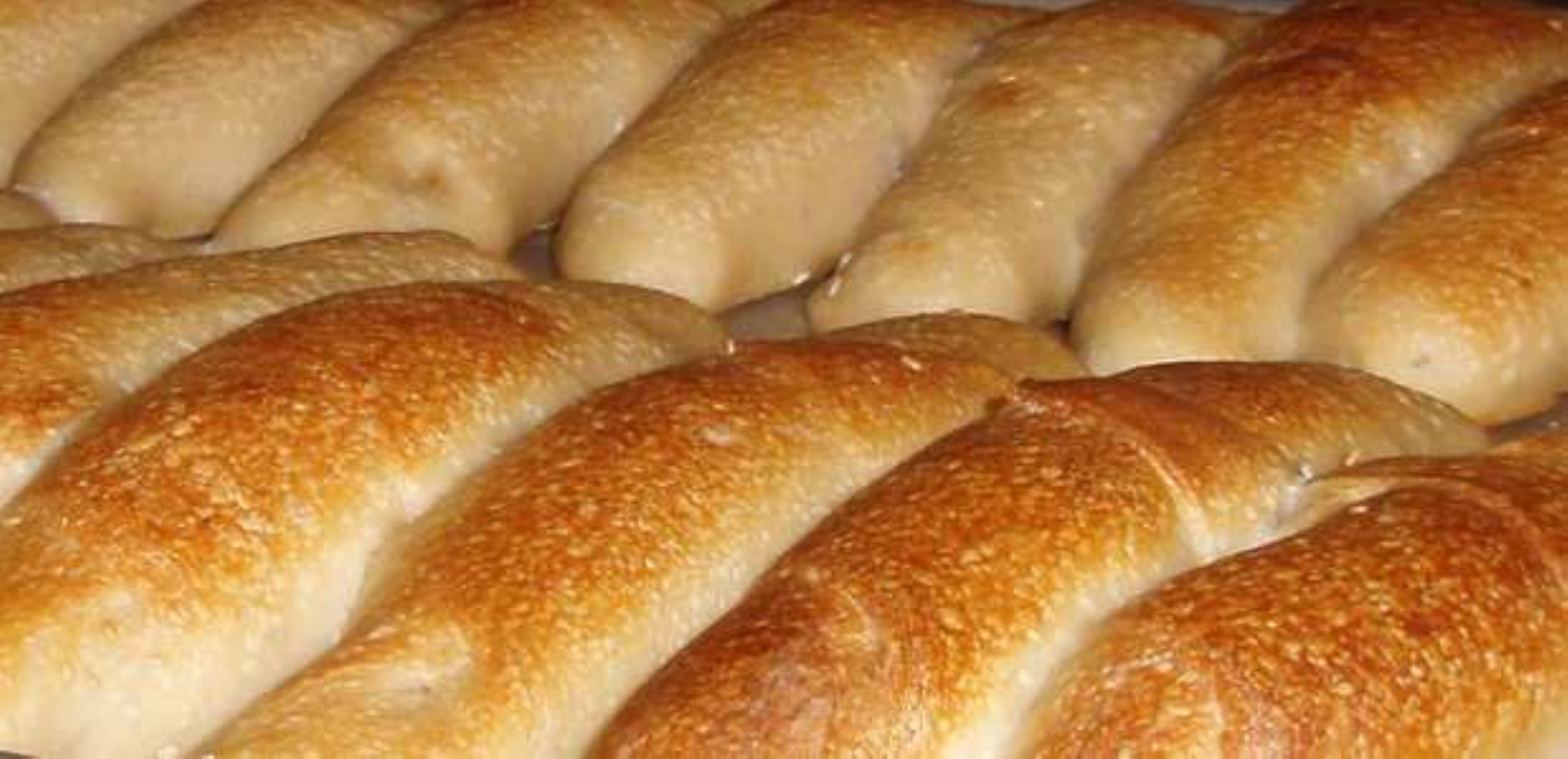,width=0.6in}
}
\caption{\protect\small
Left to right, hanging cloth, part of a sweater, cooked and slices beef, bok choy, toasted bread, leaves,
wood, vase, foam of coffee, sunset, brown banana and bread. 
\label{motivate}}
\vspace{-0.15in}
\end{figure*}

Figure \ref{motivate} shows {\em exemplar} images in which a single attribute varies in the object or its image formation
 leading to
variations in reflectance.  
The hanging cloth simply reflects smooth-surface shading variations (i.e., surface normal variations), 
the sweater conveys surface-roughness variations (i.e., normal variations), the roast beef reflects different
levels of cooking, the bok choy reflects different levels of Chlorophyll 
at the leaves and stems, the toasted bread conveys different levels of 
browning, the leaves show different levels of Chlorophyll withdrawals, the wood conveys different micro-structure variations due to growth patterns,
the vase
reflects different light transmission through glass, 
the foam reflects different coffee presence in a cappuccino drink, 
the sunset conveys variations in light transfer through clouds,
the banana shows different levels of oxidation, and the bread rolls show different levels of browning in the baking process.

While the images in Figure \ref{motivate} are visually diverse and each appears internally quite variable, 
the observation that there is a single
dominant process that caused this variability led us to hypothesize that there is a compact subspace that reflects this
single-cause variation in the image color patterns. 

The contribution of the paper is in identifying a class of seemingly complex objects and scenes that can be represented 
economically using simple parametric models. Specifically, we show that a linear model is sufficient for representing
color variations due to smooth and rough surface shading, and that cubic polynomials are sufficient for representing the
colors of objects that have undergone physical or imaging change  by a single process.

Although there are alternative methods for representing color variations of the images in Figure \ref{motivate}, 
(e.g., histograms), parametric models are more economic, well-constrained and  readily
support {\em interpolation}  and {\em extrapolation} allowing generalization from
exemplars. In contrast, histograms have no interpolation or extrapolation power. 

It is critical to note that we are considering single-process objects and images, and in many real-life images it is likely that
multiple processes operate at once. For example, we will show that while the toasted bread in Figure \ref{motivate} can be
parameterized as a cubic polynomial in RGB space, if drastically different illumination is imposed, a different
cubic polynomial may be needed to represent the colors of the same bread. We show, however, that for generic appearances
  of toasted bread (and baked goods in general) the model is reasonably robust.  It remains open for research to convey, perhaps parametrically,
how the parametric model of toasted bread changes under significantly variable imaging conditions in conjunction with normal
variations in browning. At a minimum, it is possible to represent the diverse appearances of toasted bread under different
illumination conditions as a set of polynomials which is more economic and effective than alternative representations.

\section{Approach}

All the images used in this paper were downloaded from the web and thus imaging conditions and camera 
parameters are unknown. We chose source images larger than 2MP, without blur
and little  JPEG compression degradation.

Let's assume that an exemplar image, $I$, is of an object that has an unknown but uniform {\em base} material that 
was subjected to a single unknown  process, $P$, that caused changes in its light reflectance properties. As
Figure \ref{motivate} shows there is a wide latitude in defining $P$.
In some cases it is an absorption/scattering phenomenon (e,g., the vase and sunset), surface geometry (e.g, hanging cloth, sweater)
while 
in others it is actual micro-structure material variations (e.g., the cooked and baked objects,  bok choy, leaves, bananas).

The image of an object reflects a combination of the impact of the process $P$ and the imaging process.
Unfortunately, these processes may be poorly understood or quantitatively challenging to model. However, qualitative descriptions 
derived from exemplar images may still be useful. It is critical to note that while some processes create
simple sub-spaces of appearance, others do not and in this paper we show conforming processes only. 
We assume single-process variations under illumination
that does not significantly alter the intra-object or inter-image appearance of an object.
 If multiple processes are involved the color appearance space remains open for research.

We treat RGB values as representatives of distinct spectral wavelength (i.e.,red, green and blue), although cameras, in fact, normally output multi-spectral responses.
The color sub-spaces within RGB may be as simple as a single point,  line or curve or as complex as disjoint point clouds
that require rich representations. We focus on line and curve sub-spaces and show that 
they capture  the appearance space of certain classes of objects.


\subsection{Preprocessing exemplar images }

Determining the dimensionality of the RGB data (represented as a vector of 3D points)
is done through eigenspace analysis to uncover if the
data is one or two dimensional within the RGB space (i.e., a line or a plane, respectively).
If the largest eigenvalue captures a very high percentage of the variation in the data then the data is considered
to fit a straight line defined by the respective eigenvector in RGB space.
Similarly, if the largest two eigenvalues
capture most of the variations then the data lies on a 2D plane that is defined by the respective eigenvector directions
 in the RGB space.
Let a {\em Planarity Measure} (PM)=$[v_1,v_2]$ 
express the amount of variation captured by the two largest
eigenvalues, respectively.
A value near 100\% for $v_2$ indicates perfectly  planar data.

\begin{figure}
\centerline{
\psfig{figure=figs/train5.eps,width=1.0in}
\psfig{figure=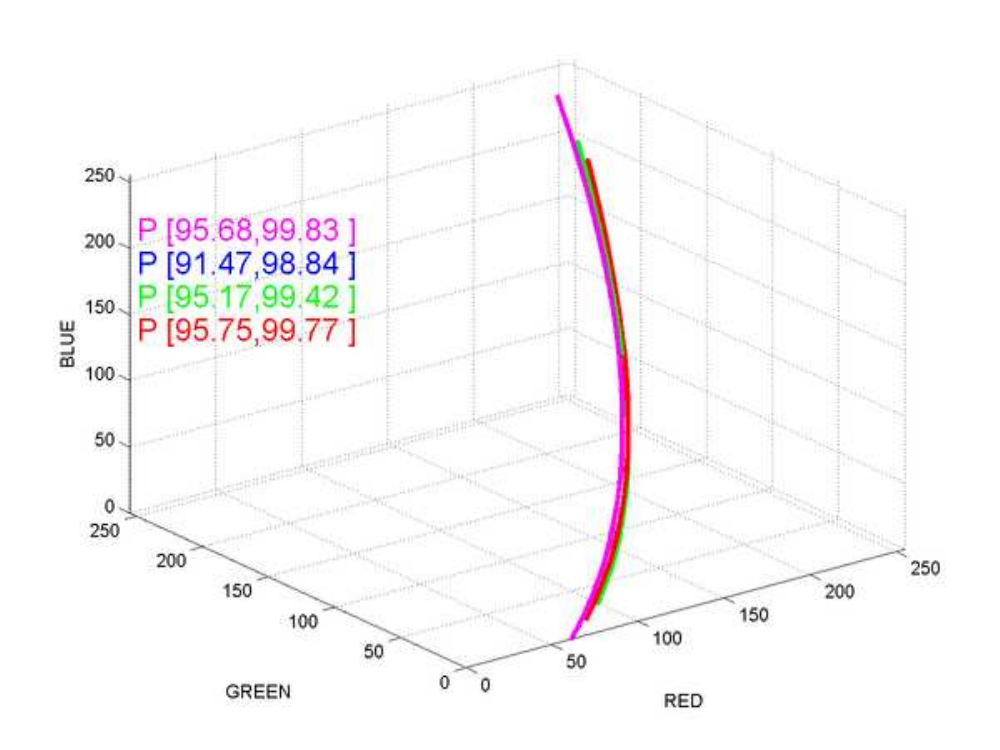,width=1.9in}
}
\caption{\protect\small
PMs and fitted cubic polynomials for quantizations of the toasted bread image using
Minimum Variance Quantization (red), Median Cut (green), octree (blue) and K-means (magenta).
}\label{QUANT}
\end{figure}

An image typically consists of hundreds of thousands of distinct colors
that create a point cloud in RGB space that in a raw form is not revealing. Instead, a reduction of the colors to
a small representative sample improves our ability to assess the color variations of objects and scenes.
There are numerous approaches to color reduction via quantization.
We perform color quantization of the exemplar image  into 256 colors using
four approaches for evaluation purposes:
Minimum Variance Quantization (part of Matlab), Octree Quantization, Median Cut, and K-Means.  
We found no visual differences in the quantization results. 
The computed 256 colors in each quantization were assessed for planarity and
fitted as a cubic polynomial.
Figure \ref{QUANT} shows the PMs and the fitted polynomials for the four approaches applied to the toasted bread image. 
Note that the octree curve is completely covered by the others. The planarity measure is very similar for all 
quantization methods, the data points are clearly planar since the largest two eigenvalues capture upwards of 99\%
of the variation. Therefore, we established that the planarity of the quantization is an attribute of the data and
not a side effect of the quantization algorithms.
In the rest of the paper the Minimum Variance Quantization
approach is used to quantize exemplar images into 256 colors.

\begin{figure*}
\centerline{
\psfig{figure=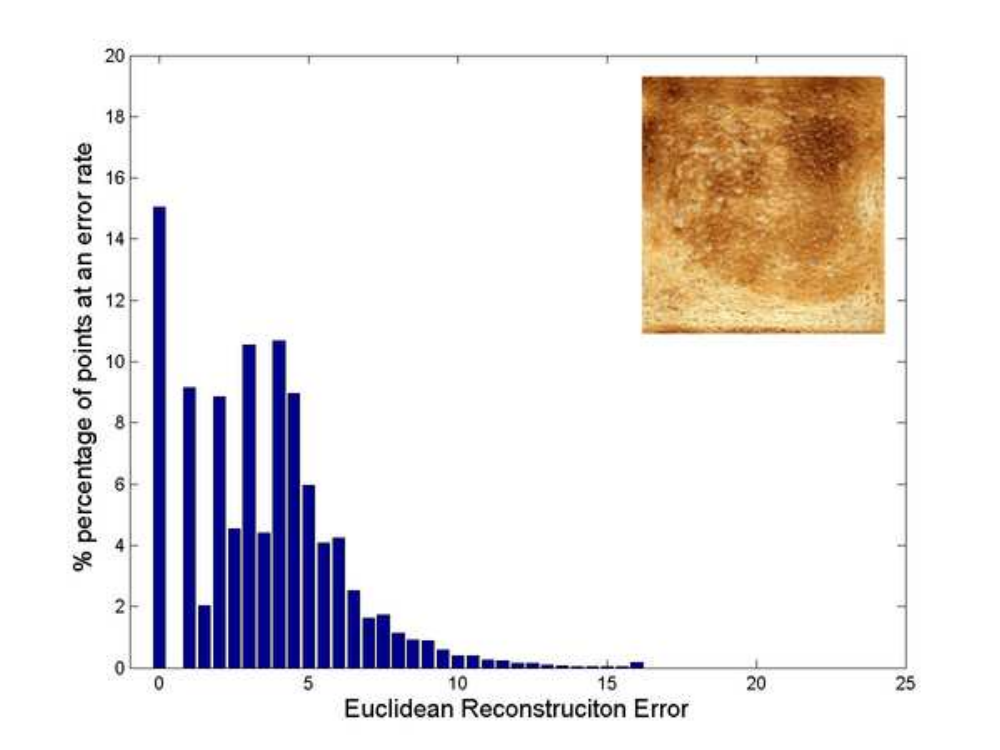,width=1.4in}
\hspace{-0.25in}
\psfig{figure=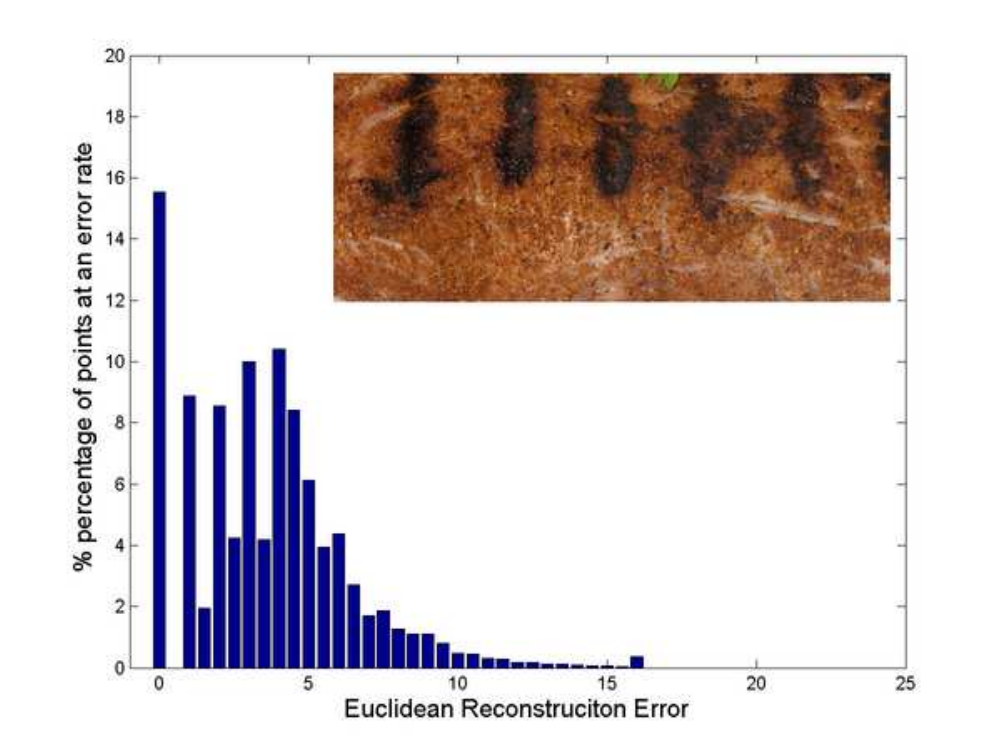,width=1.4in}
\hspace{-0.25in}
\psfig{figure=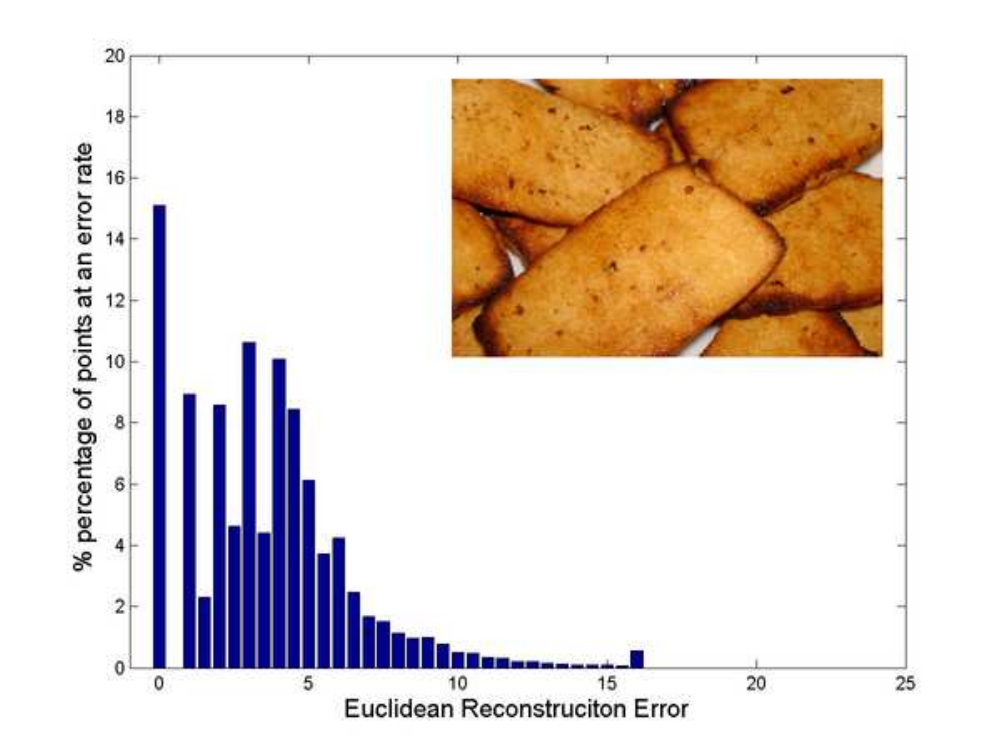,width=1.4in}
\hspace{-0.25in}
\psfig{figure=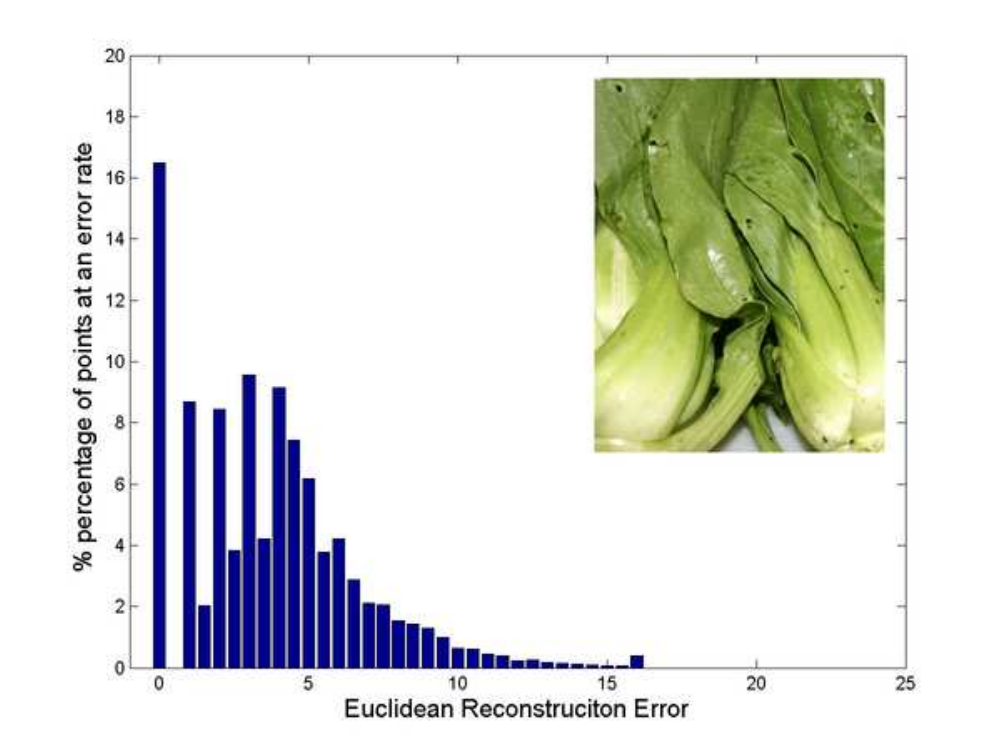,width=1.4in}
\hspace{-0.25in}
\psfig{figure=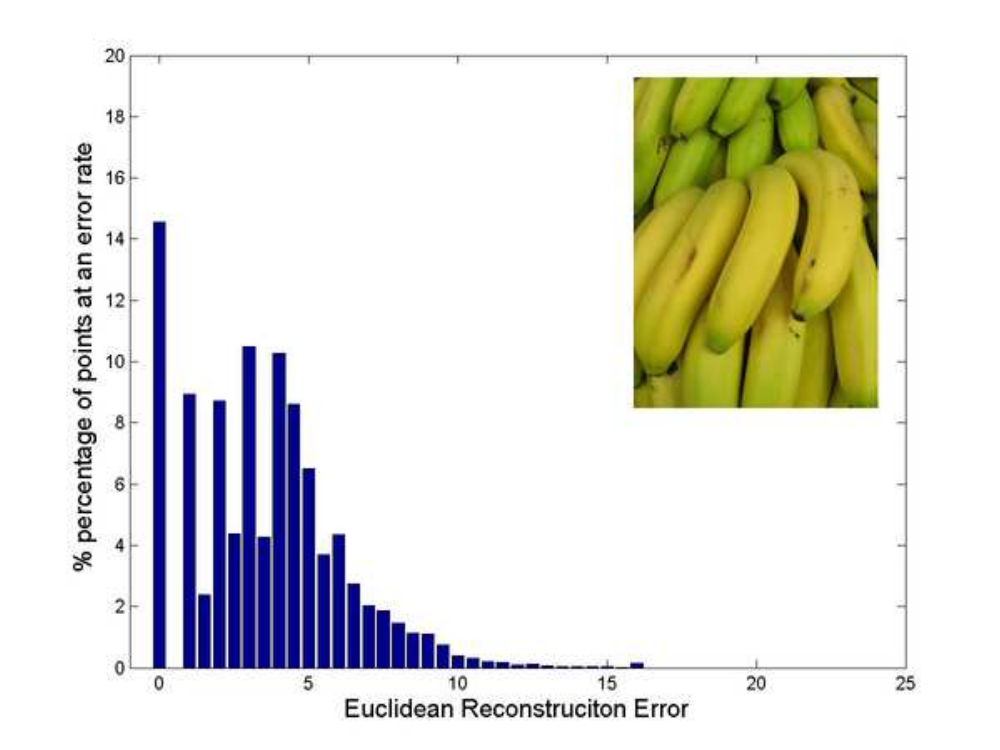,width=1.4in}
\hspace{-0.25in}
\psfig{figure=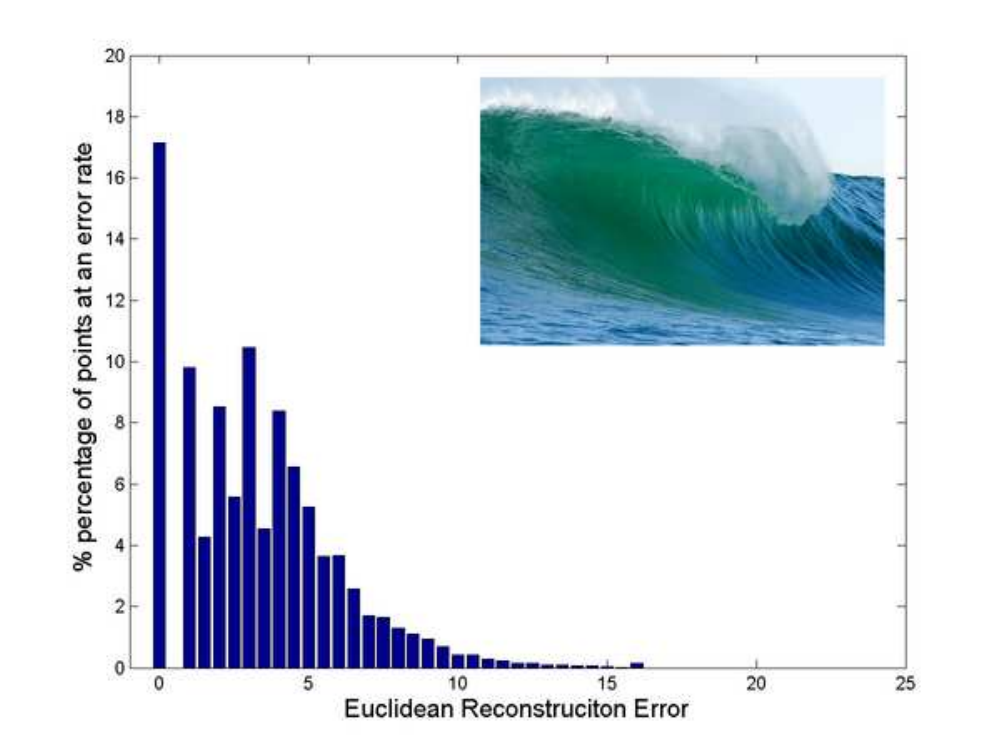,width=1.4in}
}
\caption{\protect\small
Minimum Variance Quantization error for a set of images. The percentage of points at each  Euclidean error distance
between the original and approximated images are  shown  in a form of  histograms.
}\label{MVQ}
\end{figure*}

Figure \ref{MVQ} shows the Euclidean error histogram 
distributions for a set of quantized images. The error measures the reconstruction error
between the input and the quantized RGB values. The percentage of points (with respect to the total number of pixels) 
as a function of the error in Euclidean distance (in RGB space) is shown. The overwhelming majority of the data is reconstructed within
less than 10 units of error. 

Given the quantized exemplar images, we investigate the 3D RGB space that these points span. 
Specifically, we are interested in two geometric sub-spaces. The possibility of points lying on a 3D plane in
RGB space, and the more desirable property of the points lying on a curve on this plane. We consider
only relatively simple cubic  polynomial curves since they proved sufficient to fit the data.
 
Note that only exemplar images of semantically homogeneous materials are quantized. 
Detection and segmentation examples use non-quantized images since the scenes can be 
arbitrarily complex. 

\subsection{The space of shading variations}

Assume an ideally diffusing surface under a distant point source and that light is distributed uniformly independent
of the viewing direction (i.e., Lambertian). 
Let $N_L$ represent a unit vector of the light direction, $F$ represent the flux density of the light source, and
$N_S (x,y)$ be the unit surface normal to the smooth surface at a 3D point that is the source point of (x,y) in
image $I$.
The intensity image, $I(x,y)$ is:  
\begin{equation}
\scriptsize
I(x,y)= F \times  (N_S(x,y) \cdot N_L ) \times  R(x,y)\label{srcSR}
\end{equation}
where $\cdot$ is the dot product and  $ R(x,y)$ is the reflectance of the surface (i.e., albedo in each color channel). This equation can be rewritten as
\begin{equation}
\scriptsize
I(x,y)= L(x,y) \times  R(x,y)\label{basicSR}
\end{equation}
where $L(x,y)$ is the incident illumination also known as shading.
Equation \ref{basicSR} can be rewritten using the 3 RGB wavelengths as,
\begin{equation}
{\scriptsize
\begin{array}{rlc c}
RED(x,y) &= & L(x,y) & F_{red}(x,y)\\
GREEN(x,y) &= & L(x,y) & F_{green}(x,y)\\
BLUE(x,y) &= &  L(x,y) & F_{blue}(x,y)\\
\end{array} \label{EQ_RGB}
}
\end{equation}
where $RED(x,y),GREEN(x,y)$ and  $BLUE(x,y)$ are the R,G, and B wavelength images and $F_{red}(x,y), F_{green}(x,y)$ and $F_{blue}(x,y)$ are
the spectral reflectance of pixels for each channel. The shading image is the same in the three equations since shading is
independent of reflected wavelengths (i.e.,
scene material reflects all wavelengths in the same direction).
As a result we have three equations with four unknown images. 

If we assume an image of a uniform material in terms of reflectance (e.g., the hanging cloth, but not the toasted bread), 
the reflectance functions reduce to 
unknown albedo scalars  ${\alpha}_{red}, {\alpha}_{green}$ and ${\alpha}_{blue}$ that can be determined up-to a scale parameter. 
It is clear from Equation \ref{EQ_RGB} that it should pass through the origin $(0,0,0)$ regardless of albedo values, therefore,
equation \ref{EQ_RGB} reduces to 
an equation of a line in 3D that passes through the origin:
\begin{equation}
\scriptsize
{{RED(x,y) -0}\over{{\alpha}_{red}}} = {{GREEN(x,y) -0}\over{{\alpha}_{green}}} = {{BLUE(x,y) -0}\over{{\alpha}_{blue}}}. 
 \label{EQ_RGB_LINE}
\end{equation}

\begin{figure}
\centerline{
\psfig{figure=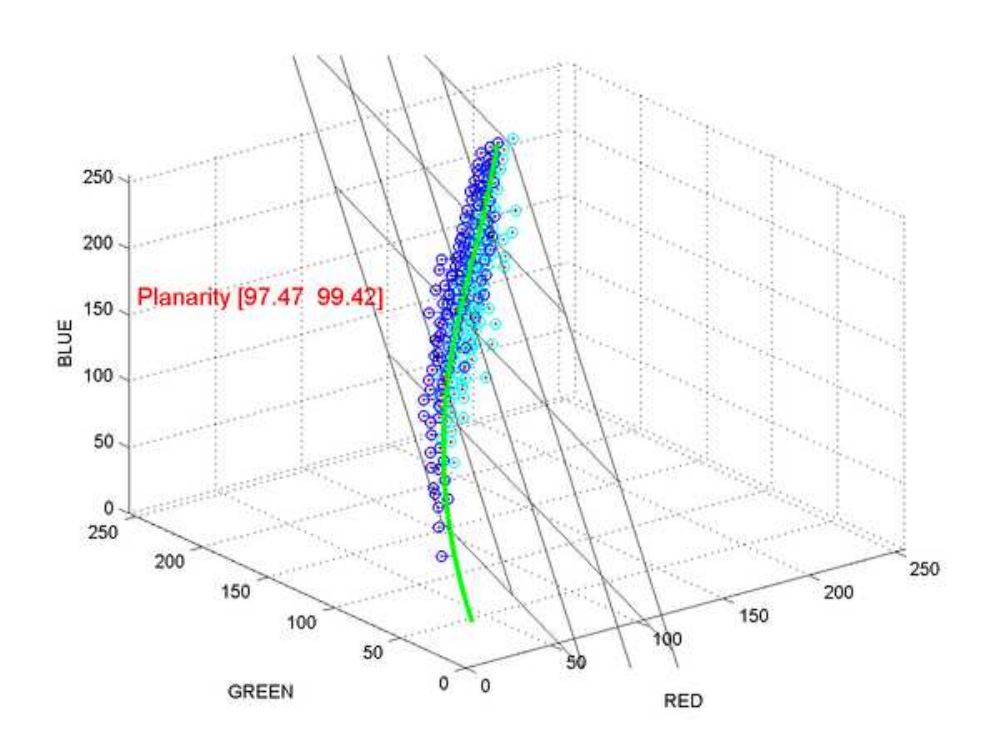,width=2.0in}
\hspace{-0.5in}
\psfig{figure=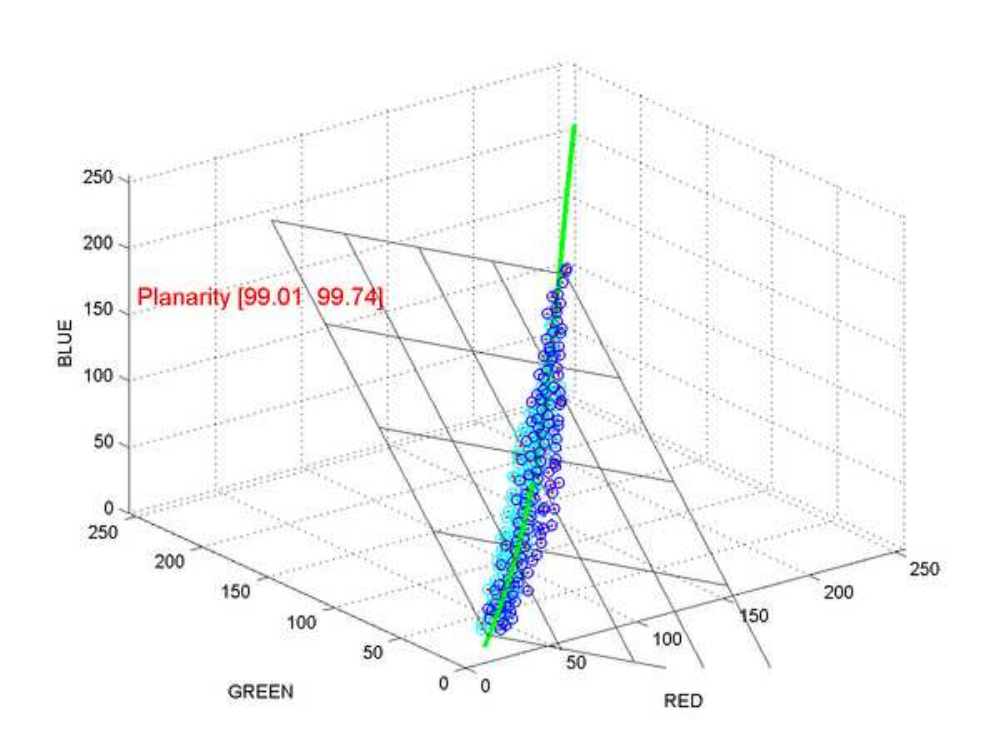,width=2.0in}
}
\caption{\protect\small
The quantized RGB values of the hanging cloth (left) and sweater (right)
 that reflect  single variables, smooth surface shading (i.e., surface normal) and roughness shading which lie on
a curve that can be approximated as a line. 
\label{cloth}}
\vspace{-0.15in}
\end{figure}

This can be observed in Figure \ref{cloth} (left) for the hanging cloth. Similar analysis  can be applied to
objects that have shading variations due to surface roughness (Figure \ref{cloth} (right)), 
since they involve a different pattern of change in the surface normal
(i.e., non smooth change over local spatial areas).  The curves in Figure \ref{cloth} are fit as cubic polynomial
curves and not  lines and therefore bending is visible.
For the hanging cloth PM=[97.47,99.42] and the sweater 
 PM=[99.01,99.74] suggesting a single vector is sufficient for describing the data.

\begin{figure}
\centerline{
\psfig{figure=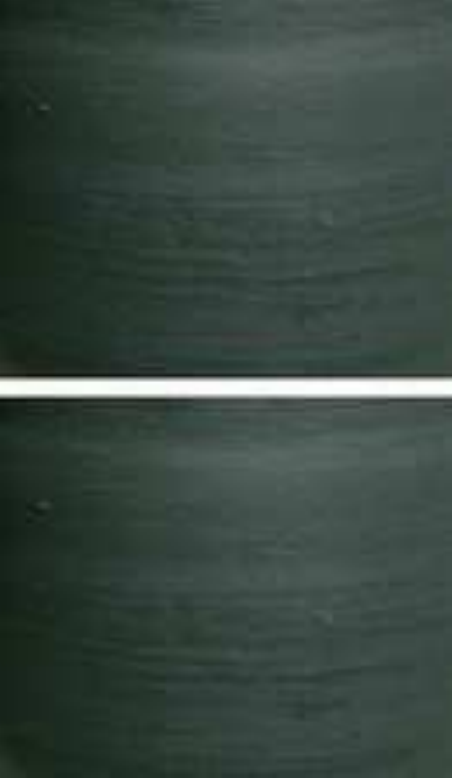,width=0.6in}
\psfig{figure=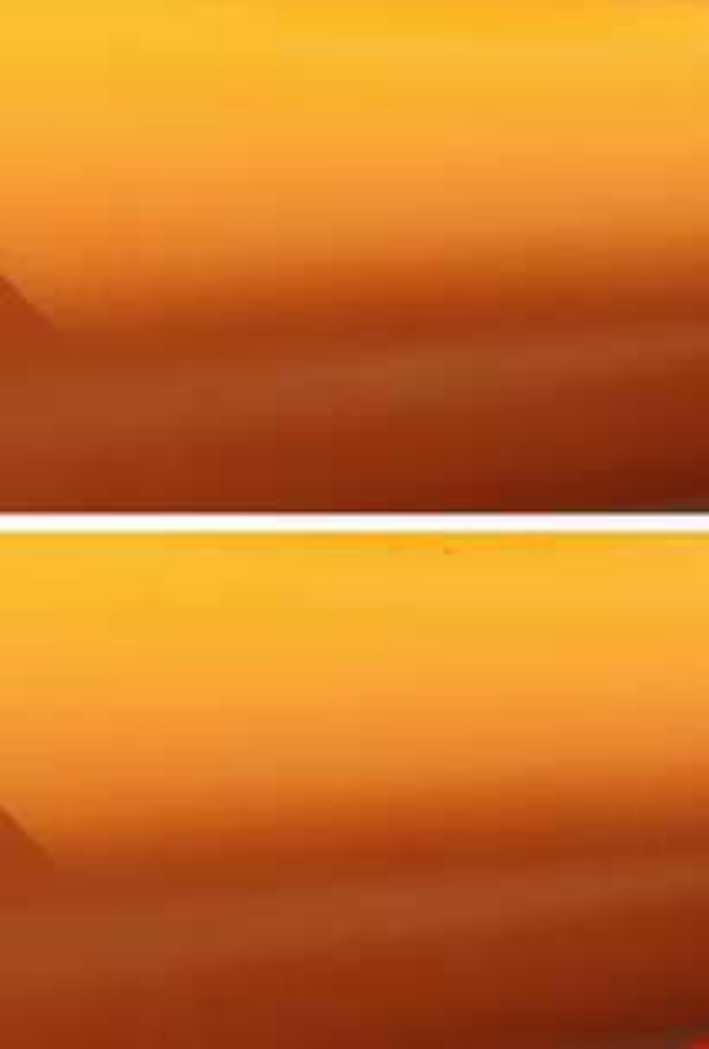,width=0.7in}
\psfig{figure=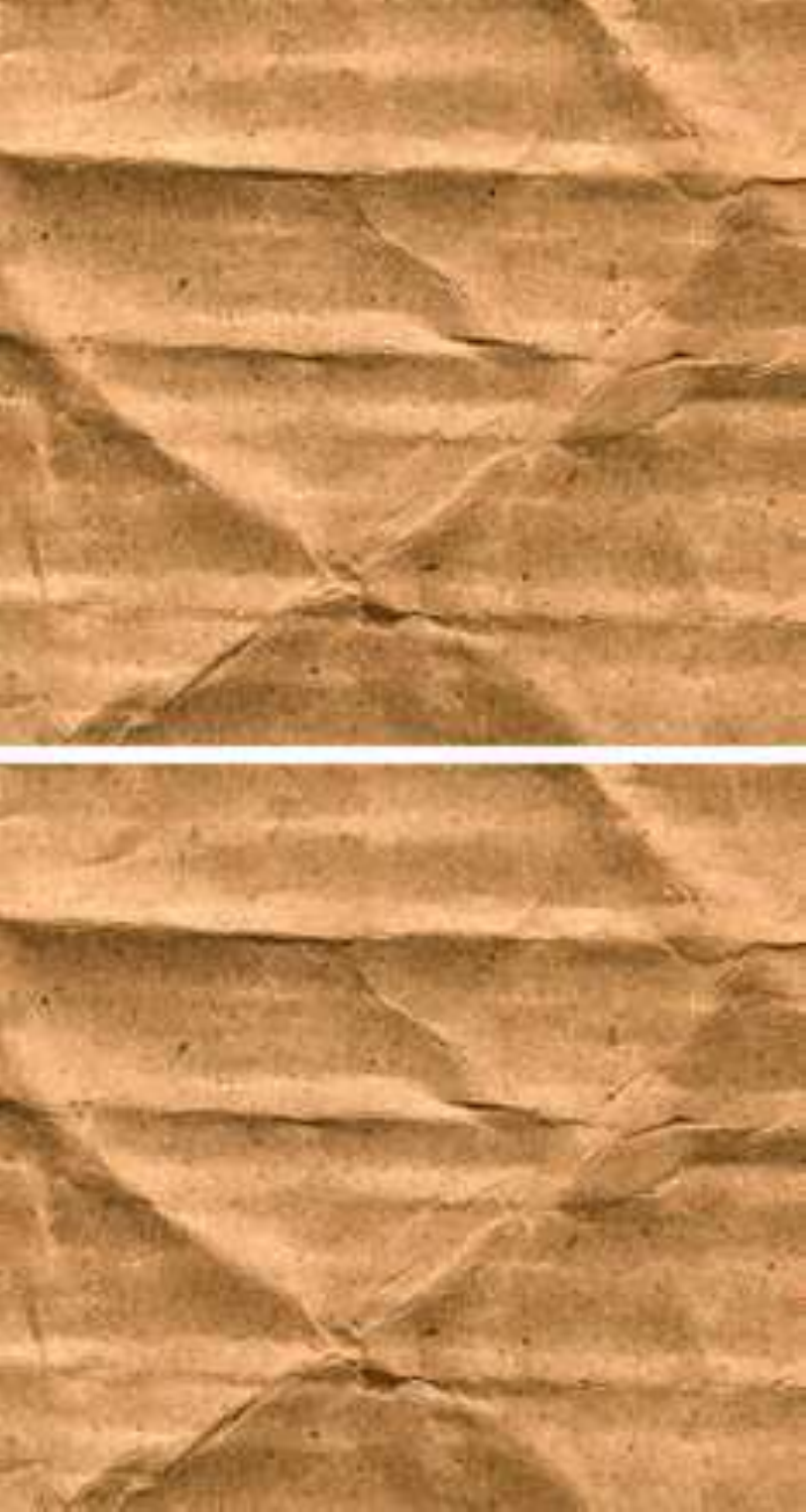,width=0.55in}
\psfig{figure=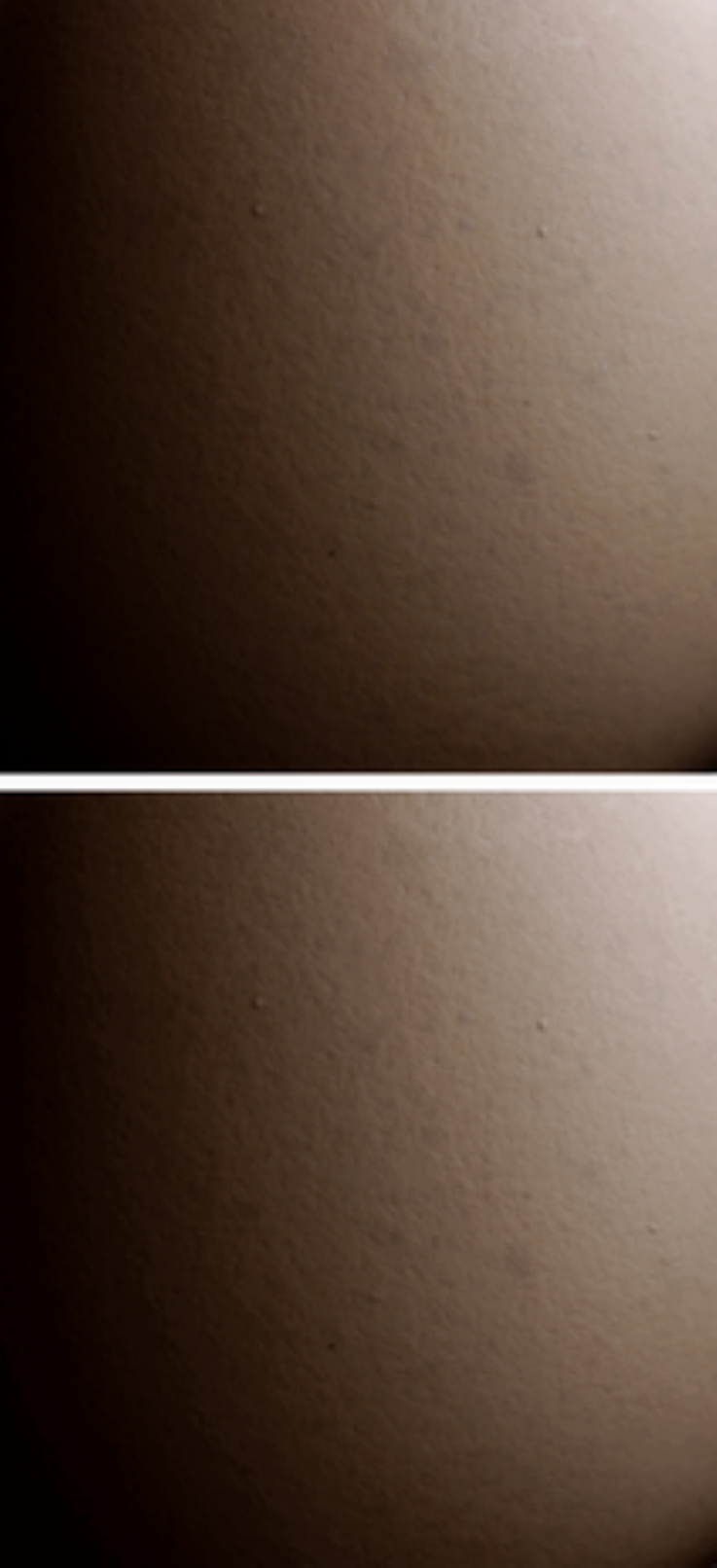,width=0.5in}
\psfig{figure=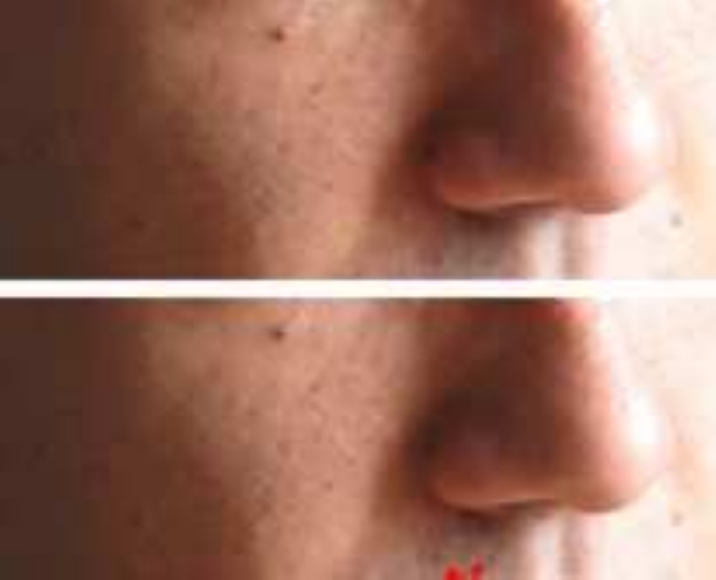,width=0.7in}
}
\centerline{\tiny PM[98.18,99.71]    \hspace{0.1in} PM[96.37,98.74] \hspace{0.1in} PM[97.23,99.57] \hspace{0.1in} PM[99.54,99.98] \hspace{0.1in} PM[97.91,99.89]}
 \centerline{
\psfig{figure=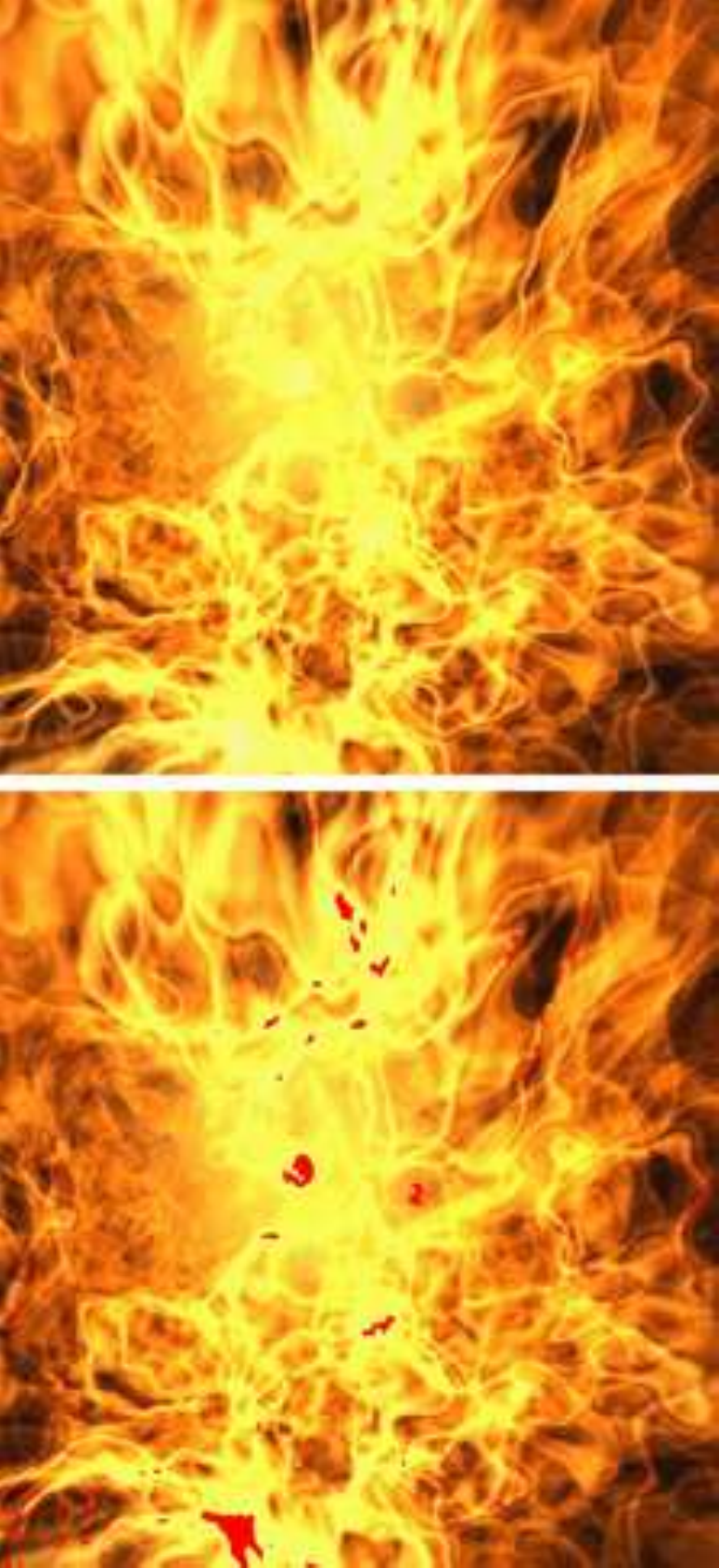,width=0.6in}
\psfig{figure=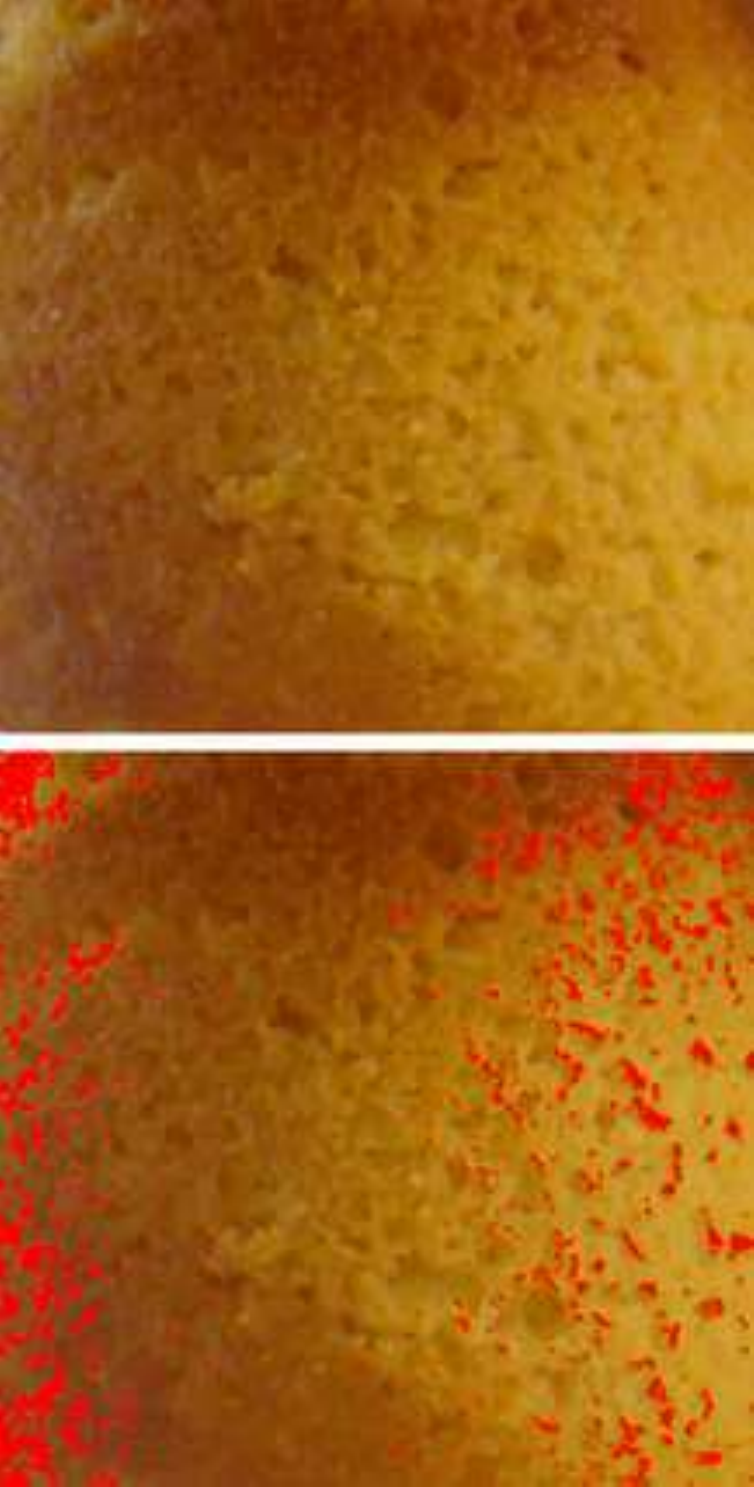,width=0.65in}
\psfig{figure=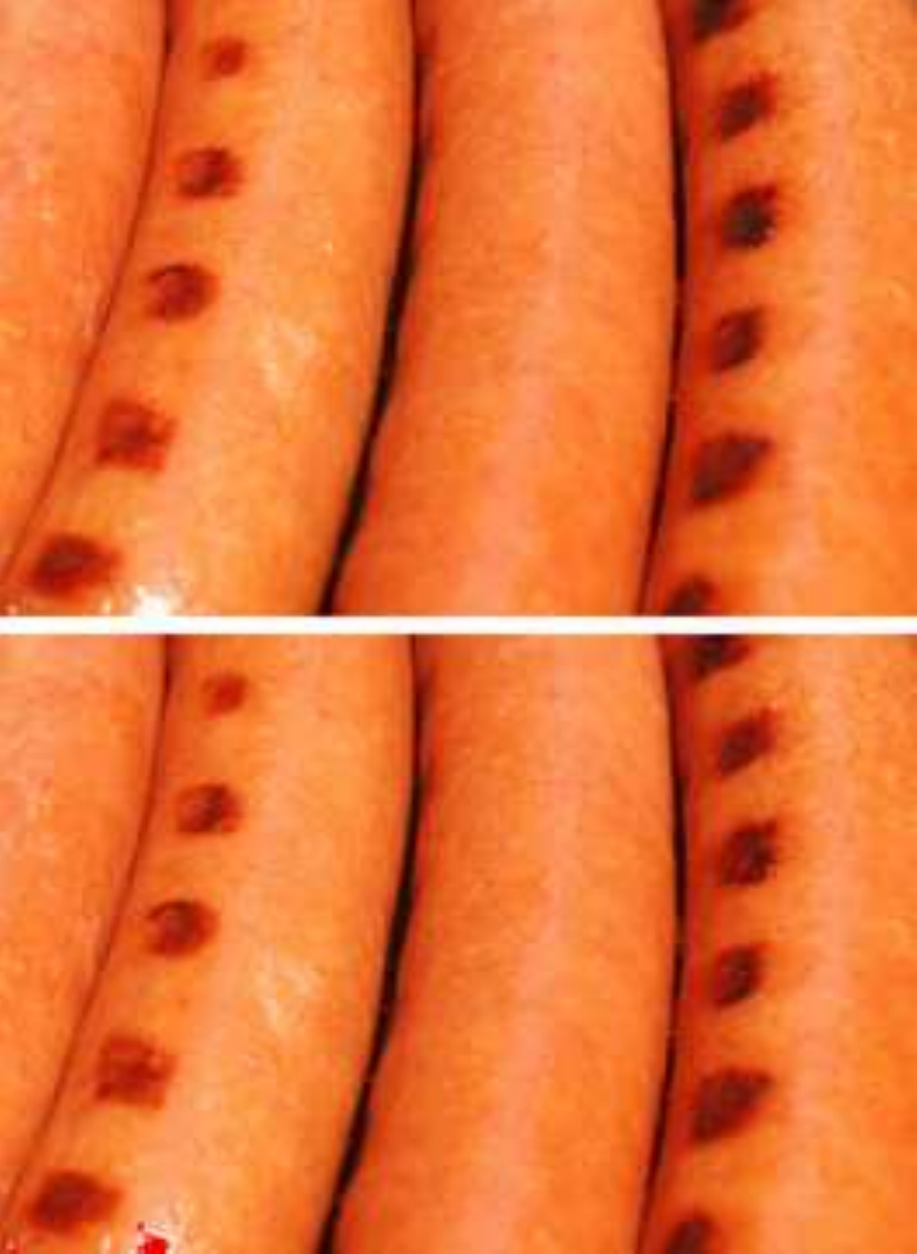,width=0.8in}
\psfig{figure=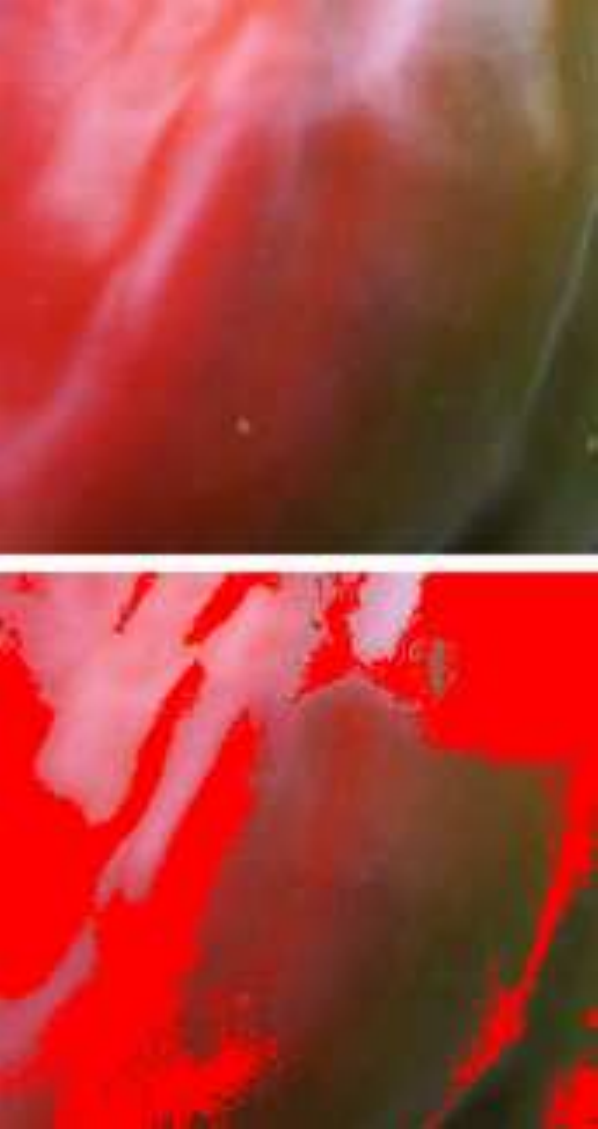,width=0.7in}
}
\centerline{\tiny PM [88.26,98.64]    \hspace{0.3in} PM [76.41,98.86] \hspace{0.3in} PM [85.30,98.9] \hspace{0.3in} PM [82.90,99.01]}
\centerline{
\psfig{figure=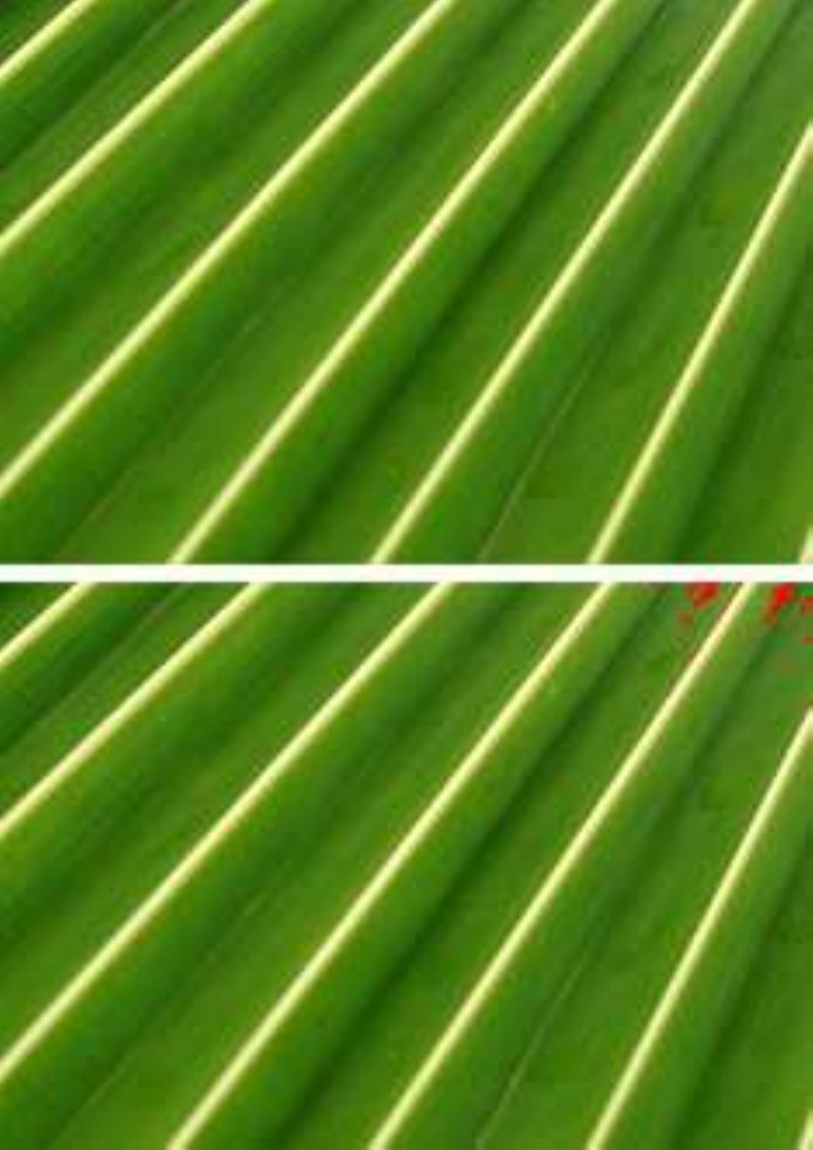,width=0.8in}
\psfig{figure=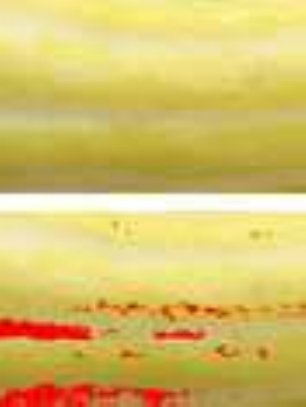,width=0.8in}
\psfig{figure=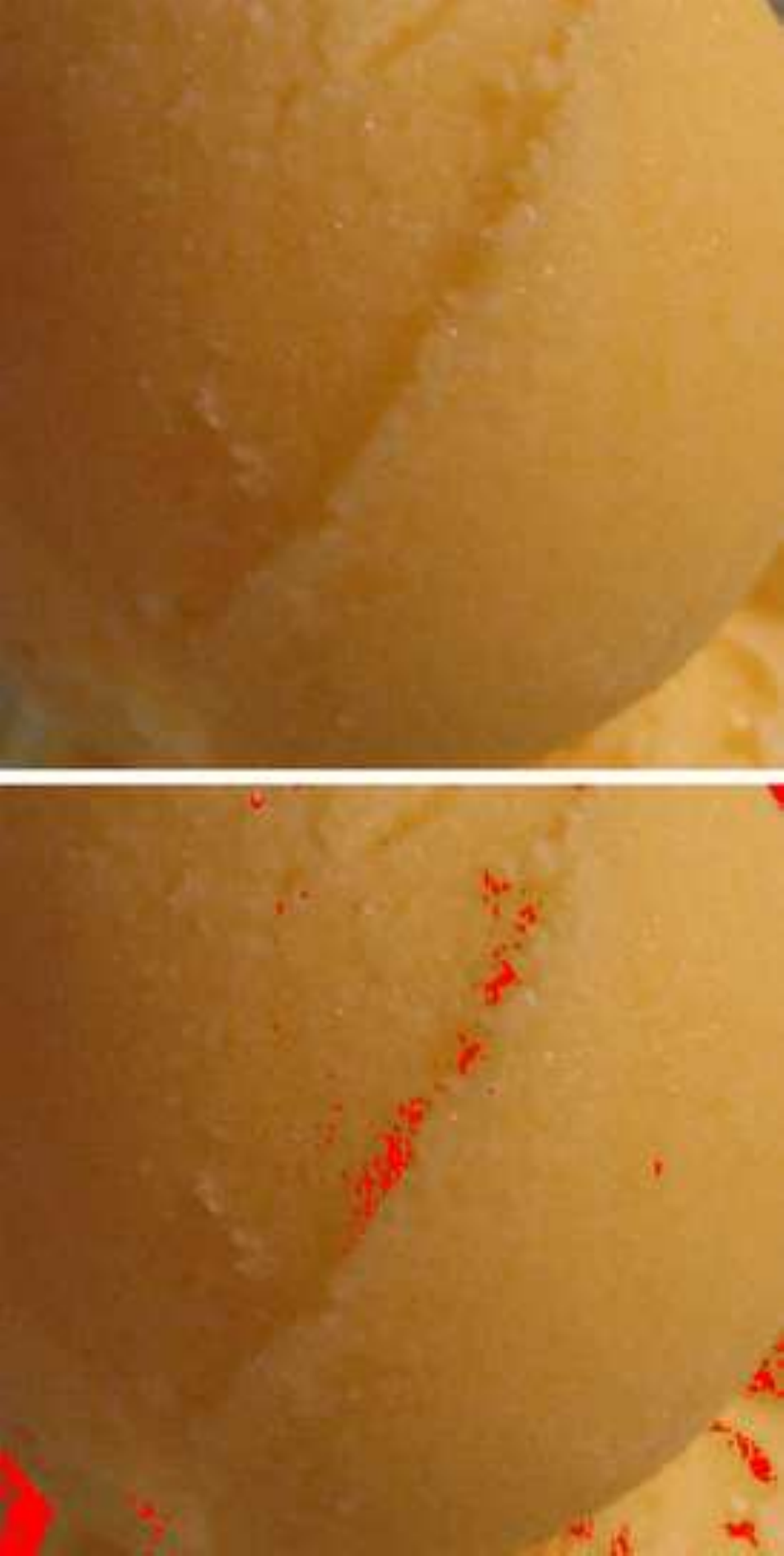,width=0.6in}
\psfig{figure=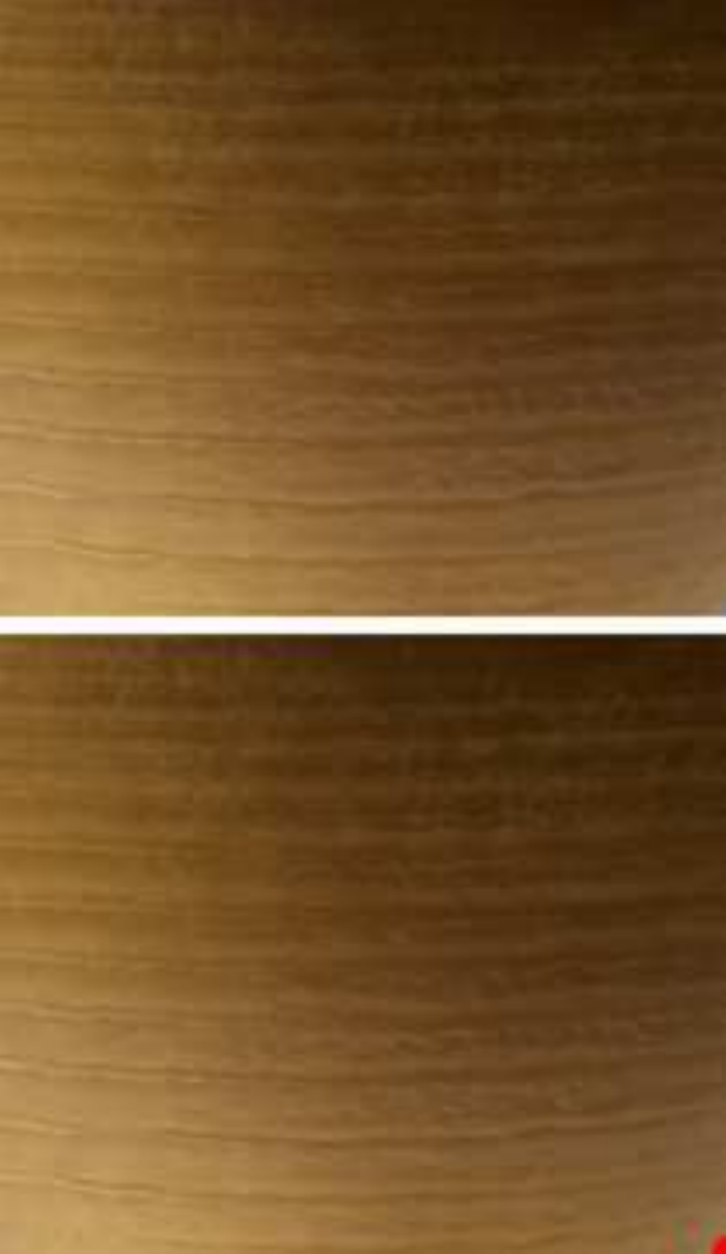,width=0.7in}
}\centerline{\tiny PM [96.52,99.66] \hspace{0.2in} PM [90.9,99.84] \hspace{0.2in} PM [91.28,99.34] \hspace{0.2in} PM [96.74,99.82]}
\caption{\protect\small
PMs and outlier pixels for images that are shading-change or reflectance-change.
Top row, shading-change examples for a clay cylinder, metal cylinder, cardboard, eggshell and human face.
Middle row, fire, cake, grilled hot dogs and green/red pepper.
Bottom row, palm leaf, yellow/white squash, sorbet ice cream and conic coral.
\label{shading_sample}}
\vspace{-0.15in}
\end{figure}

The linear nature of shading colors suggests that it can be used to classify image patches into shading-change
versus reflectance-change induced images \cite{land}. Figure \ref{shading_sample} shows planarity measures and the outlier
pixels (marked in red) for different objects and imaging conditions. 
The top row shows color variations due to shading-change. 
From left to right, a gray clay cylinder, orange painted metallic cylinder, dented cardboard, egg-shell and part of a face (the last
two are under strong illumination).
The egg-shell matte surface leads to near perfect linearity with the largest eigen-vector capturing 99.54\% of the variation in the data.
The other objects also show similar behavior despite some interference of specular reflections (orange-painted metal object) and skin
markings (for the face). The middle row shows examples where reflectance-change is the dominant factor. From left to right, images of
fire, cake, grilled hot-dogs and green-red pepper. The PMs  suggest that the colors are not linear since the top
eigenvector captures less than 90\% of the variations. The bottom row shows examples in which the data is more ambiguous.
The left-most image shows part of a palm-leaf, and although the yellow-lines
are presumed to result from different reflectance, the data is strongly linear and the colors appear to lie on the shading-change 
line
as well.  The yellow/white squash shows reflectance change, however it exhibits some linearity since the white pixels 
can also be interpreted as shading change of the yellow surface. 
The mango sorbet is expected to show just shading-change but due to specular reflections of the ice
the first eigen-vector captures only 91.28\% of the variations. Finally, the conic coral surface is genuinely ambiguous, on one hand the
shading change is evident in the first eigenvector capturing 96.74\% of the variations, but the coral could also have experienced
reflectance change (we could not determine the correct interpretation from the source image).

The examples in Figure \ref{shading_sample} illustrate that diversity of visual interpretations may occur when context and 
auxiliary information (e.g., light direction) are absent. Nevertheless, it is clear that the likelihood of correct interpretation of shading
versus reflectance changes is high and therefore it can be used as a basis for bottom-up image interpretation.

\subsection{Browned and cooked foods}

\begin{figure*}
\centerline{
\psfig{figure=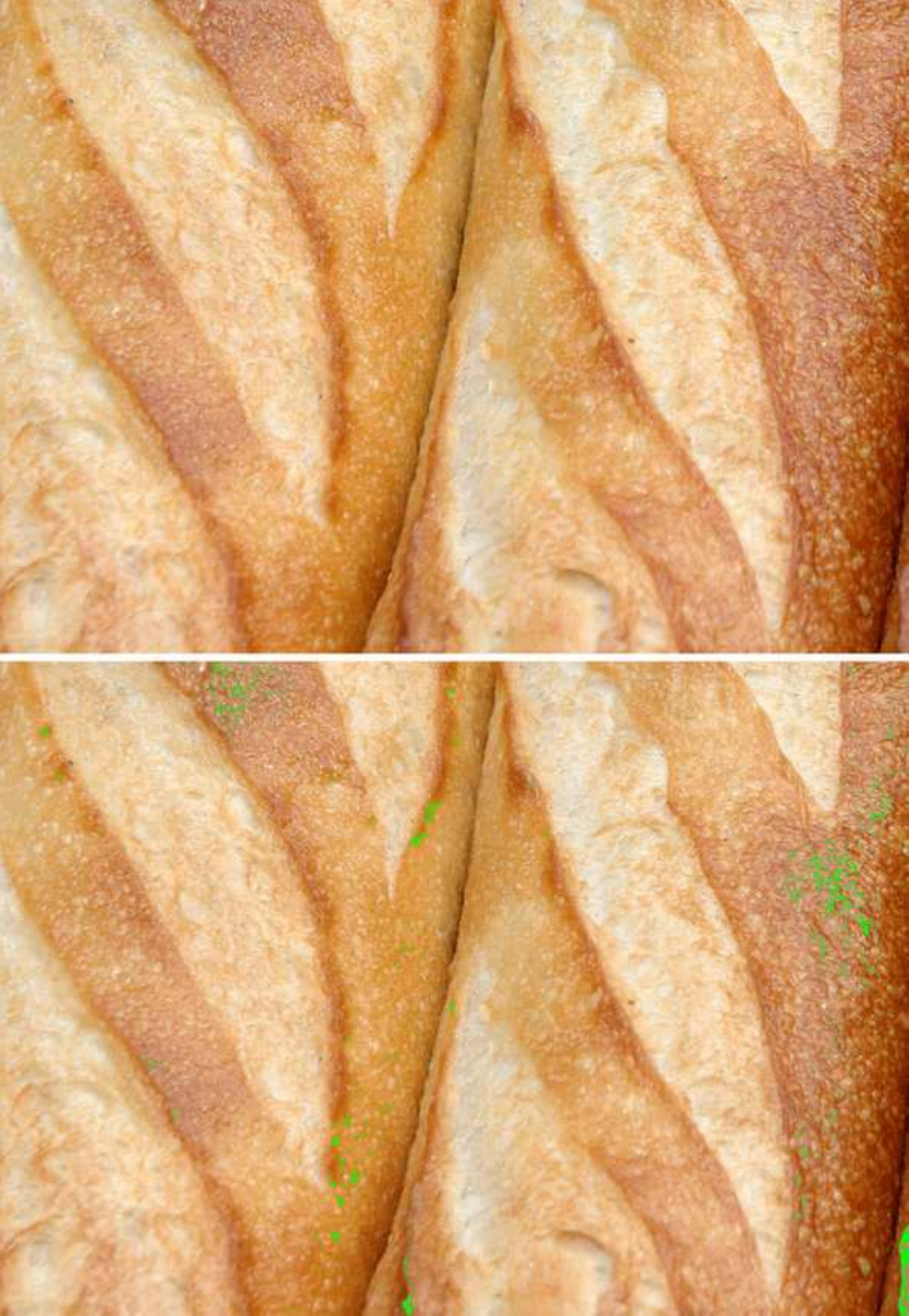,width=0.6in}
\psfig{figure=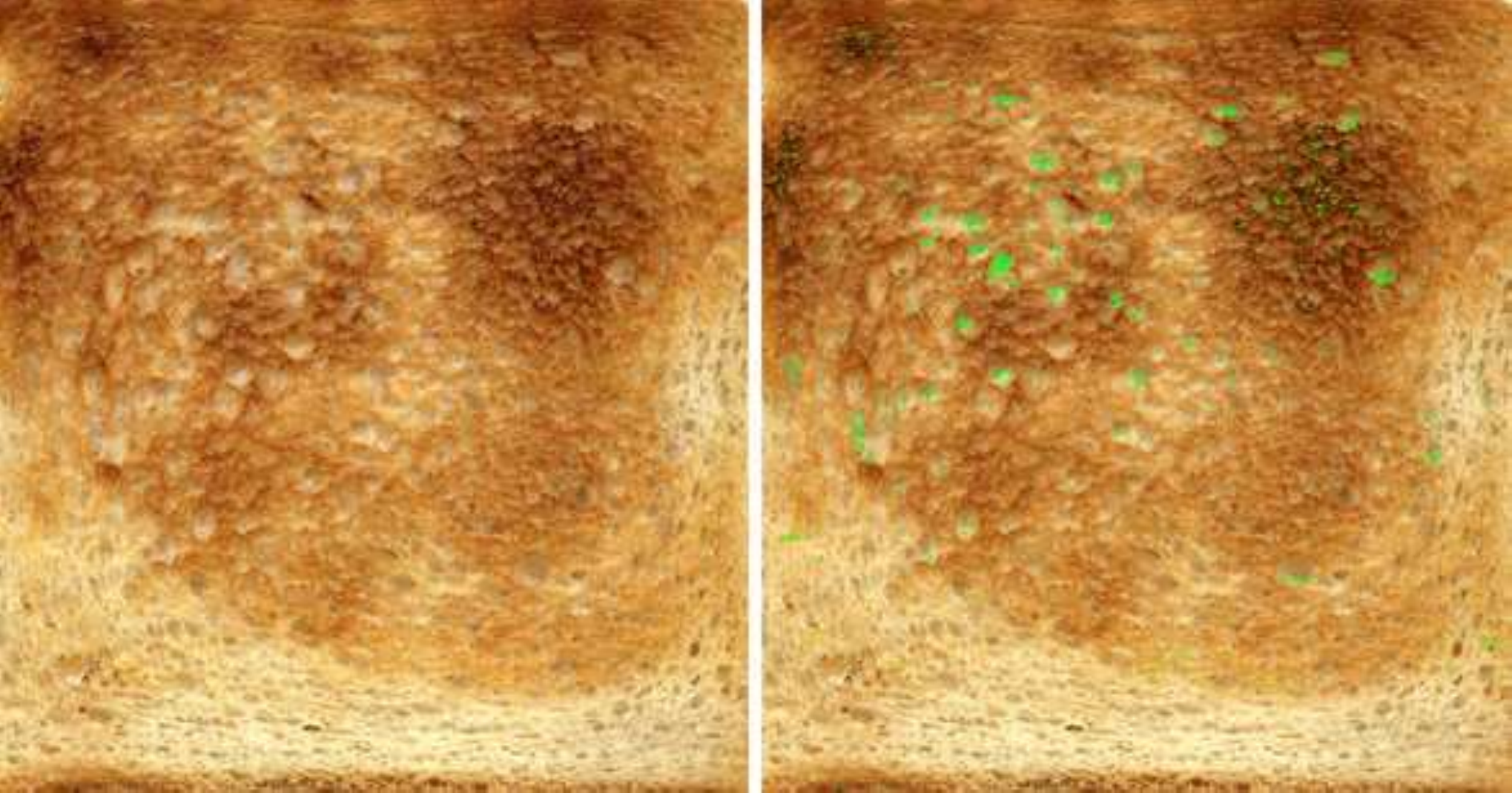,width=1.7in}
\psfig{figure=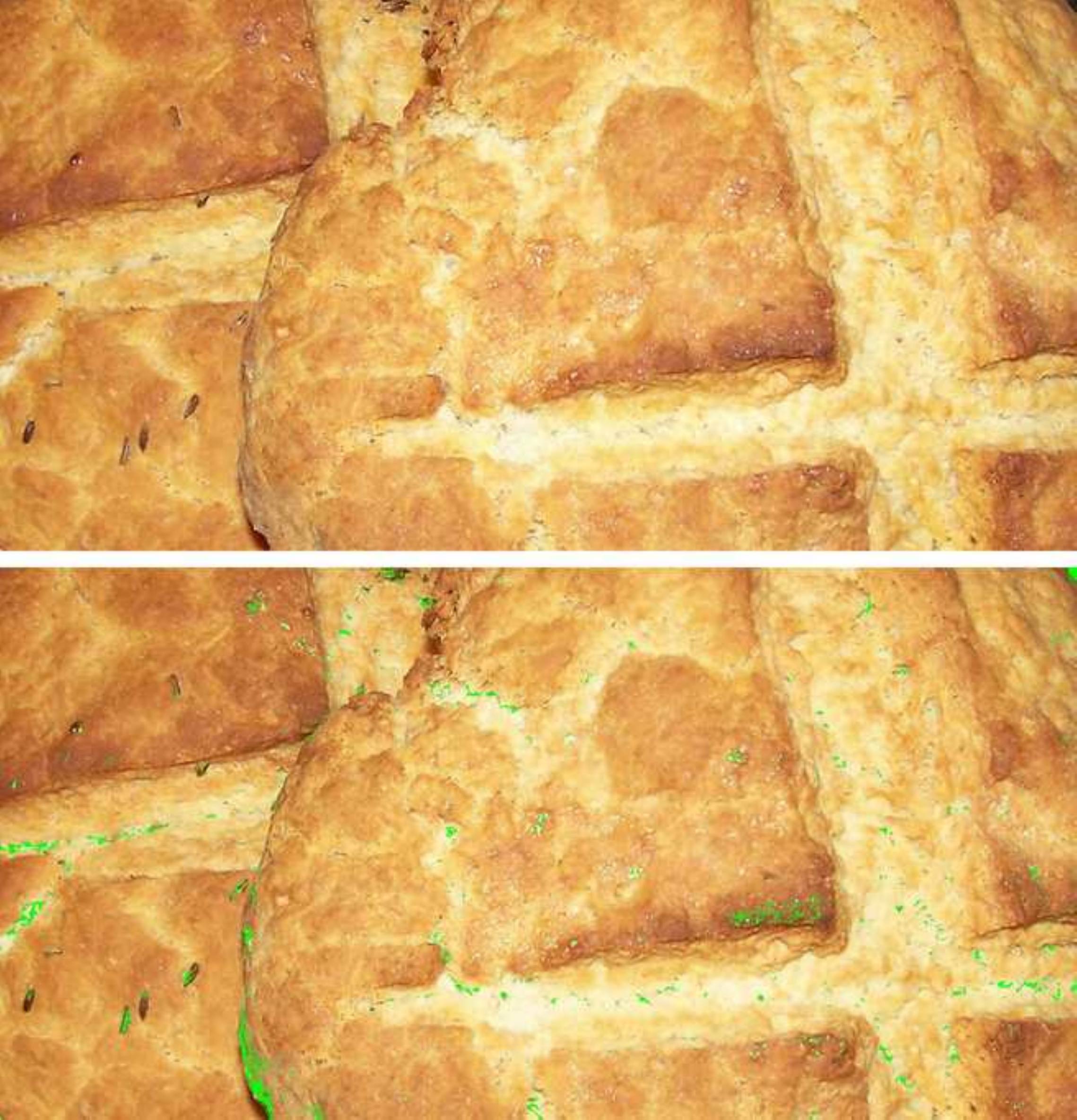,width=0.9in}
\psfig{figure=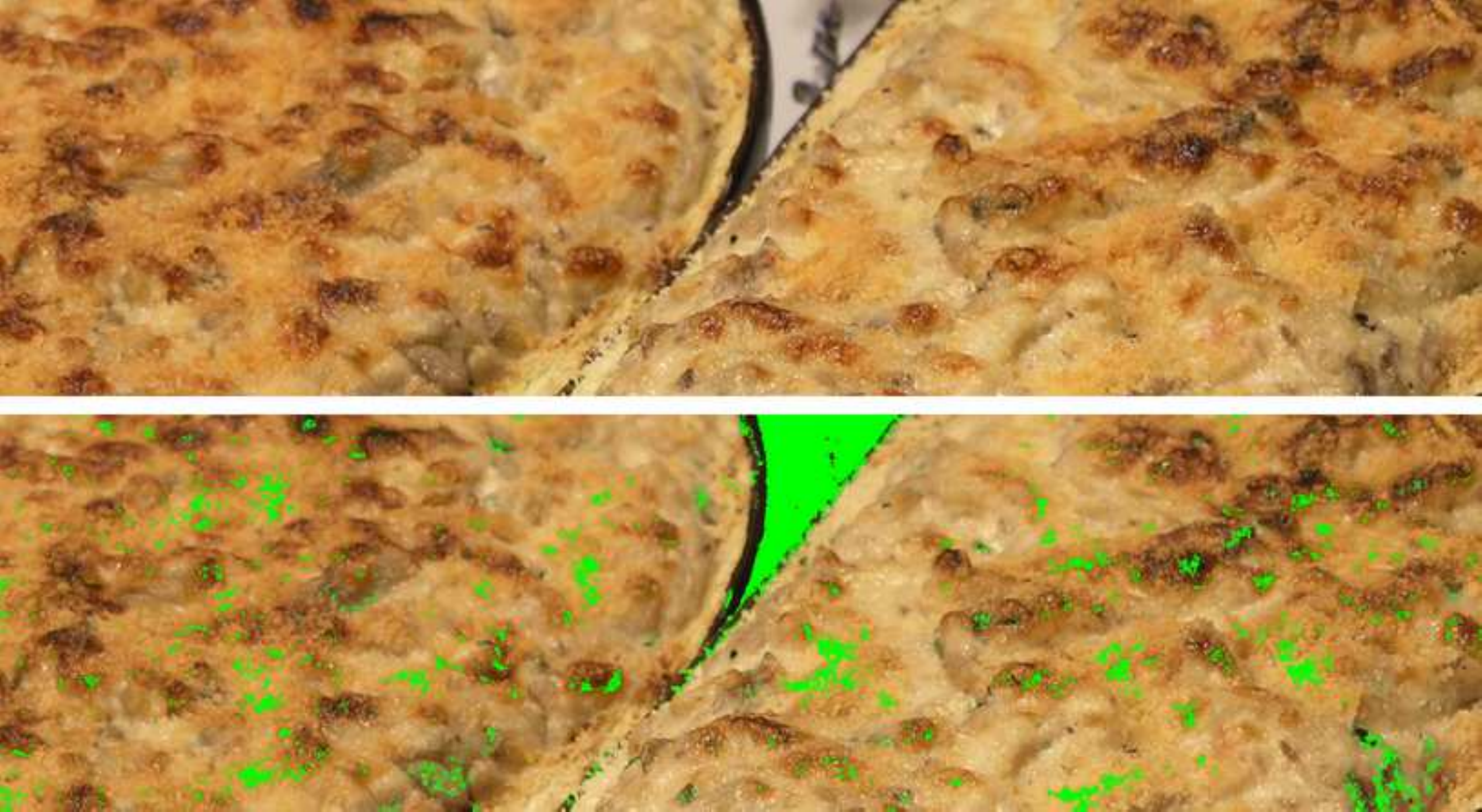,width=1.1in}
\psfig{figure=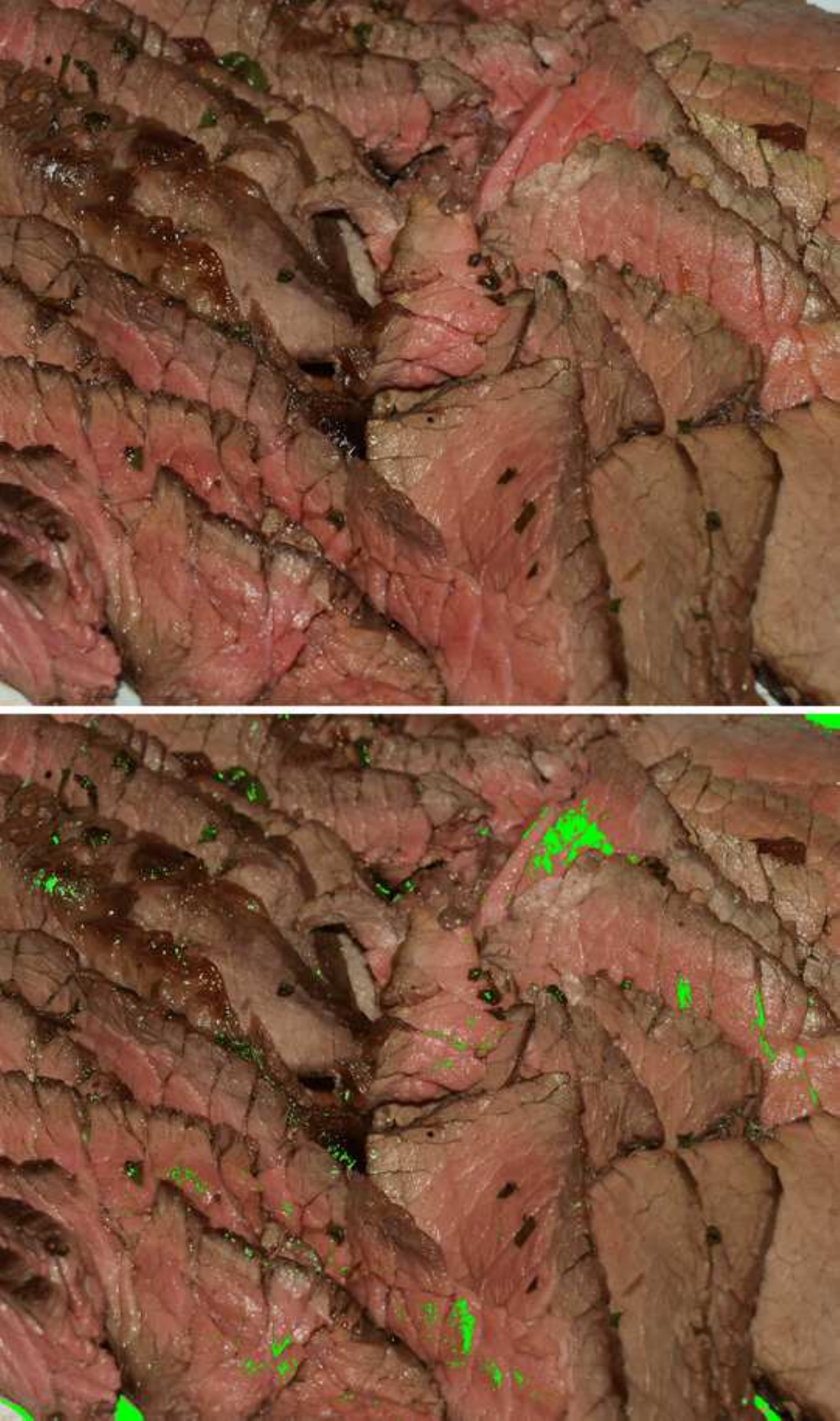,width=0.6in}
\psfig{figure=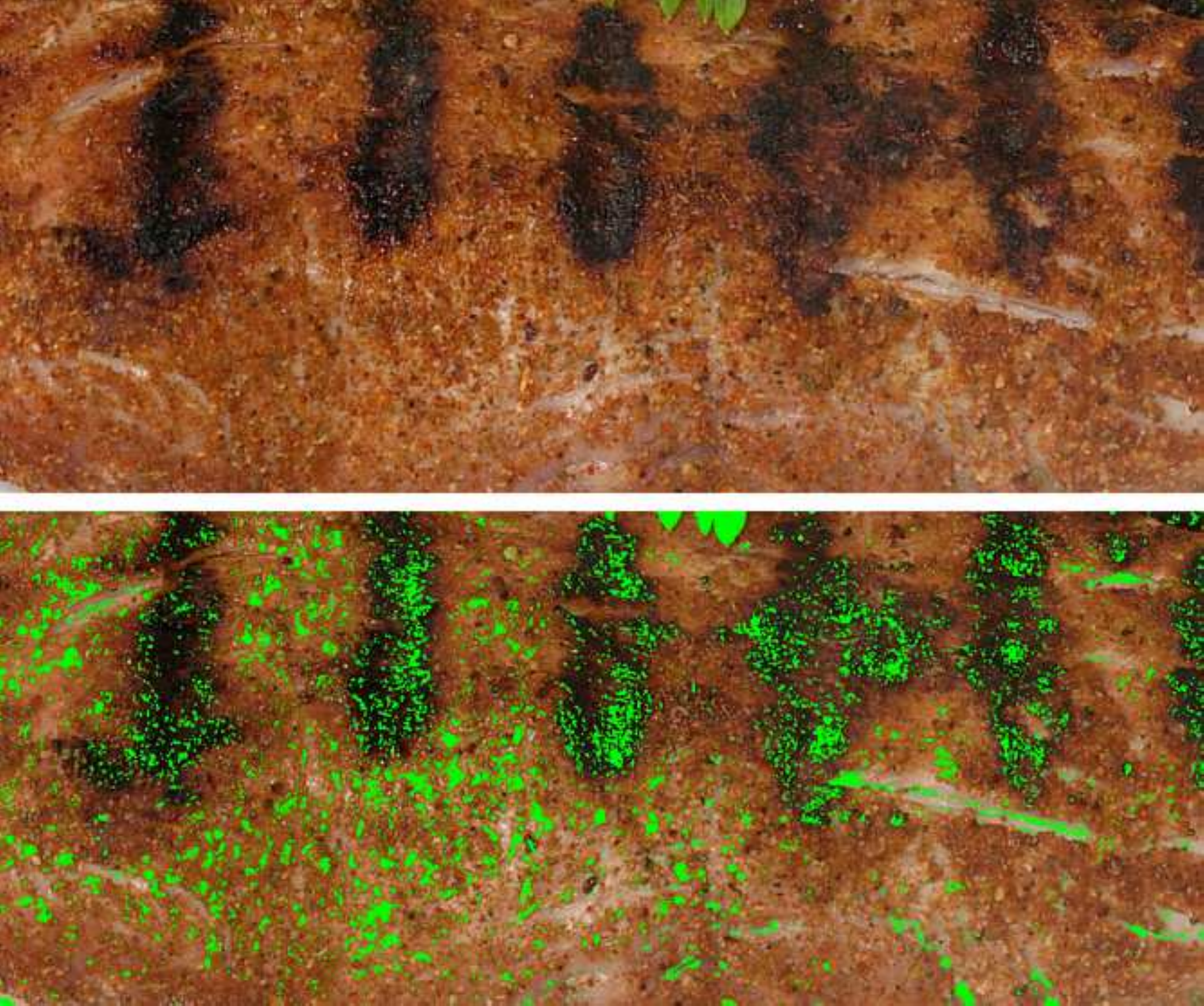,width=1.0in}
\psfig{figure=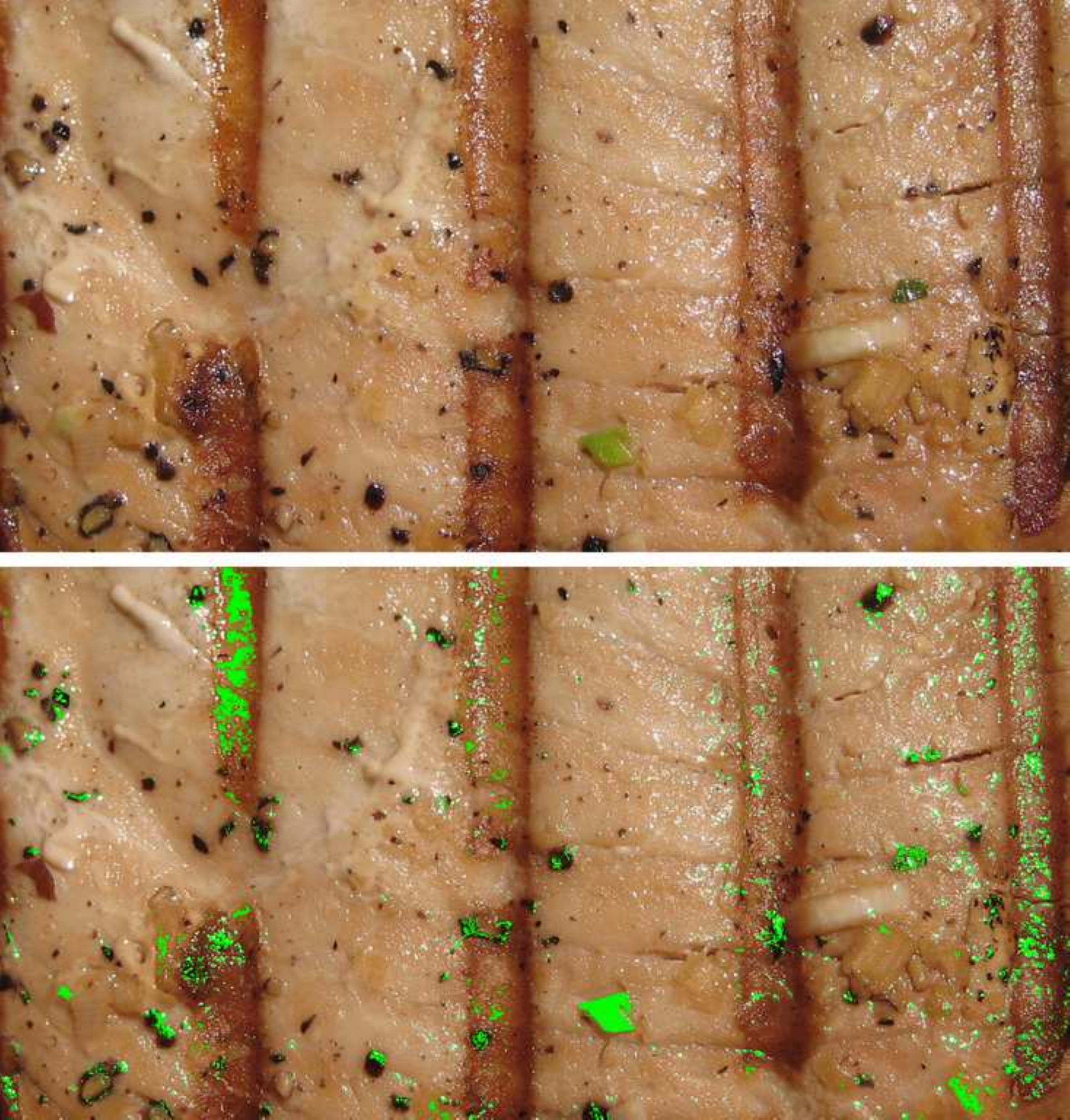,width=1.0in}
}
\centerline{\tiny \hspace{-0.4in} PM[94.52,99.48] \hspace{0.9in} PM[95.75,99.77] \hspace{0.4in} PM[93.03,99.11]   \hspace{0.4in} PM[89.17,99.54] \hspace{0.6in}  PM[94.32,99.47]    \hspace{0.4in} PM[86.82,98.37] \hspace{0.4in} PM[94.83,99.34]}
\centerline{
\psfig{figure=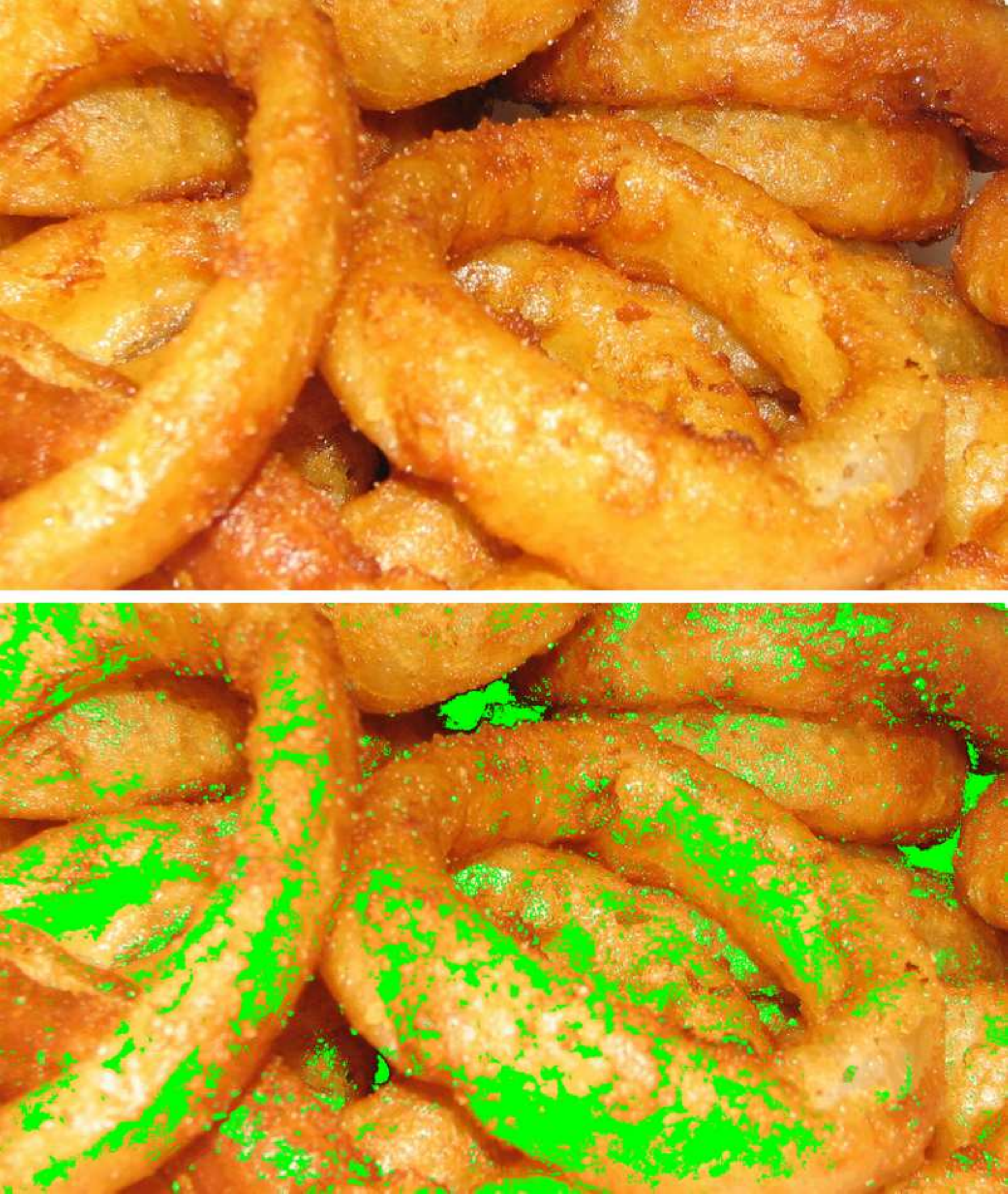,width=0.6in}
\psfig{figure=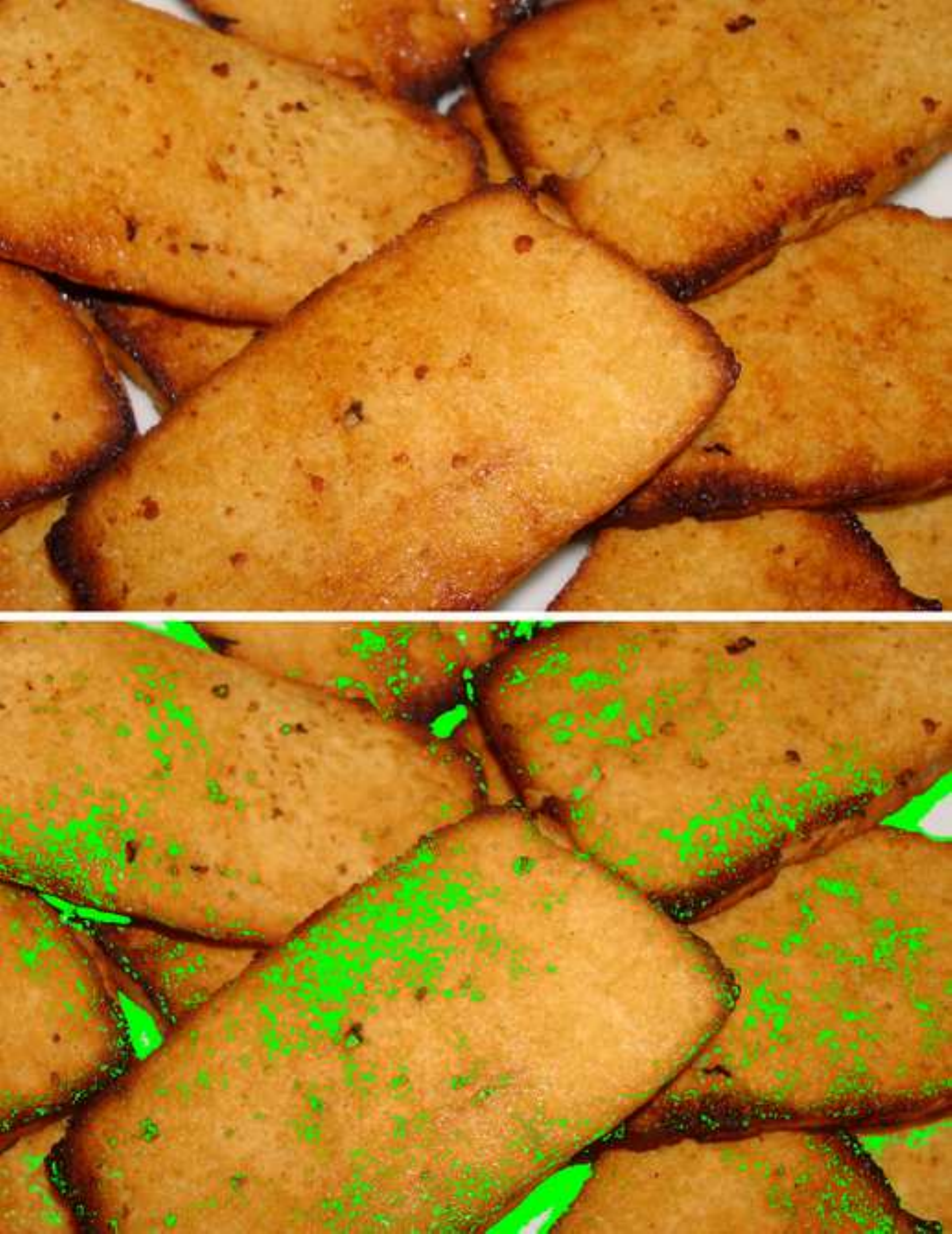,width=0.6in}
\psfig{figure=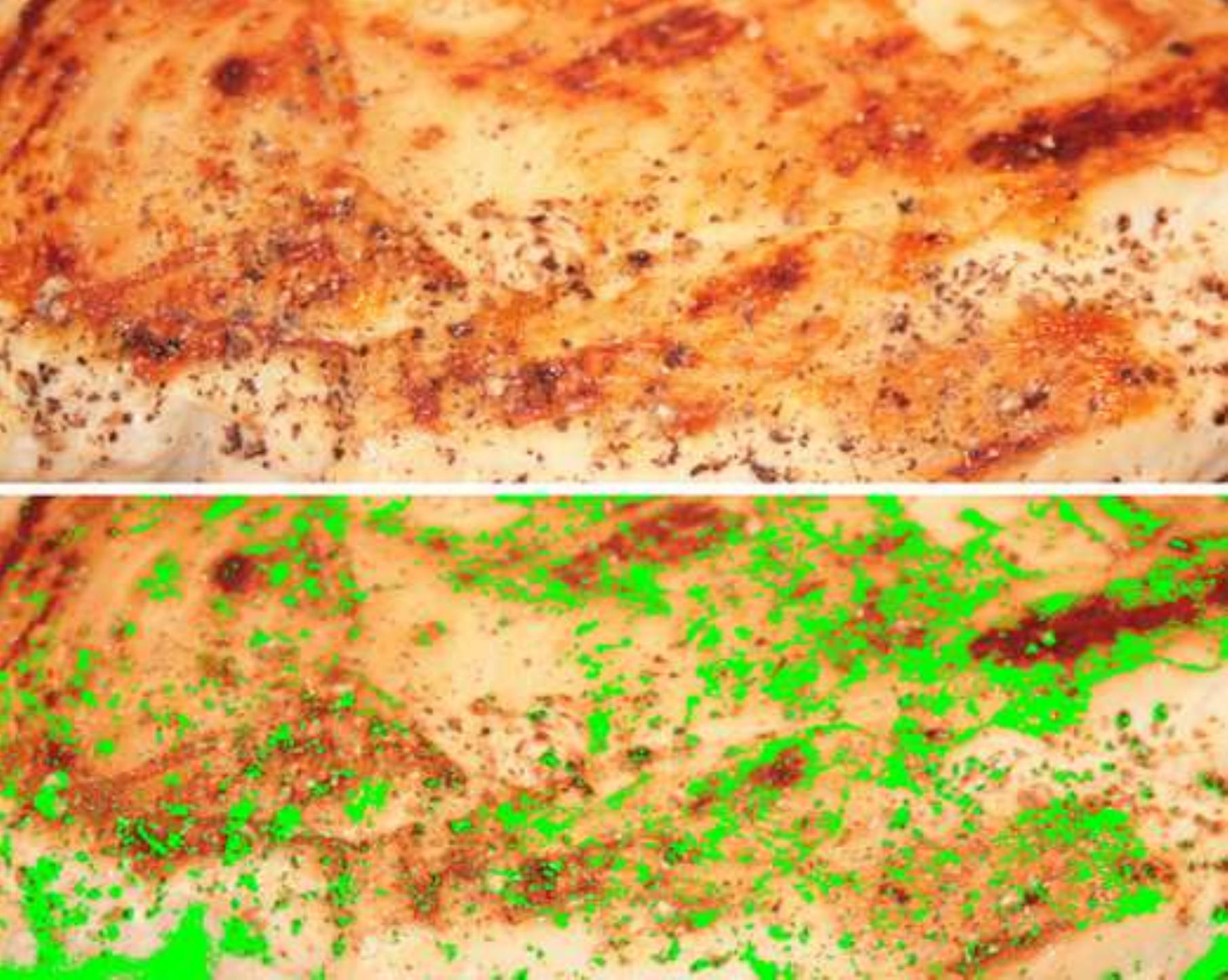,width=0.8in}
\psfig{figure=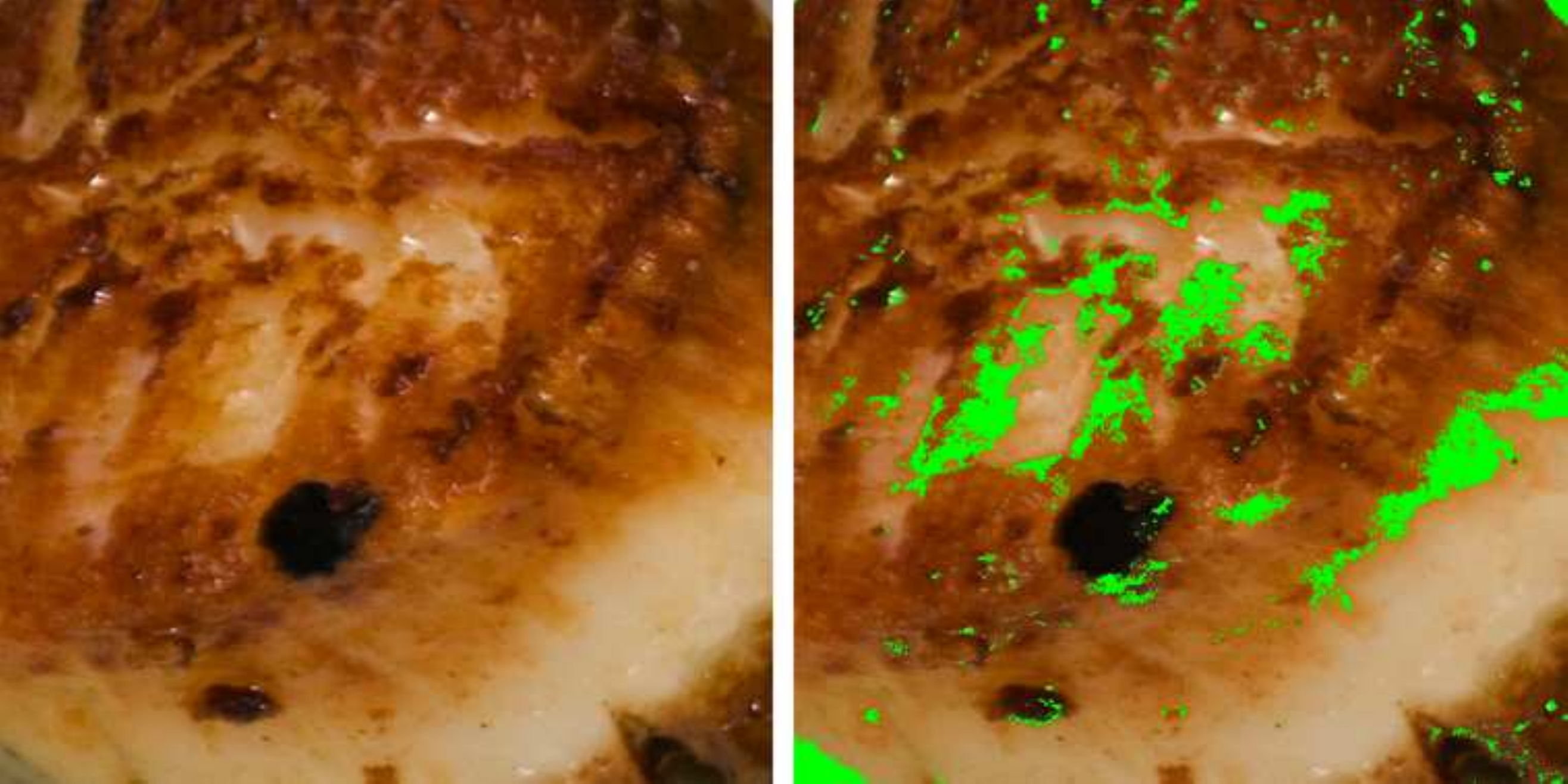,width=1.1in}
\psfig{figure=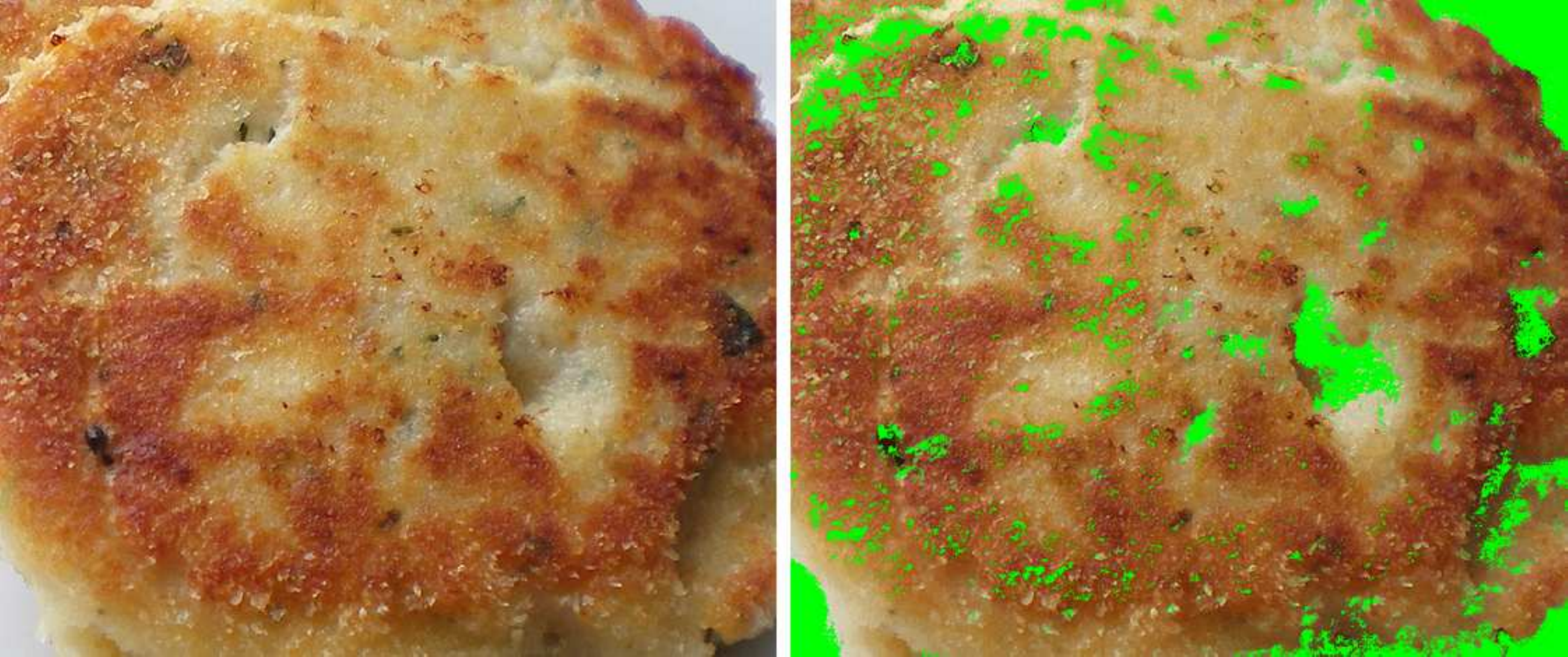,width=1.2in}
\psfig{figure=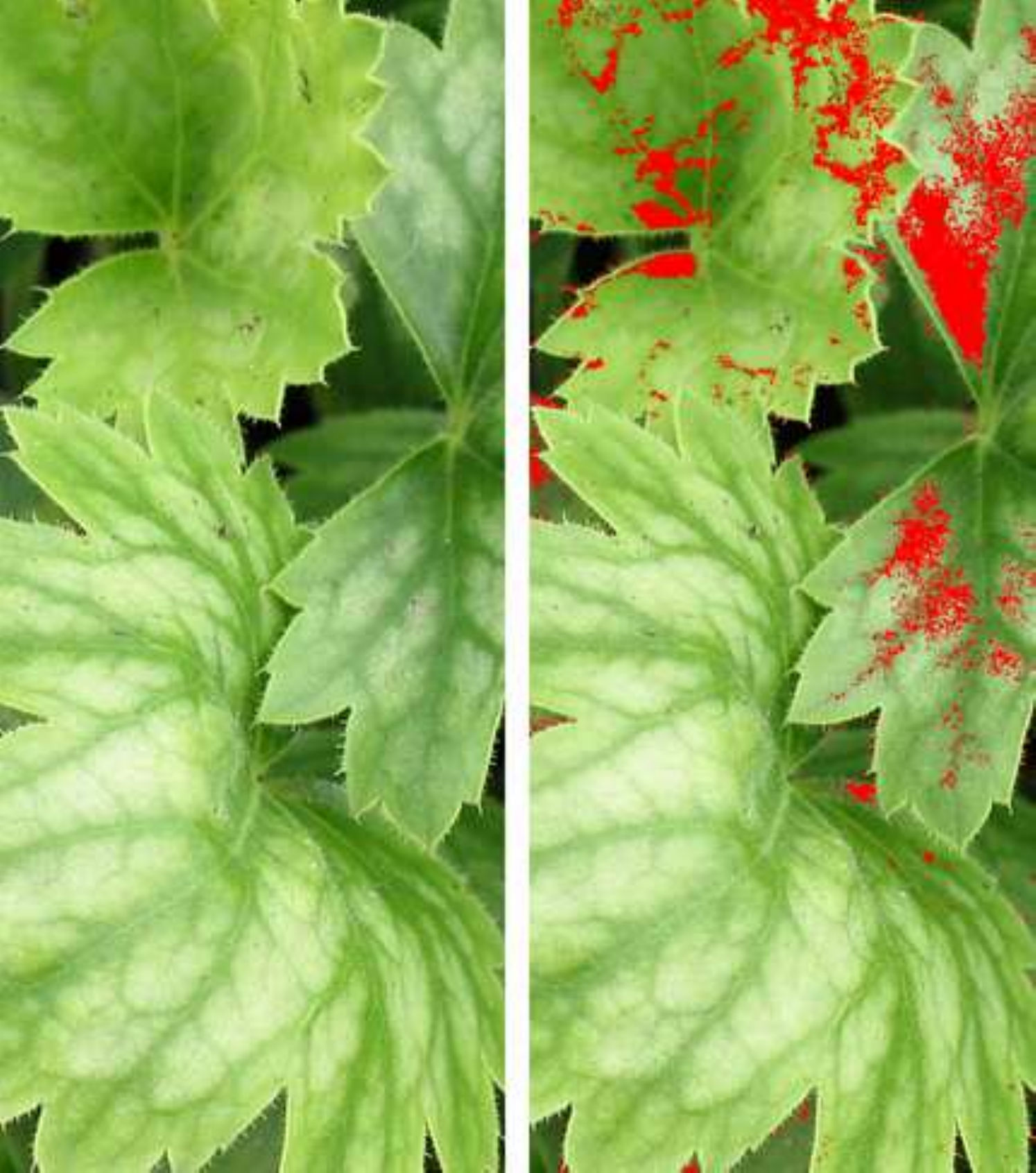,width=0.65in}
\psfig{figure=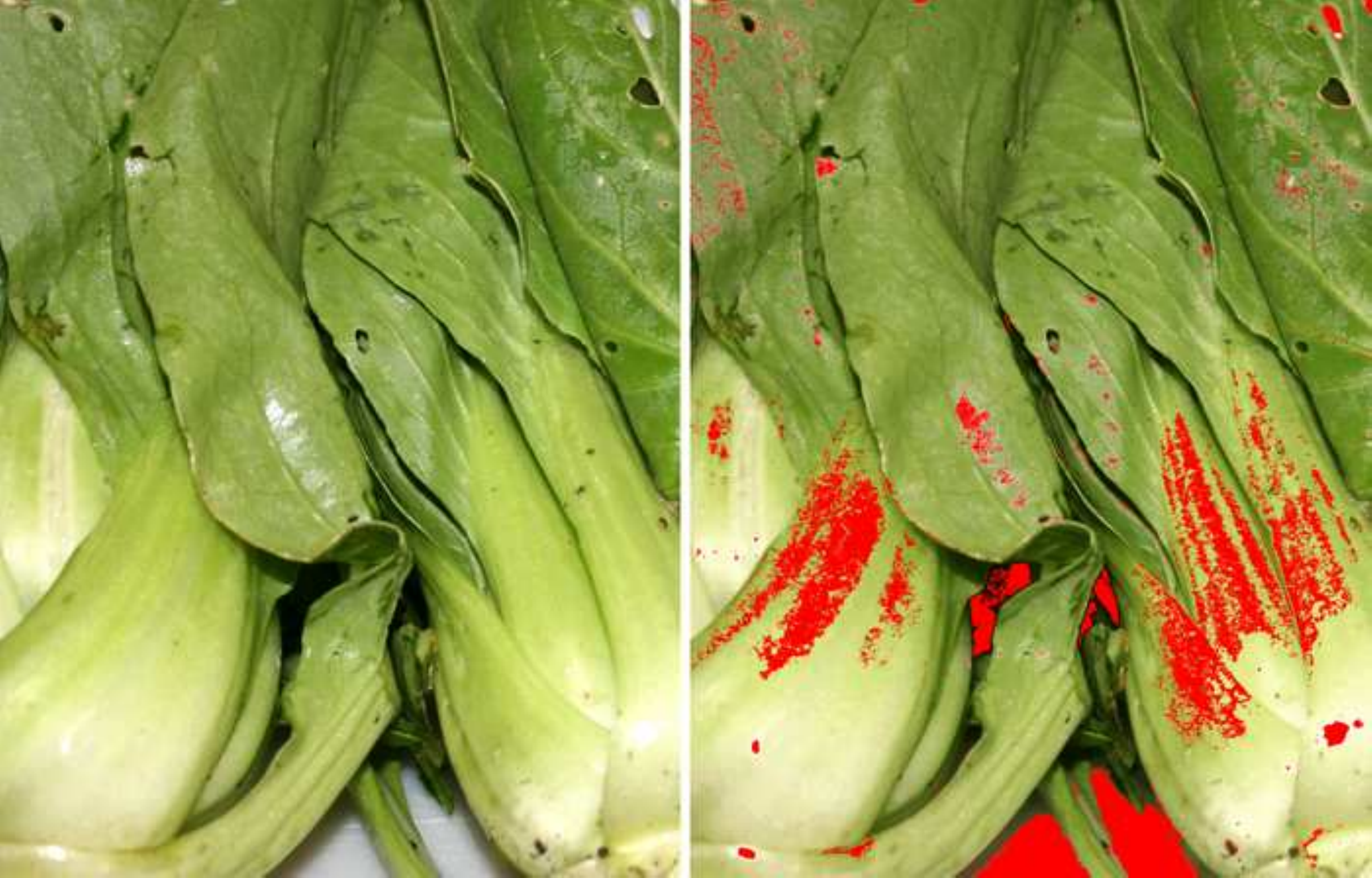,width=1.0in}
\psfig{figure=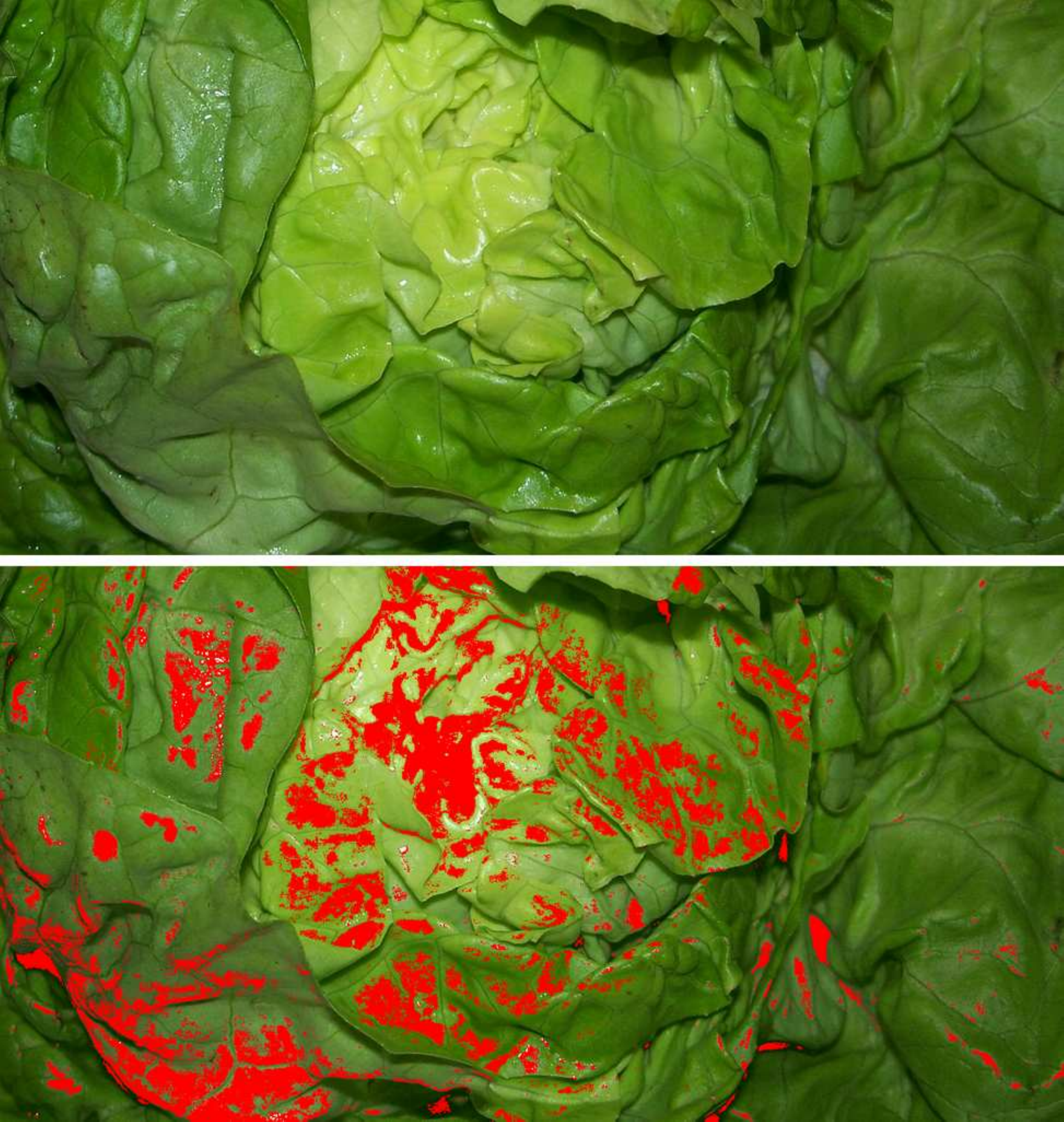,width=0.75in}
\psfig{figure=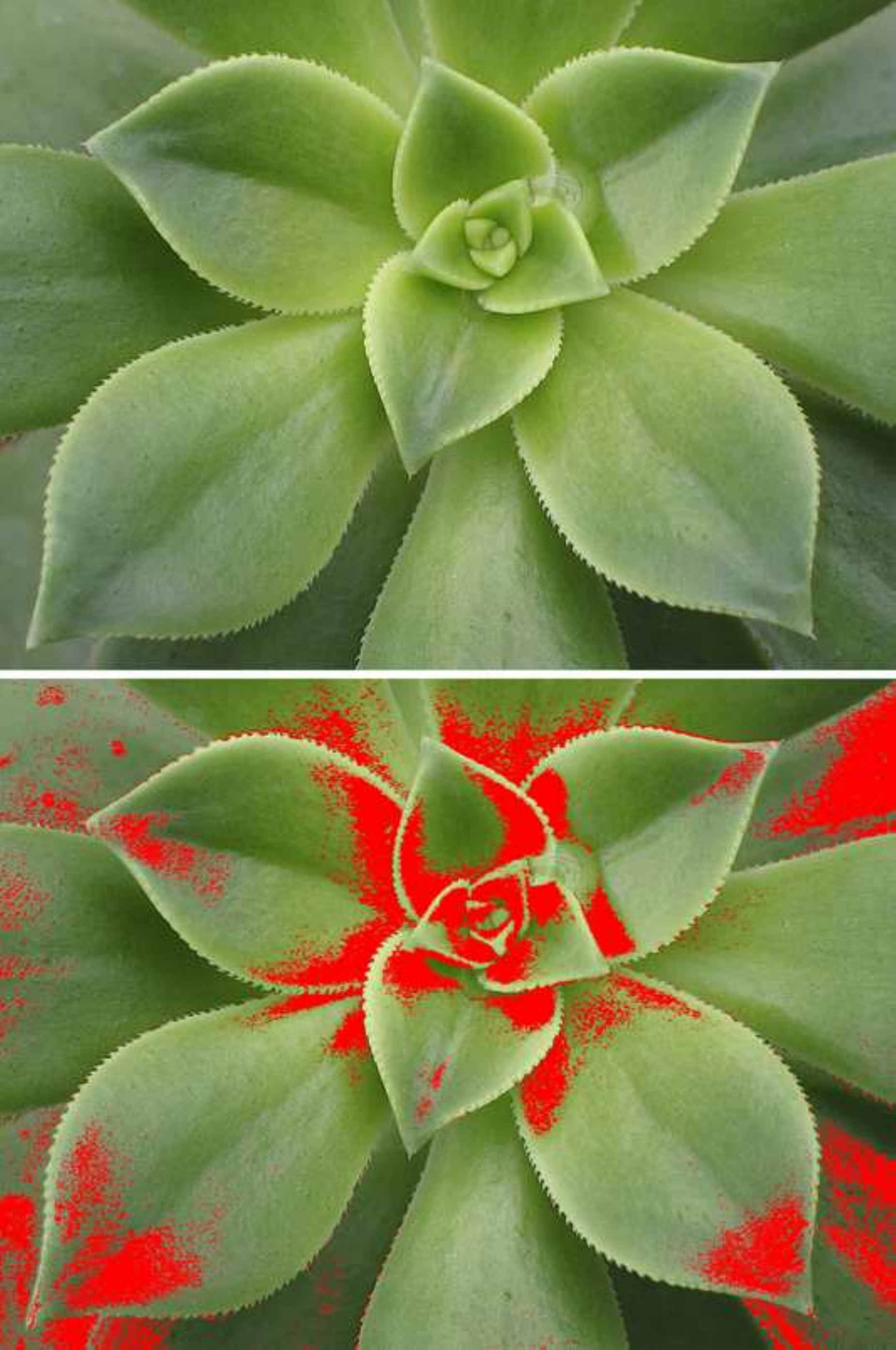,width=0.5in}
}
\centerline{\tiny PM[82.21,98.18]    \hspace{0.1in} PM[81.02,98.87] \hspace{0.3in} PM[84.07,98.42]  \hspace{0.3in}  PM[87.89,99.37]    \hspace{0.6in} PM[88.65,98.35] \hspace{0.7in} PM[94.94,99.66]    \hspace{0.4in} PM[94.23,99.60]\hspace{0.5in} PM[91.11,99.46]    \hspace{0.2in} PM[91.66,99.75]}\centerline{
\psfig{figure=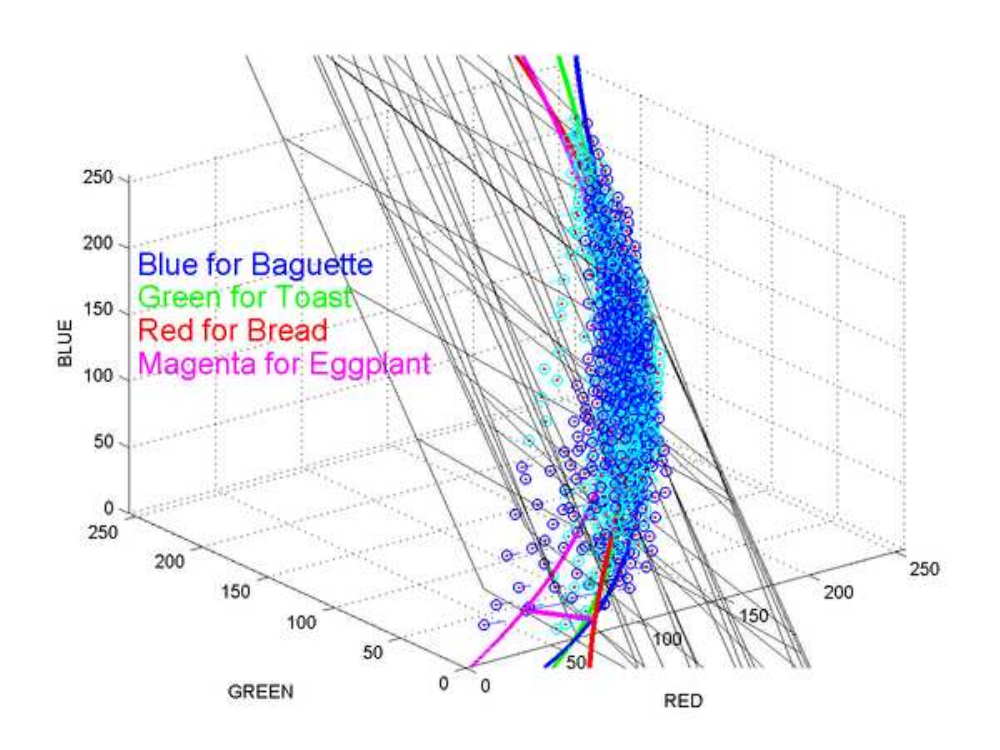,width=2.0in}
\hspace{-0.4in}
\psfig{figure=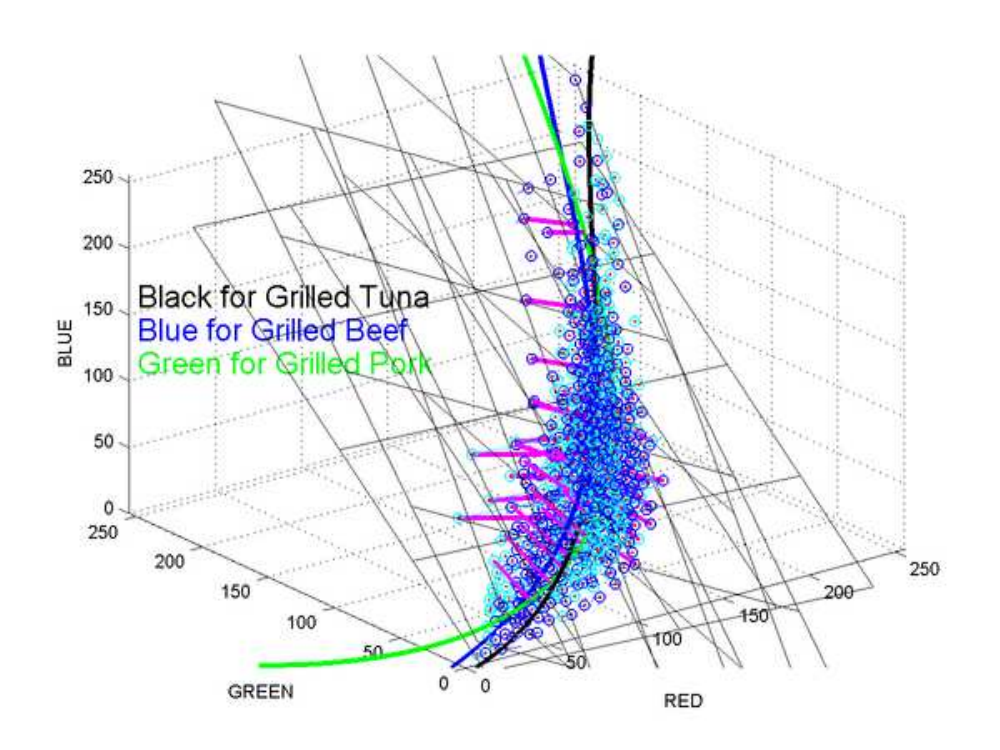,width=2.0in}
\hspace{-0.4in}
\psfig{figure=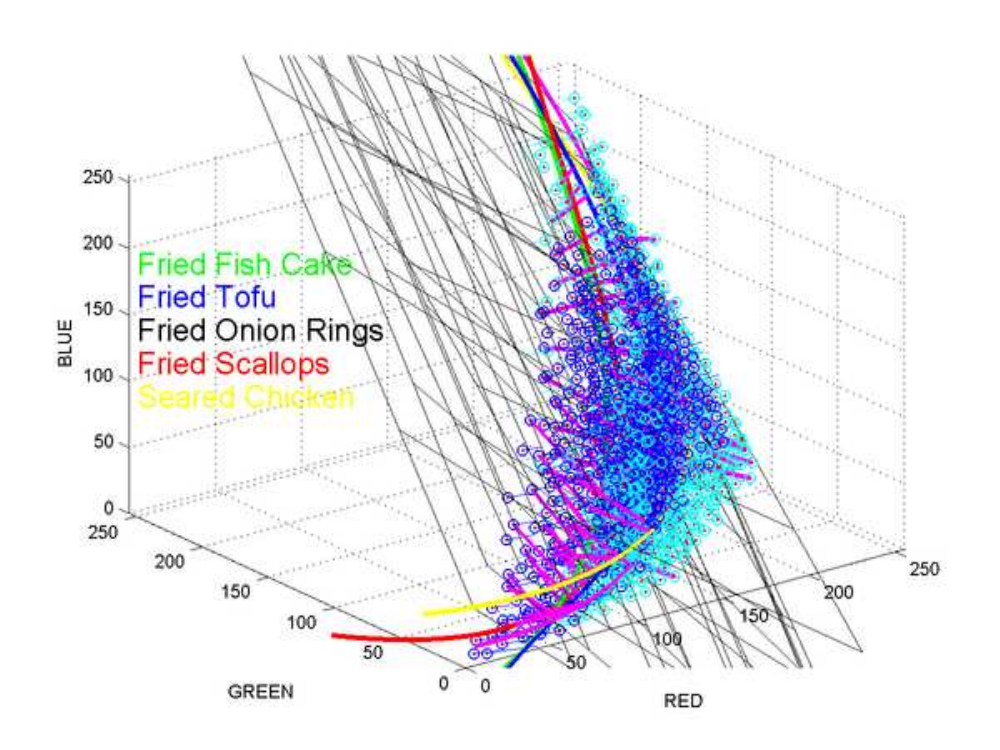,width=2.0in}
\hspace{-0.4in}
\psfig{figure=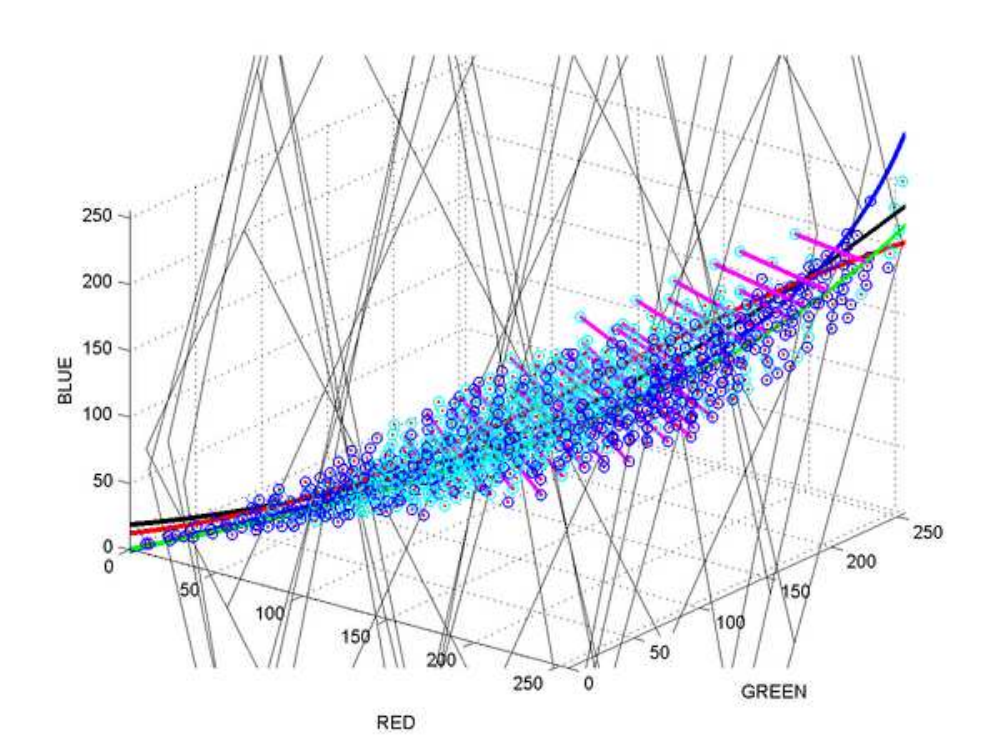,width=2.0in}}\centerline{\tiny Baked Joint PM[90.29,98.81]    \hspace{0.7in} Grilled Joint PM[91.30,98.32]  \hspace{0.7in} Fries Joint PM[78.08,96.69]  \hspace{0.7in} Plants Joint PM [90.37,98.13] }
\caption{\protect\small Baguette, toast, bread, breaded eggplants, beef, pork and tuna, fried onion rings, tofu, seared chicken, seared scallops and fish cake followed by plants. Outliers shown in Red and Green as well as PMs , joint PMS, fitted planes and polynomials.}
\label{bake_sample}
\vspace{-0.15in}
\end{figure*}

Browned foods are materials that were exposed to heat that altered their 
reflectance properties from
initial uniform color to multiple shades of brown.  
Figure  \ref{bake_sample} shows several examples of browned foods:
baked foods (baguette, toasted bread, baked soda bread, baked breaded eggplants), grilled  meats  
(roasted beef,  grilled pork meat and grilled tuna steak), and 
 food exposed to hot oil (fried onion rings, seared chicken, scallops and fish cake).
Clearly, a diverse appearance can result from this food processing. Humans easily recover
important attributes of the scene such as the: initial material, process of cooking involved,  quality  of the
food, freshness, etc. Some of these attributes are derived from prior experiences and therefore they are subjective. However, we conjecture that there are unique  patterns to the appearance of such objects.

Figure \ref{bake_sample} (bottom row) shows also plots of the quantized exemplars, 
their best-fit 3D plane and cubic polynomial curves fitting. 
It is apparent that materials that share a similar browning process tend to cluster in a relatively small
subspace of the RGB space. The 3D planes are geometrically very close but do not fully overlap. 
In all cases the first 2 eigenvalues capture
at least 98\% of the variations in the data.
As a result, near planarity is strongly evident despite the presence of some outlier patches.
The joint planarity measure is computed for each group of images and the data is planar by this measure as well
(e.g., all the baking images colors are considered as a single image).

Planarity suggests fitting a 2D curve to the point projections on the 2D plane. Empirical analysis suggested
that a cubic polynomial is sufficient to minimize fitting error. Higher degree polynomials tend to be less 
numerically stable and their contribution to error reduction is minimal. 
We show in green or red colors 
the points that are farther than a distance threshold $d_t=25$ (fixed for the rest of the paper) from the fitted 
polynomials. This is the Euclidean distance between the input RGB points (not the quantized data)
 and the 3D curve.
This threshold exceeds the typical quantization reconstruction errors shown in Figure \ref{MVQ} (typically
less than 10) since the distances in Figure \ref{MVQ} apply to within-bin quantization while $d_t$
applies to the cubic polynomial which naturally requires a higher threshold to accommodate reflectance and shading variations.

In Figure \ref{bake_sample} the baked-food polynomials are close since the basis-material (flour) is likely similar in the four images.
The grilled and roasted meats have different basis-materials so the polynomials are somewhat aligned due to the basic
red-tone of beef, pork and tuna. The fried materials are more diverse and as a result the polynomials
are more spread out.

It is worth noting that the notion that the color space of baked food can be reduced to a curve in RGB space has recently been
observed in Food Inspection Science. Abdullah \cite{Abdullah} quantified the color appearance of baked crackers under controlled
illumination. The objective was to determine the optimal baking temperature and duration as a function of color changes on
the surface of baked goods. It was suggested that the process of baking takes on a range of colors that appear to lie on a curve that
begins with the raw material, moves to near-white color zone and then descends through light brown to dark brown and eventually
to black if burning is allowed to take place.  The curves described in \cite{Abdullah} look similar to the ones we computed.


\subsection{Plants}

Plants are distinguished by their use of Chlorophyll to convert light to energy. The amount
of Chlorophyll varies in different parts due to material properties, growth patterns and function
for each part of the plant. The amount of Chlorophyll determines the reflected wavelengths, so
that low concentrations lead to reflecting all light wavelengths, 
while high concentrations reflect mostly the wavelengths around green color.
We observe that Chlorophyll acts as a single process that controls the 
reflectance of light, and therefore it is an example of a constrained color space. Figure \ref{bake_sample}
illustrates the planar and curve qualities for four sample images. Note that shadows, shading variations and
specular reflections affect the analysis. The red pixels indicate points farther than $d_t$ from the polynomial fit.
Nevertheless, the planarity of colors in each image is strong, and the 
joint planarity of all colors holds despite the diversity in the shades of green across the images.


\subsection{Miscellaneous objects and scenes}

\begin{figure*}
\centerline{
\psfig{figure=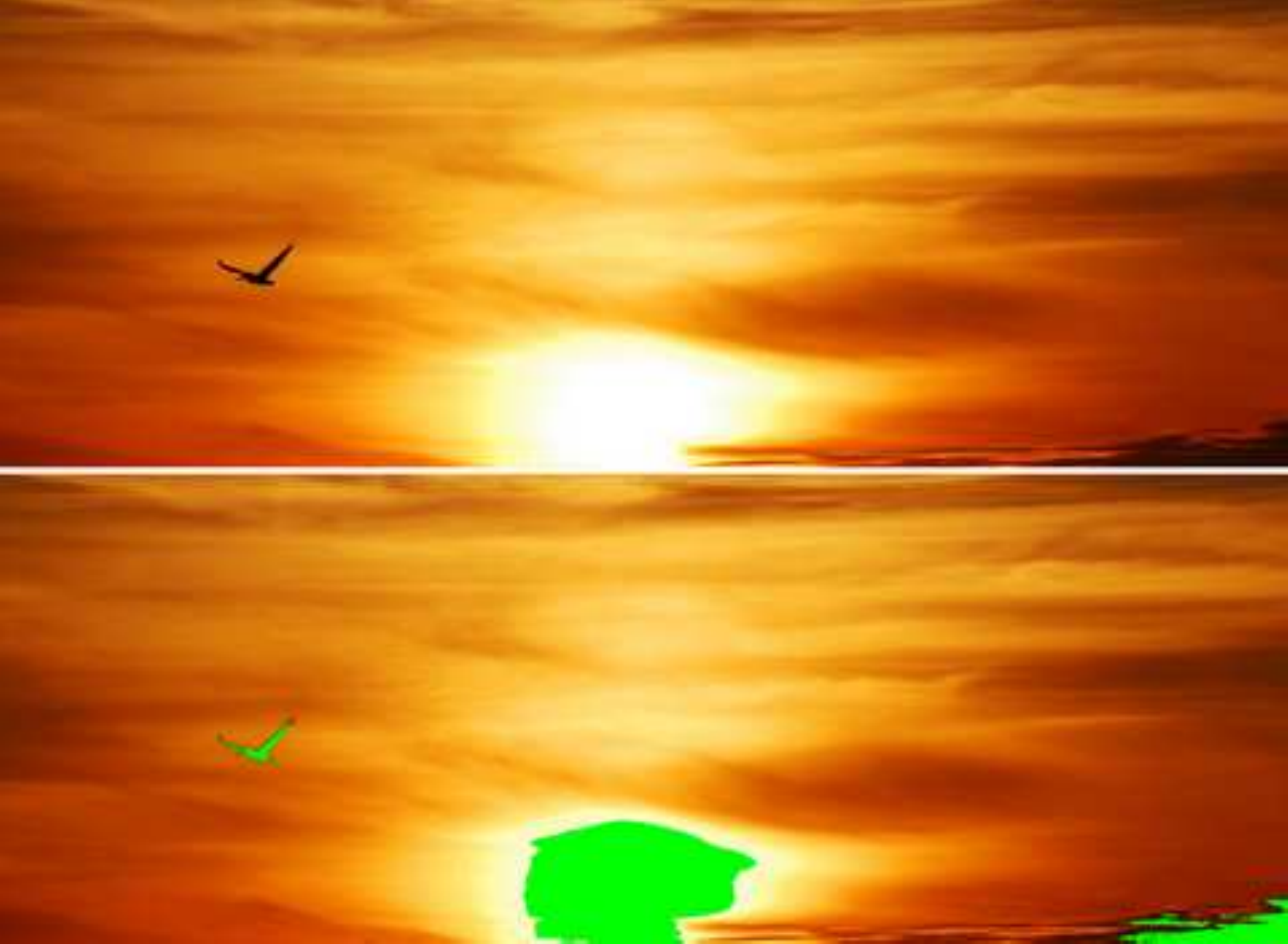,width=1.1in}
\hspace{0.1in}
\psfig{figure=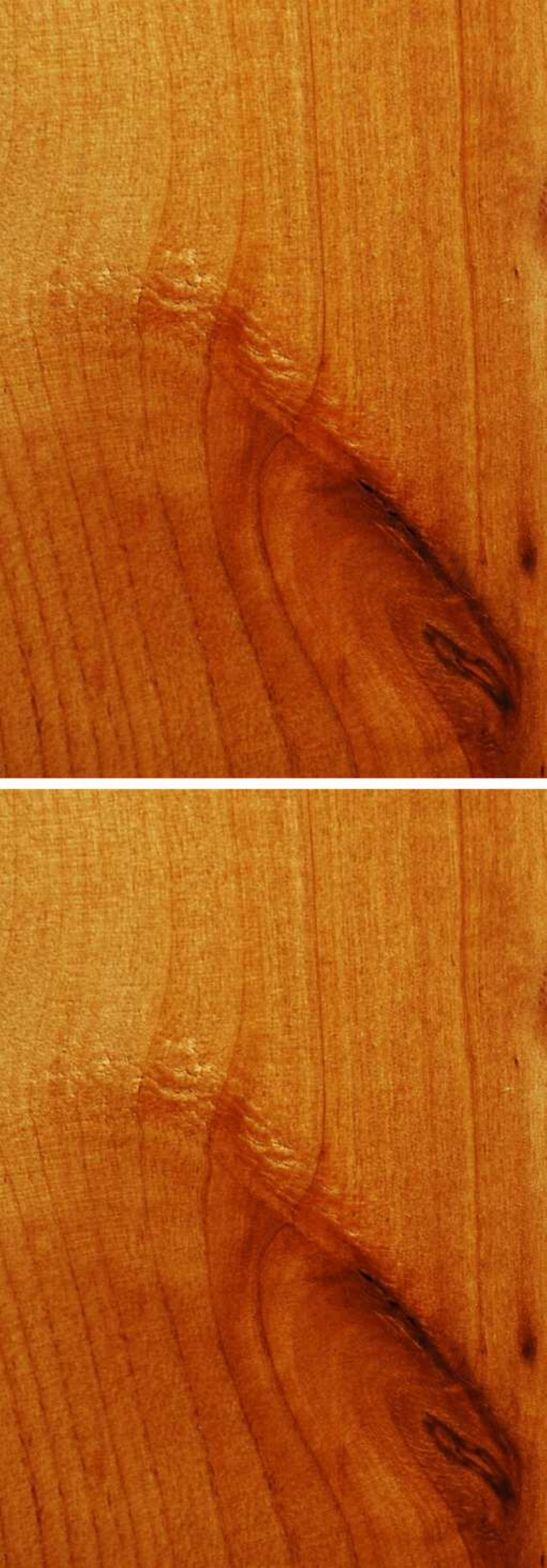,width=0.4in}
\hspace{0.1in}
\psfig{figure=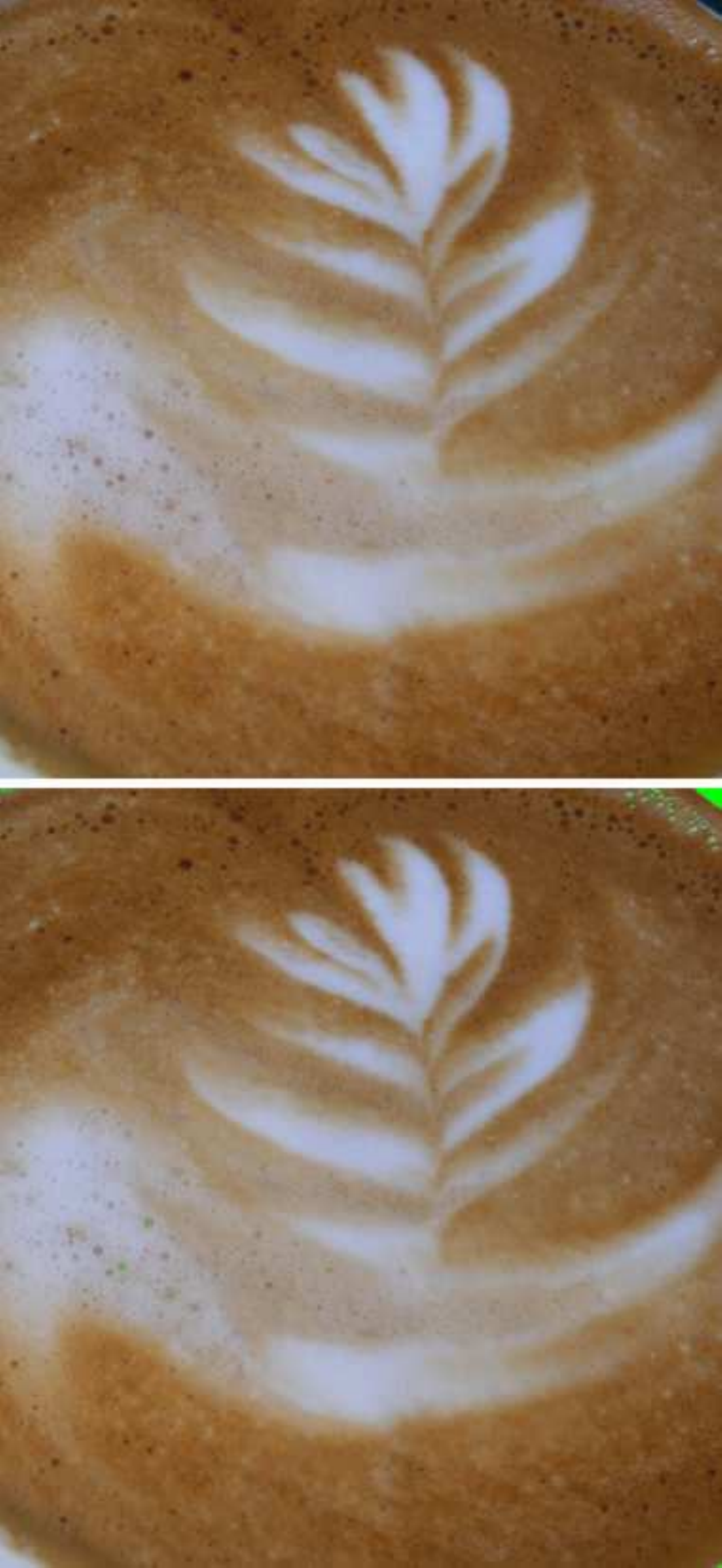,width=0.5in}
\hspace{0.1in}
\psfig{figure=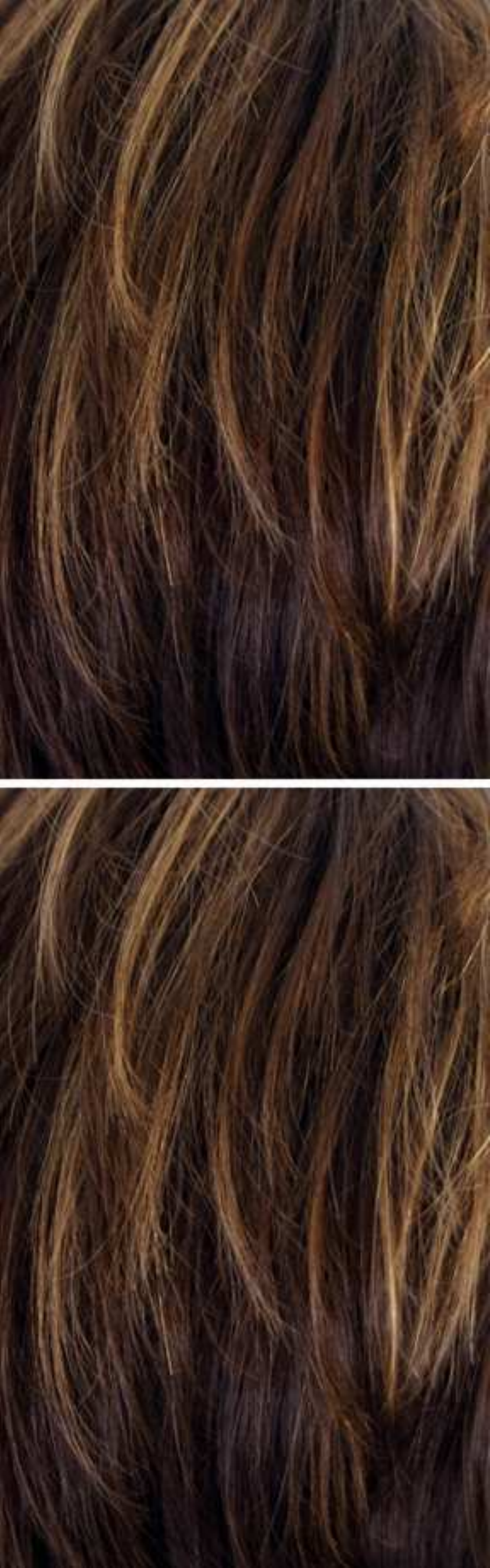,width=0.4in}
\hspace{0.1in}
\psfig{figure=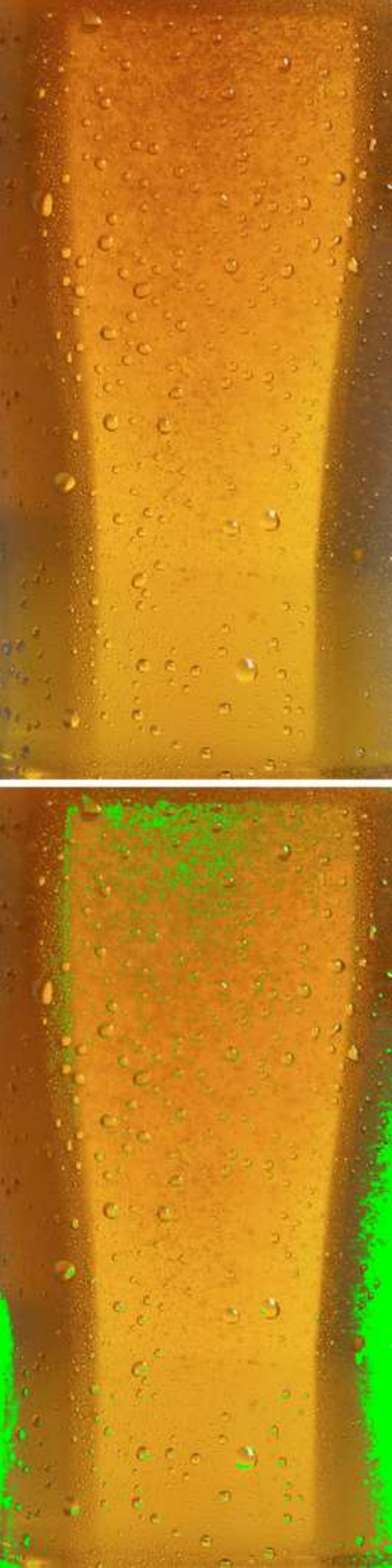,width=0.3in}
\hspace{0.1in}
\psfig{figure=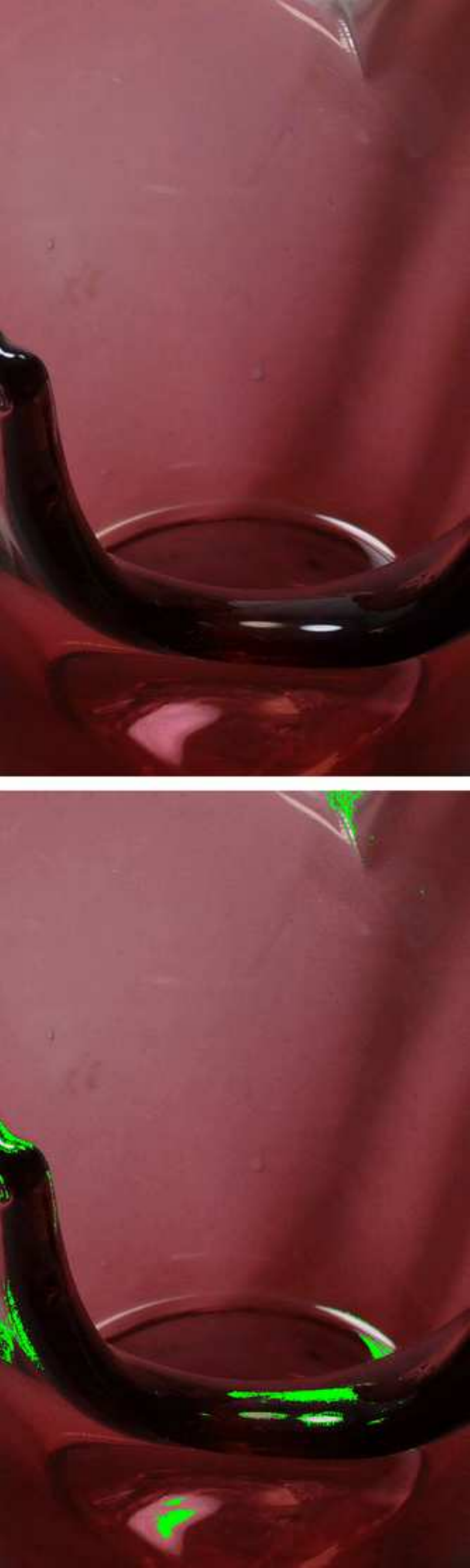,width=0.4in}
\hspace{0.1in}
\psfig{figure=figs/COMB_train347.eps,width=0.5in}
\hspace{0.1in}
\psfig{figure=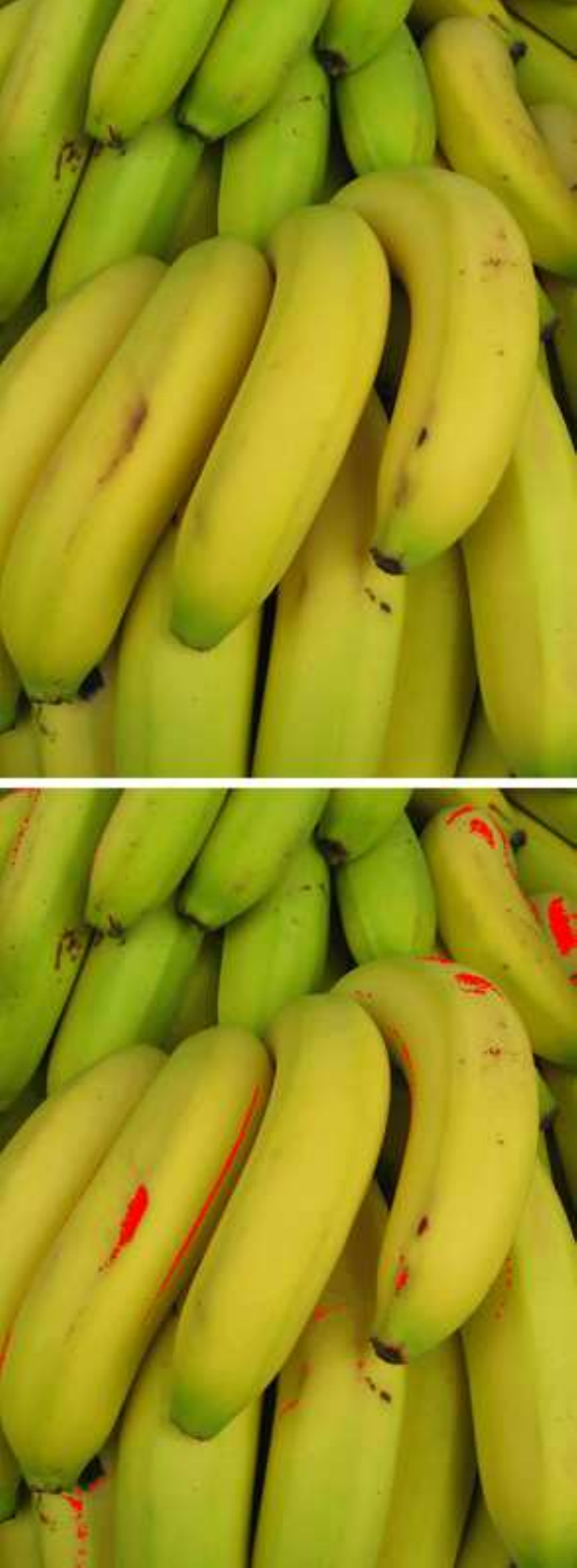,width=0.45in}
\hspace{0.1in}
\psfig{figure=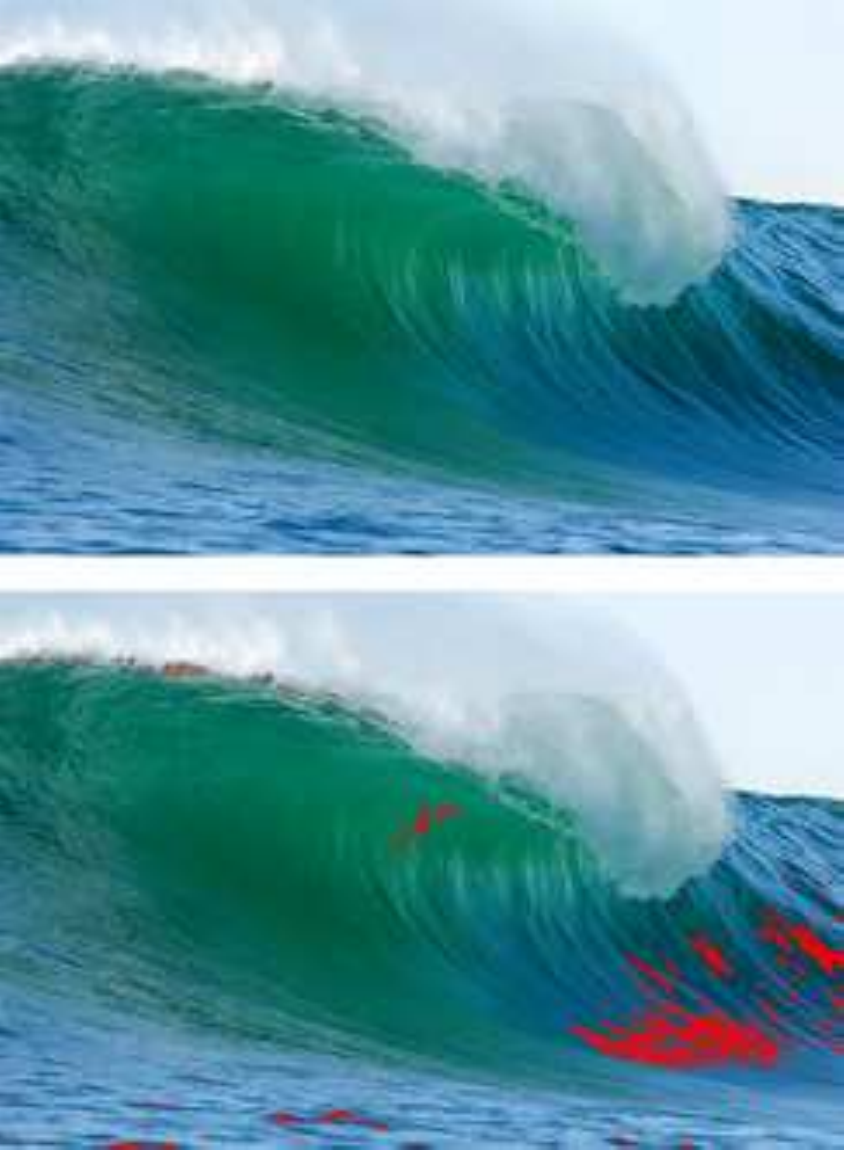,width=0.8in}
}
\centerline{\protect\tiny PM[85.23,98.5]    \hspace{0.5in} PM[92.96,98.74] \hspace{0.1in}  PM[92.60,99.85] \hspace{0.1in} PM[97.11,99.71] \hspace{0.1in} PM[80.76,97.97] \hspace{0.1in} PM[94.19,99.85] \hspace{0.1in} PM[88.26,98.64] \hspace{0.1in} PM[93.09,97.61] \hspace{0.1in} PM[95.35,98.88]}
\centerline{
\psfig{figure=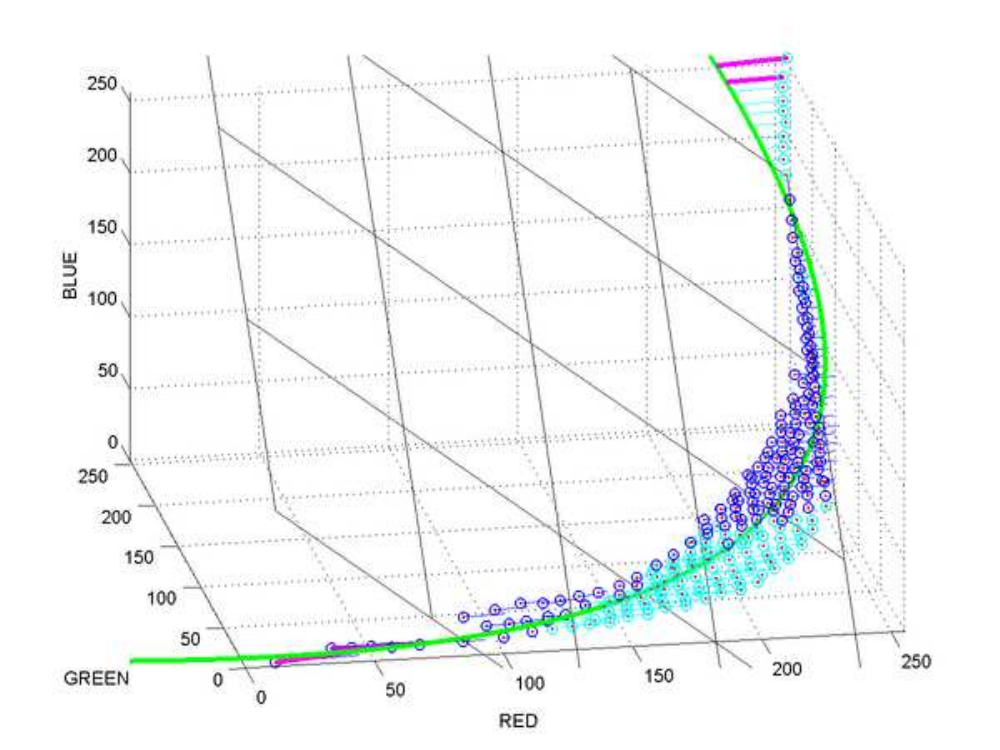,width=0.9in}
\hspace{-0.16in}
\psfig{figure=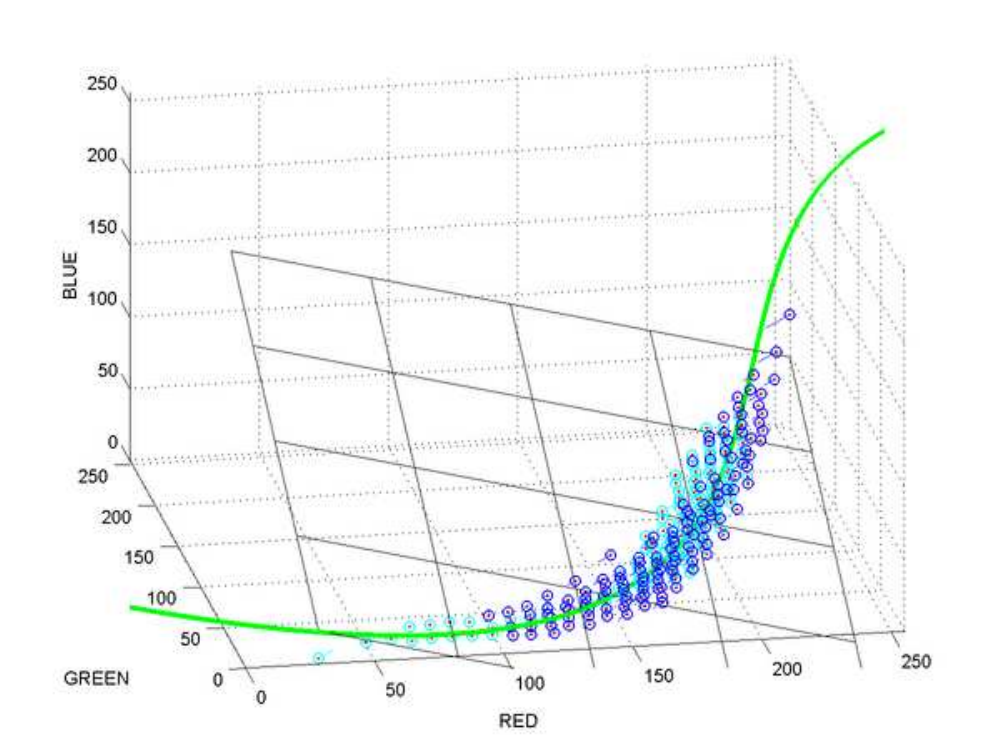,width=0.9in}
\hspace{-0.16in}
\psfig{figure=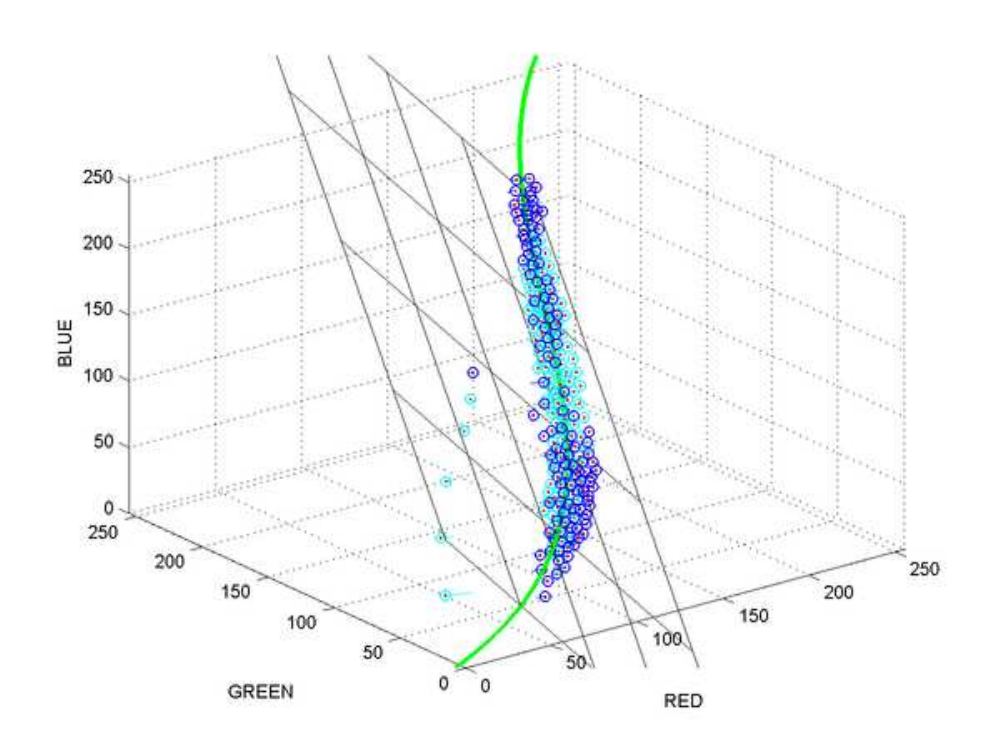,width=0.9in}
\hspace{-0.16in}
\psfig{figure=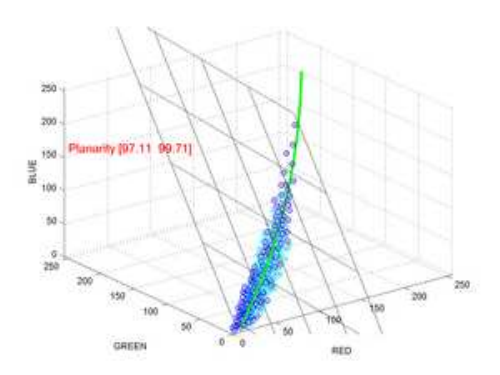,width=0.9in}
\hspace{-0.16in}
\psfig{figure=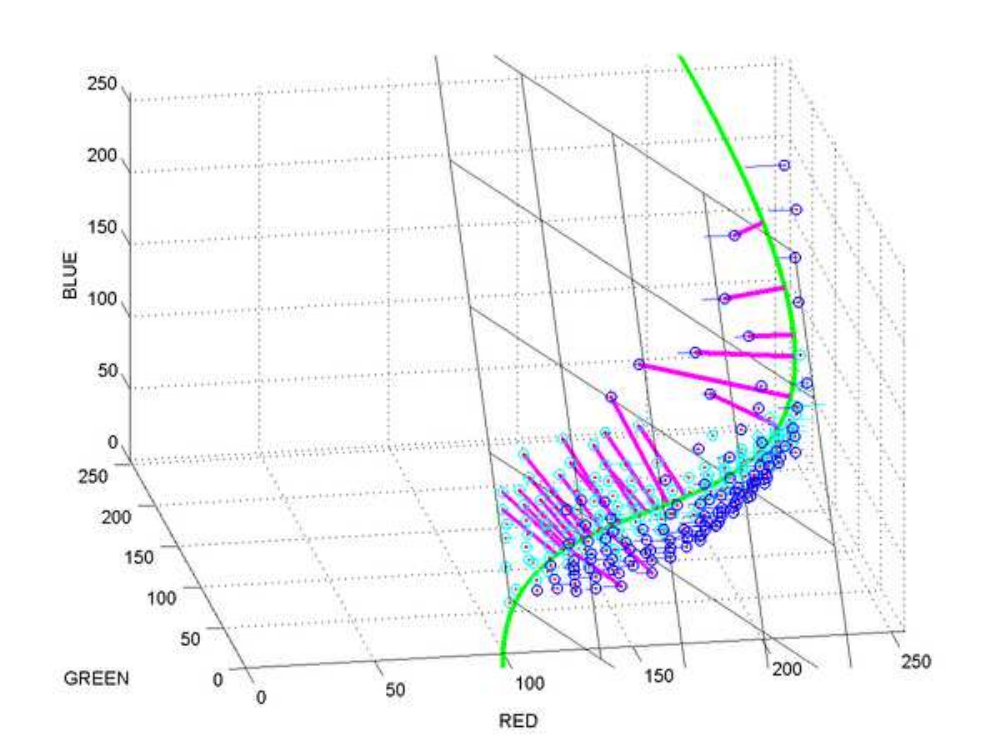,width=0.9in}
\hspace{-0.16in}
\psfig{figure=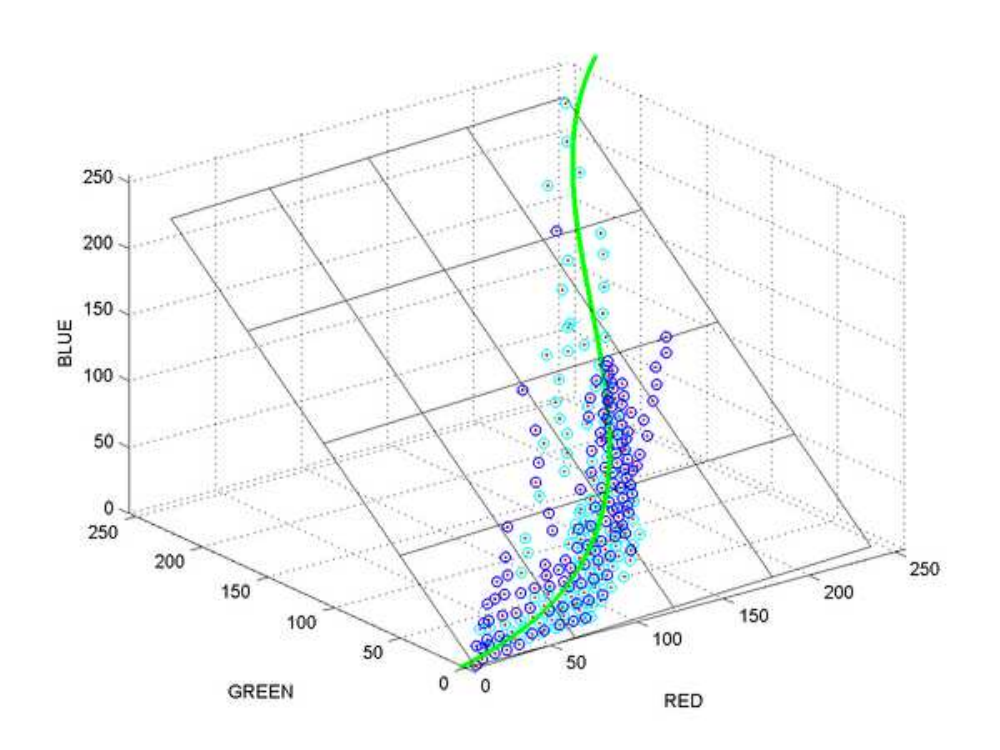,width=0.9in}
\hspace{-0.16in}
\psfig{figure=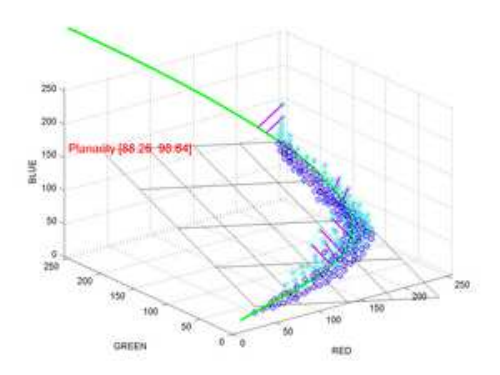,width=0.9in}
\hspace{-0.16in}
\psfig{figure=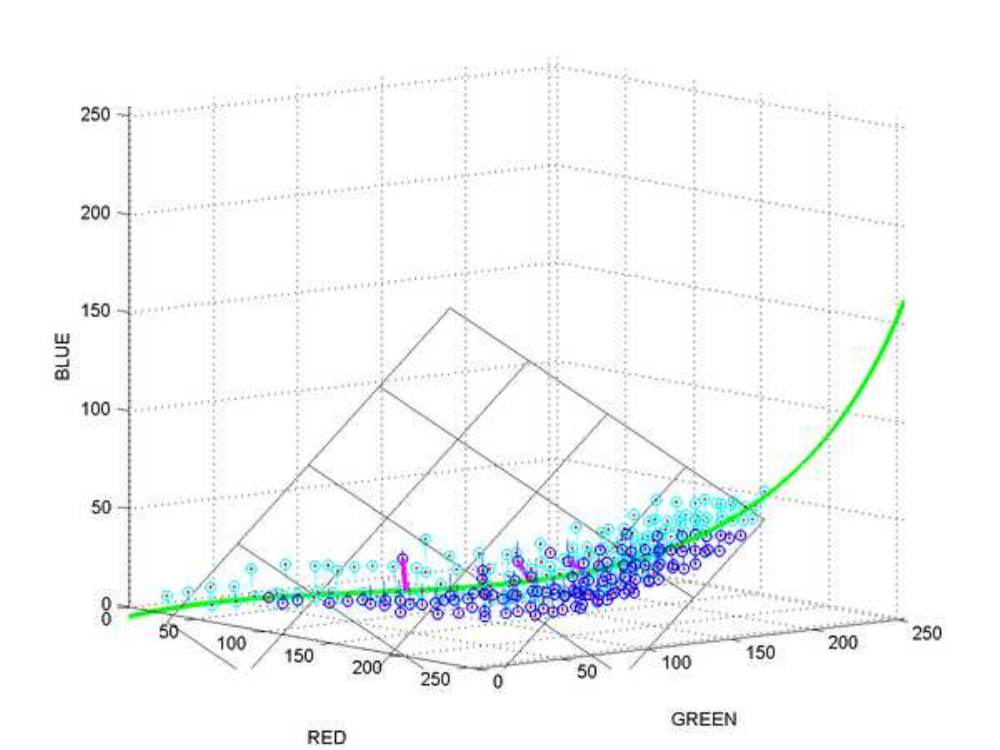,width=0.9in}
\hspace{-0.16in}
\psfig{figure=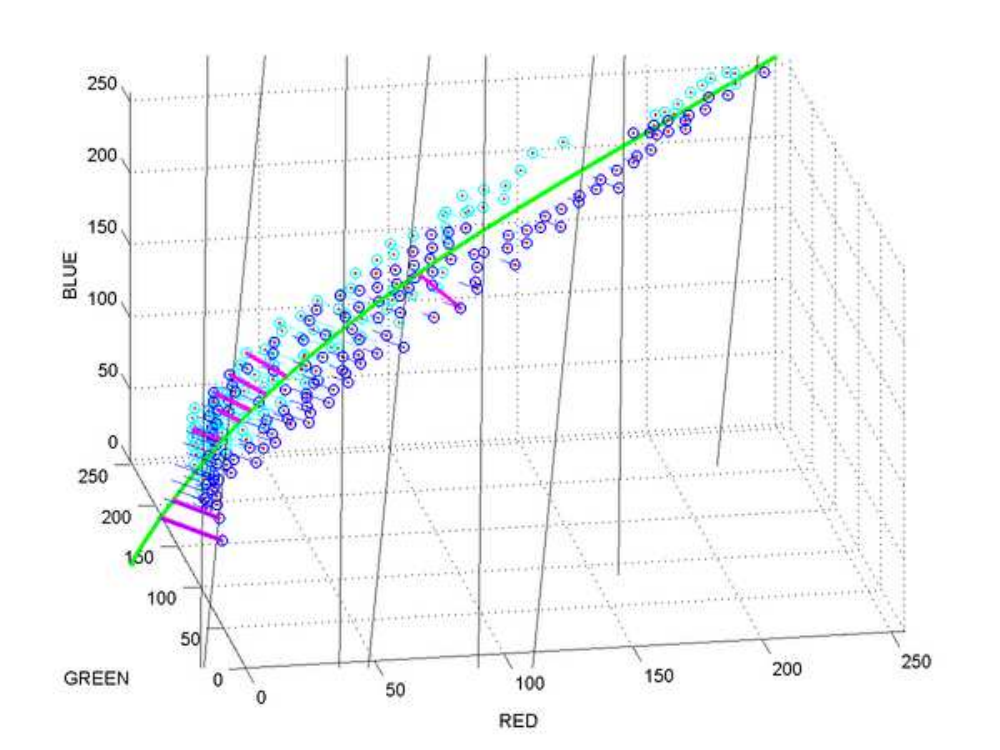,width=0.9in}
}
\caption{\protect\small
Sunset, wood, cappuccino, dyed hair, beer in a class, vase, fire, bananas (greenish and browning), and an ocean wave.
The respective curve fit for colors in image image. In the bottom image the outlier points to the curve (measured by a fixed
distance threshold equal for all images) are shown in green or red colors.
\label{mix_sample}}
\vspace{-0.15in}
\end{figure*}

Figure \ref{mix_sample} illustrates a set of images that we presume that a single process
is involved in creating the diversity of colors. The PMs suggest that the colors
lie on a planar subspace. Polynomial fitting shows different degrees of adequacy. Images have few outlier
pixels that are far from the curves so overall it appears that these curves provide a good abstraction to the
color space in the images.

The sunset image is formed due to variable density of cloud cover and consequent absorption/scattering of light. The
foreground regions (e.g., bird) are detected as outliers and the Sun is also an outlier since its distance
is greater than $d_t$. The wood, cappuccino foam, and dyed hair
show almost no outliers. The beer in the glass shows multiple processes at play.
The primary process is the glass-thickness and beer volume
 that the light is passing through (observe light artifacts at the sides), other
processes include the water condensation on the surface of the glass and the image formation  causing specular reflections.
The vase outliers primarily occur at specular reflections. The fire outliers appear at extreme brightness and are due
to the inadequacy of cubic fit. The green/yellow banana outliers are  due to secondary browning process
and the wave outliers are due to poor cubic fit.

\subsection{Outliers}

Points that are farther from the curve than the threshold  $d_t$ are considered outlier points and
stem from:

{\bf Background}. Pixels that are clearly due to patches that do not belong to the object (e.g., center-top area of the baked eggplants in Figure \ref{bake_sample} belongs to the plate).

{\bf Independent processes}. Interference of independent processes from the primary process. 
For example, specular reflections and poorly illuminated pixels tend to be classified as outliers since they 
 involve a second independent process (e.g., image formation artifacts).

{\bf Boundary conditions}.
The cubic polynomial fitting imposes boundary conditions that are not perfectly aligned with the object reflectance
or image formation.
This typically occurs at the locations where the polynomial exits the RGB space
(e.g., sunset and fire images in Figure \ref{mix_sample}).

{\bf Compression}. Block Compression artifacts that accompany JPEG occasionally bias the pixel values and affect the 
distribution of colors.

 \section{Detection and segmentation}

The hypothesis that some objects and scene colors cluster in a highly reduced dimensionality is 
useful for image understanding. However, this compact representation is not
invariant to significant illumination changes in which case polynomials change shape and location in RGB space,
while maintaining their polynomial prototype. Nevertheless, an exemplar material can be used to search for similar materials in images
as long as the color variations are within $d_t=25$ with respect to the subspace represented by
the polynomial fit to the quantized colors of the exemplar.

Let $POLY$ represent the exemplar polynomial, and $(R_{x,y},G_{x,y},B_{x,y}) $ be the color value at
$(x,y)$ in the probe image.
We employ two steps for detection of color distributions that conform to the exemplar:

{\bf Step I}: A pixel in the probe image is conforming if: 
\begin{equation}
\scriptsize
D((R_{x,y},G_{x,y},B_{x,y}),POLY)<d_t\label{pdistance}
\end{equation}
where $D$ represents the closest Euclidean distance between the point and the points of $POLY$.

{\bf Step II:}
For a {\em spatially-contiguous} region $M$ in the probe image, 
the set of corresponding closest points of $POLY$ is computed and labeled as $S$. We define $M$ to conform to the exemplar
if
\begin{equation}
\scriptsize
L(S)>l_t \label{sdistance}
\end{equation}
where $L$ is the length of the segments of $POLY$ that the points in $S$ cover (a threshold $l_s=10$
is used so that only points in $S$ that have a greater number of votes than $l_s$ are used). 
In the experiments below we use $l_t=150$. Effectively, this means that only if a significant portion of the points 
of $M$ are spread-out over $POLY$ is $M$ considered conforming to the exemplar. 
The correspondence between $M$ and $POLY$ is to any part of the space defined by $POLY$ (subject to exceeding $l_t$).

\begin{figure*}
\centerline{
\psfig{figure=figs/train206.eps,width=0.8in}
\hspace{0.3in}
\psfig{figure=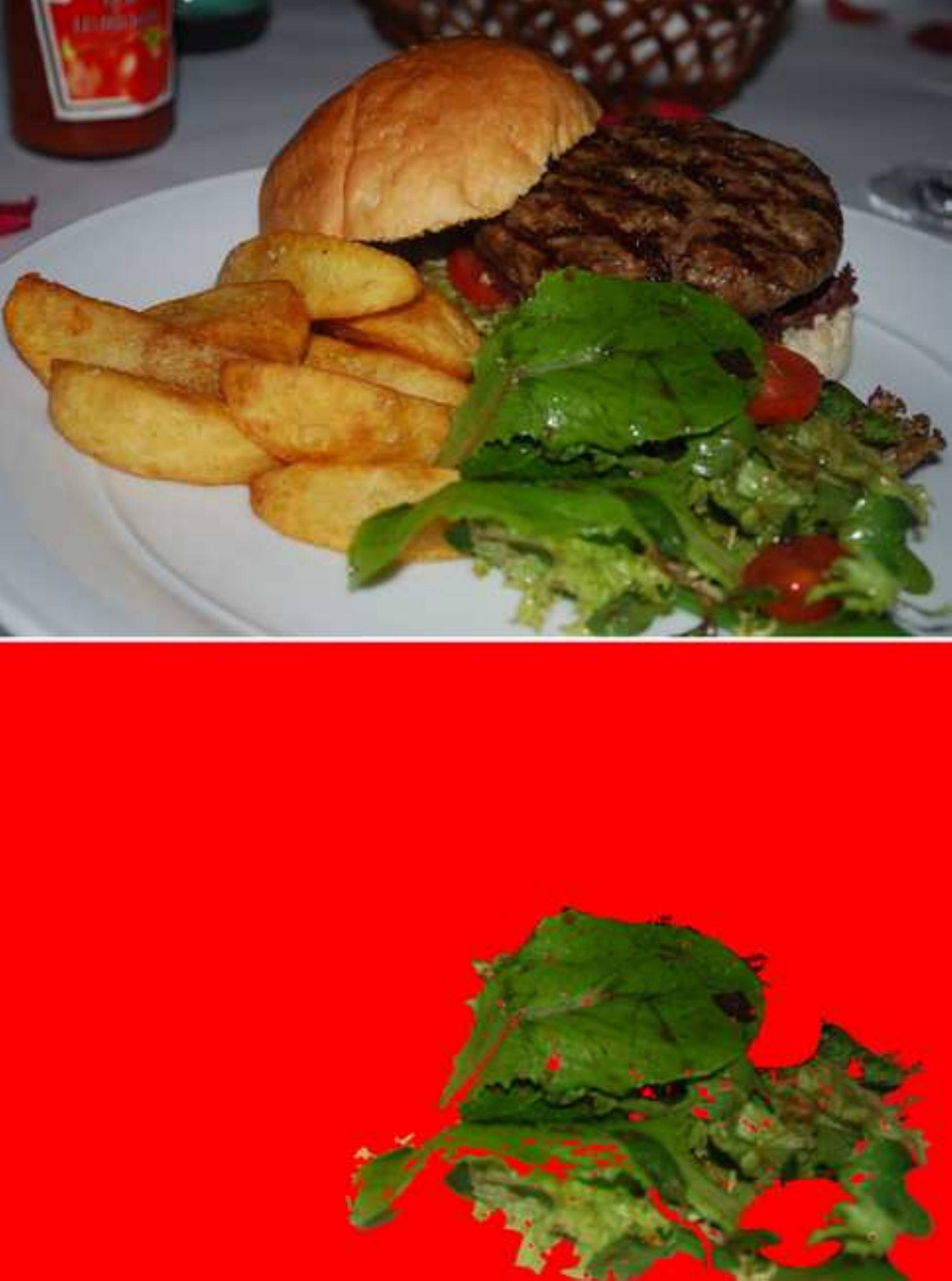,width=0.8in}
\psfig{figure=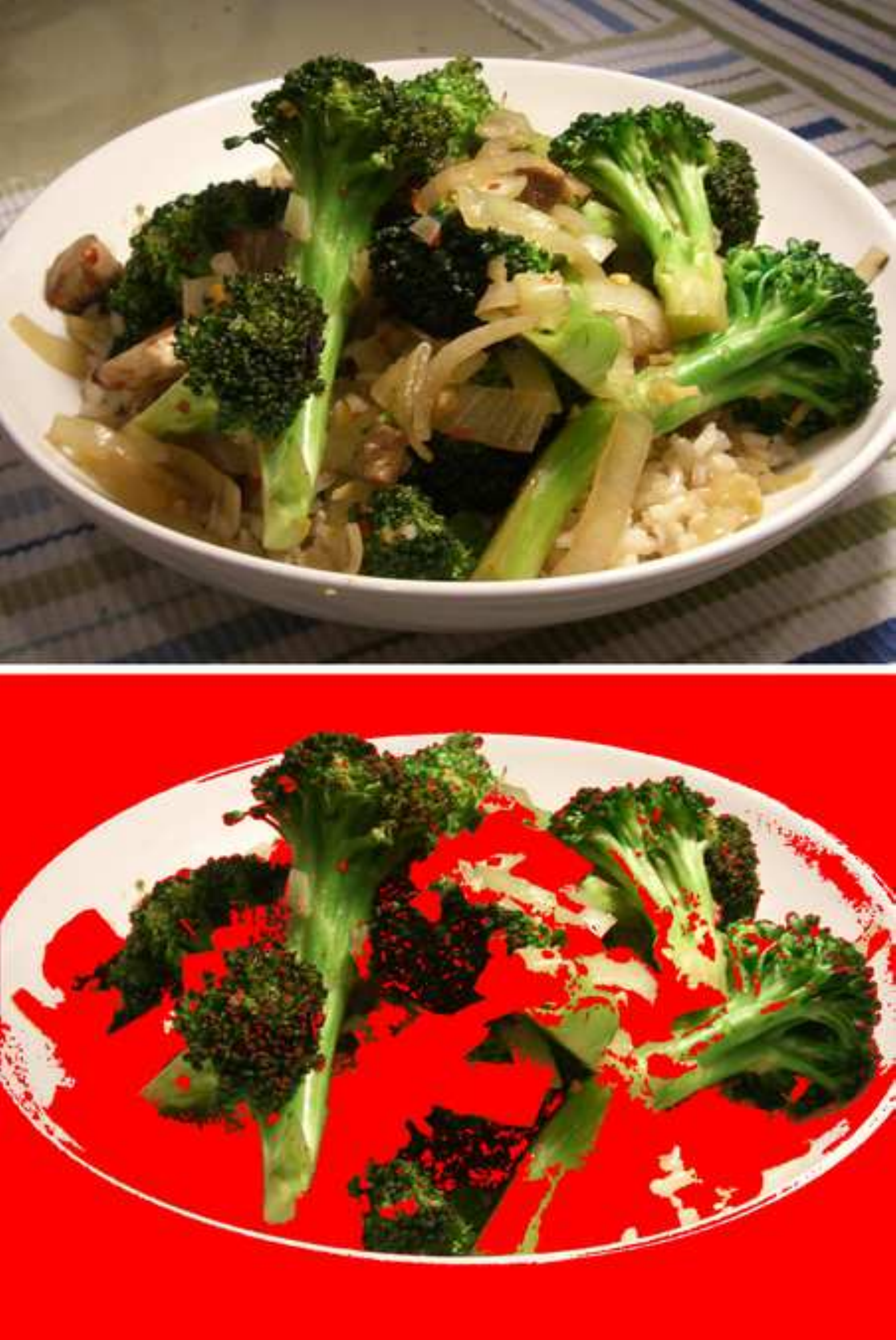,width=0.7in}
\psfig{figure=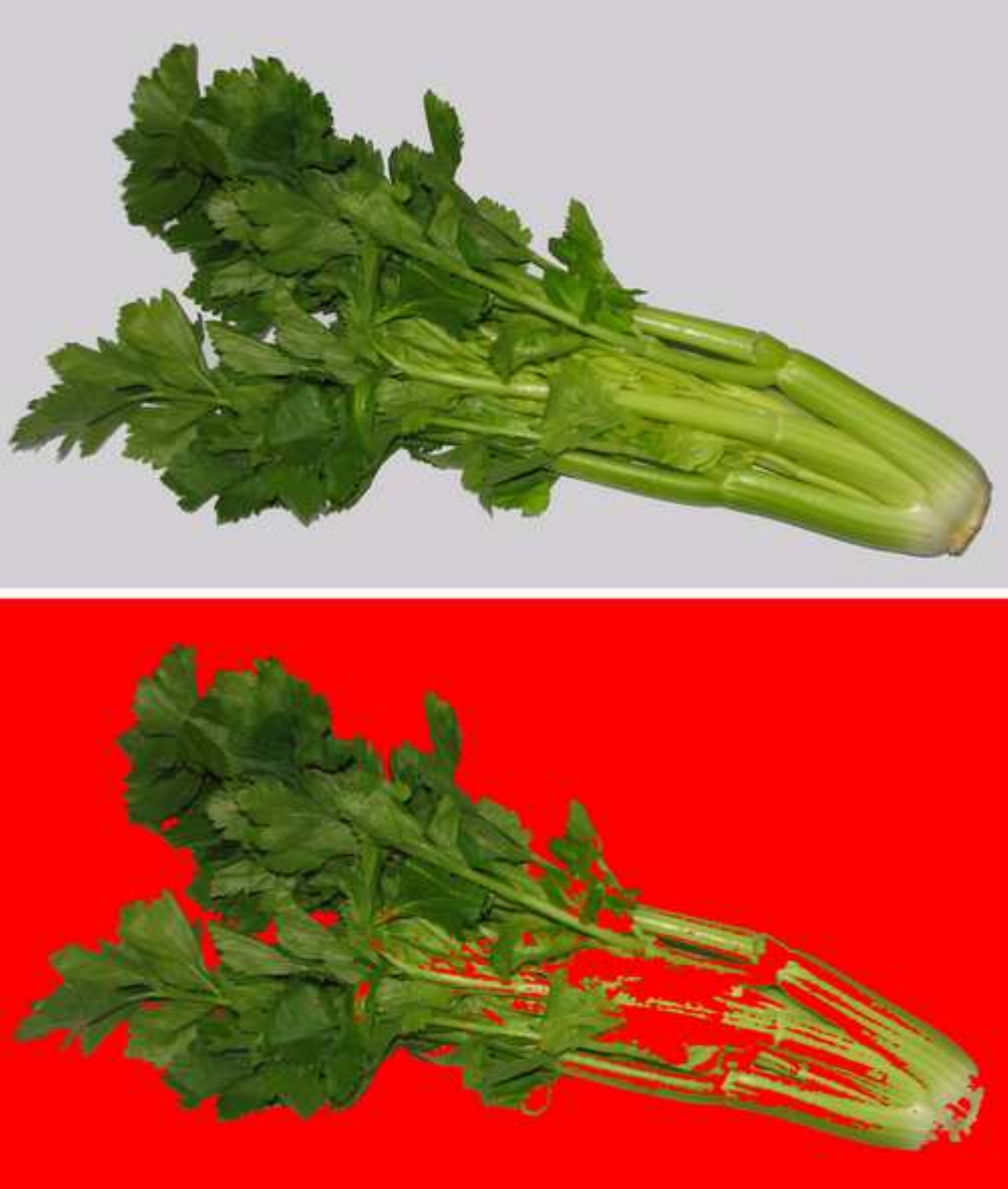,width=0.9in}
\psfig{figure=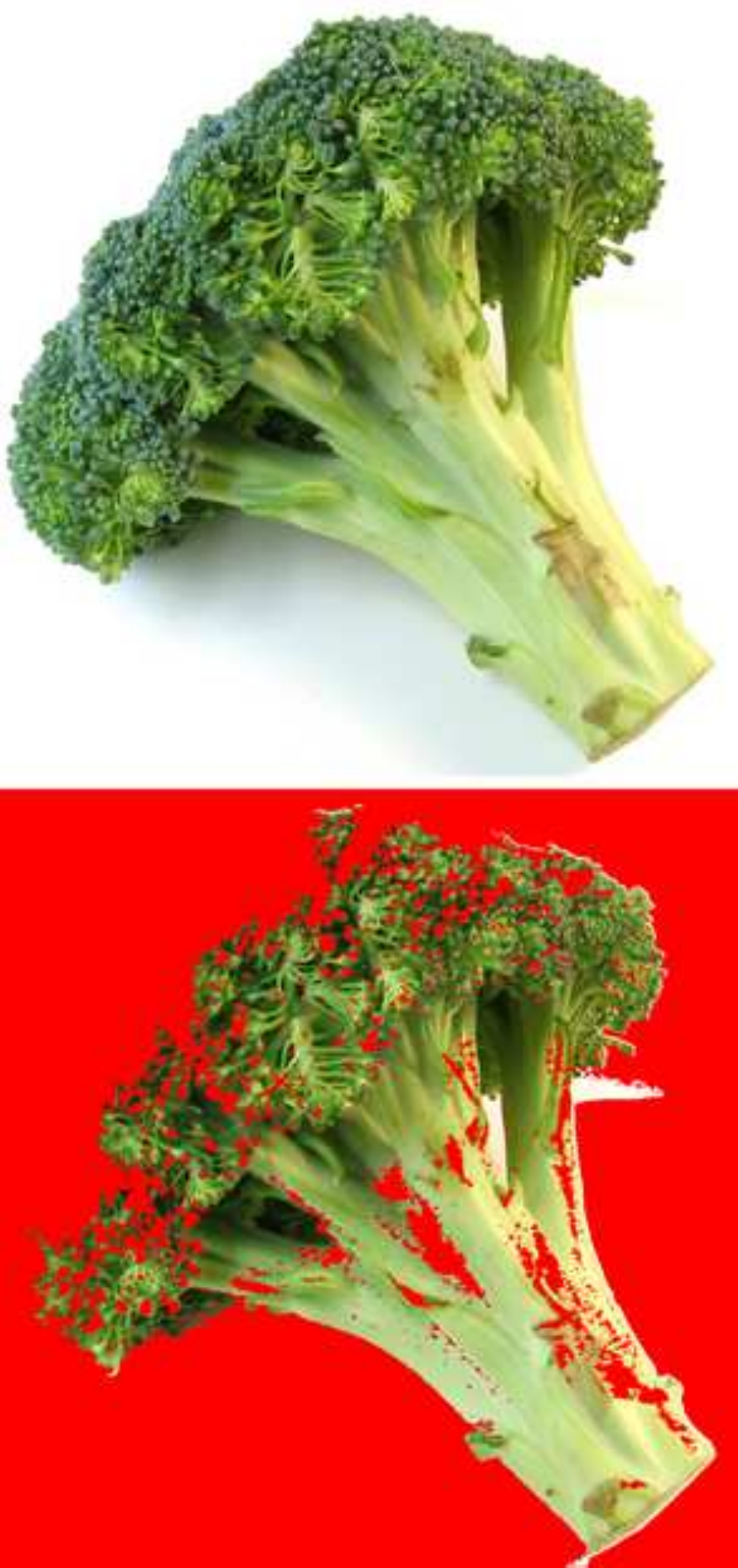,width=0.5in}
\psfig{figure=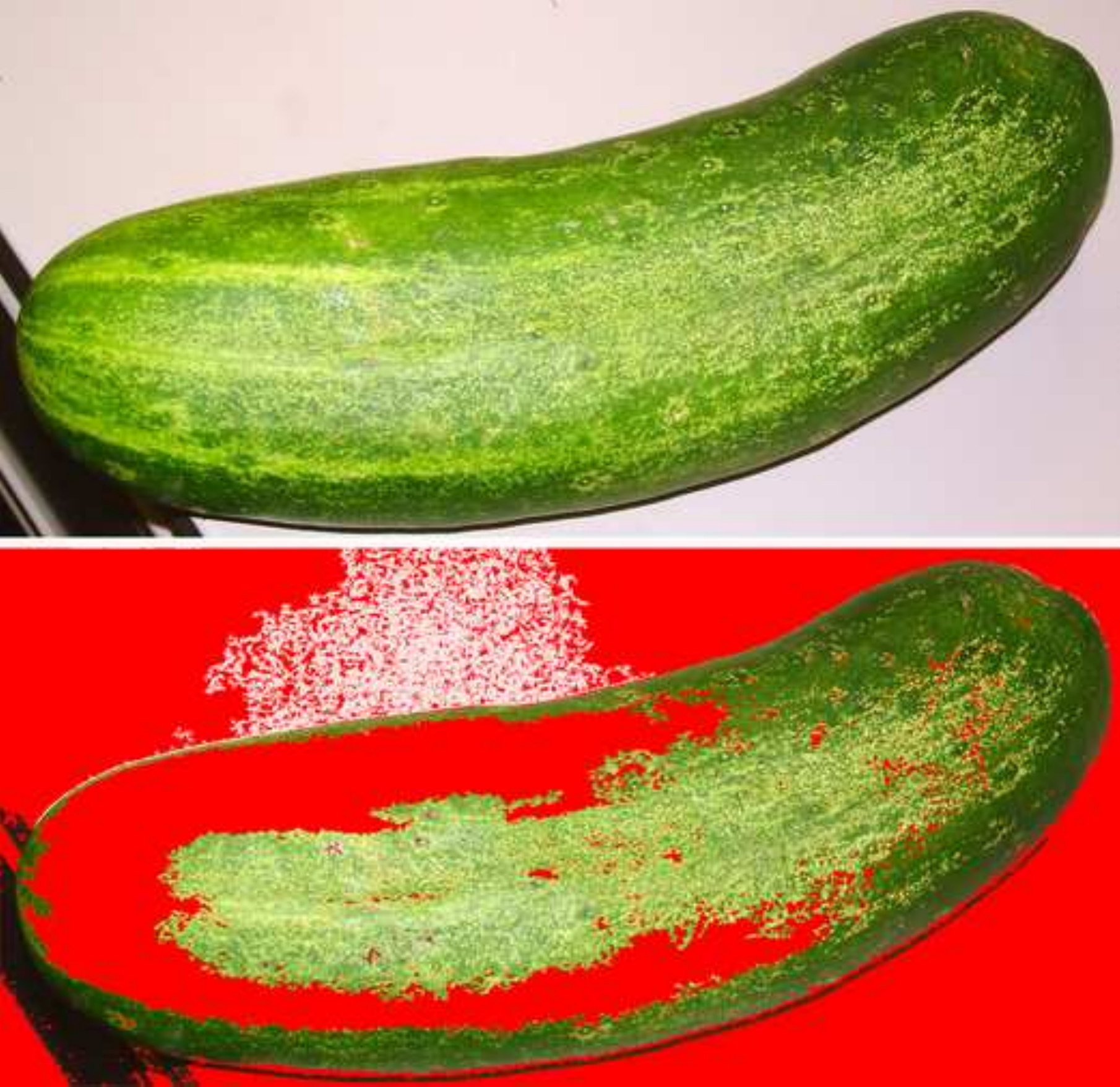,width=0.9in}
\psfig{figure=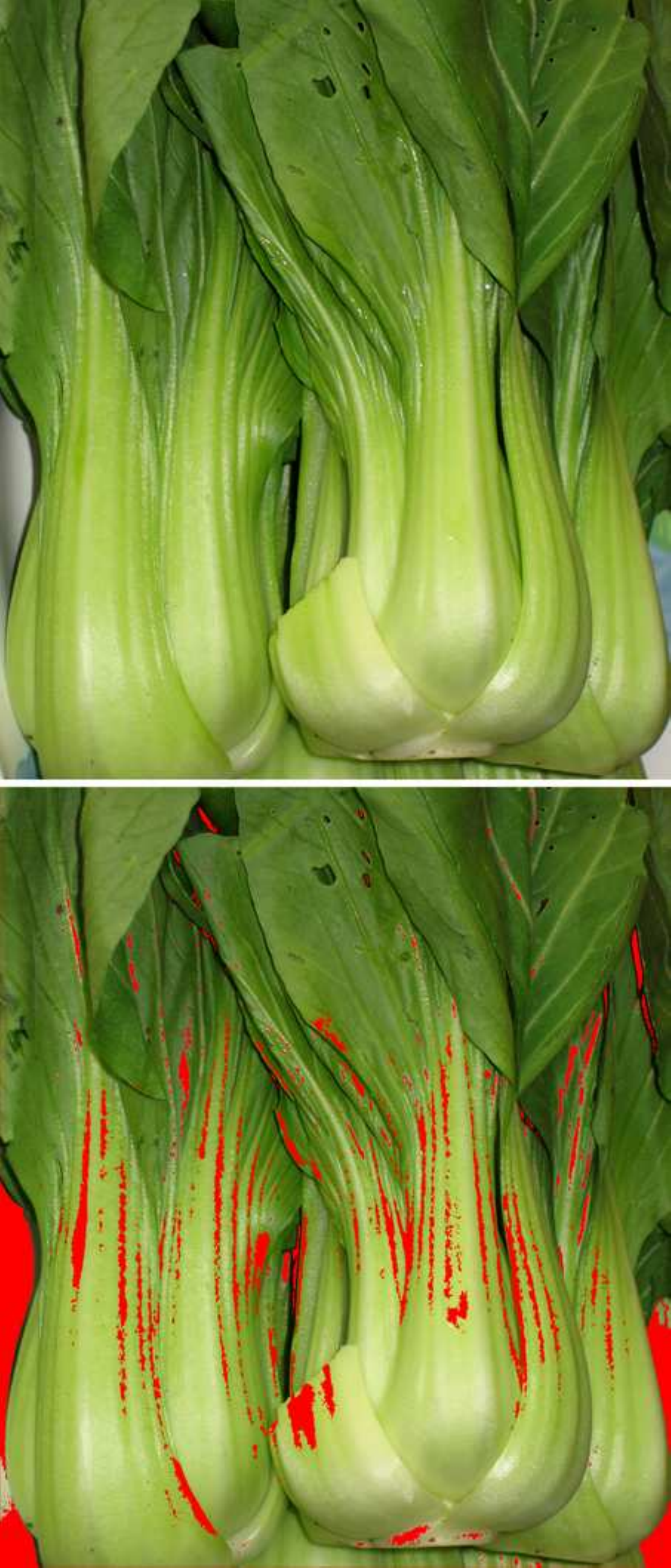,width=0.45in}
\psfig{figure=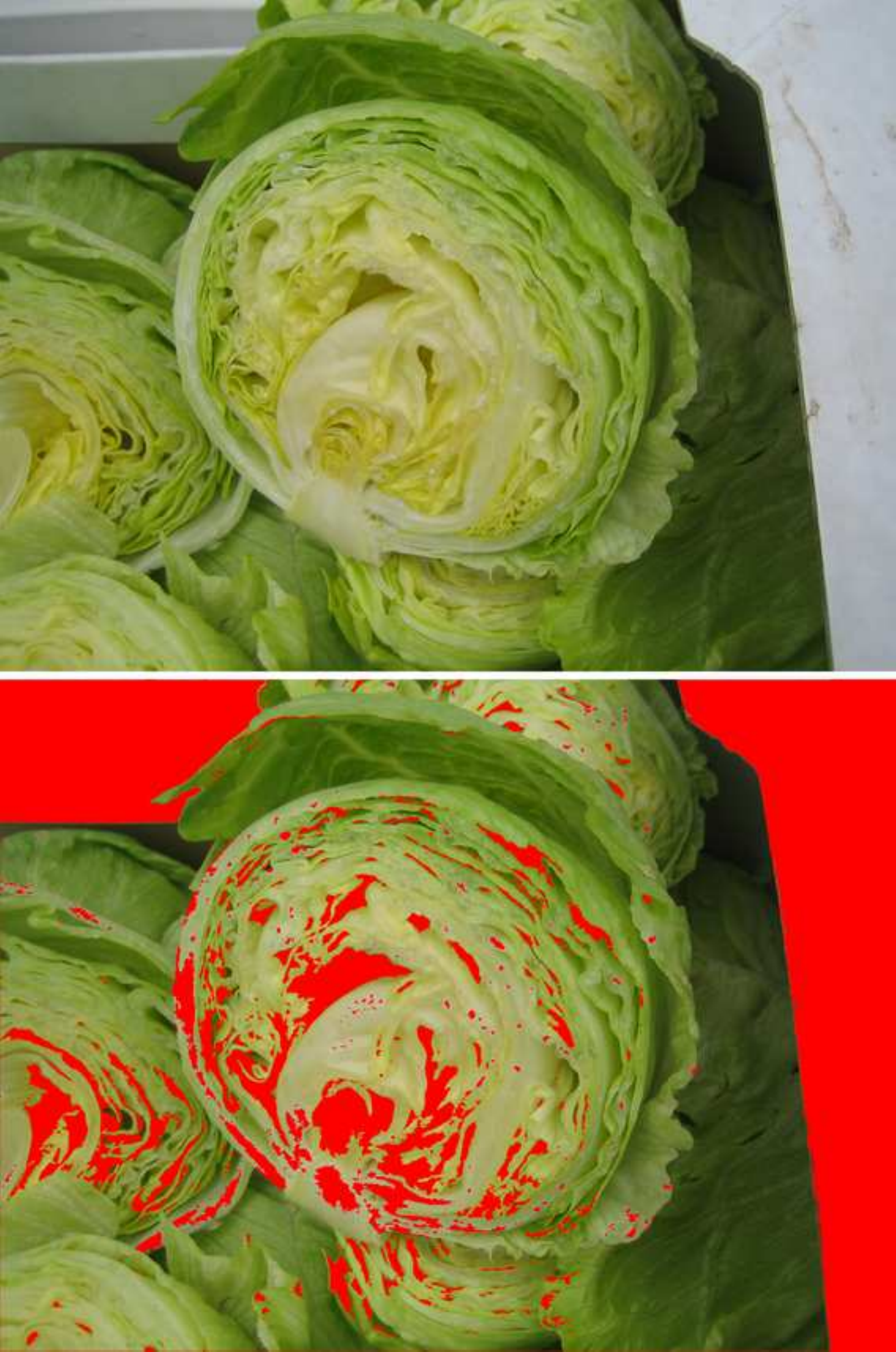,width=0.6in}
\psfig{figure=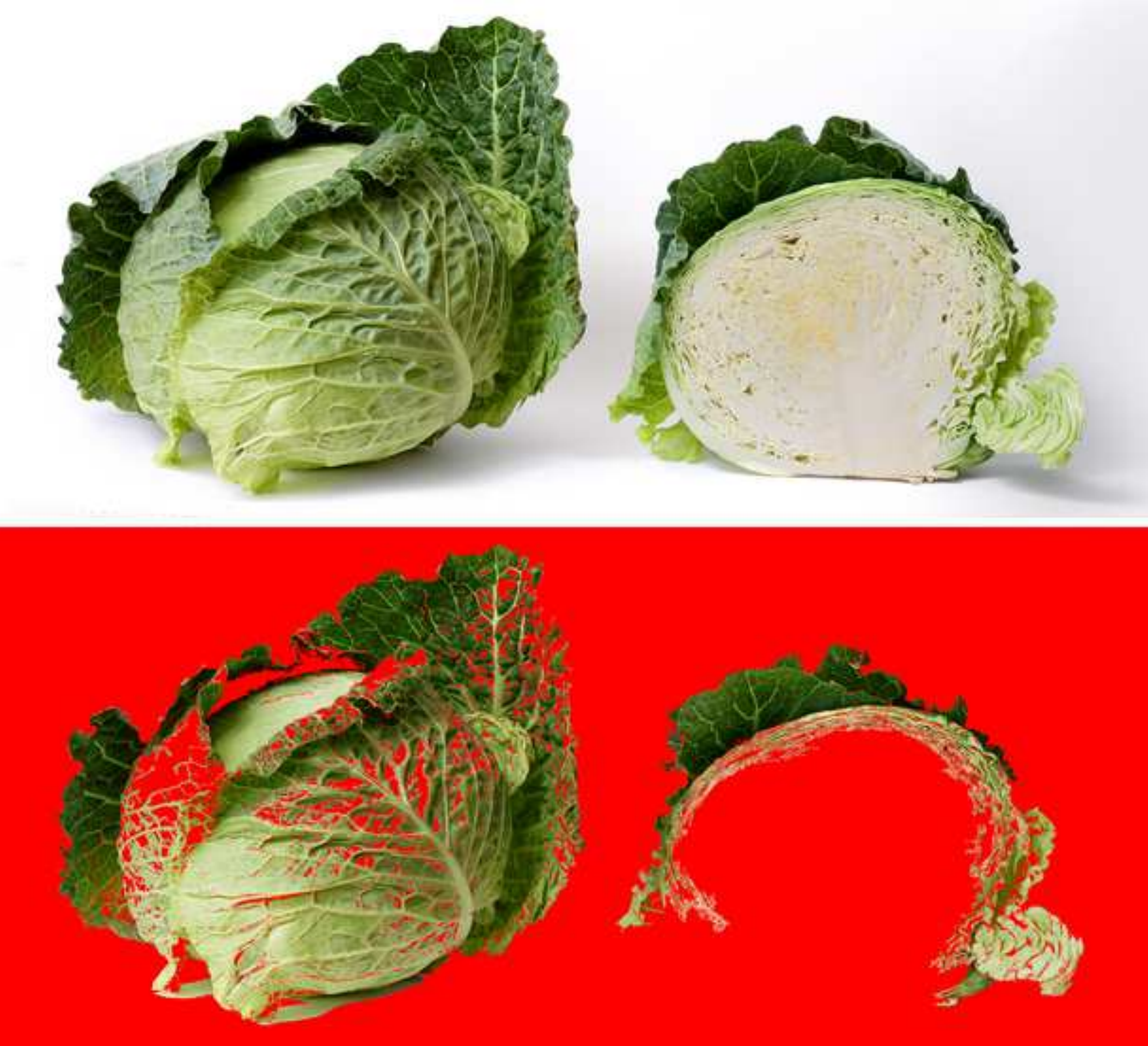,width=0.8in}
}
\centerline{
\psfig{figure=figs/train5.eps,width=0.8in}
\hspace{0.1in}
\psfig{figure=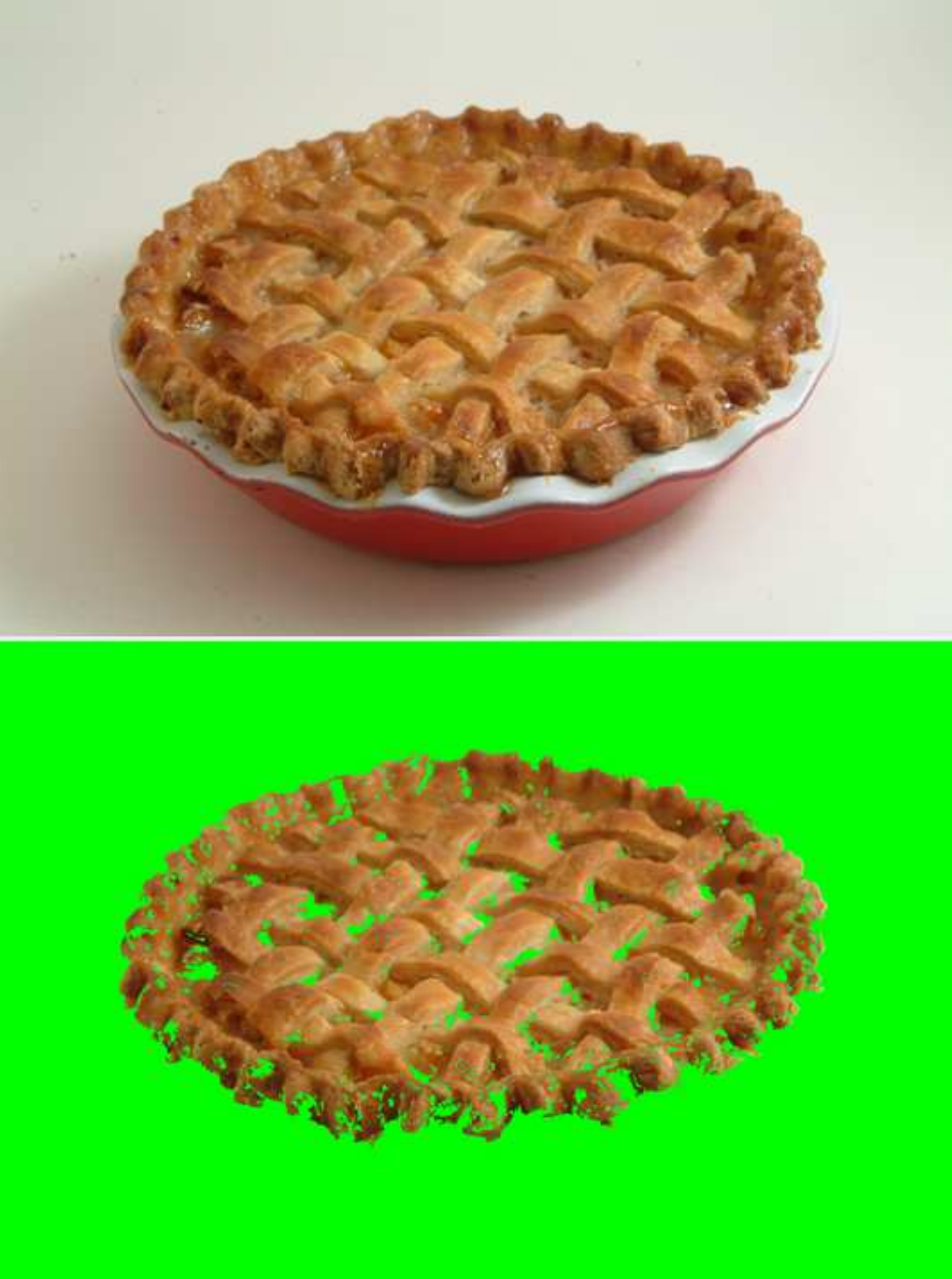,width=0.8in}
\psfig{figure=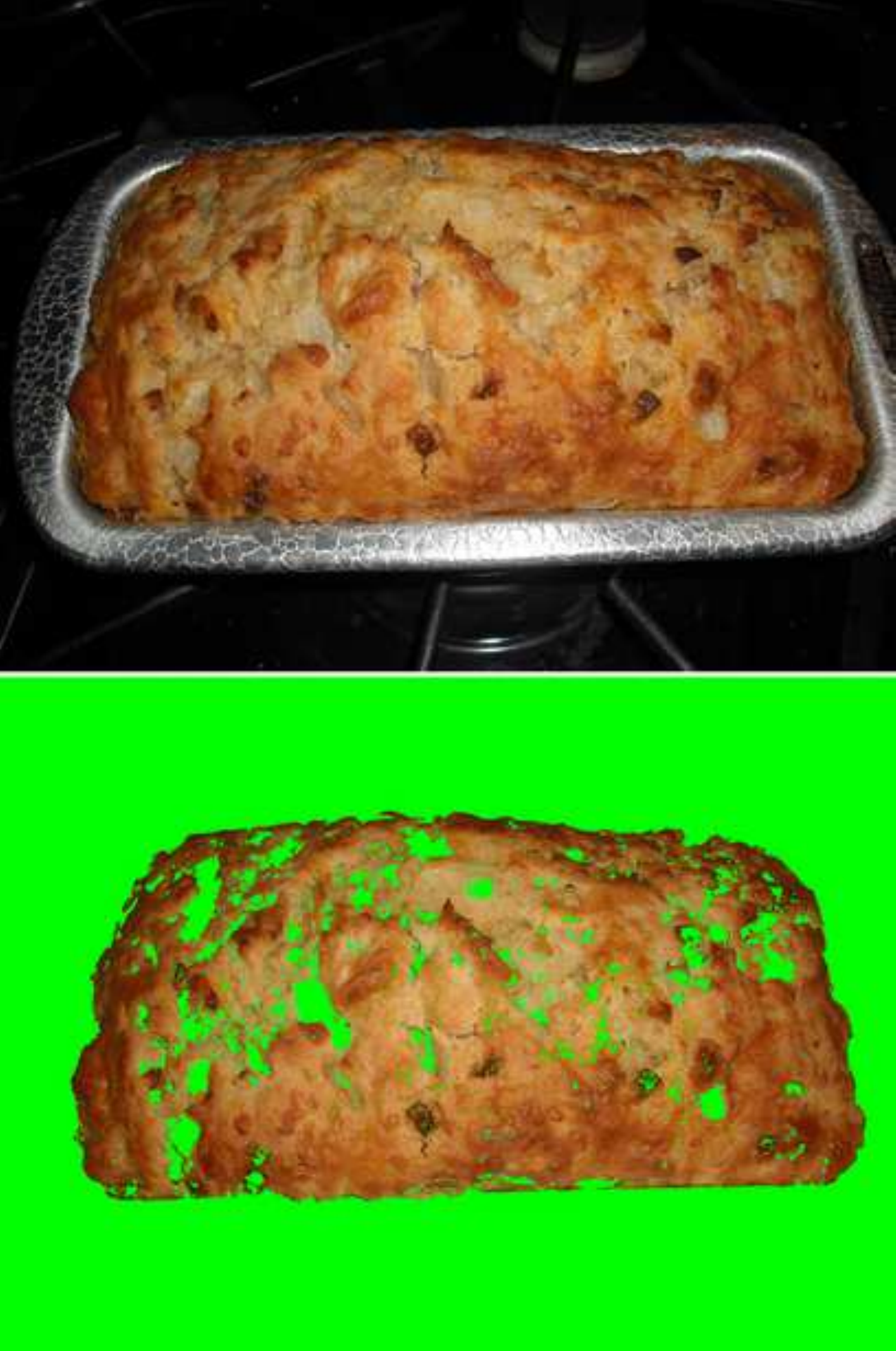,width=0.75in}
\psfig{figure=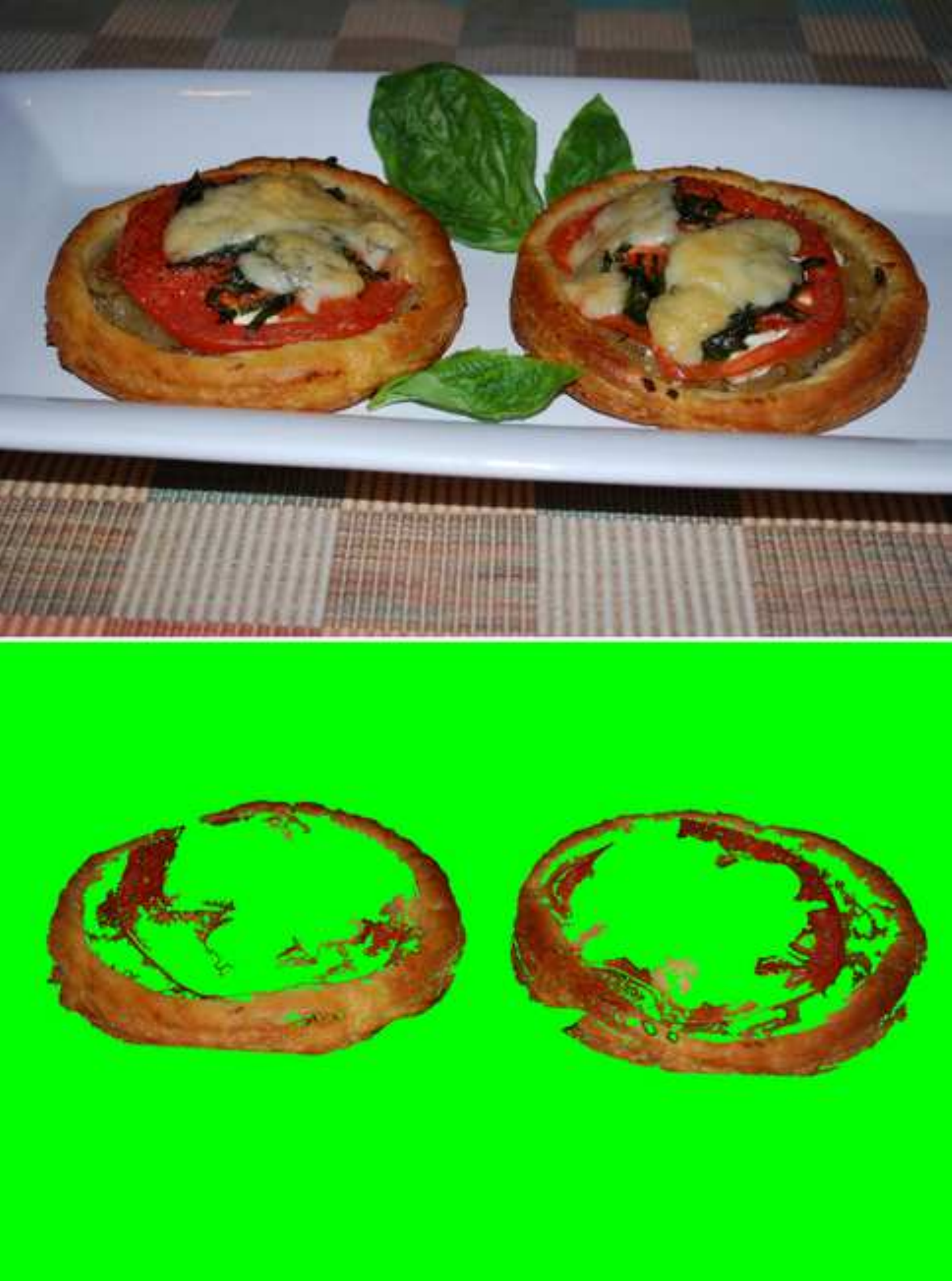,width=0.8in}
\psfig{figure=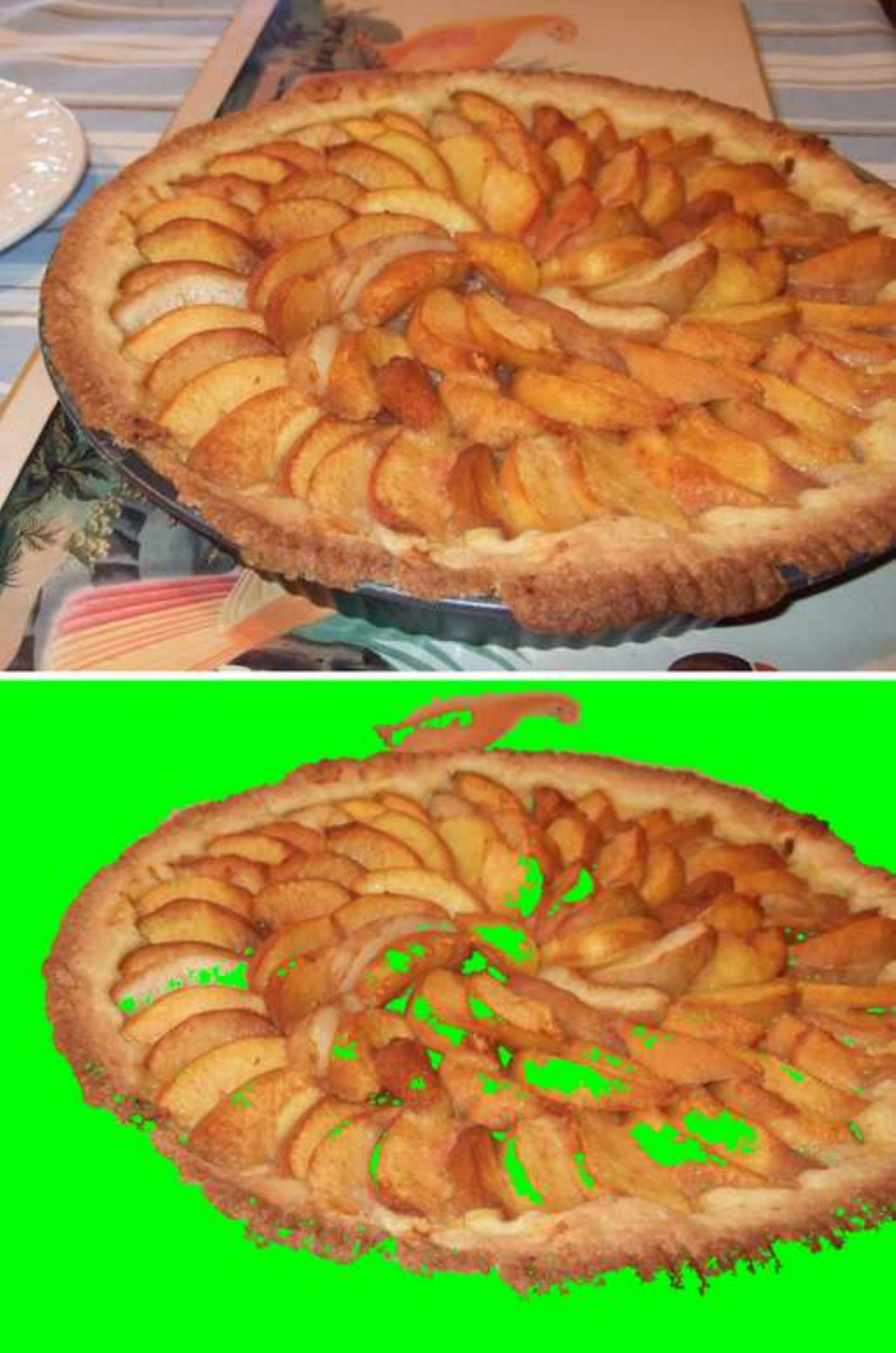,width=0.75in}
\psfig{figure=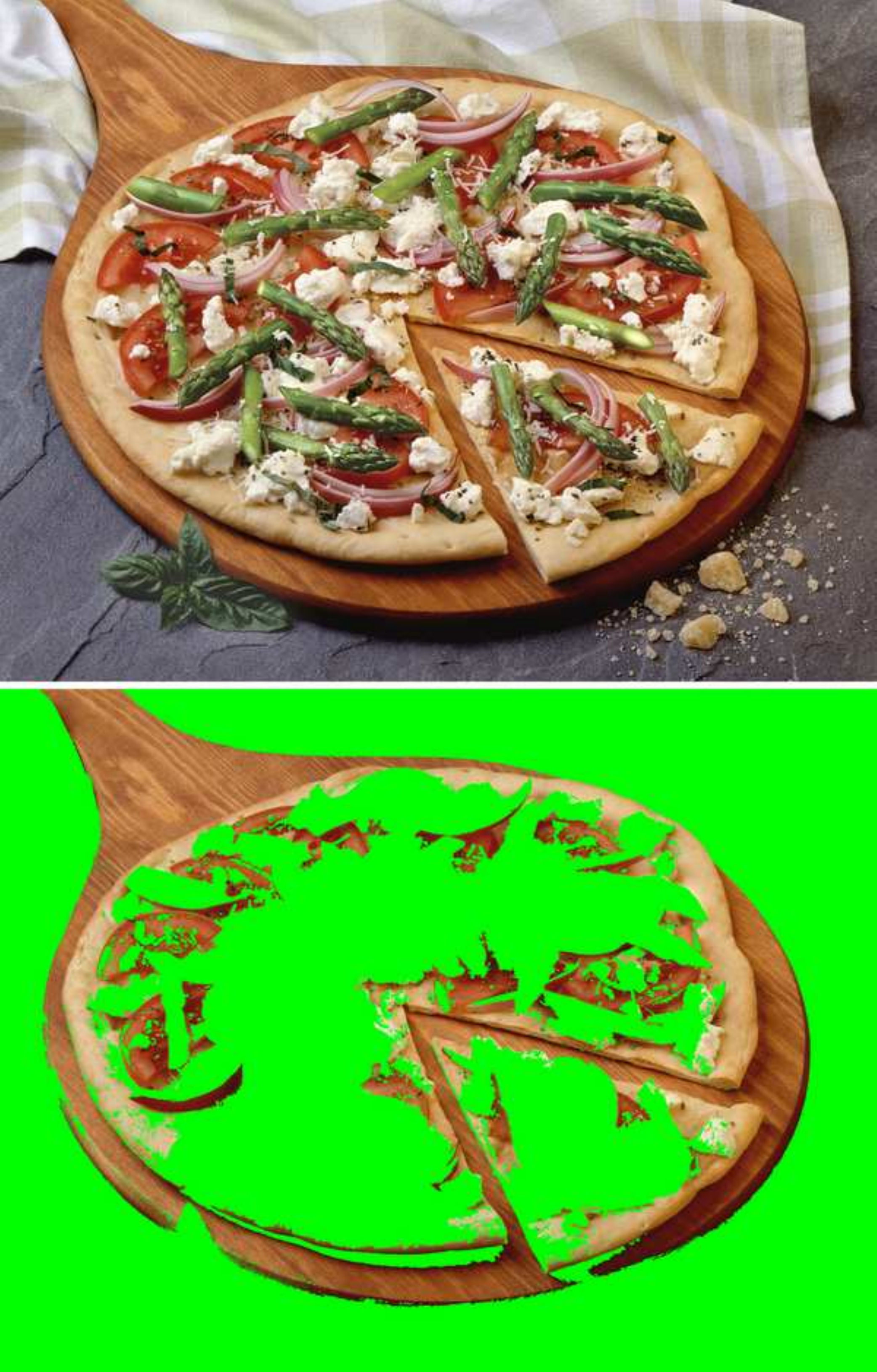,width=0.75in}
\psfig{figure=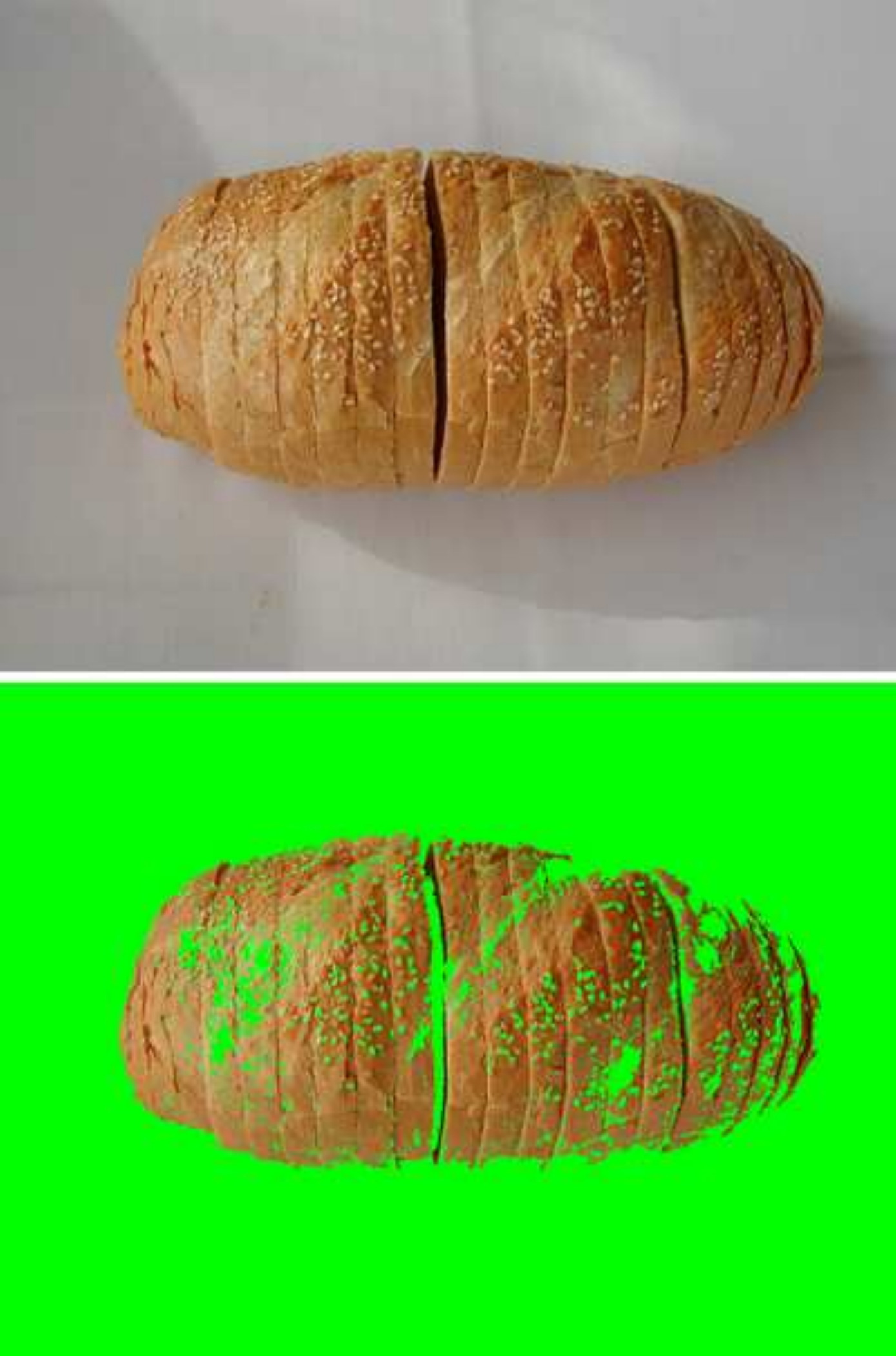,width=0.75in}
\psfig{figure=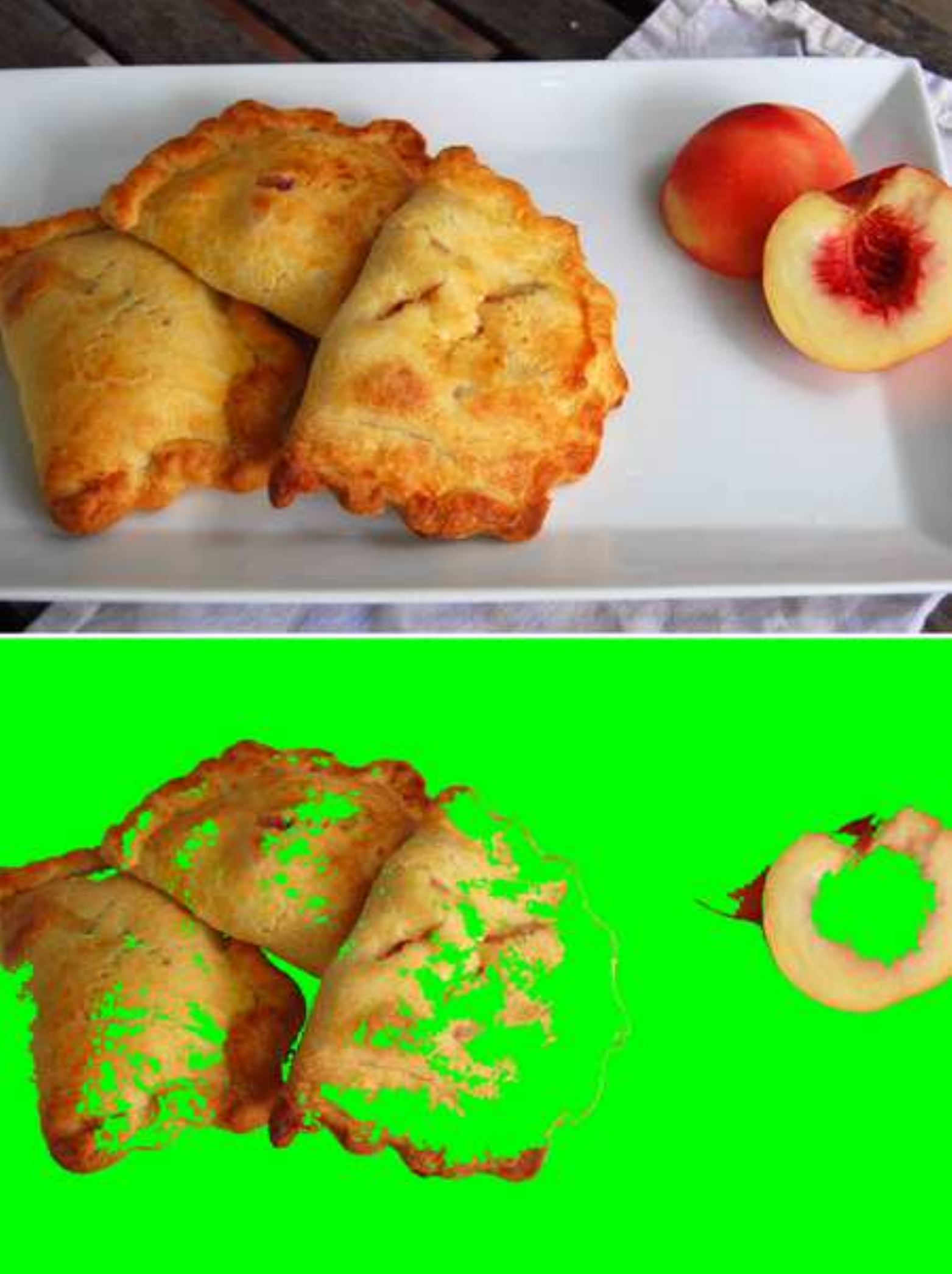,width=0.7in}
\psfig{figure=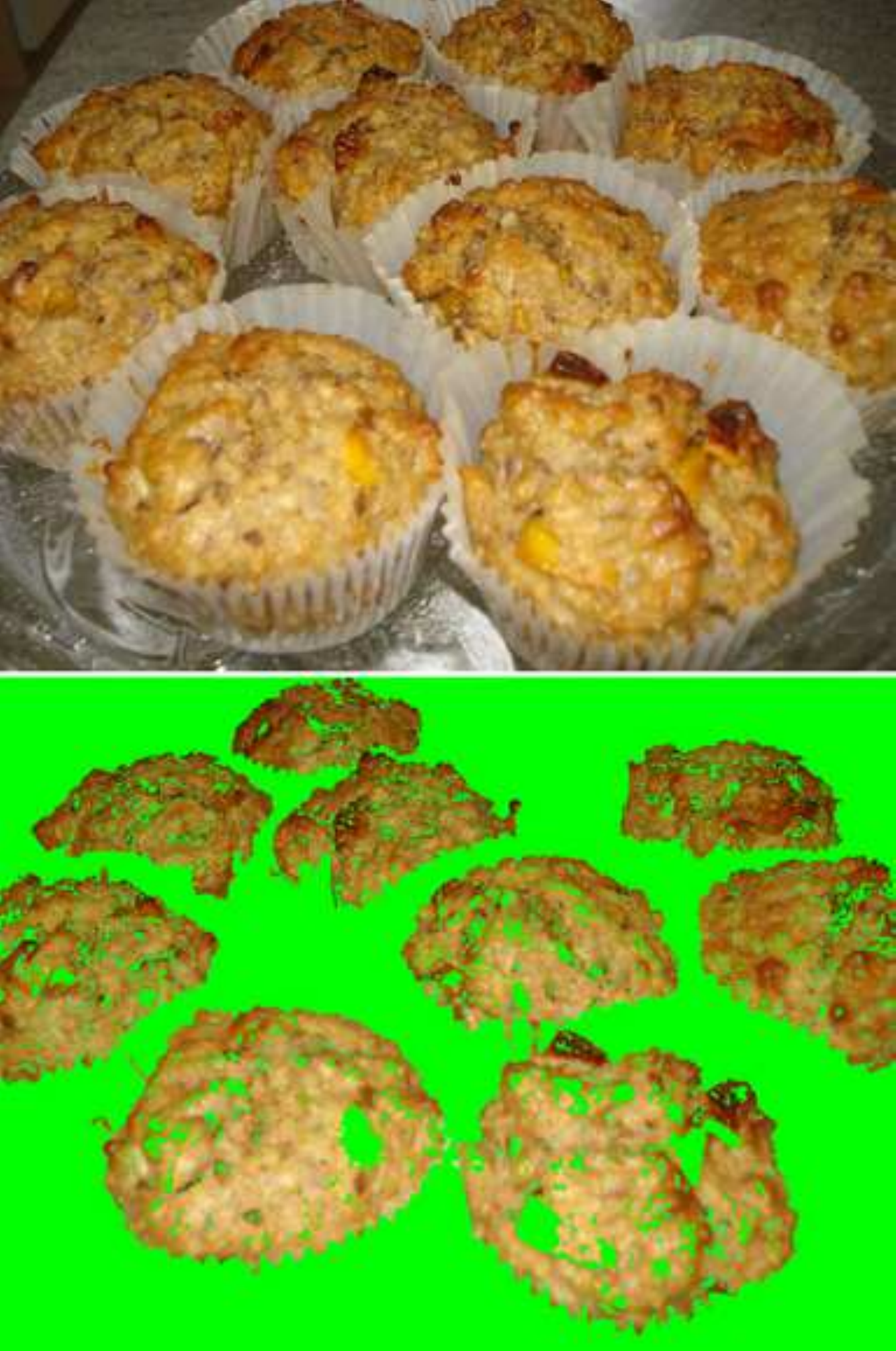,width=0.75in}
}
\centerline{
\psfig{figure=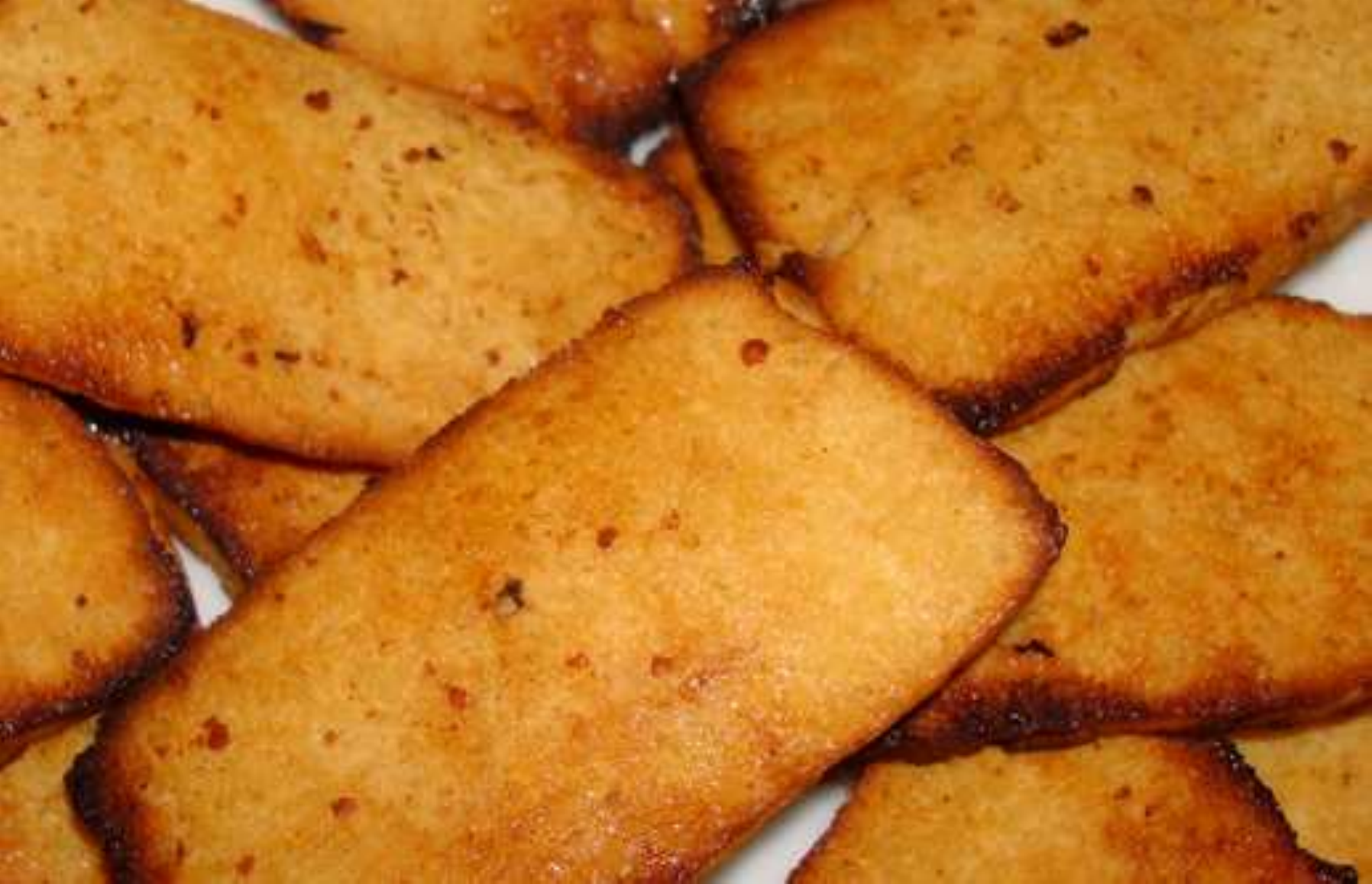,width=0.8in}
\hspace{0.1in}
\psfig{figure=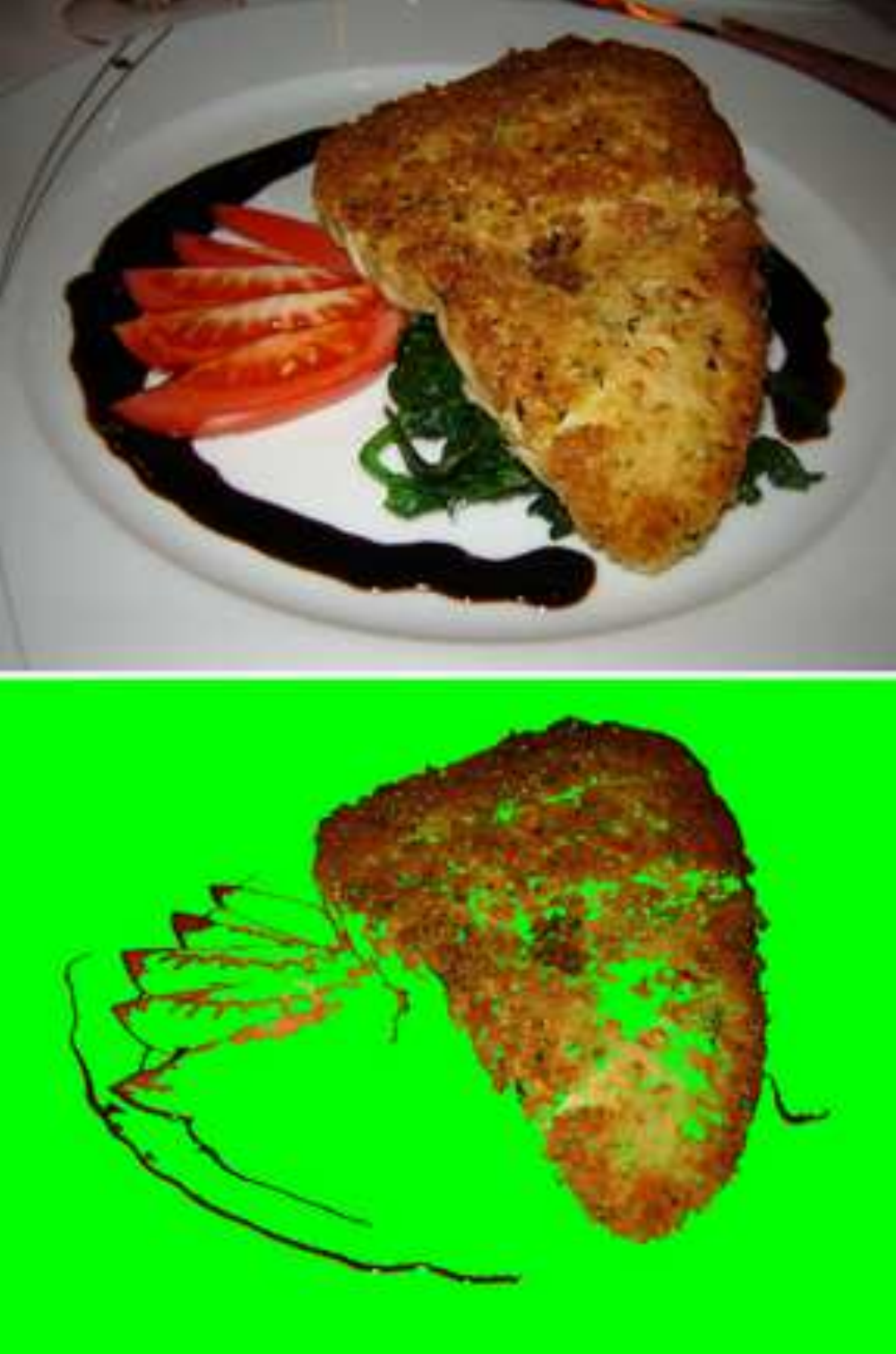,width=0.7in}
\psfig{figure=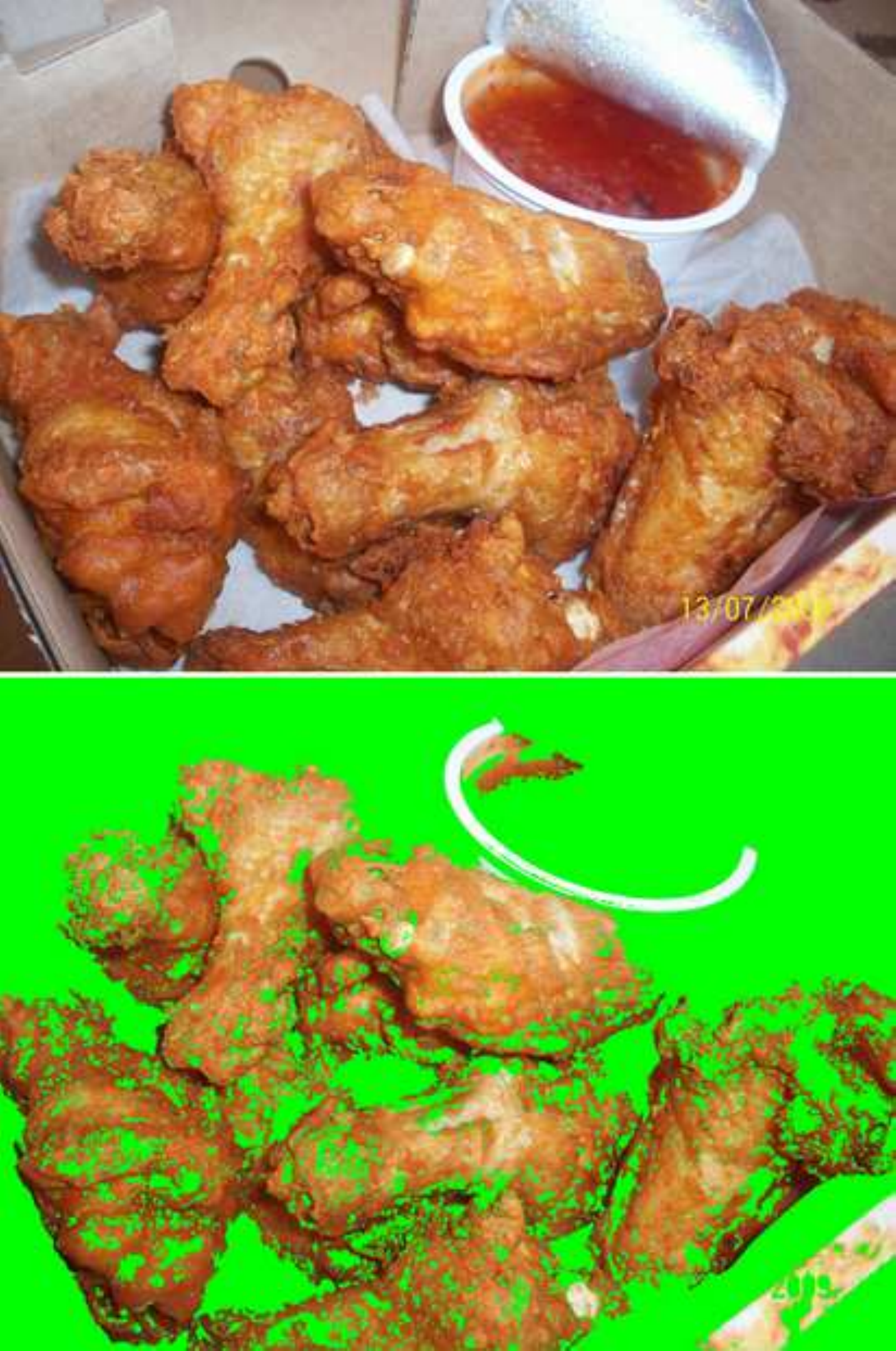,width=0.7in}
\psfig{figure=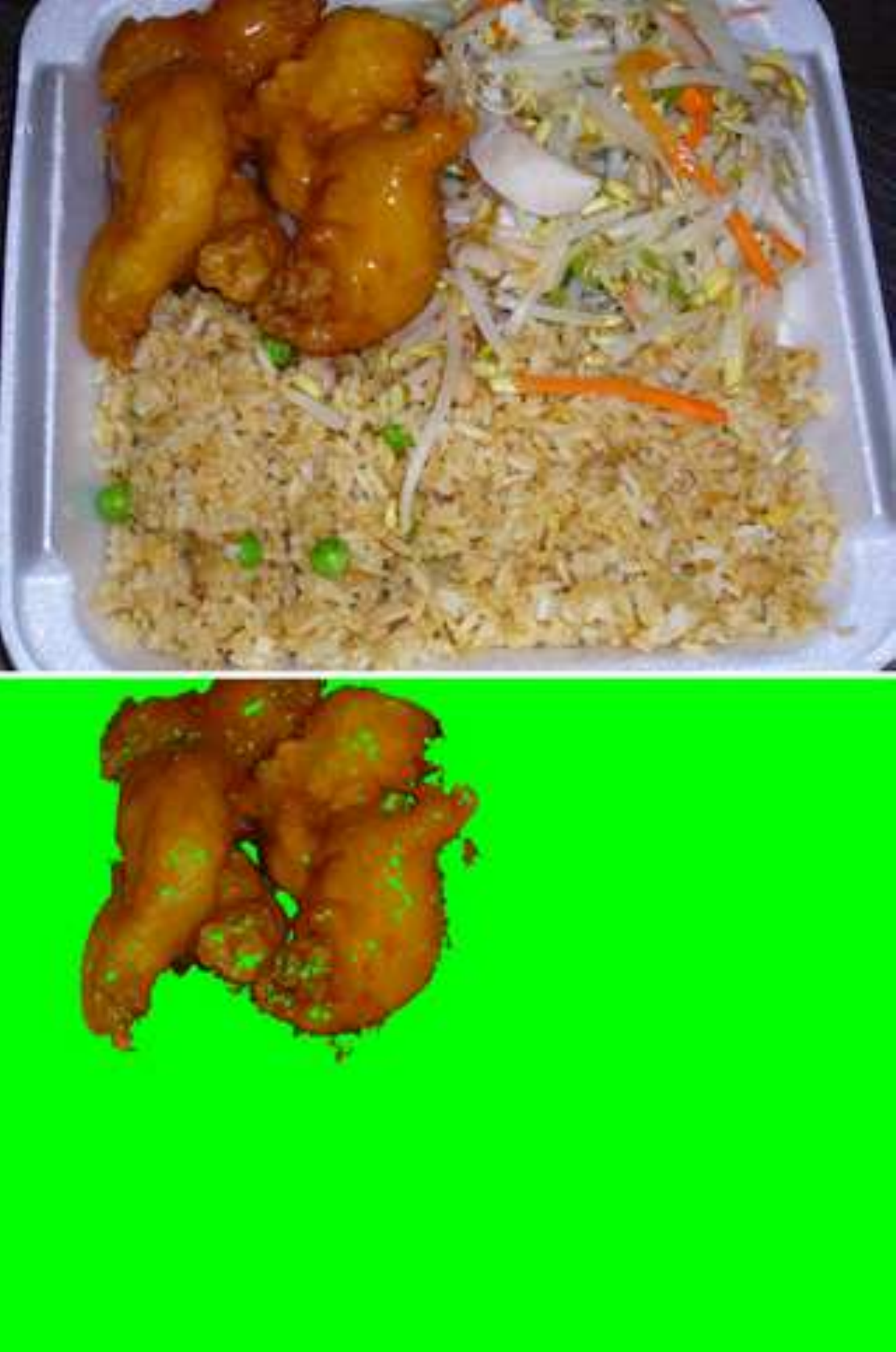,width=0.7in}
\psfig{figure=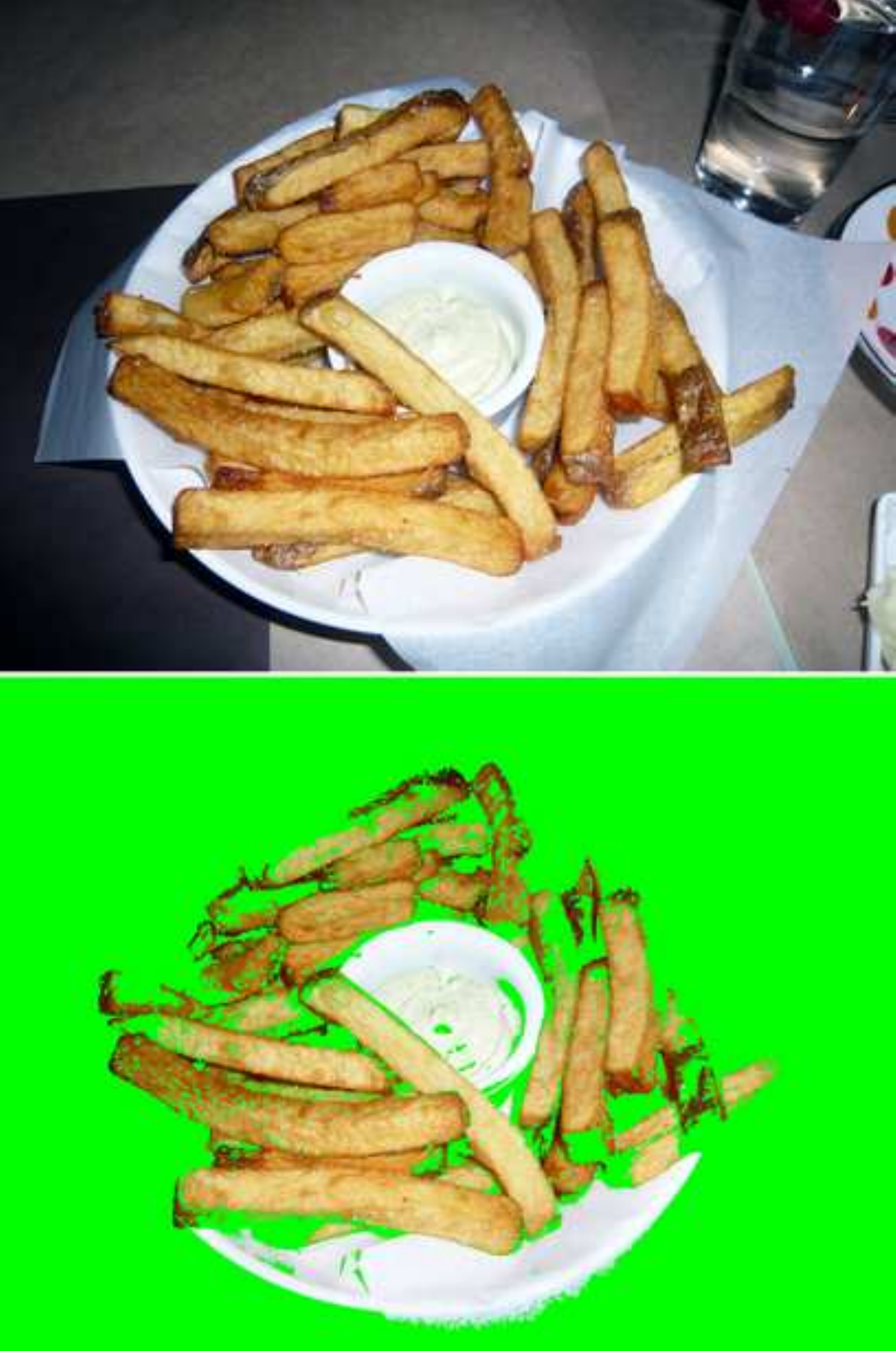,width=0.7in}
\psfig{figure=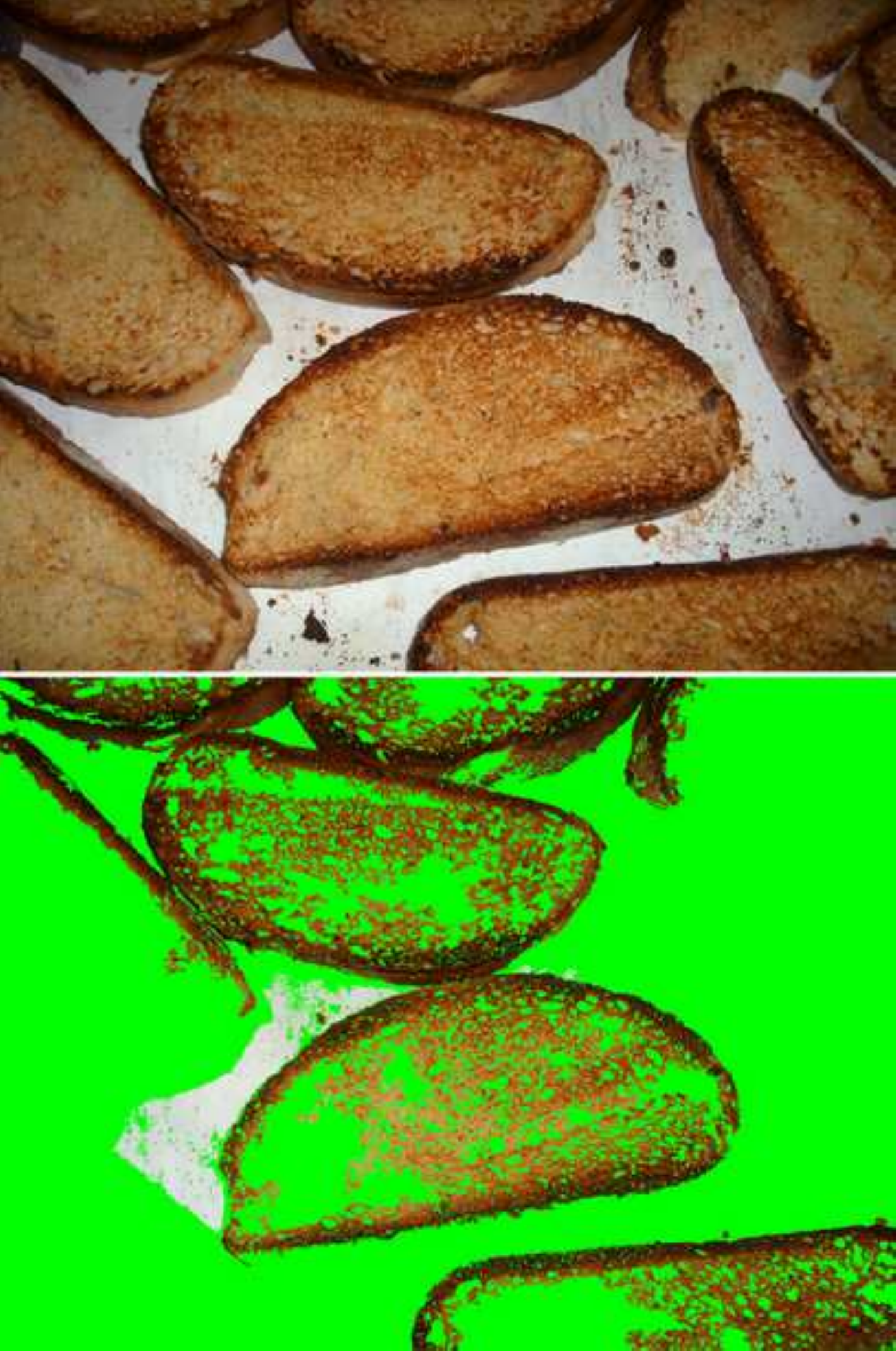,width=0.7in}
\psfig{figure=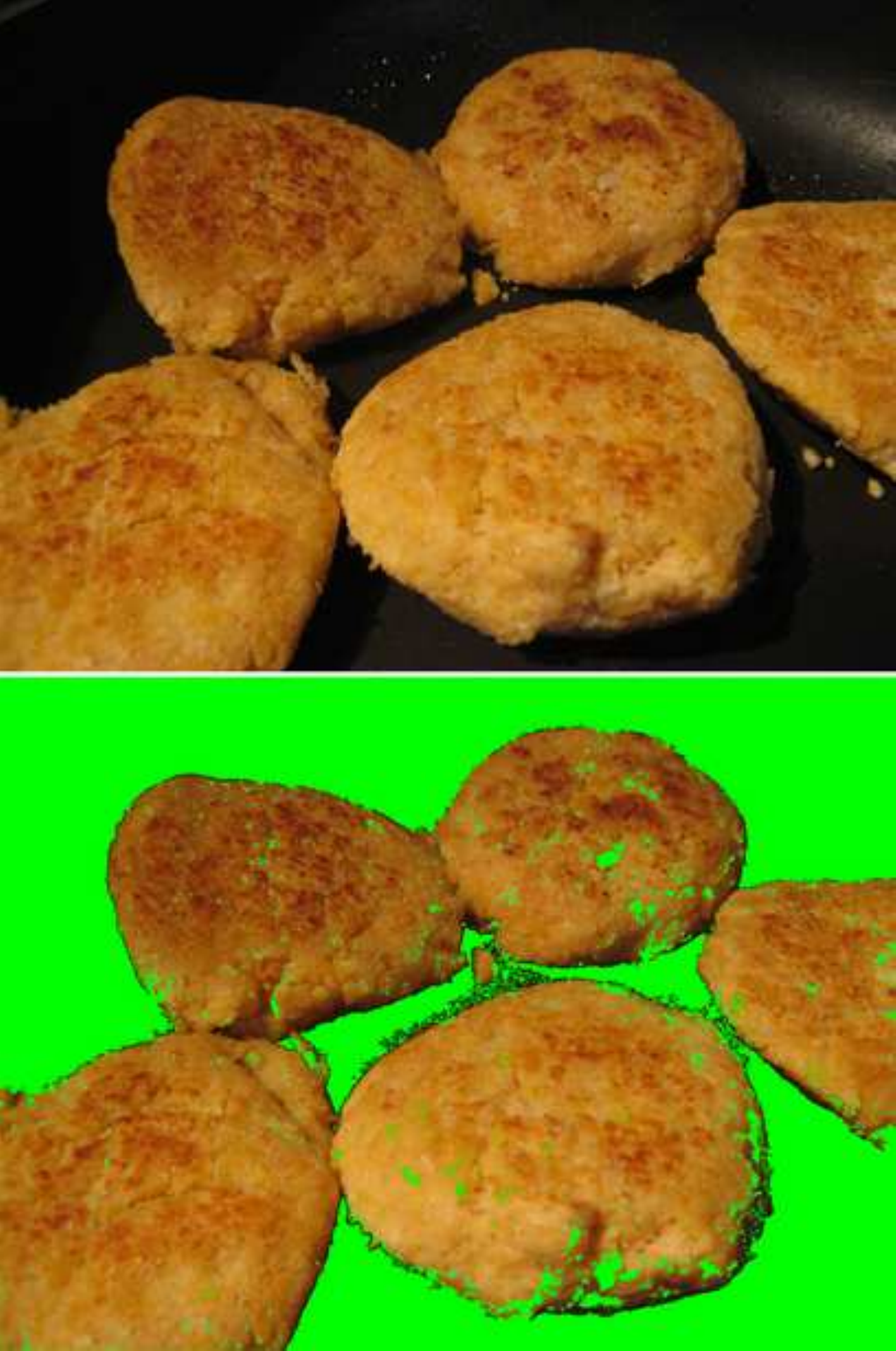,width=0.7in}
\psfig{figure=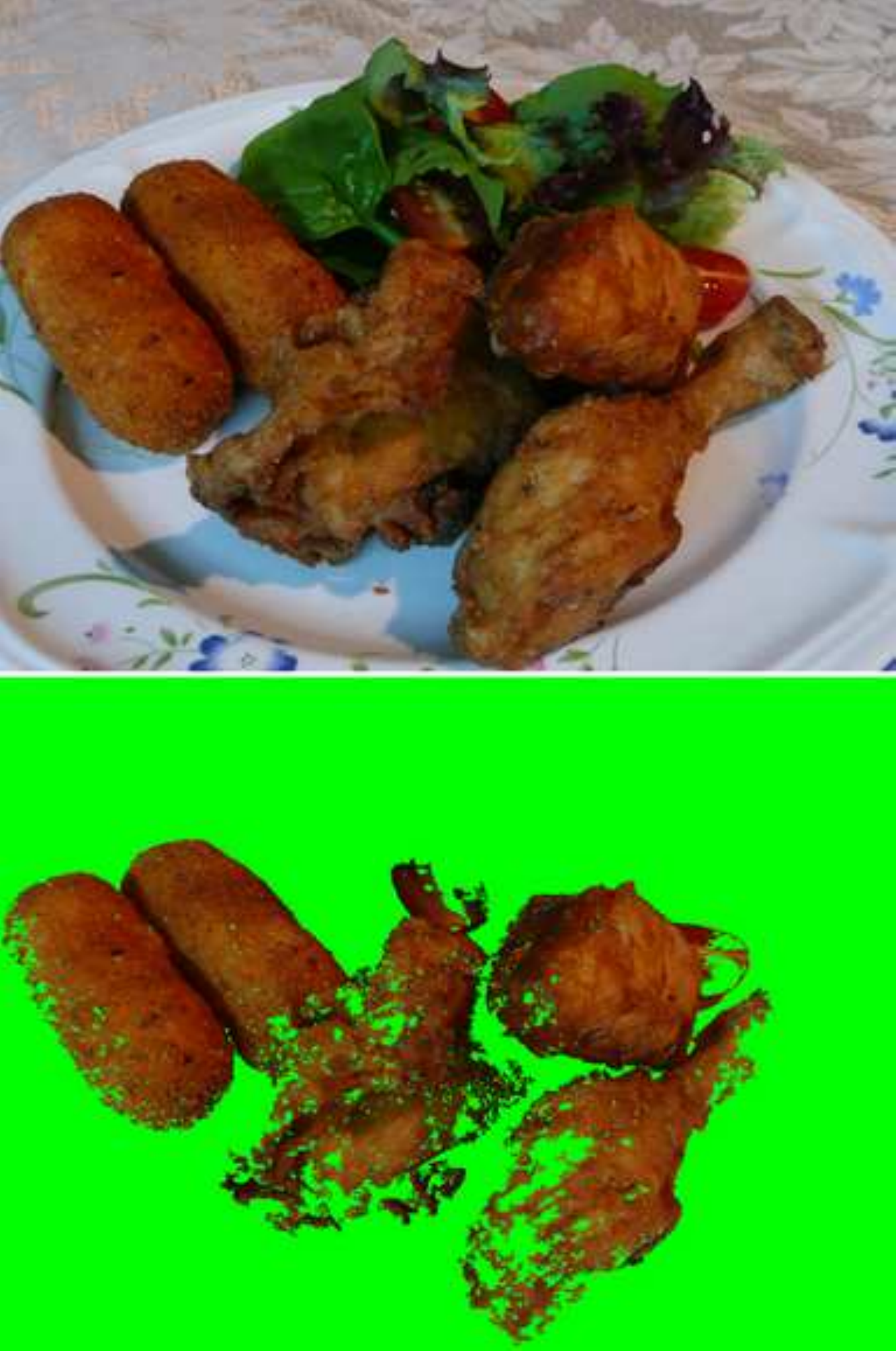,width=0.7in}
\psfig{figure=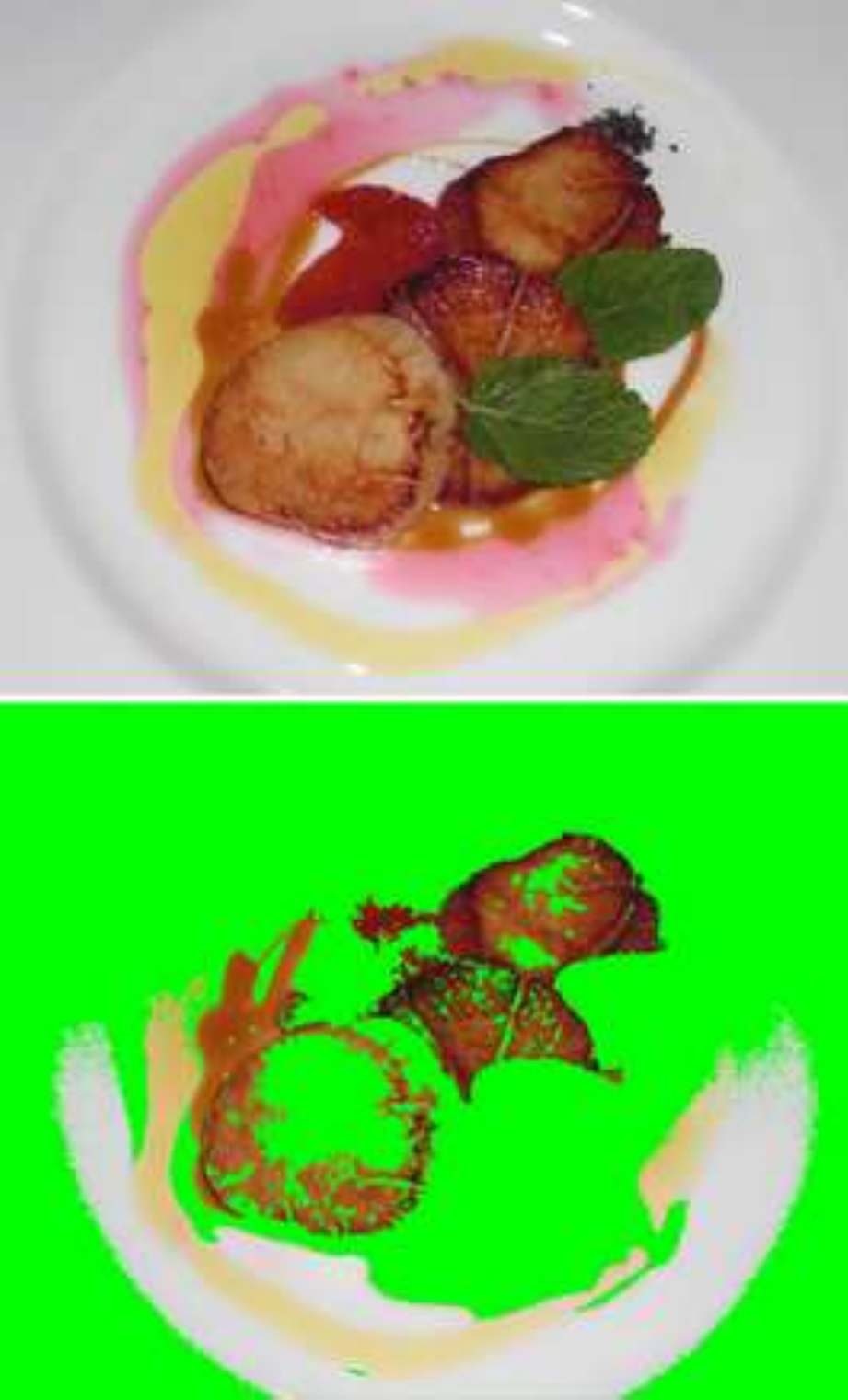,width=0.6in}
}
\centerline{
\psfig{figure=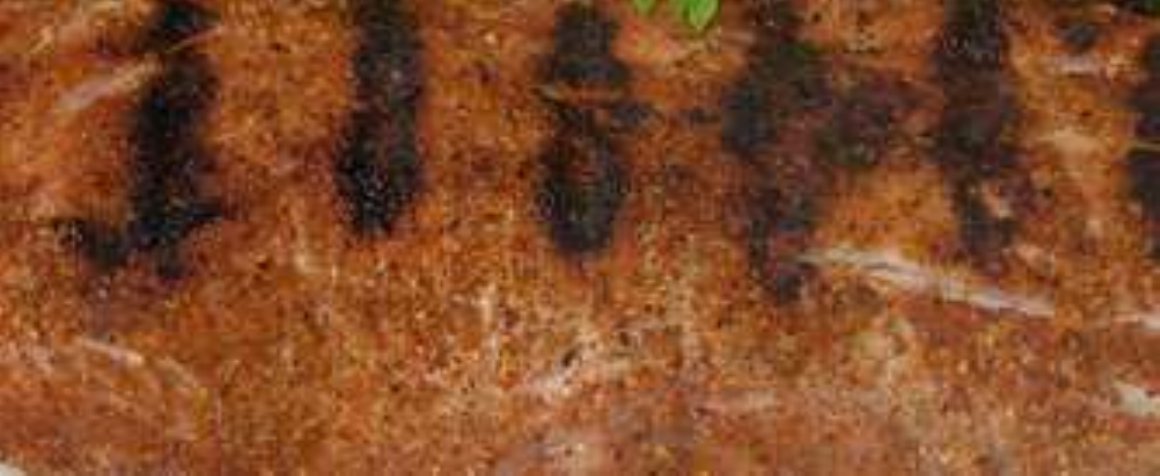,width=0.9in}
\hspace{0.1in}
\psfig{figure=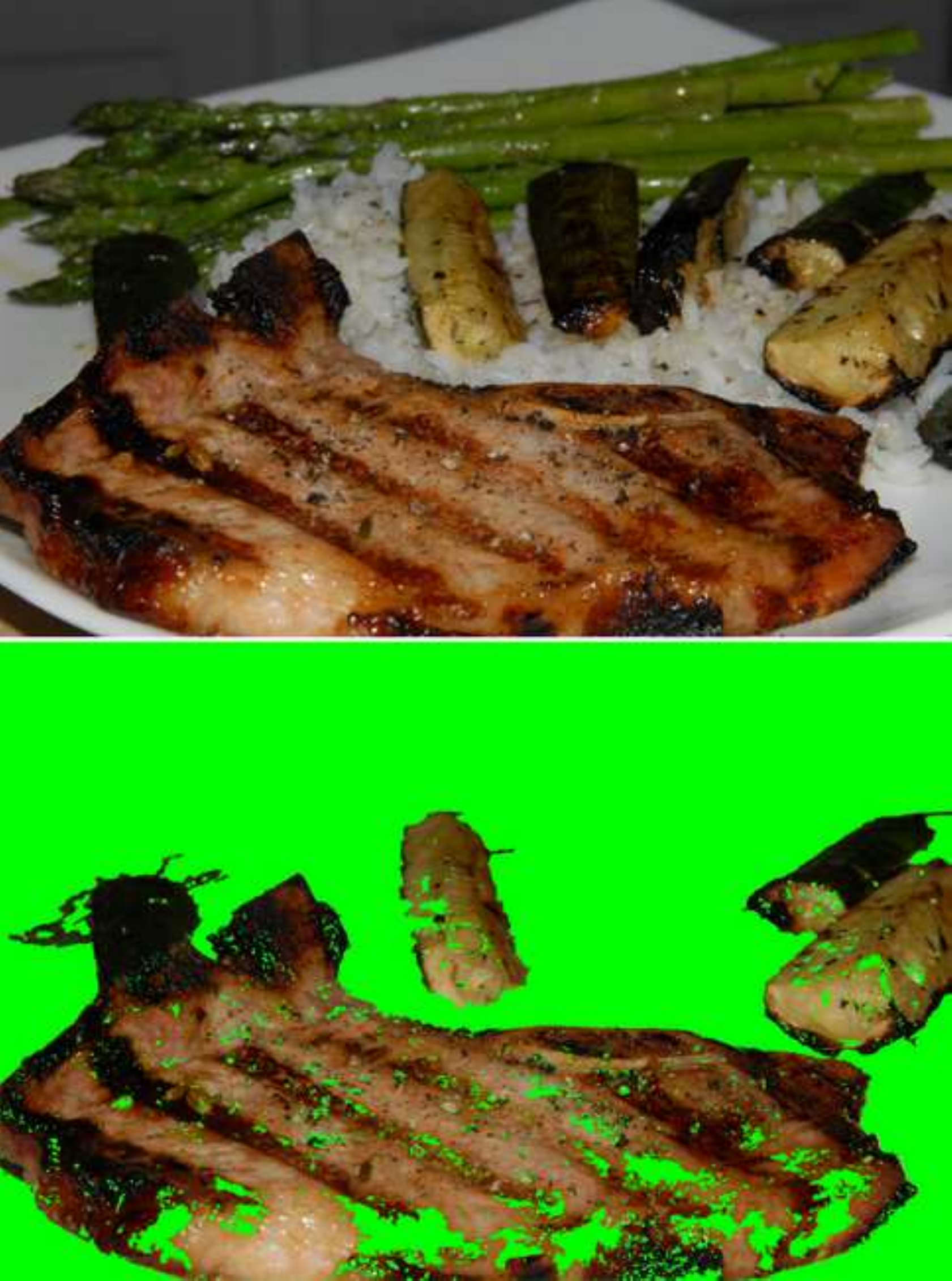,width=0.7in}
\psfig{figure=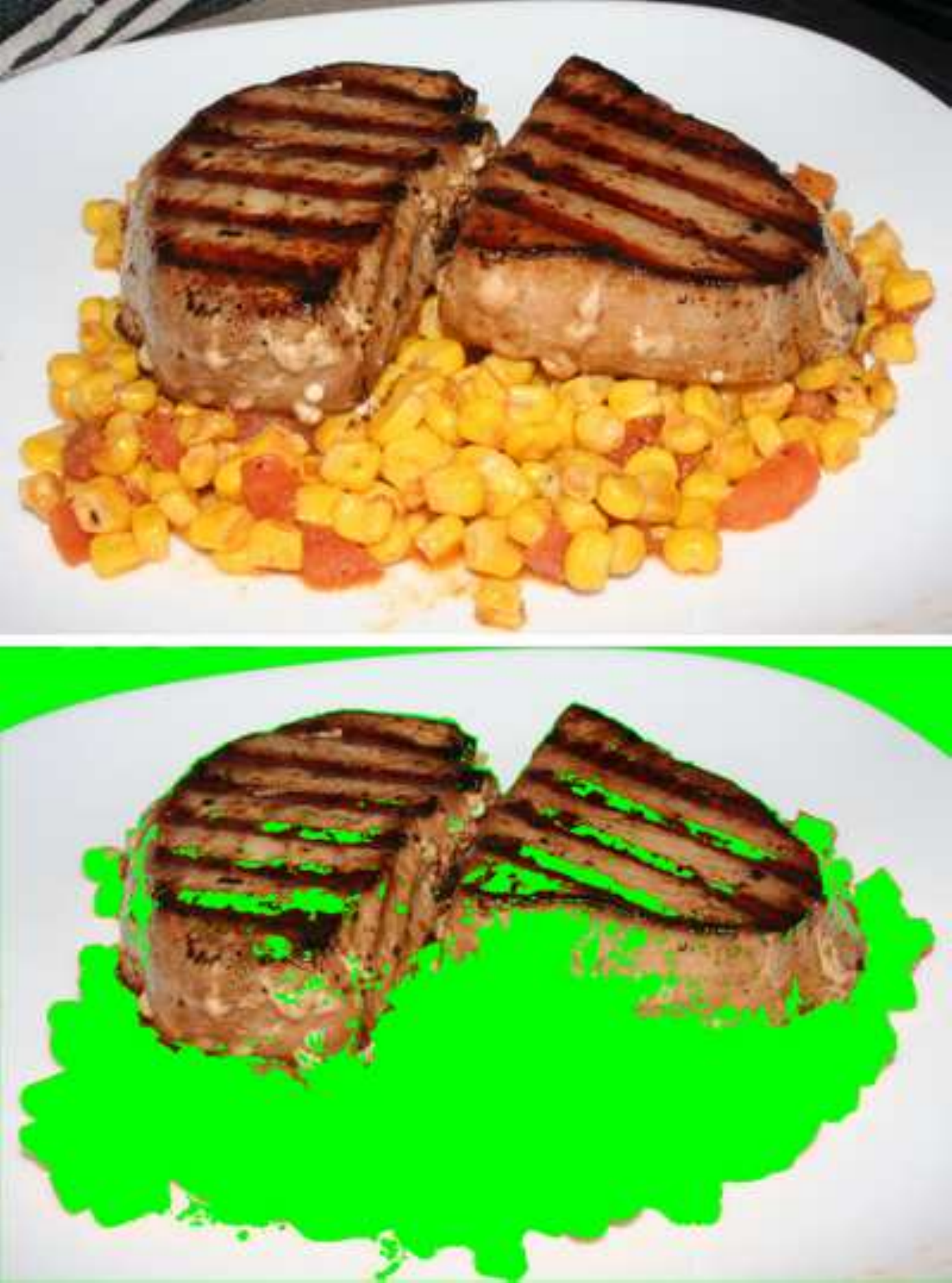,width=0.7in}
\psfig{figure=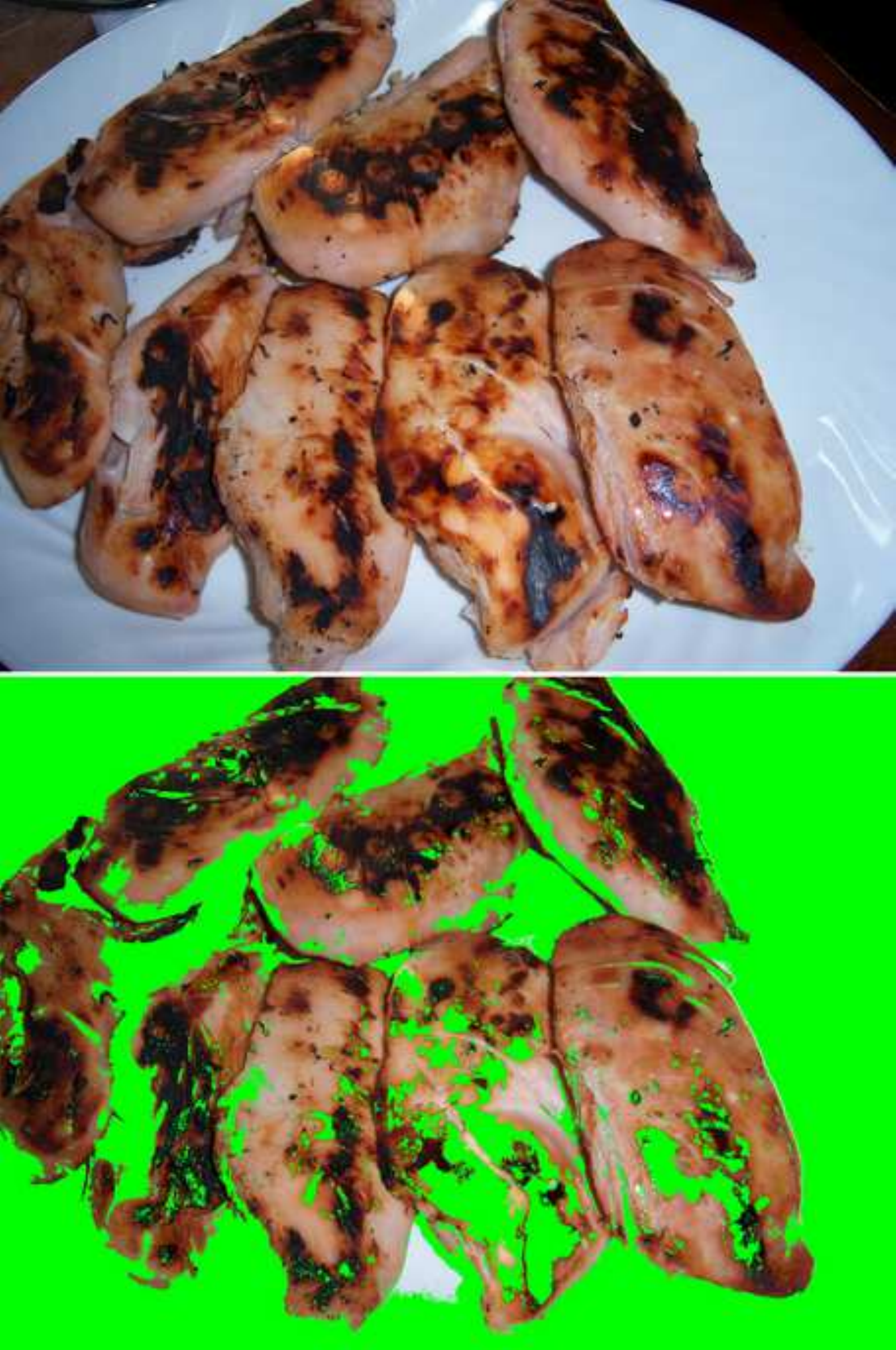,width=0.7in}
\psfig{figure=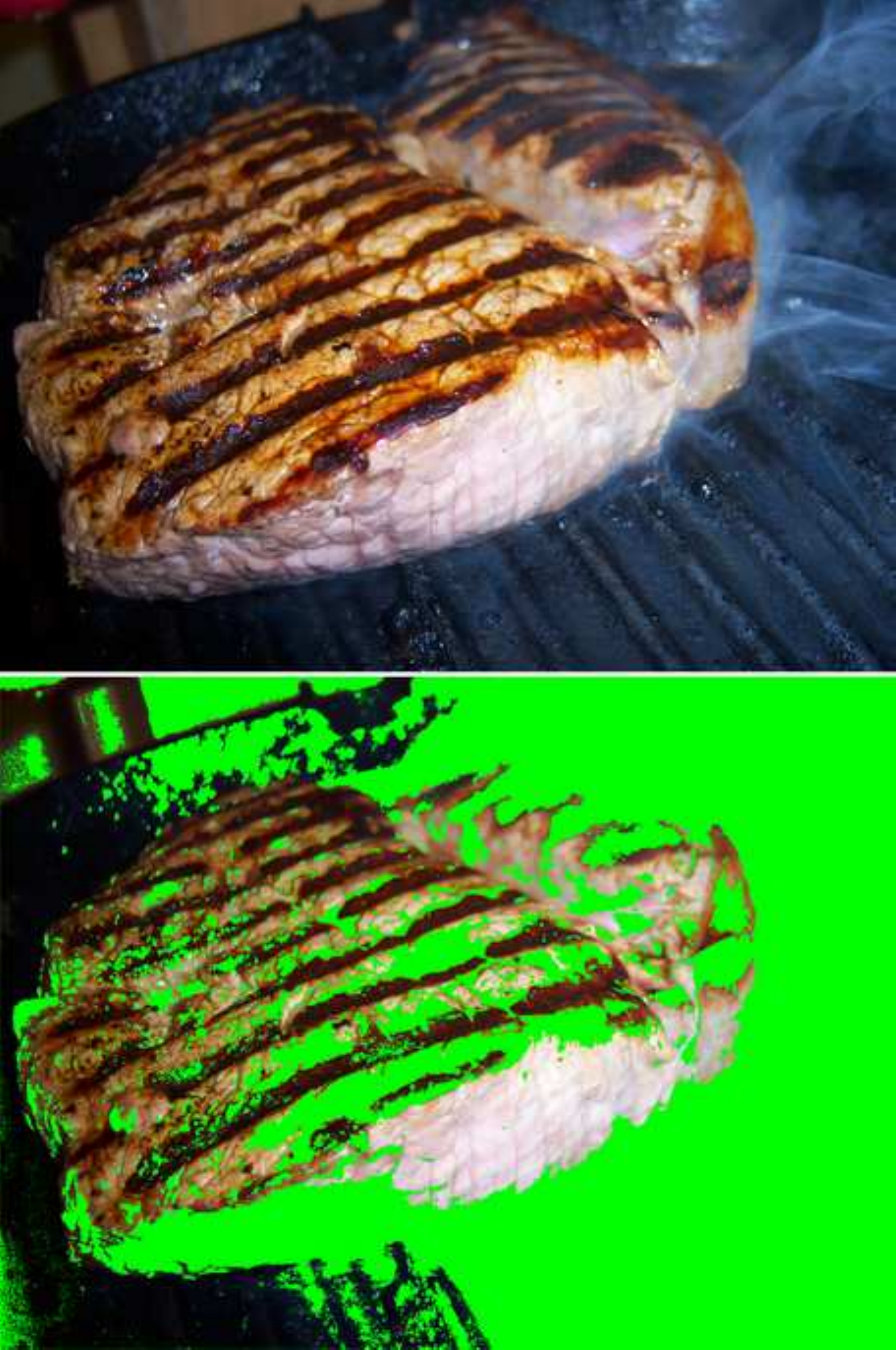,width=0.7in}
\psfig{figure=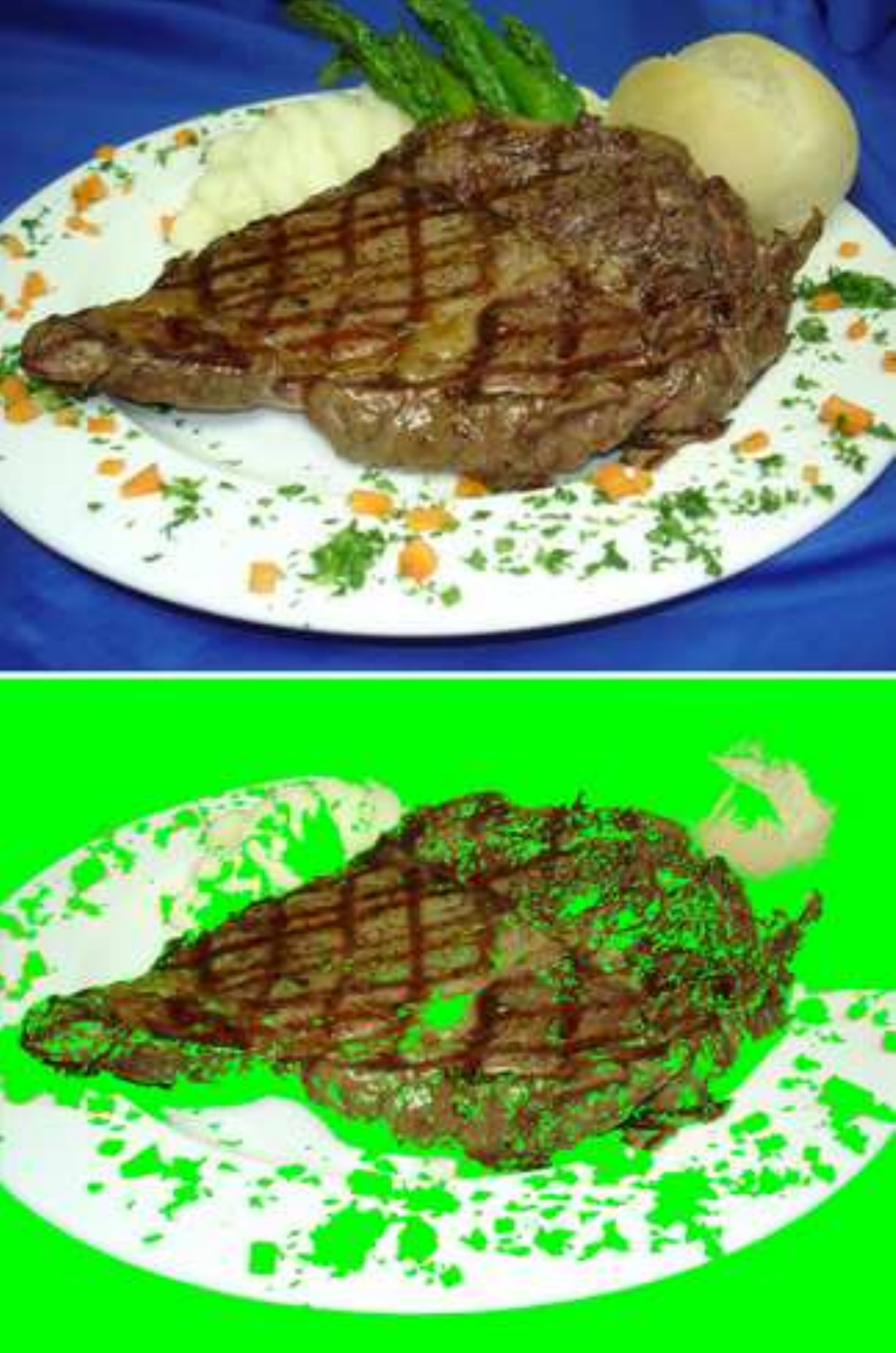,width=0.7in}
\psfig{figure=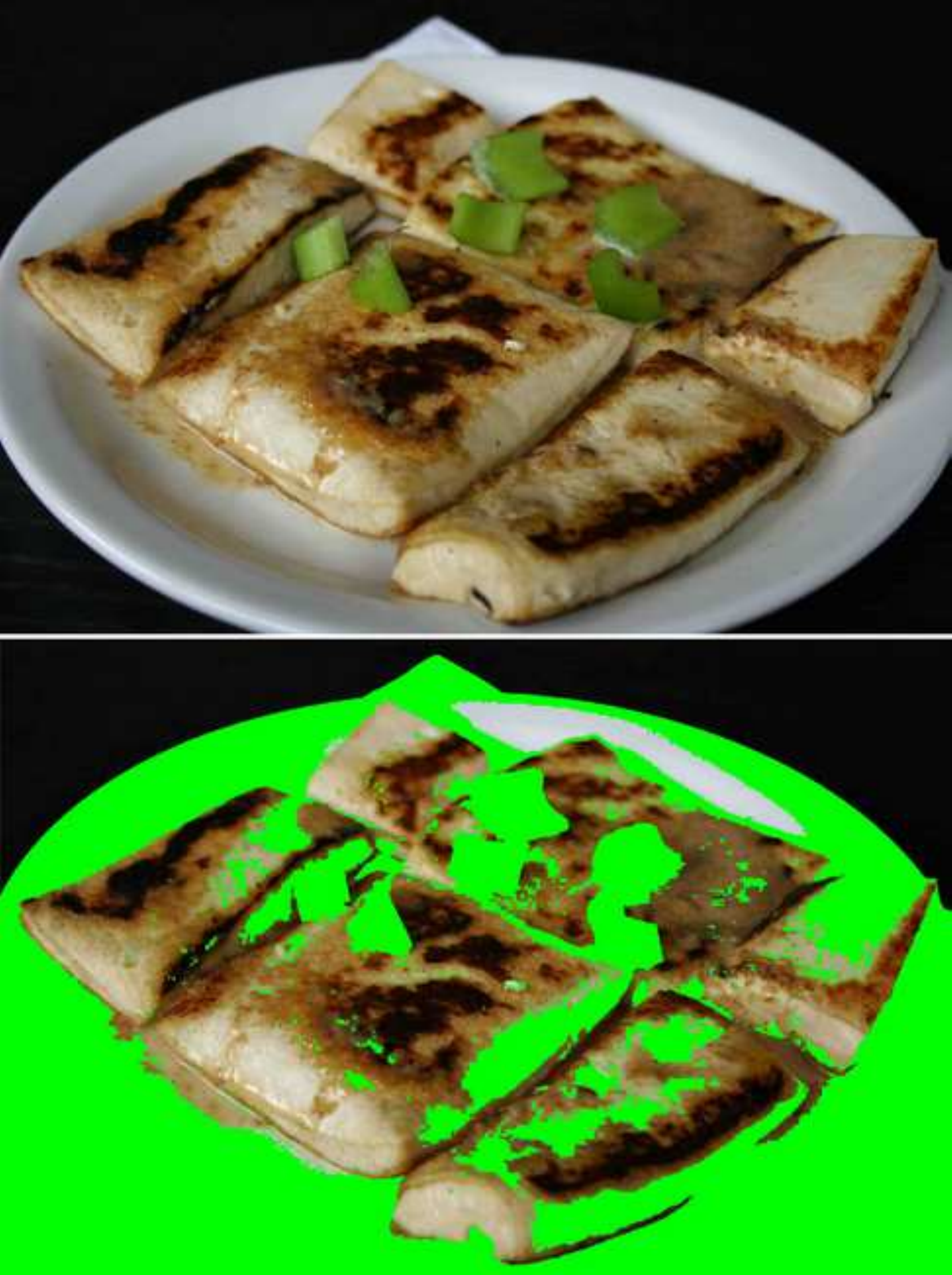,width=0.7in}
\psfig{figure=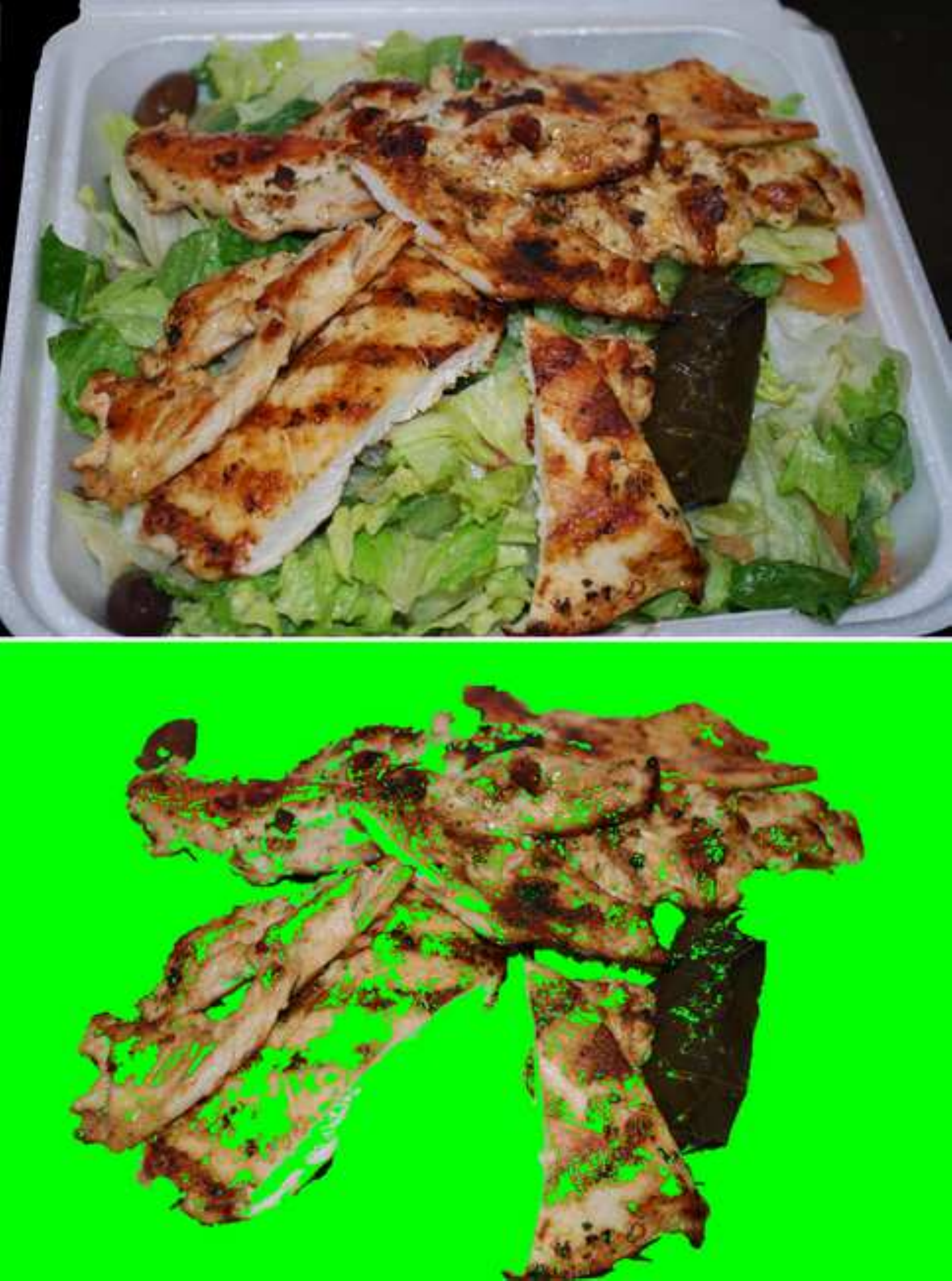,width=0.7in}
\psfig{figure=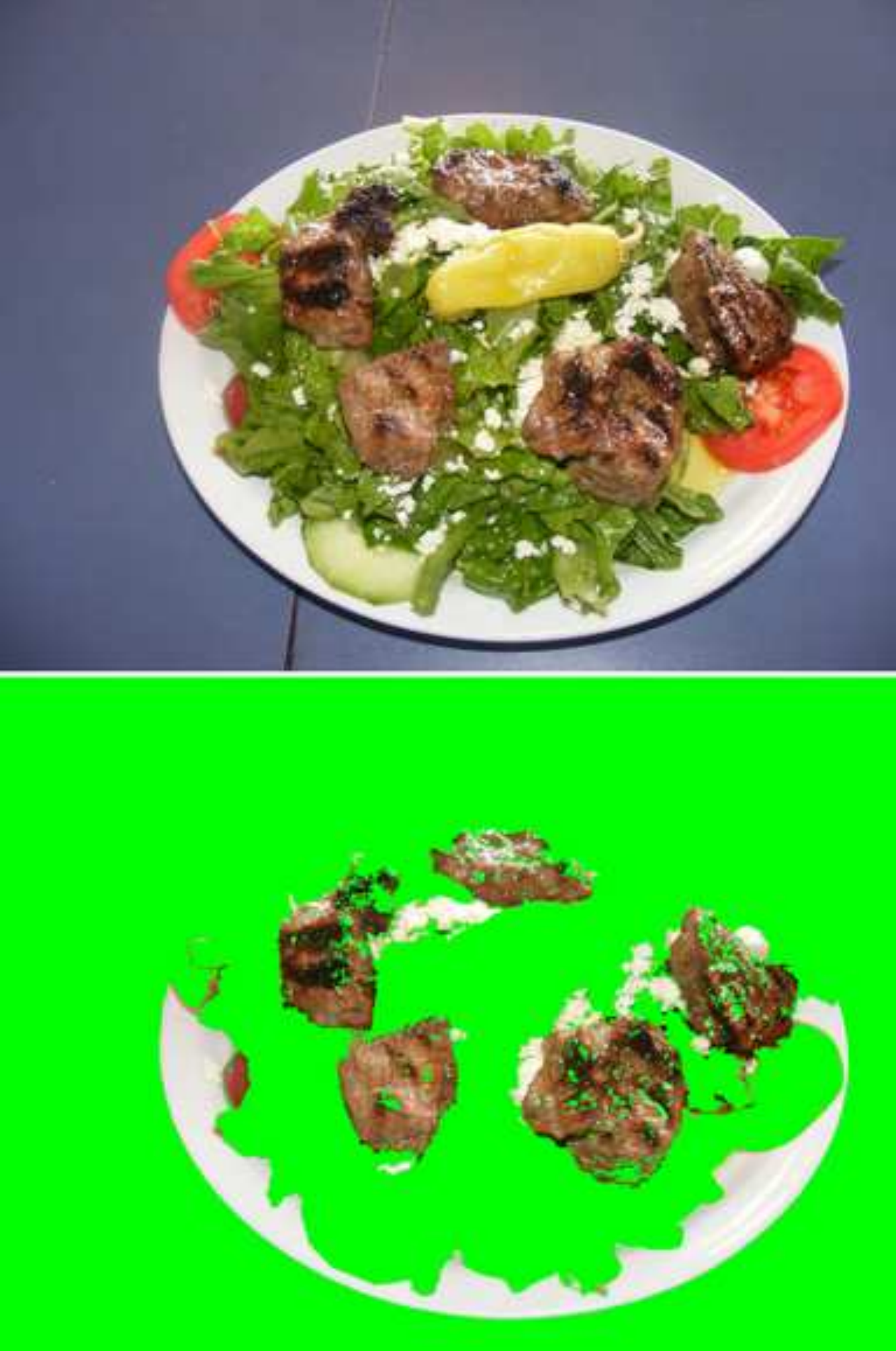,width=0.7in}
}
\centerline{
\psfig{figure=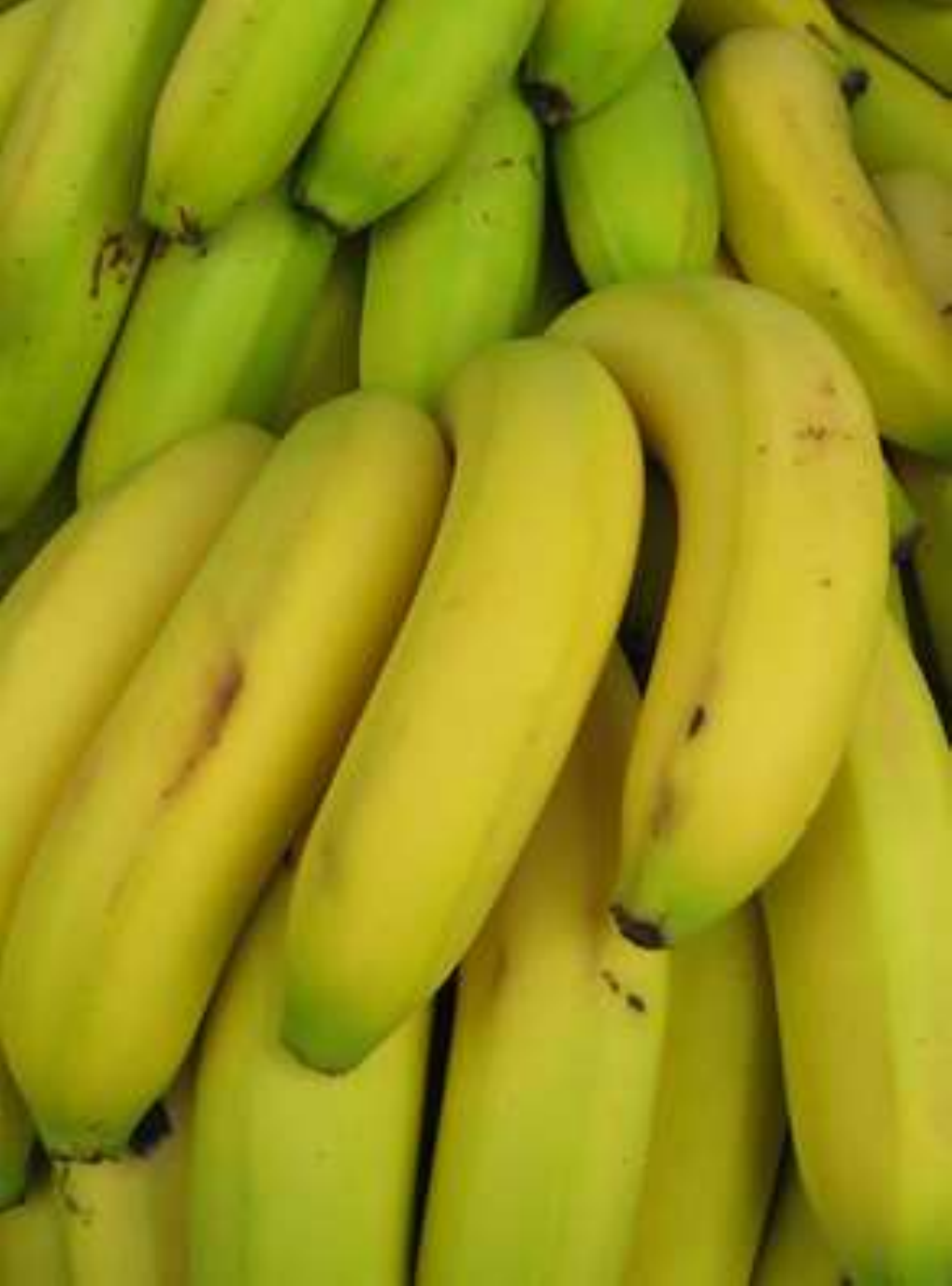,width=0.6in}
\hspace{0.4in}
\psfig{figure=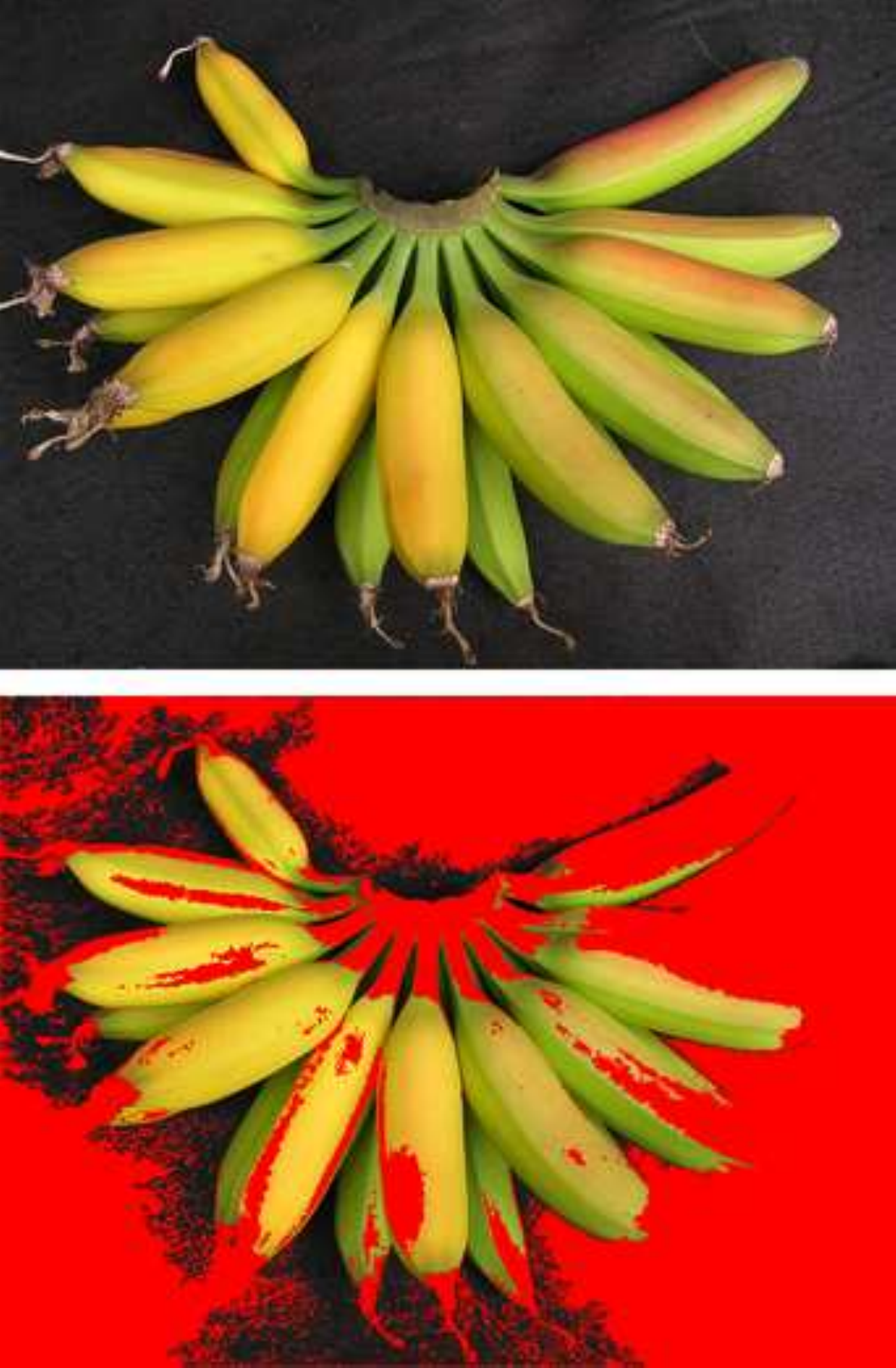,width=0.7in}
\psfig{figure=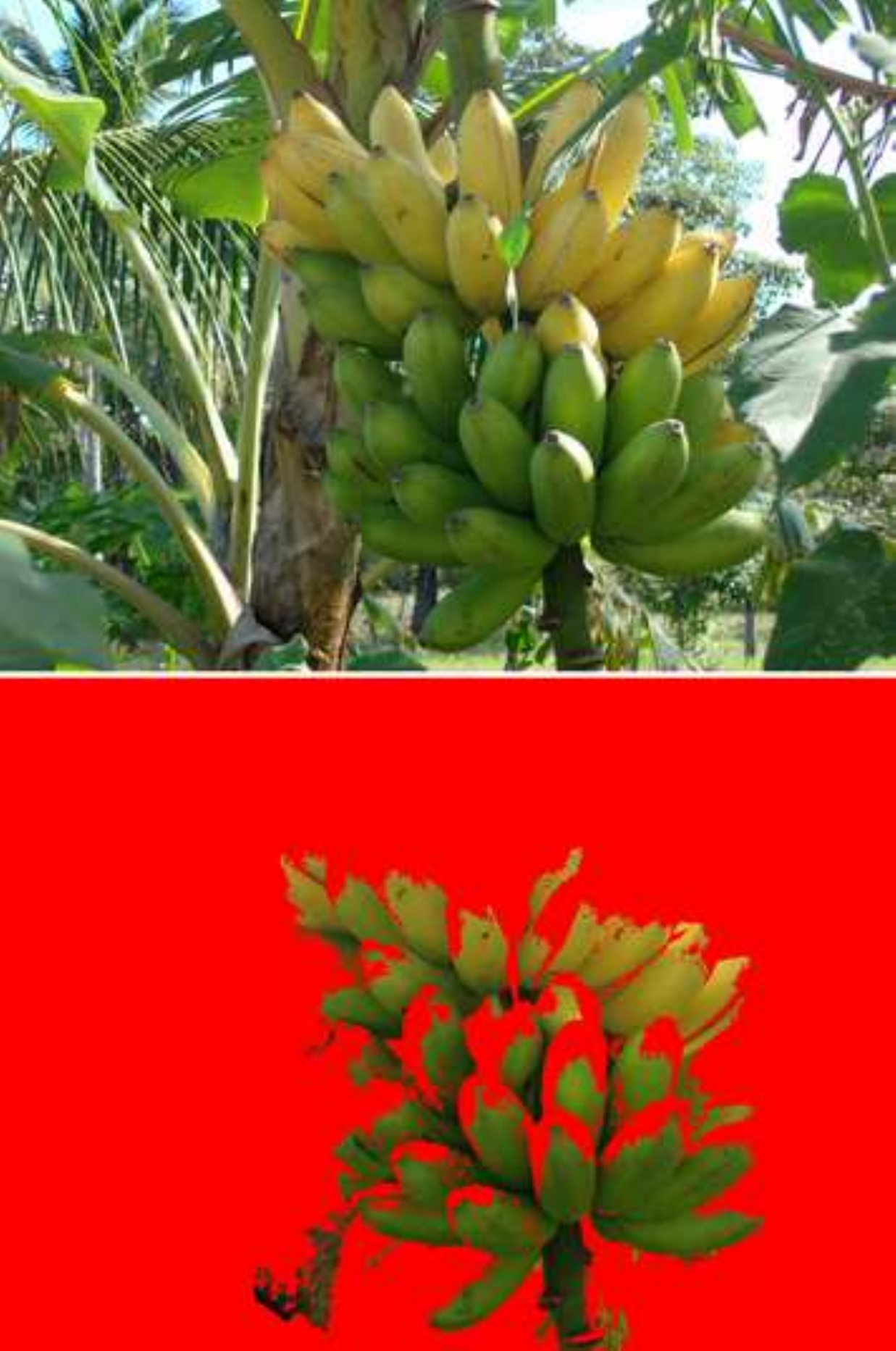,width=0.7in}
\psfig{figure=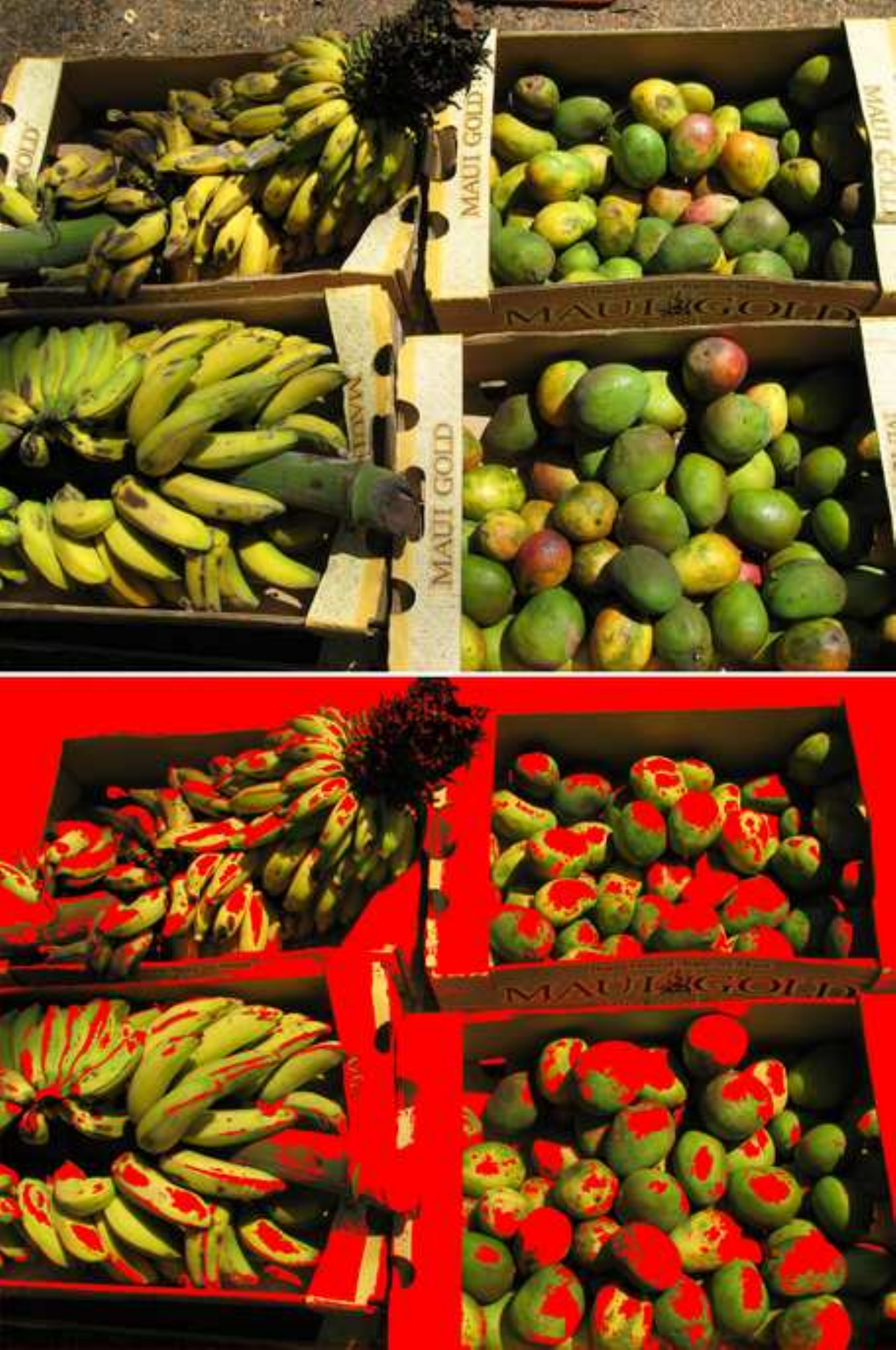,width=0.7in}
\psfig{figure=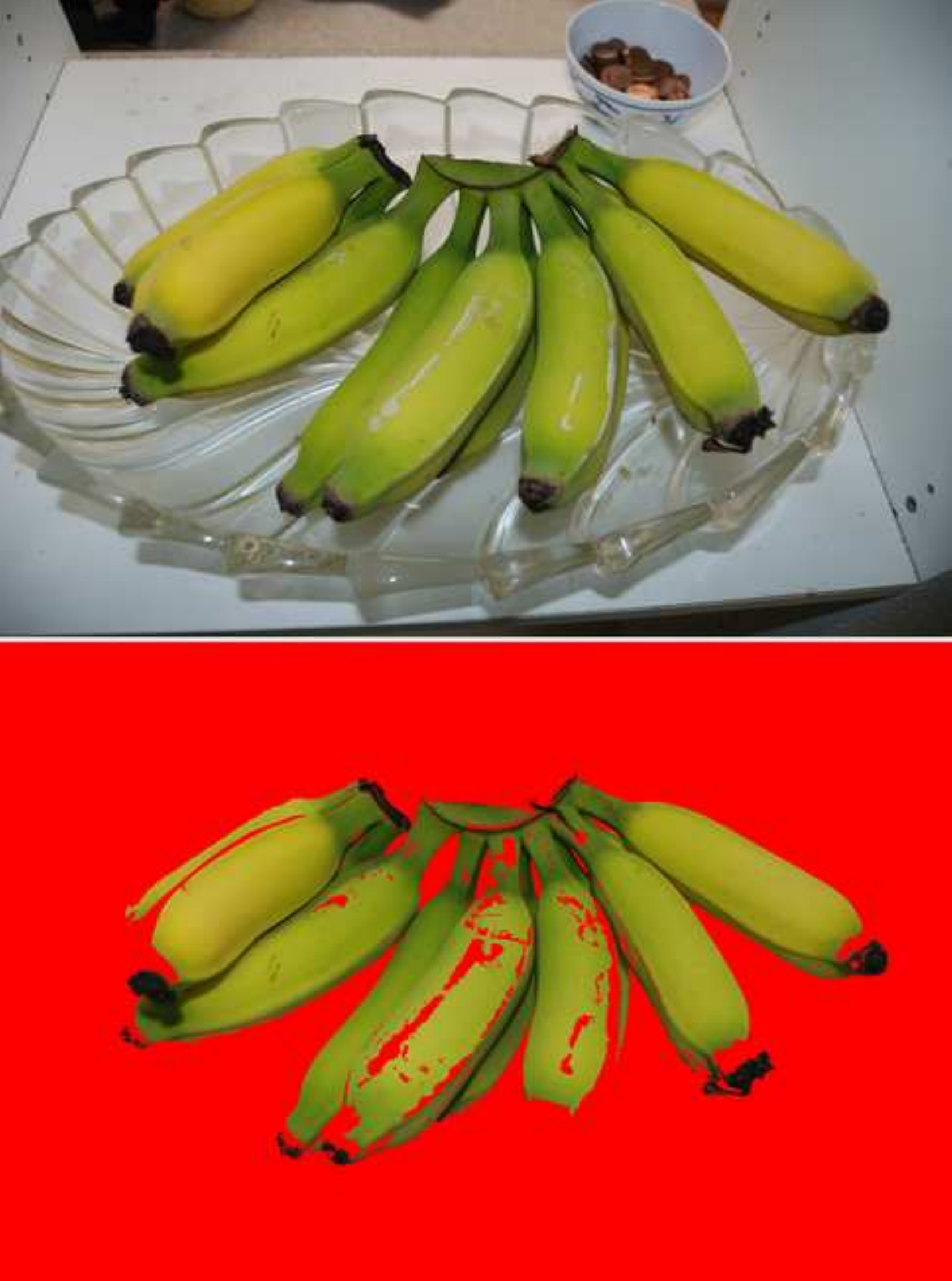,width=0.7in}
\psfig{figure=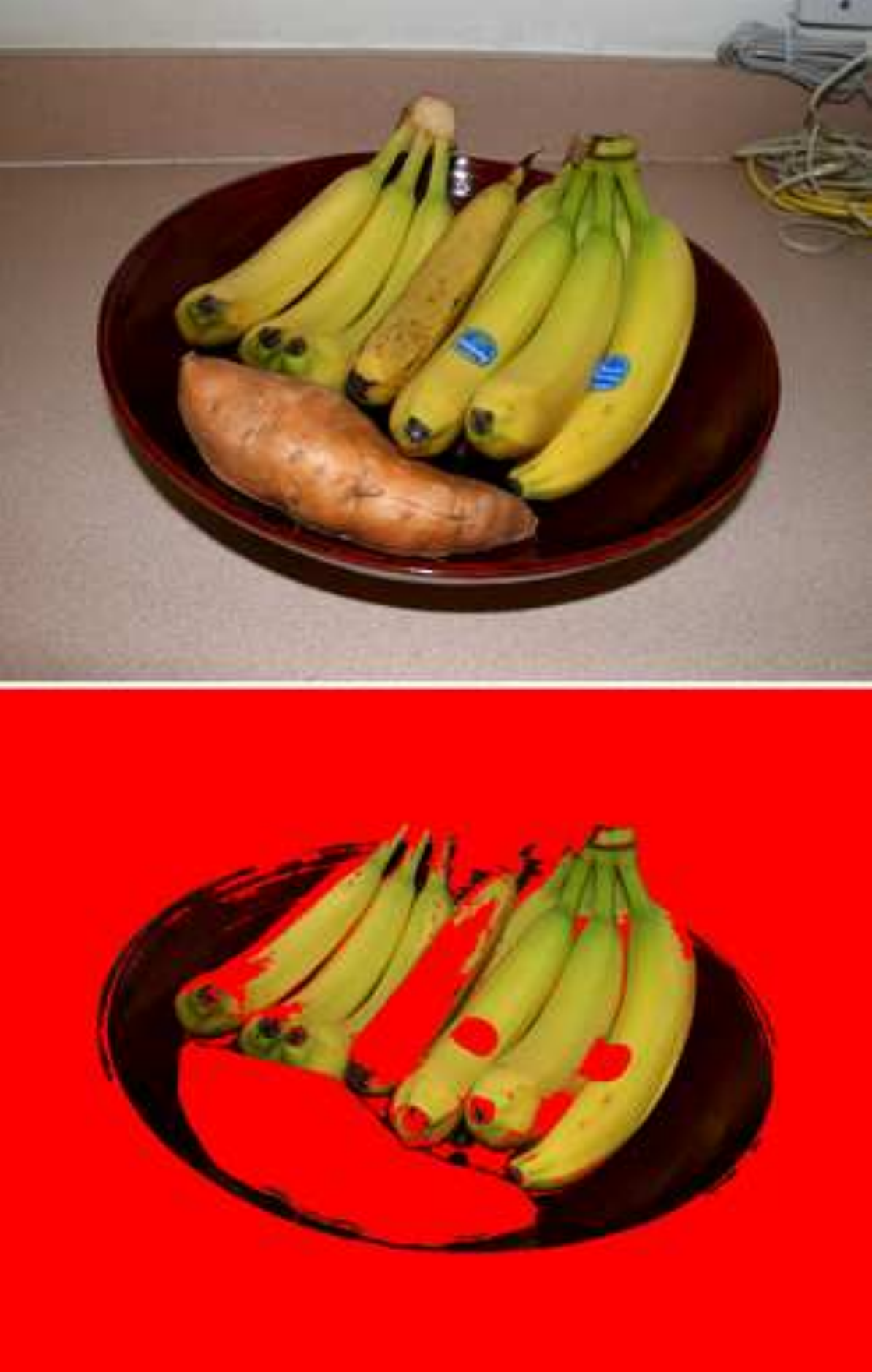,width=0.7in}
\psfig{figure=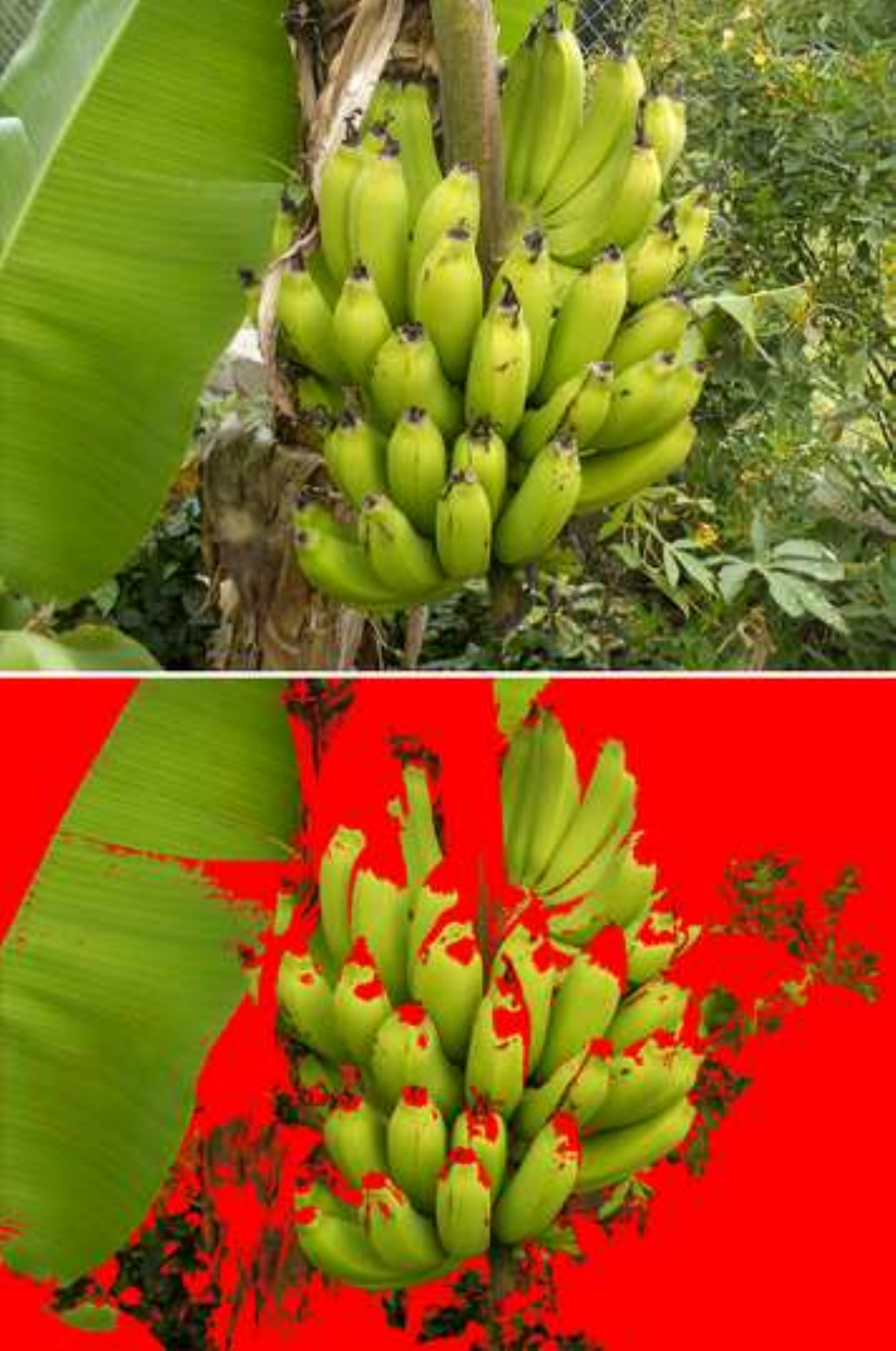,width=0.7in}
\psfig{figure=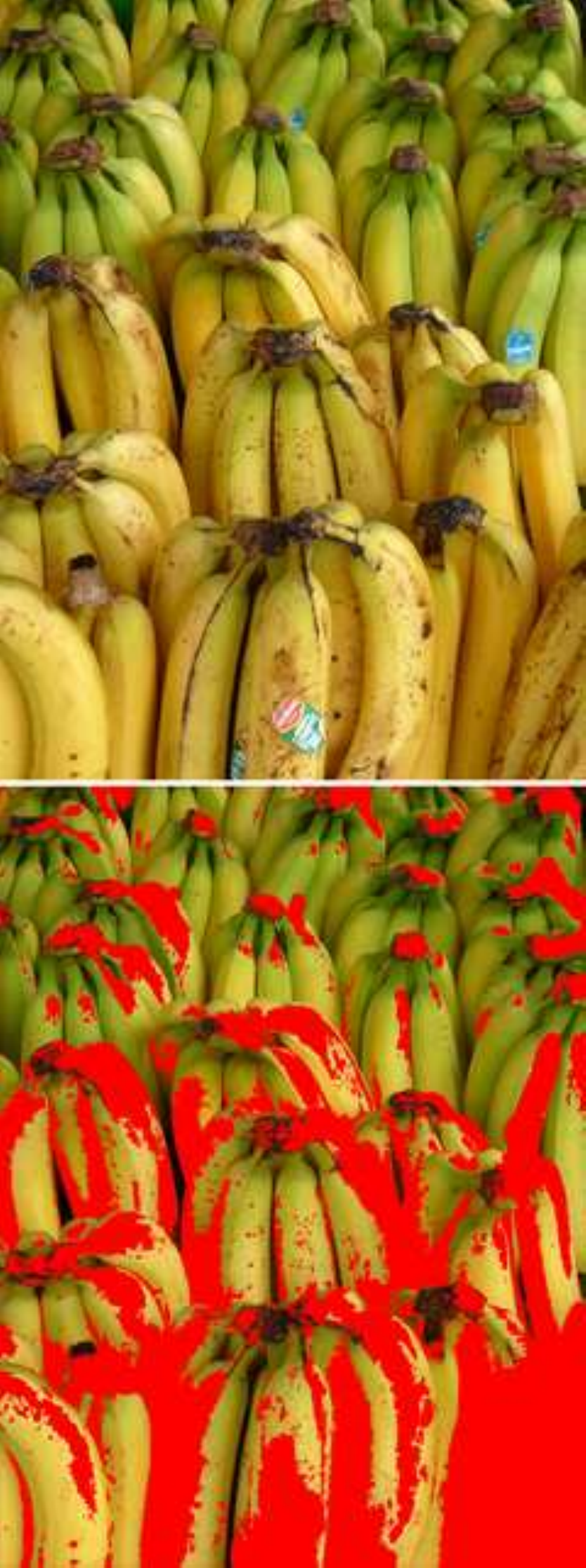,width=0.4in}
\psfig{figure=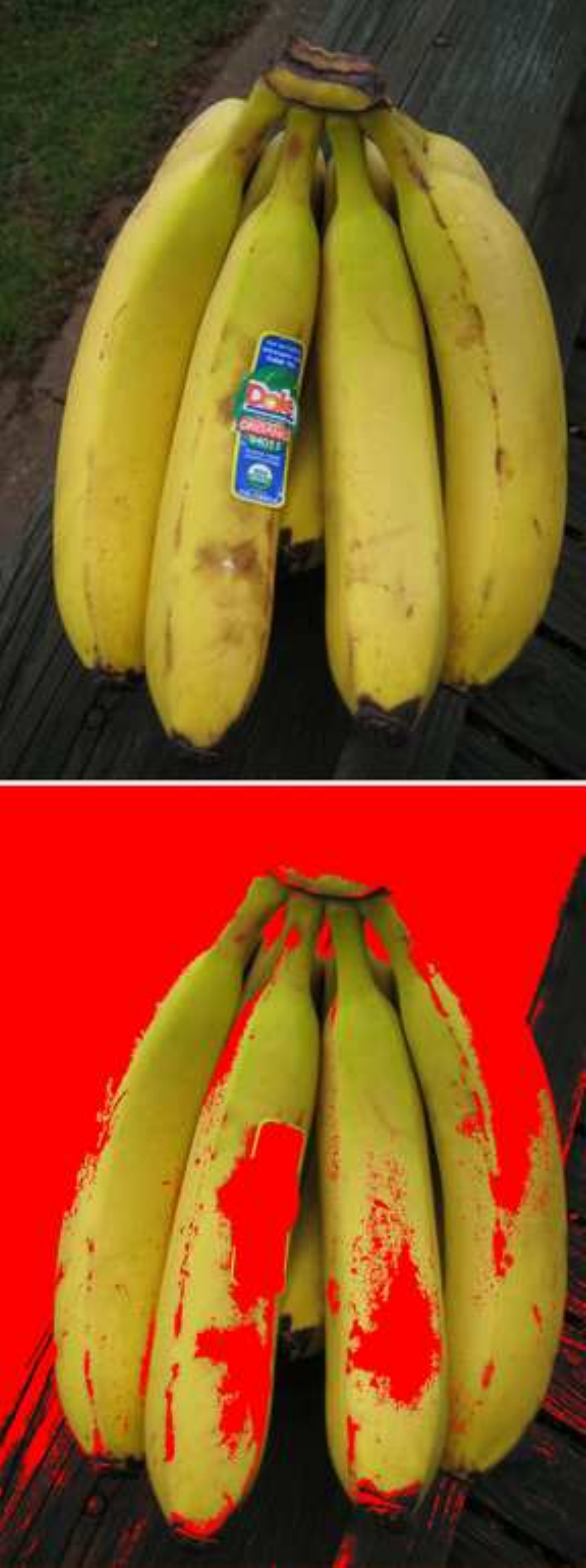,width=0.4in}
}
\centerline{
\psfig{figure=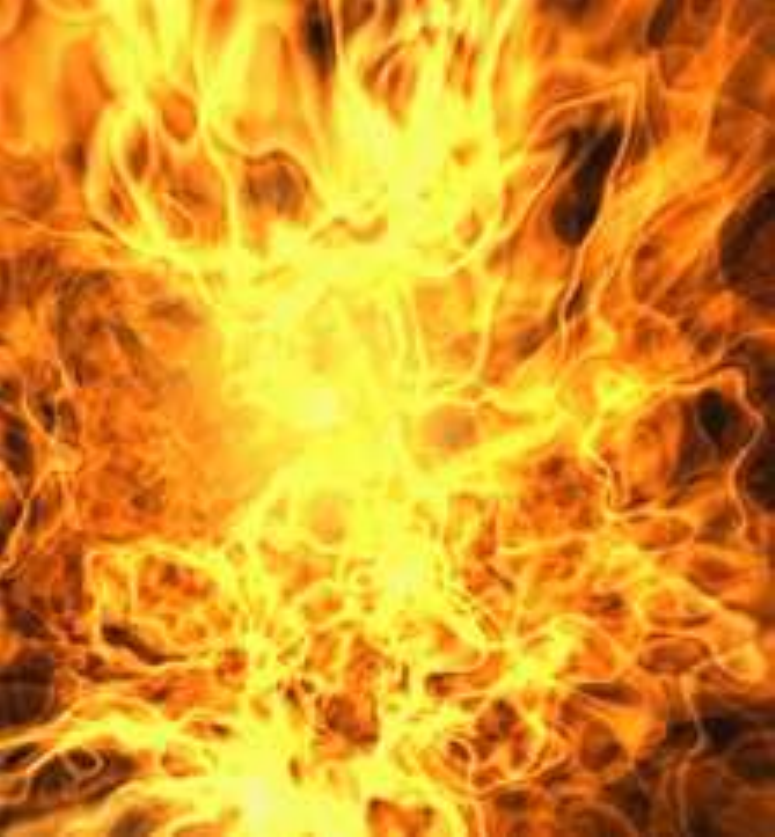,width=0.6in}
\hspace{0.4in}
\psfig{figure=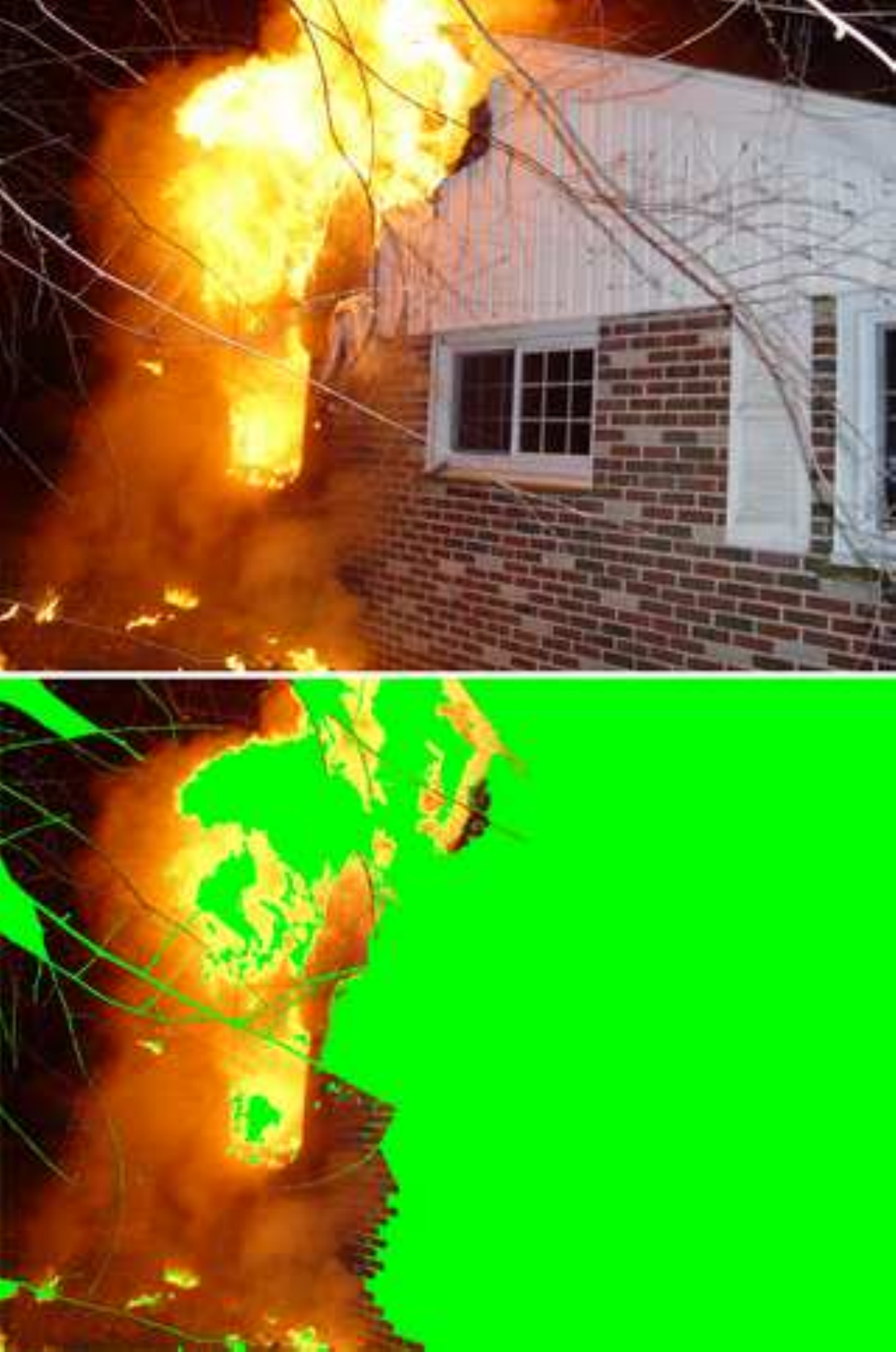,width=0.7in}
\psfig{figure=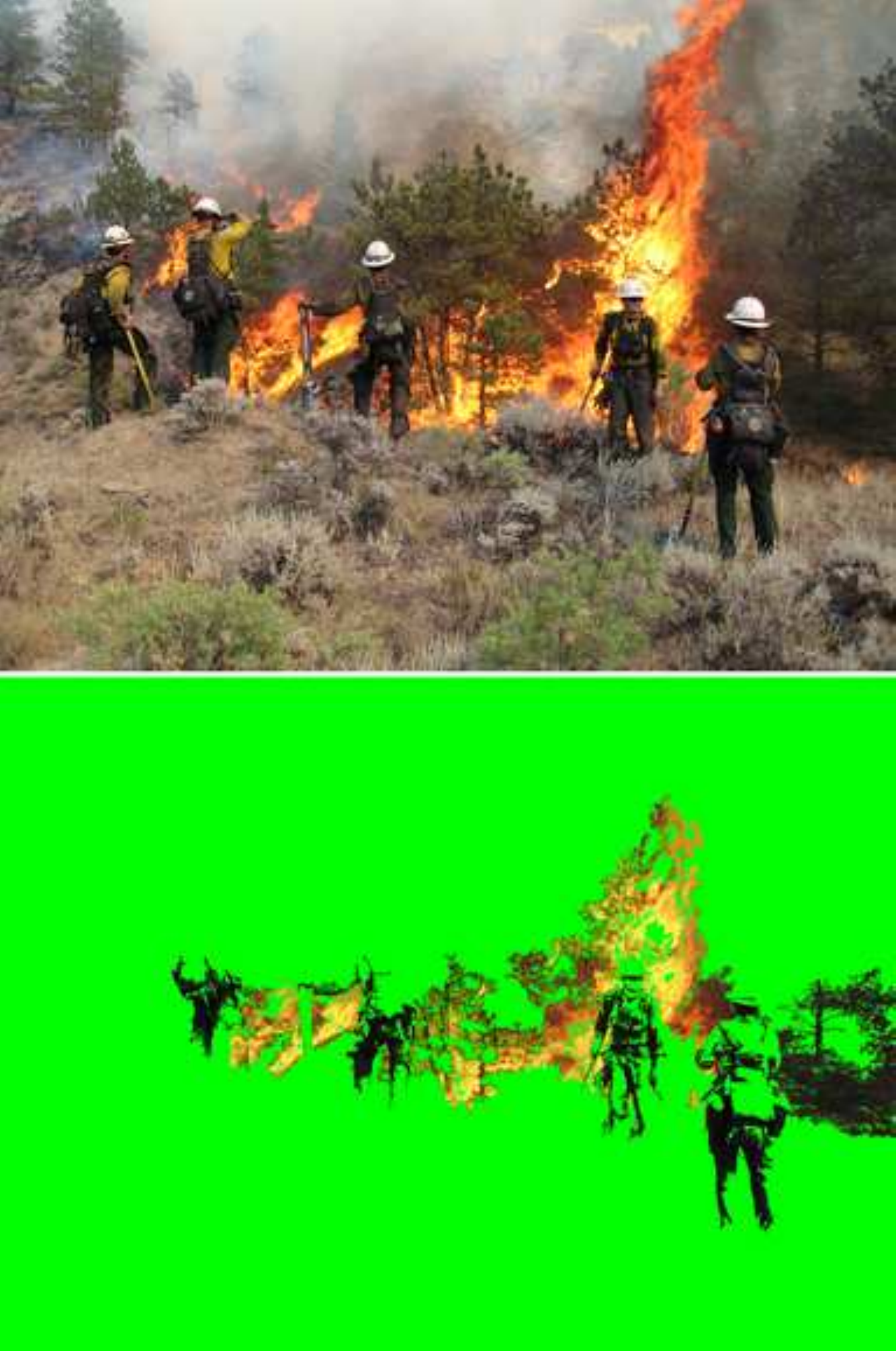,width=0.7in}
\psfig{figure=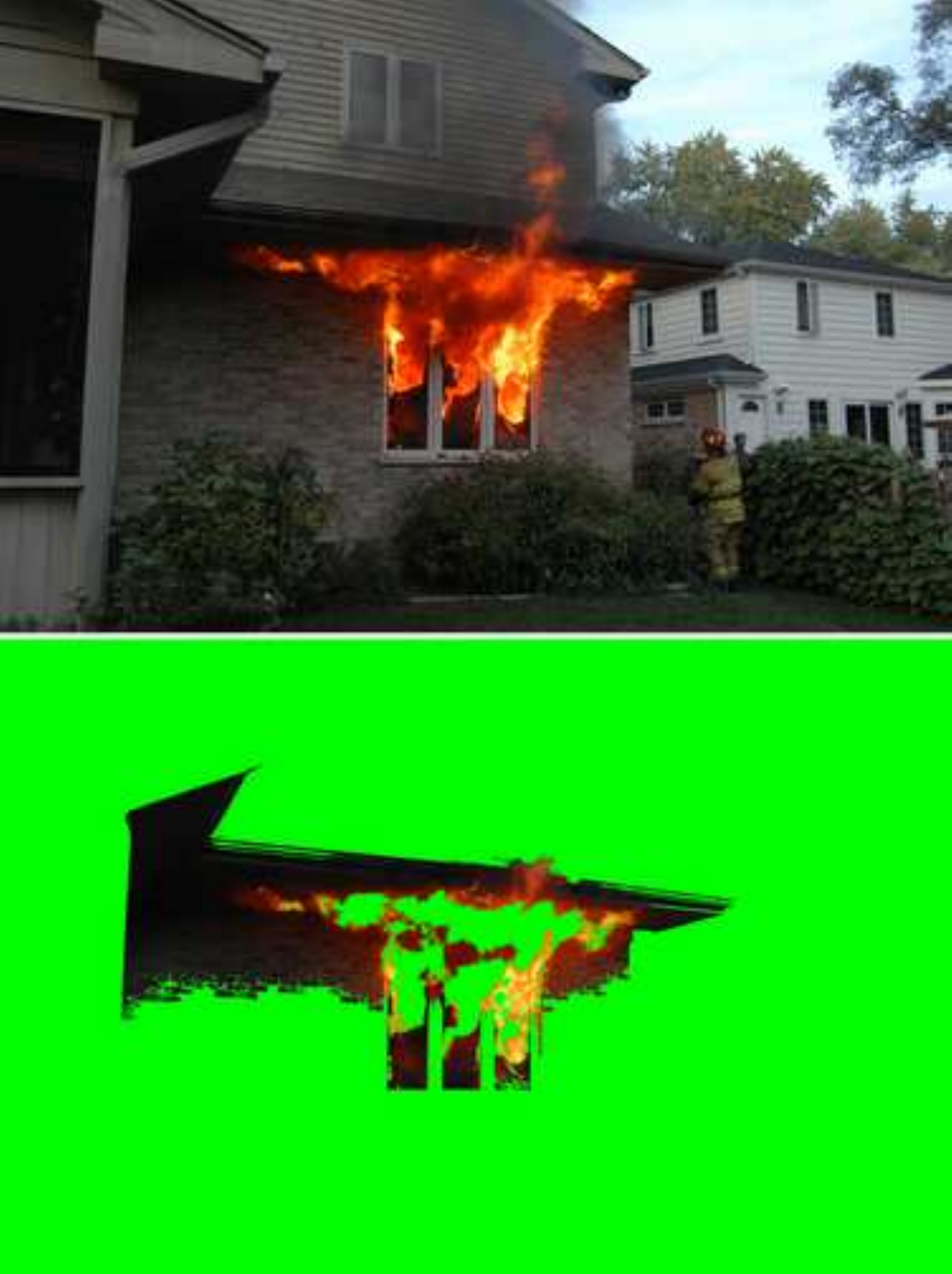,width=0.7in}
\psfig{figure=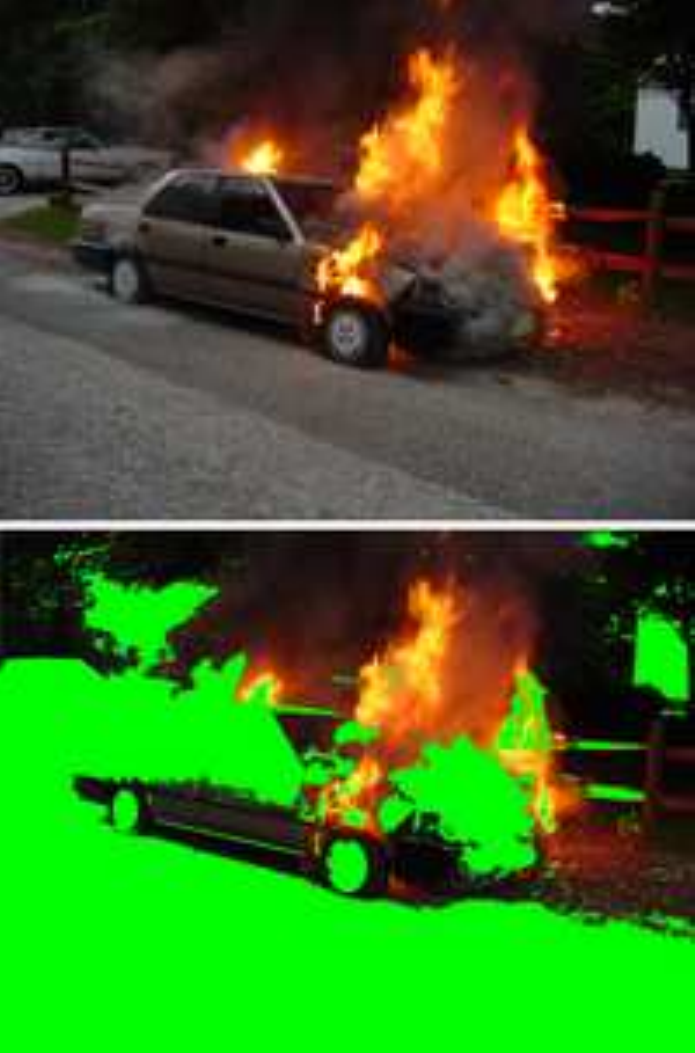,width=0.7in}
\psfig{figure=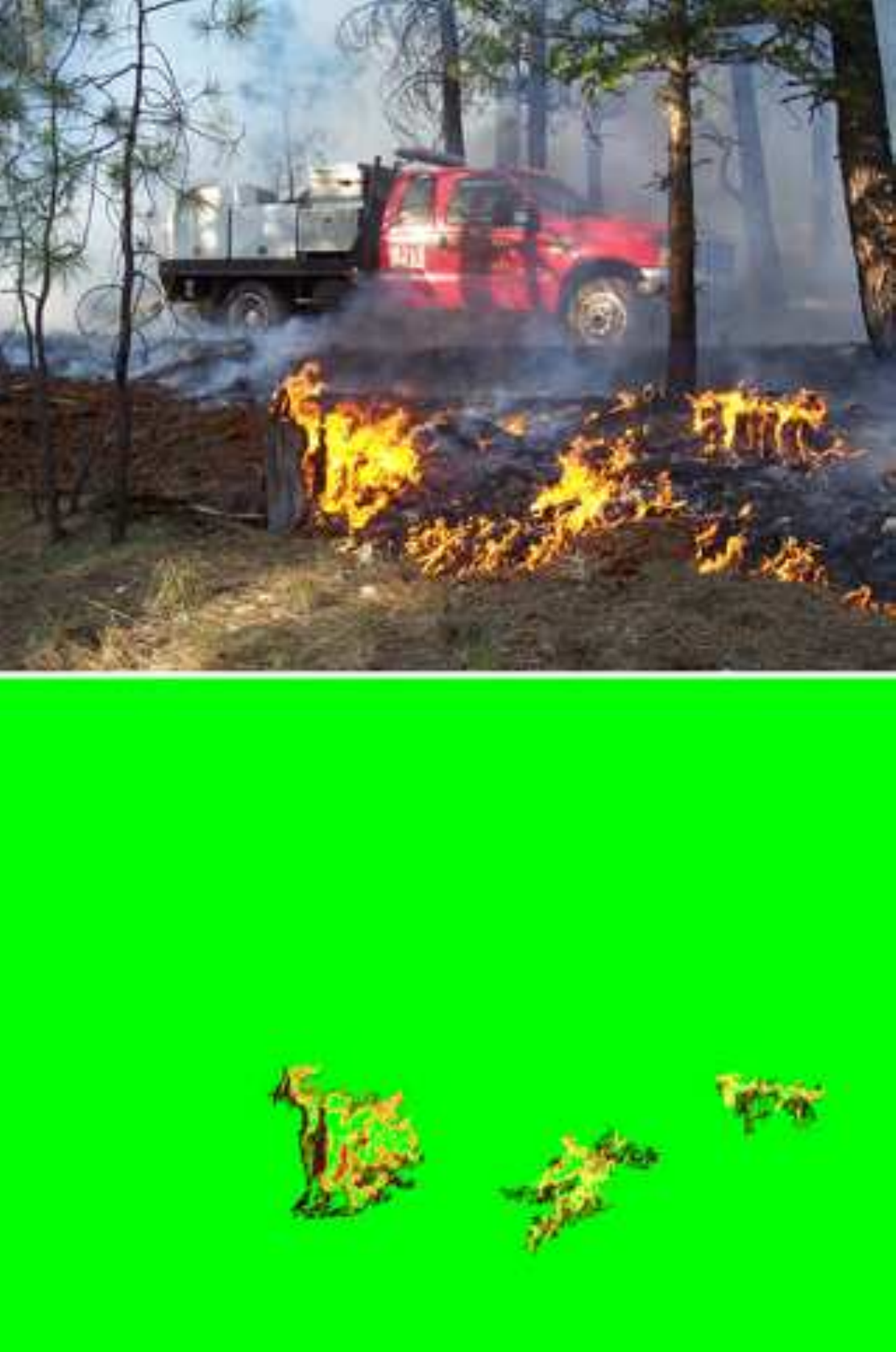,width=0.7in}
\psfig{figure=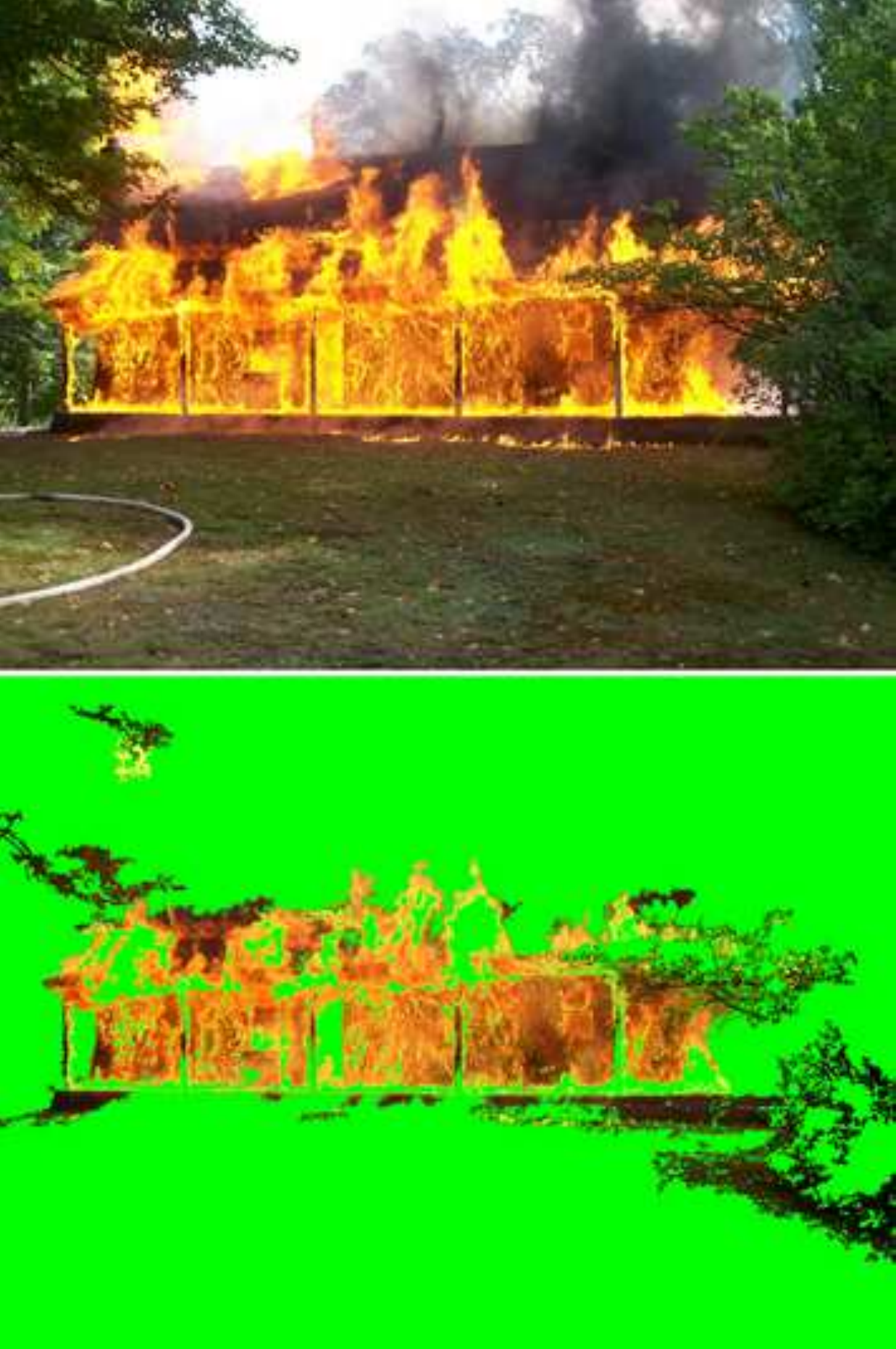,width=0.7in}
\psfig{figure=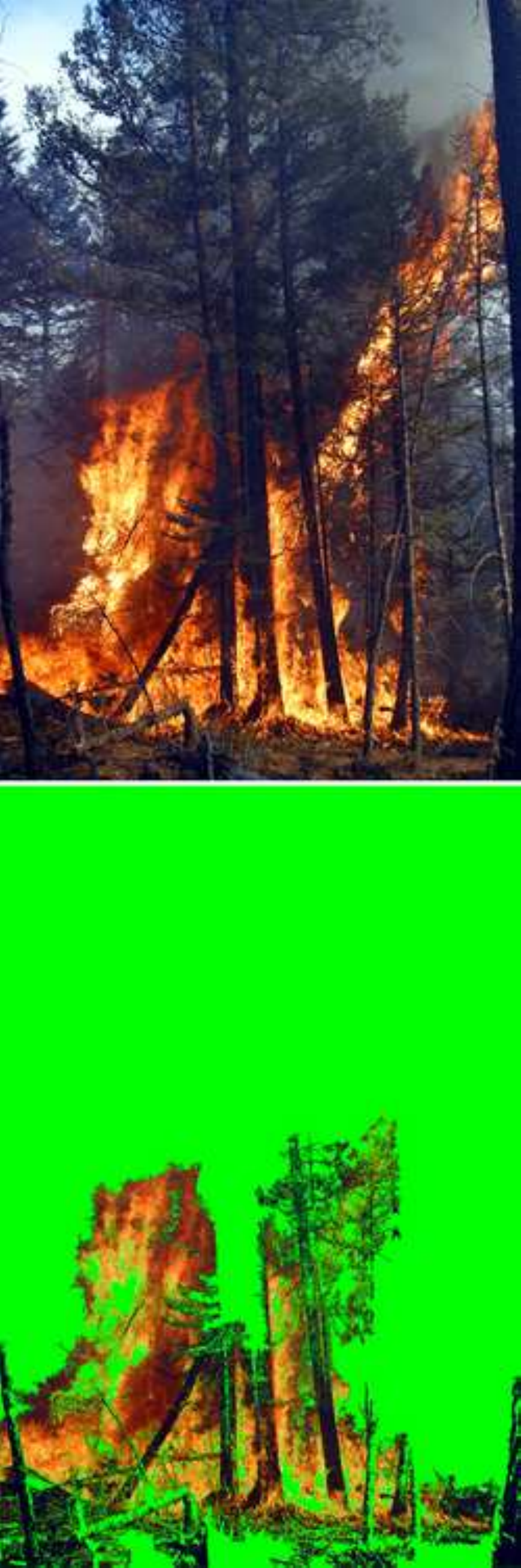,width=0.35in}
\psfig{figure=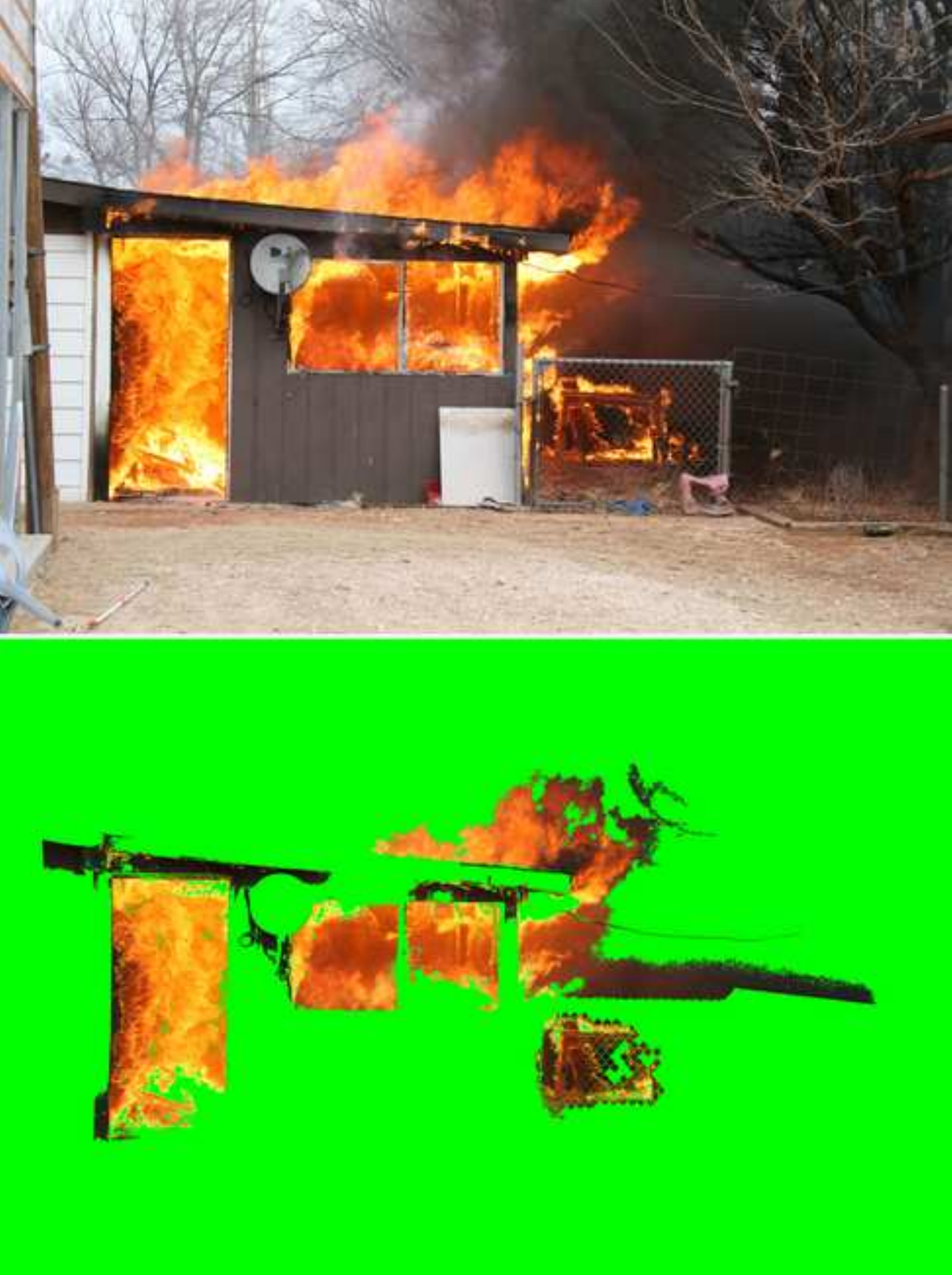,width=0.75in}
}
\caption{\protect\scriptsize
Detections using one exemplar (shown at the left of each row).  Plant colors, baked, fried and grilled food, yellowing bananas and fire.
\label{detection}}
\centerline{
\psfig{figure=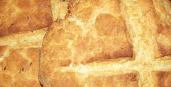,width=0.5in}
\psfig{figure=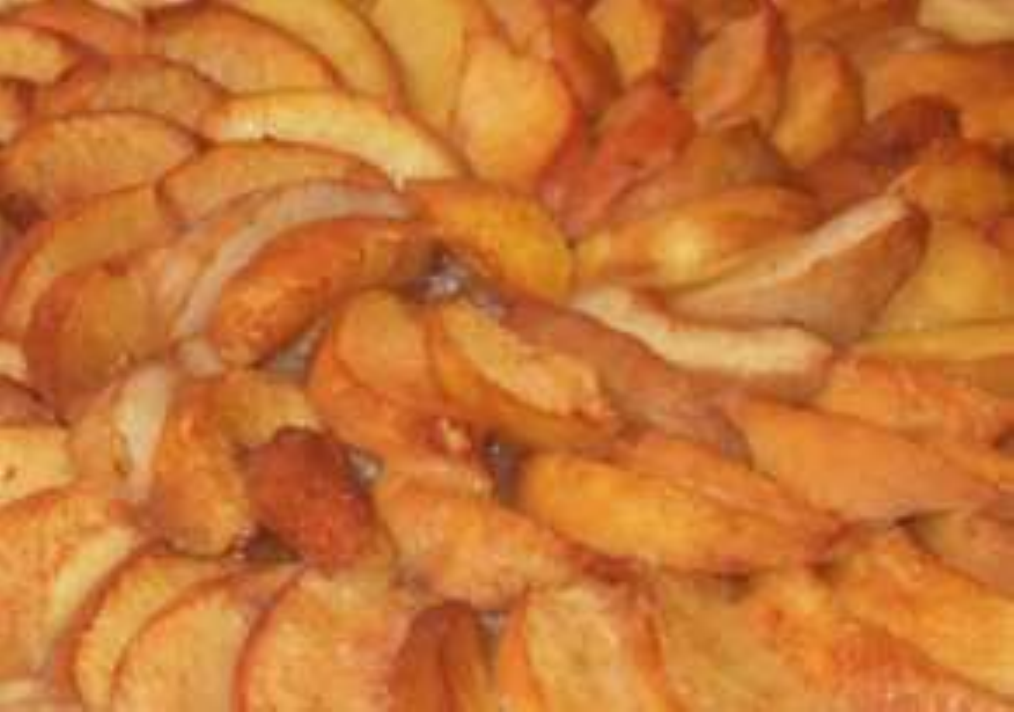,width=0.5in}
\psfig{figure=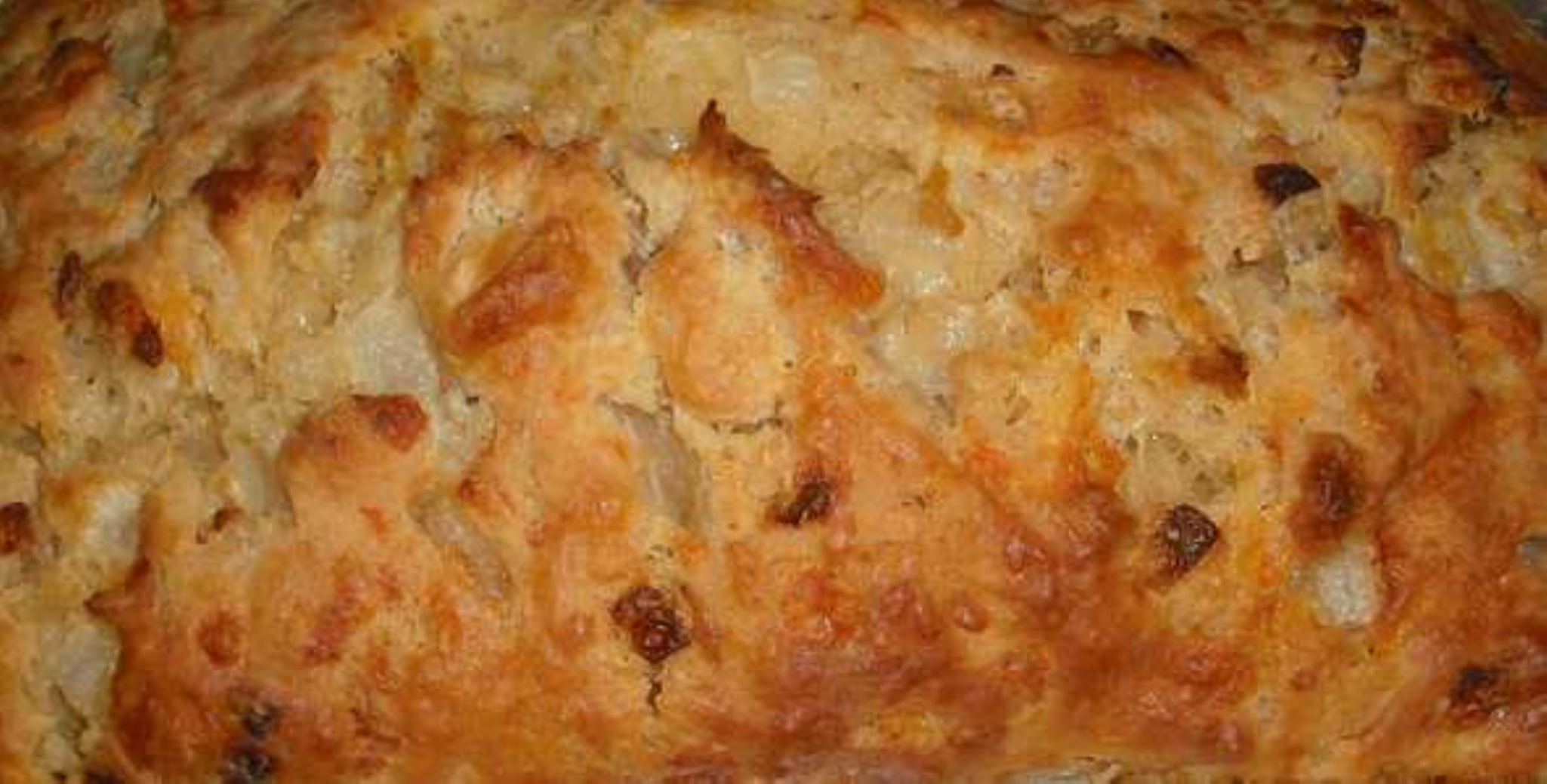,width=0.5in}
\psfig{figure=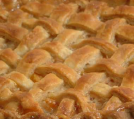,width=0.5in}
\psfig{figure=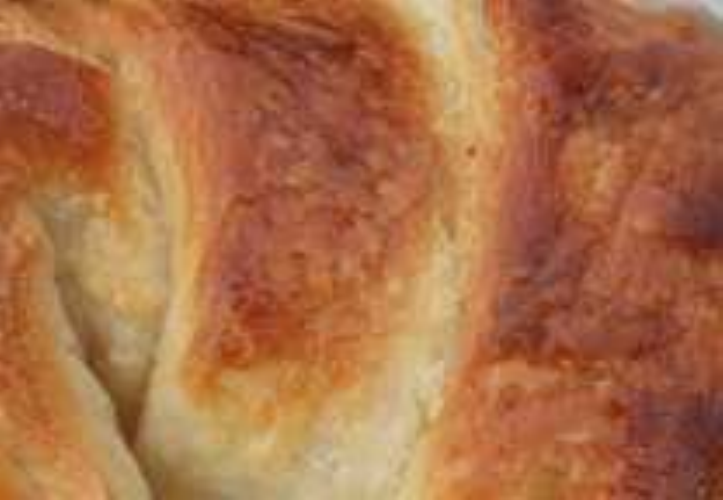,width=0.5in}
\psfig{figure=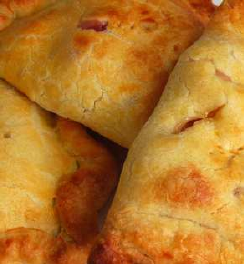,width=0.4in}
\psfig{figure=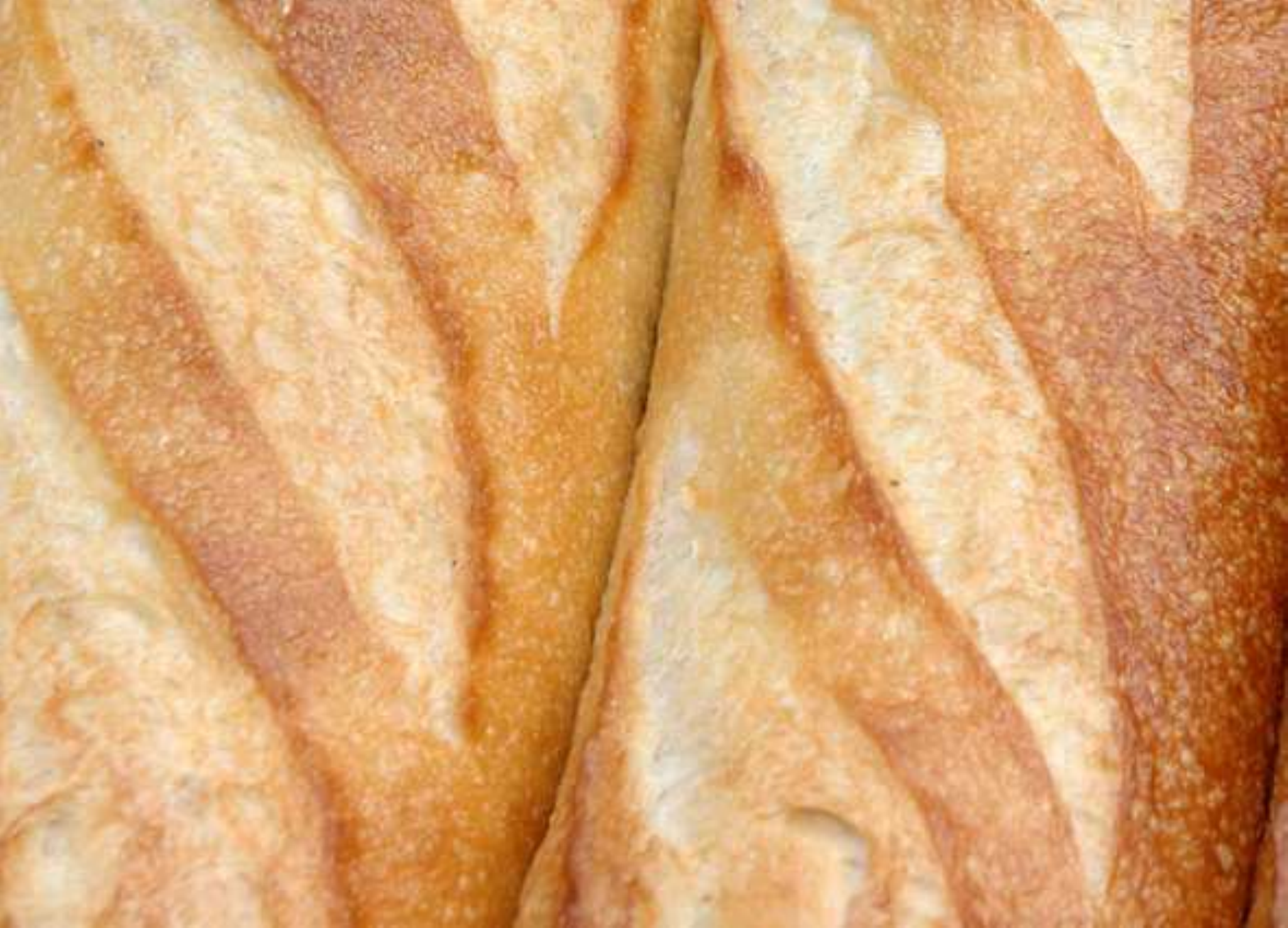,width=0.5in}
\psfig{figure=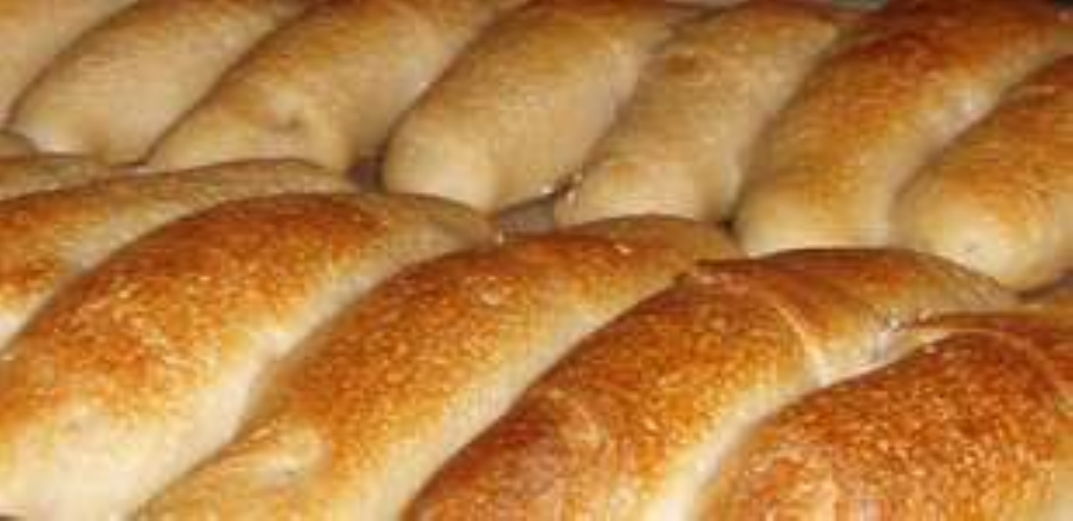,width=0.5in}
\psfig{figure=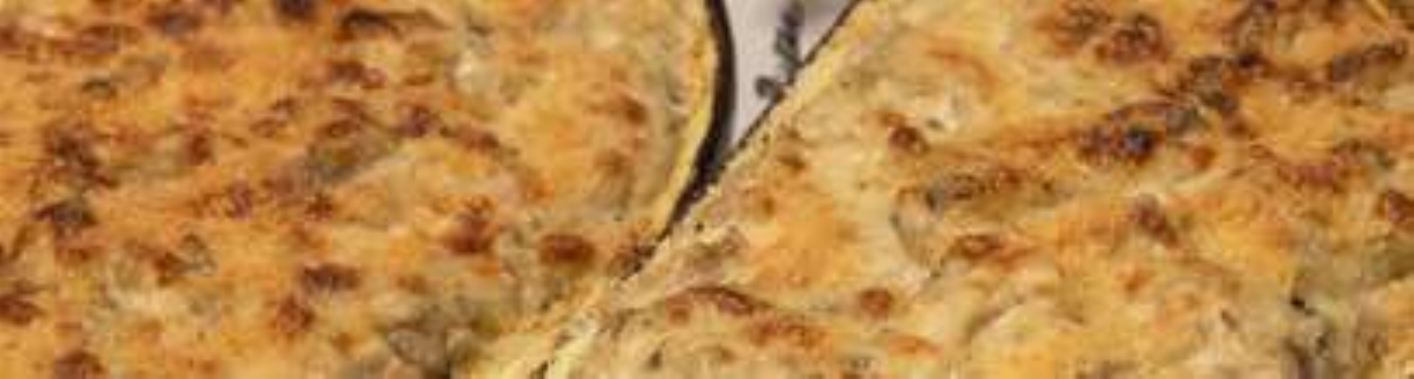,width=0.5in}
\psfig{figure=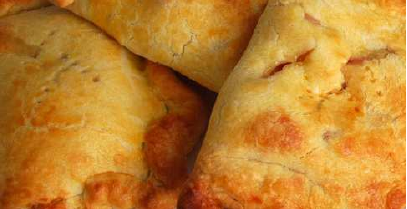,width=0.5in}
\psfig{figure=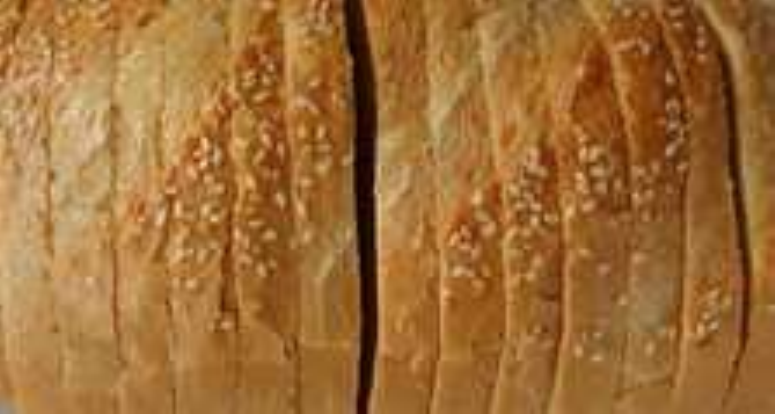,width=0.5in}
\psfig{figure=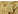,width=0.5in}
}
\centerline{\protect\tiny 148.7 \hspace{0.35in} 139.4\hspace{0.35in}   90.6 \hspace{0.35in} 104.3 \hspace{0.35in} 109.9\hspace{0.35in} 81.34\hspace{0.35in} 107.0 \hspace{0.35in} 108.1 \hspace{0.35in} 112.4 \hspace{0.35in} 107.0 \hspace{0.35in} 87.9 \hspace{0.35in}7.5}
\centerline{
\psfig{figure=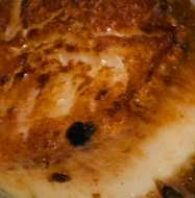,width=0.5in}
\psfig{figure=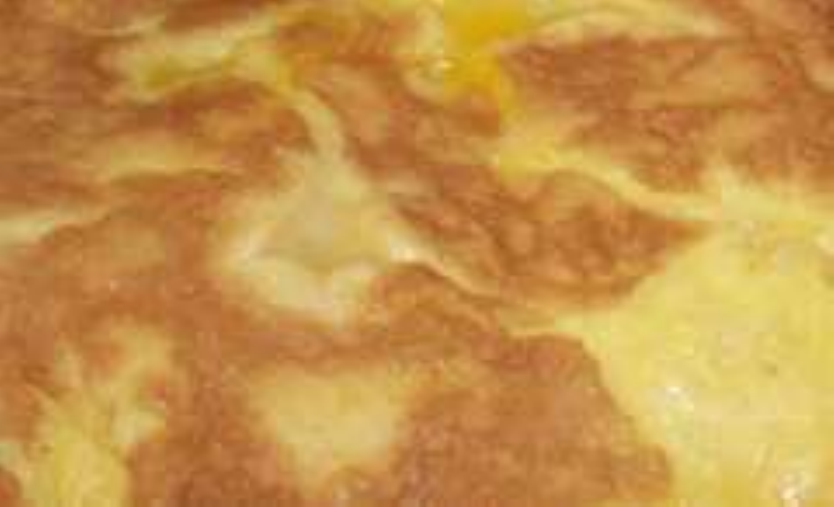,width=0.5in}
\psfig{figure=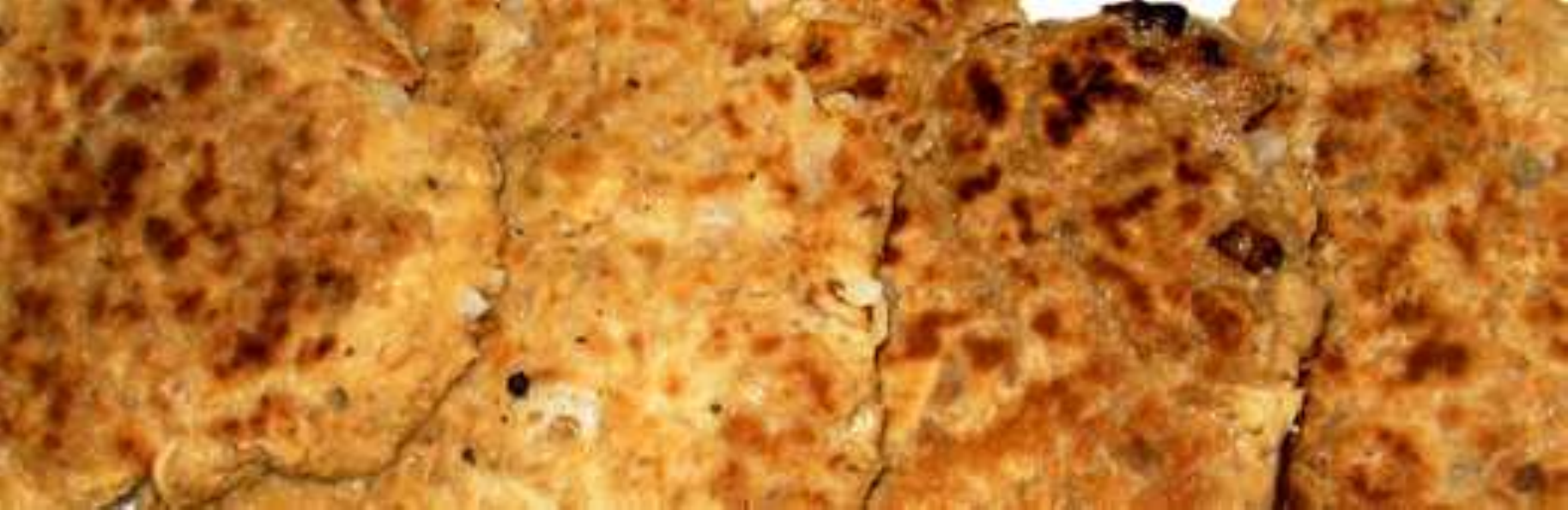,width=0.5in}
\psfig{figure=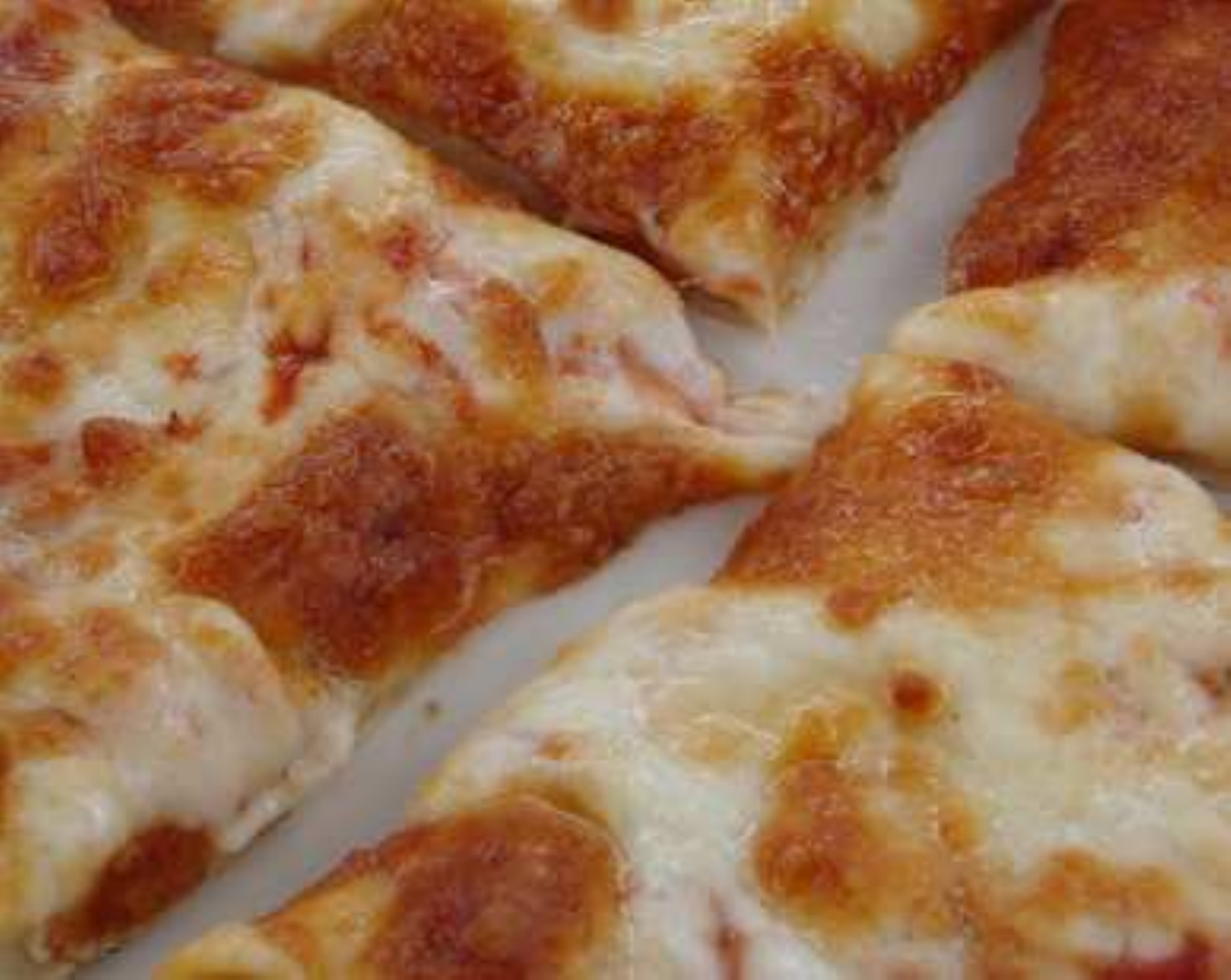,width=0.5in}
\psfig{figure=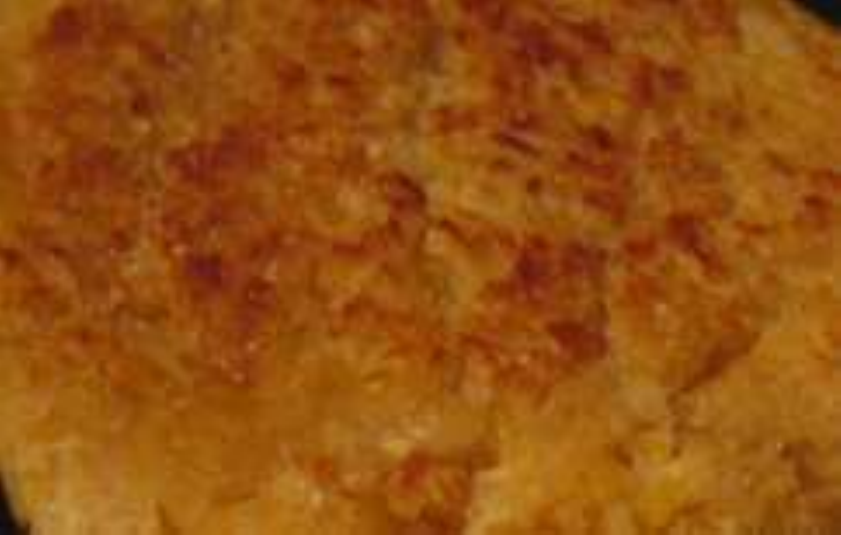,width=0.5in}
\psfig{figure=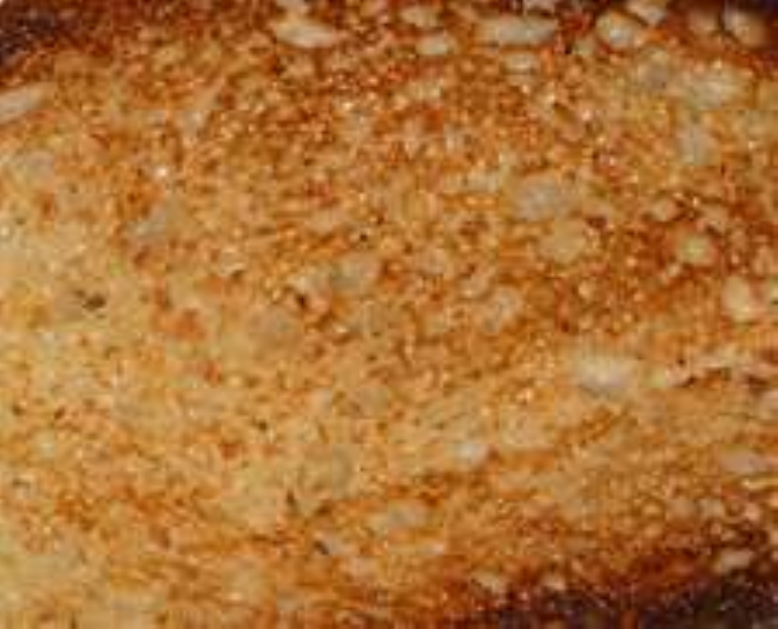,width=0.5in}
\psfig{figure=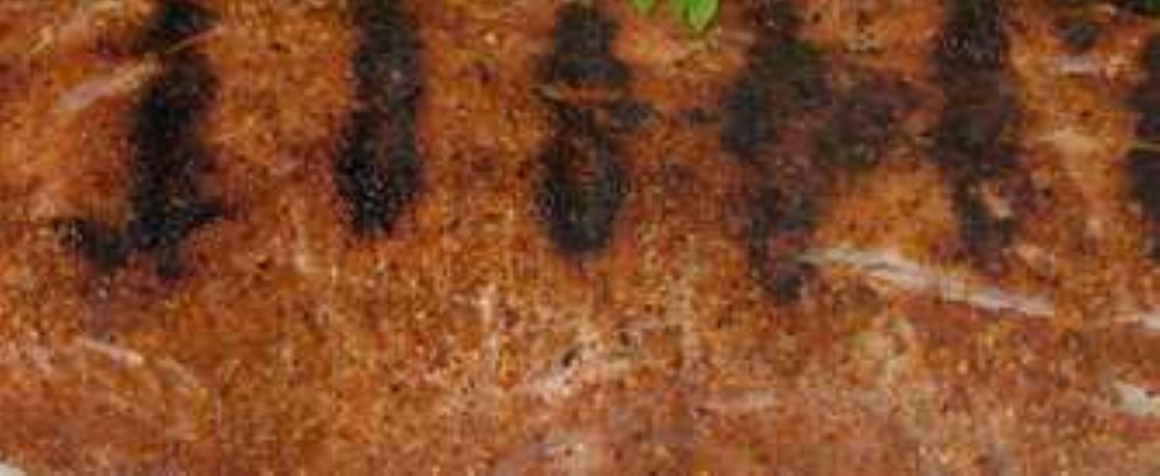,width=0.5in}
\psfig{figure=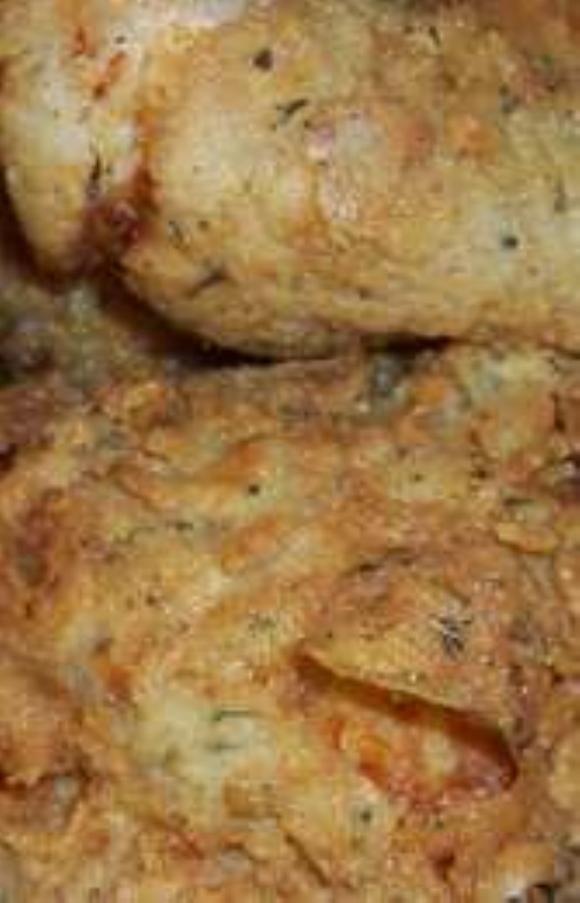,width=0.35in}
\psfig{figure=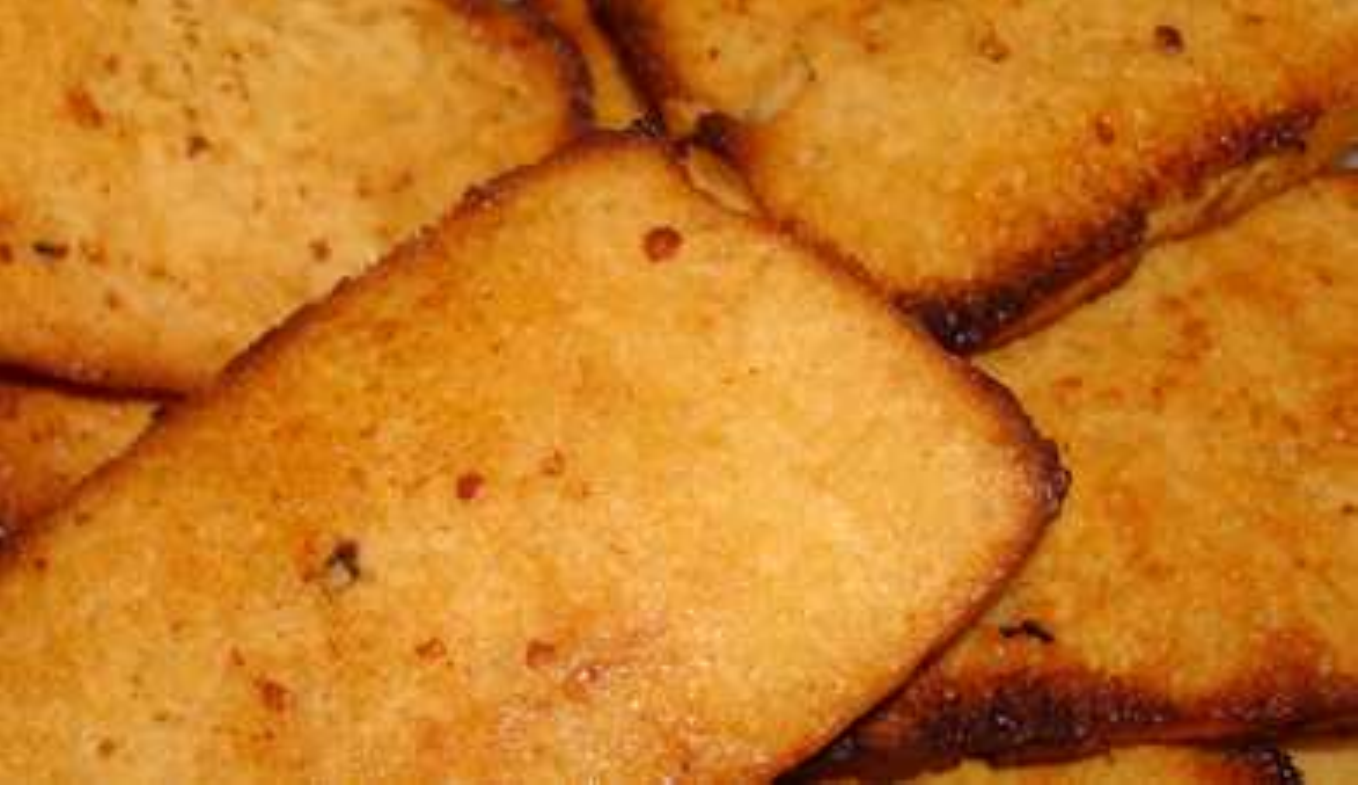,width=0.5in}
\psfig{figure=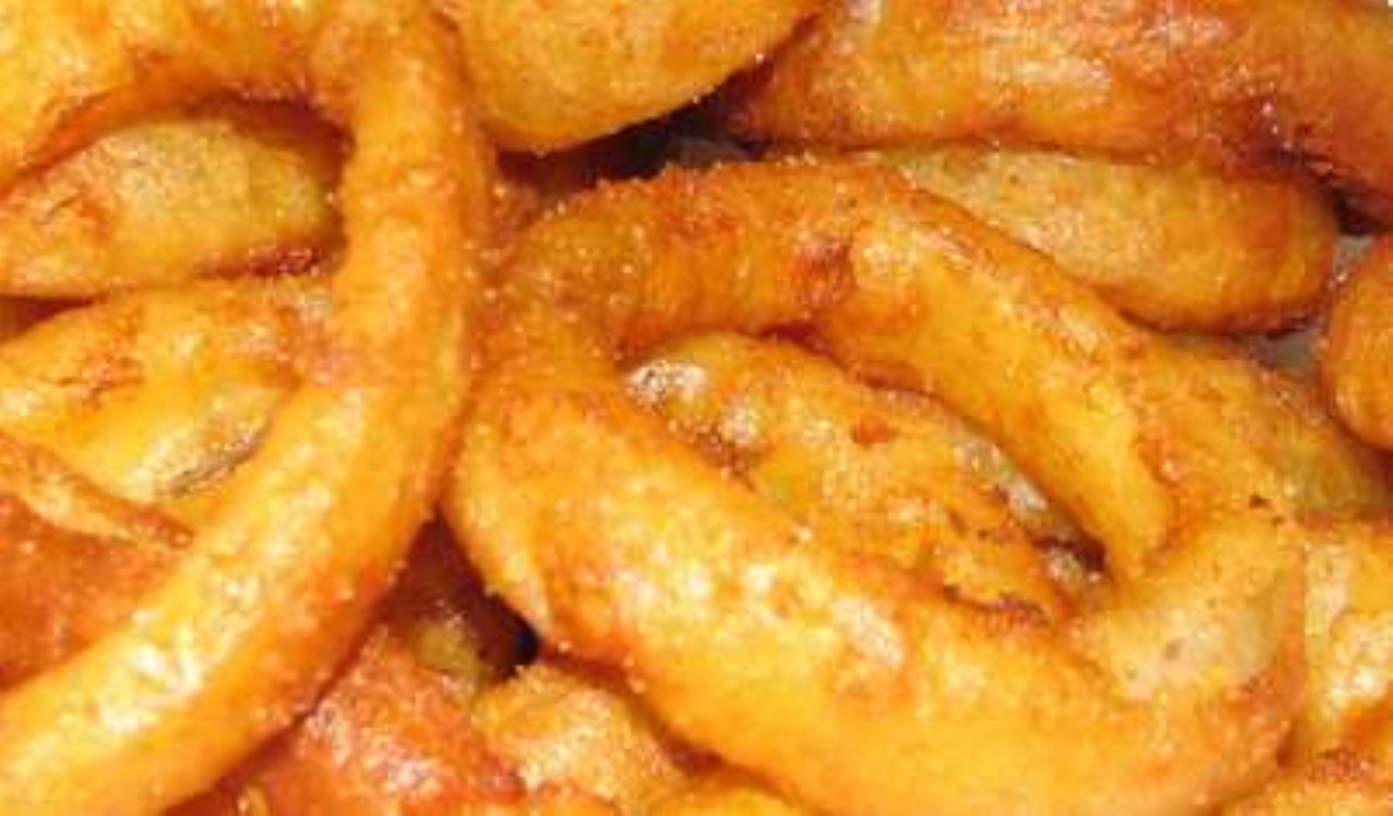,width=0.5in}
\psfig{figure=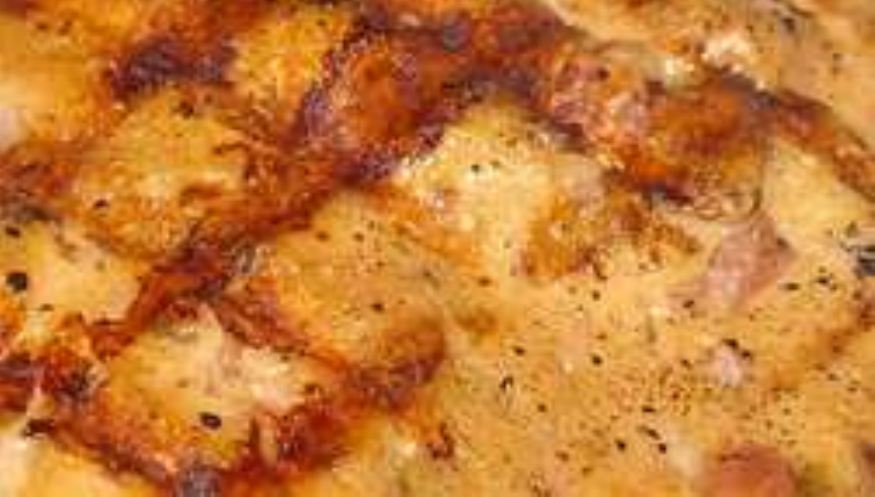,width=0.5in}
\psfig{figure=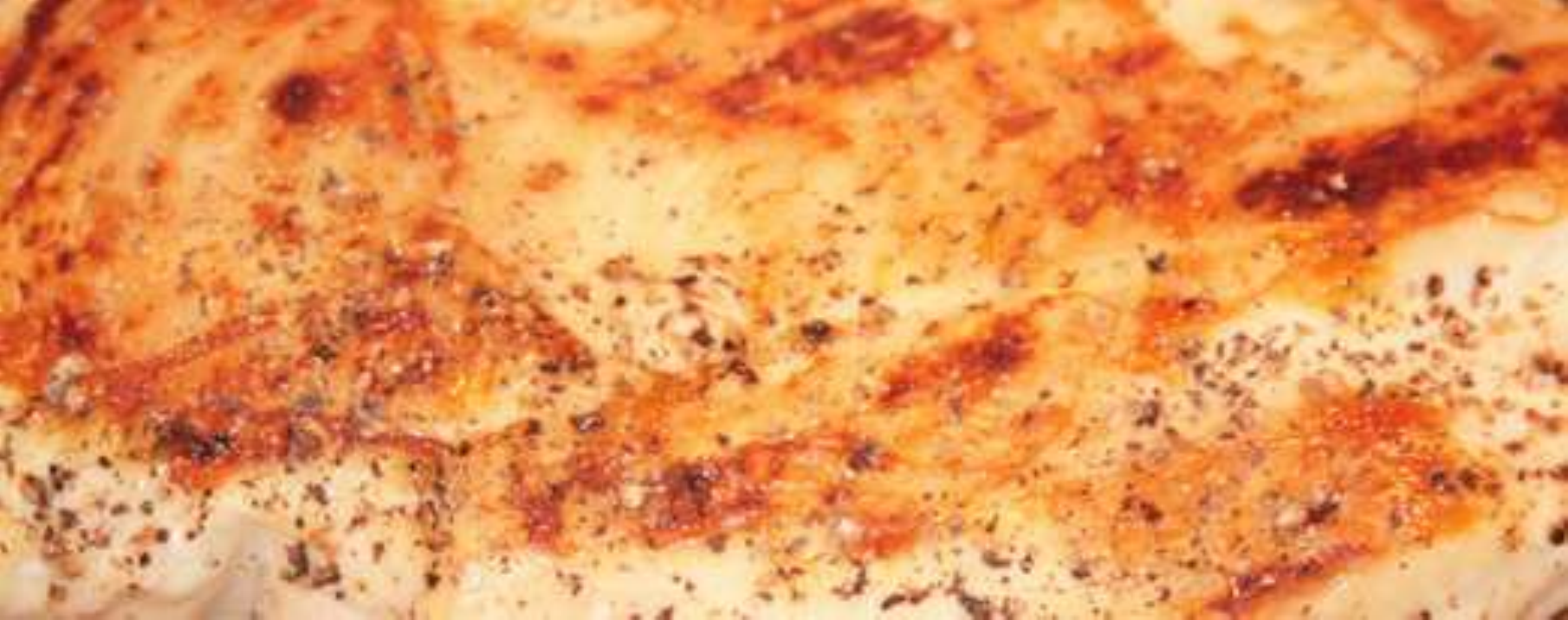,width=0.5in}
}
\centerline{\protect\tiny 127.3 \hspace{0.35in} 108.0 \hspace{0.35in} 142.0 \hspace{0.35in}  142.5 \hspace{0.35in} 88.13 \hspace{0.35in} 91.7\hspace{0.35in} 76.2 \hspace{0.35in} 0.0\hspace{0.35in} 7.66 \hspace{0.35in} 0.0 \hspace{0.35in} 9.2 \hspace{0.35in}  0.2
}
\centerline{
\psfig{figure=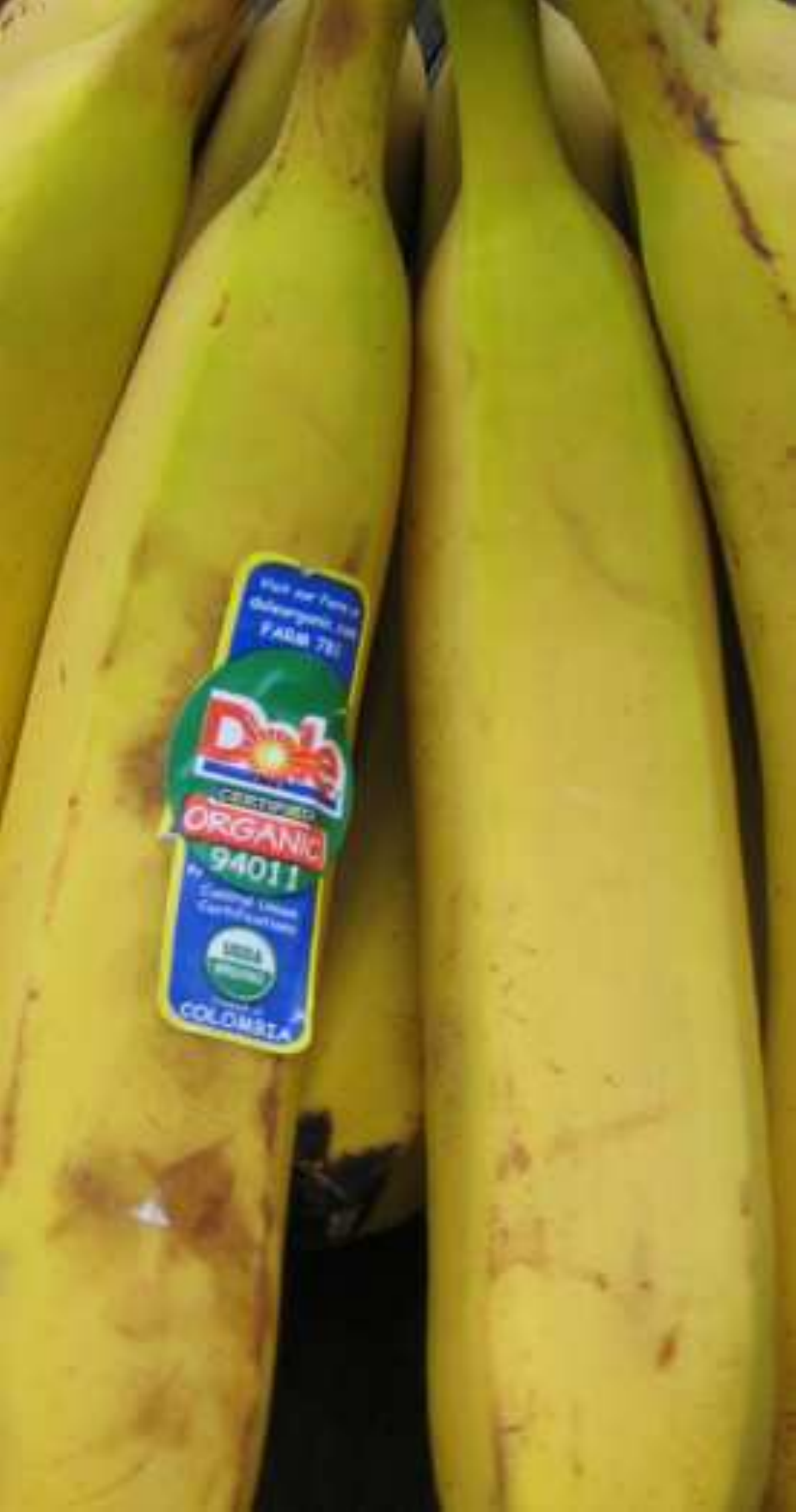,width=0.3in}
\psfig{figure=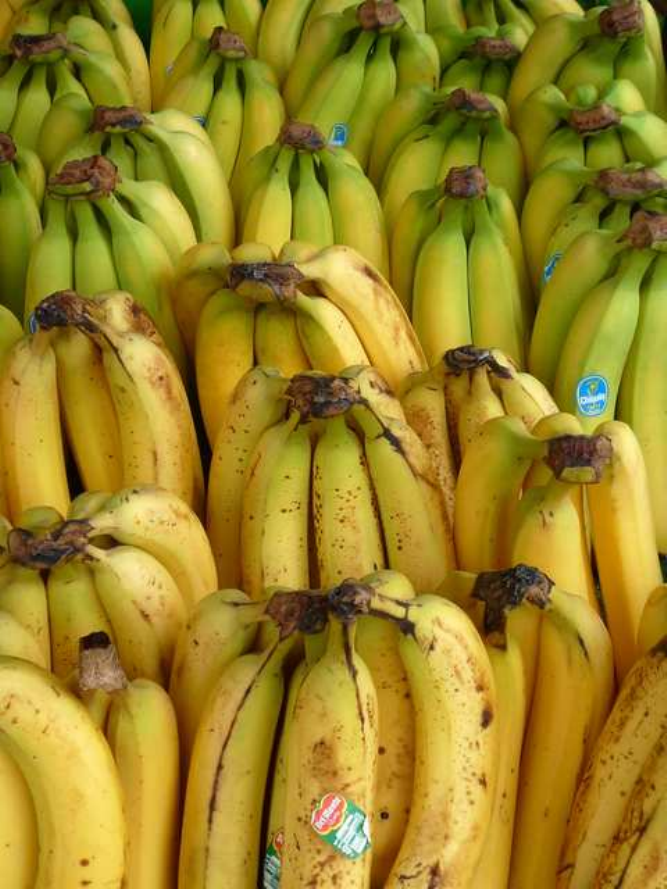,width=0.4in}
\psfig{figure=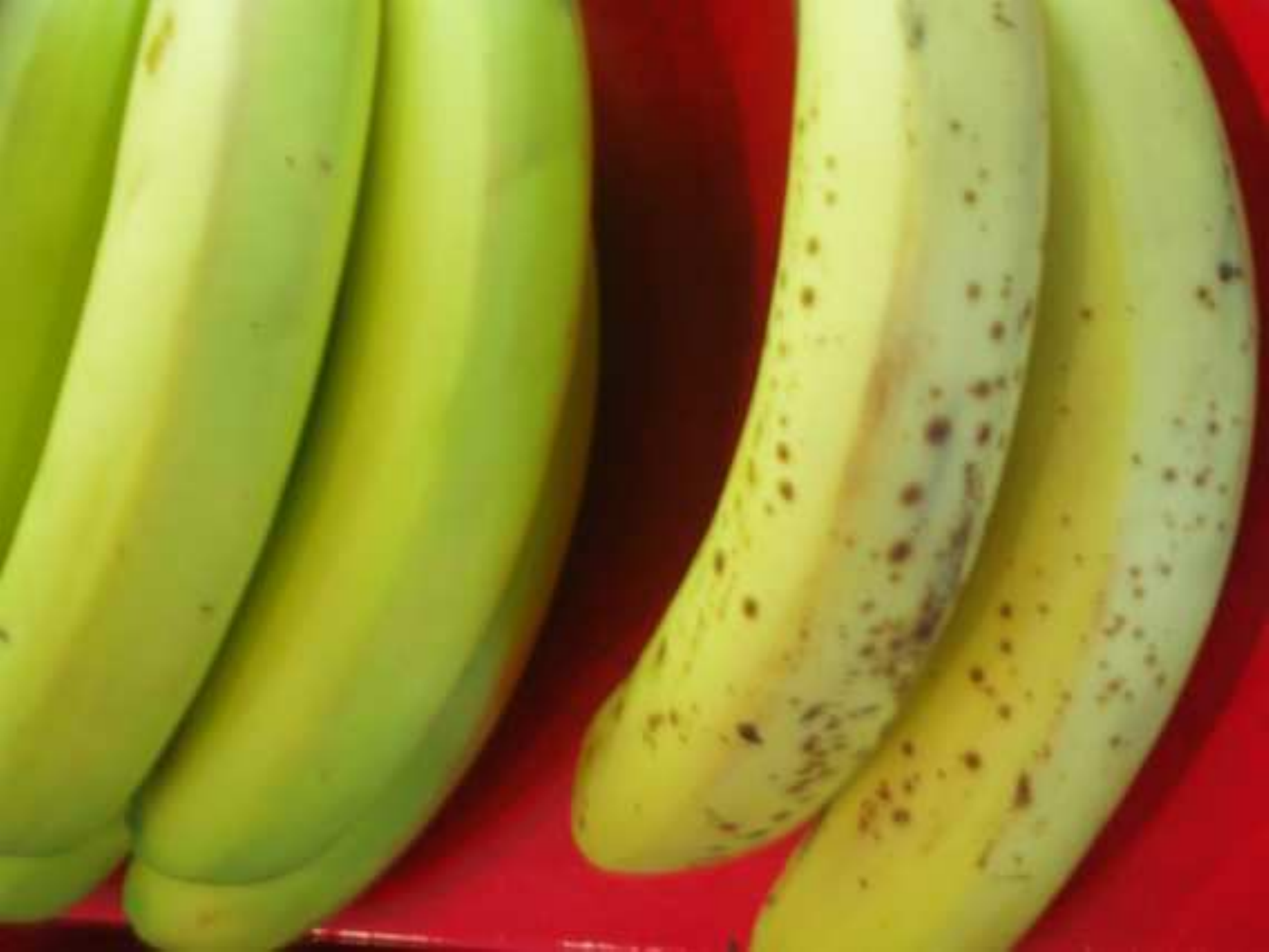,width=0.5in}
\psfig{figure=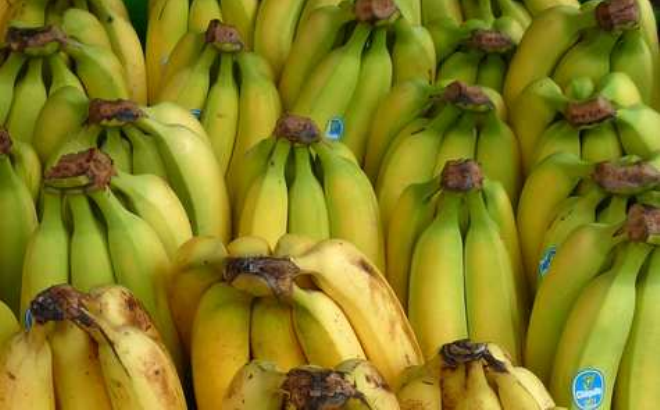,width=0.5in}
\psfig{figure=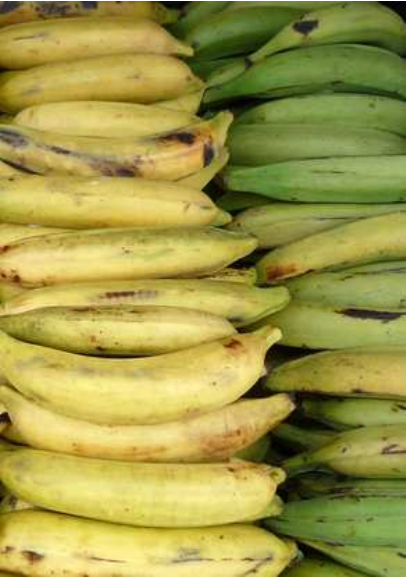,width=0.45in}
\psfig{figure=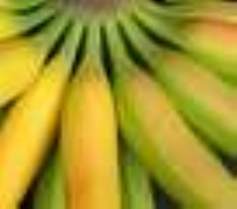,width=0.5in}
\psfig{figure=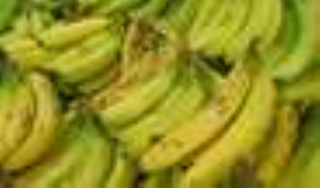,width=0.5in}
\psfig{figure=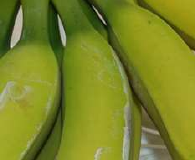,width=0.5in}
\psfig{figure=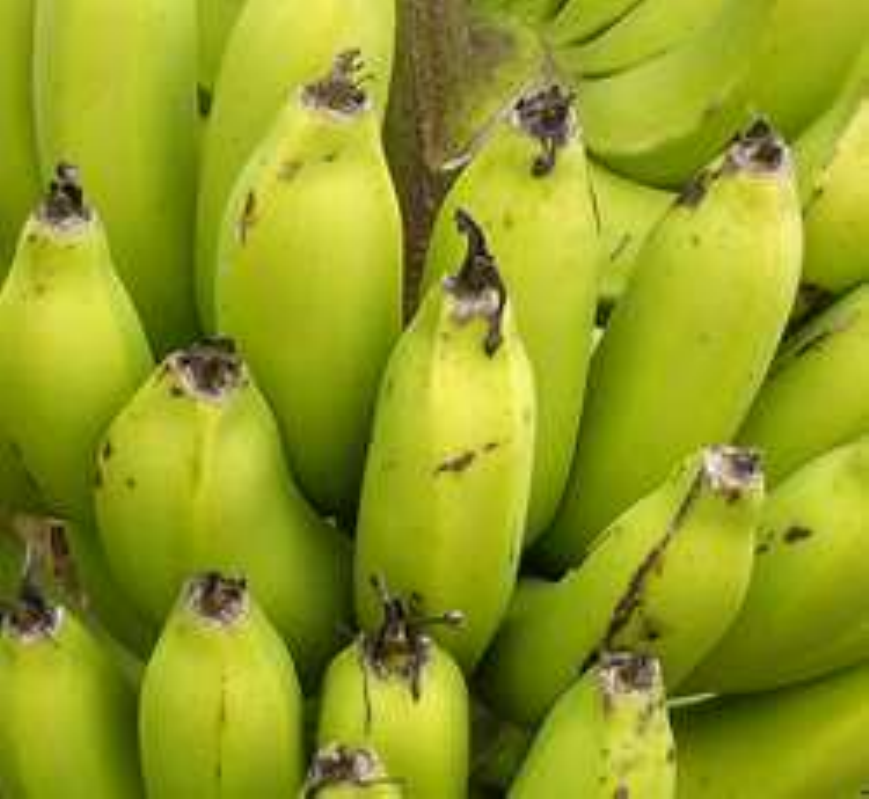,width=0.5in}
\psfig{figure=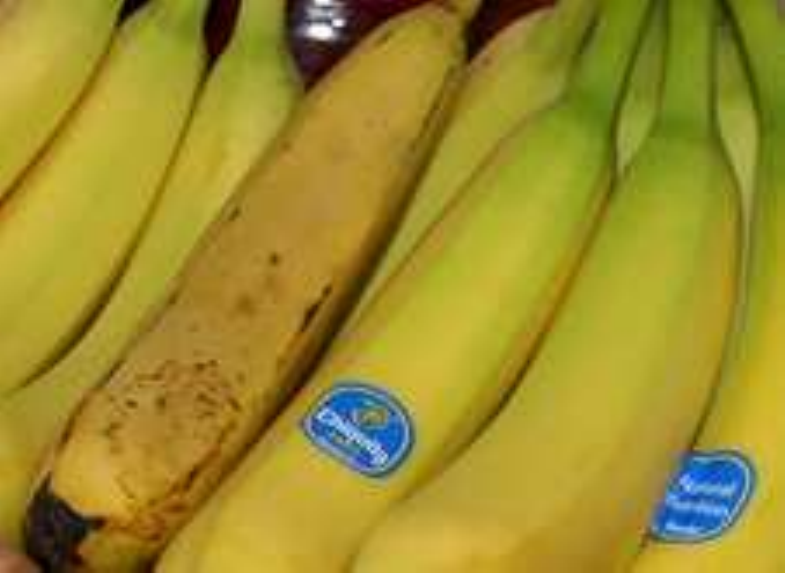,width=0.5in}
\psfig{figure=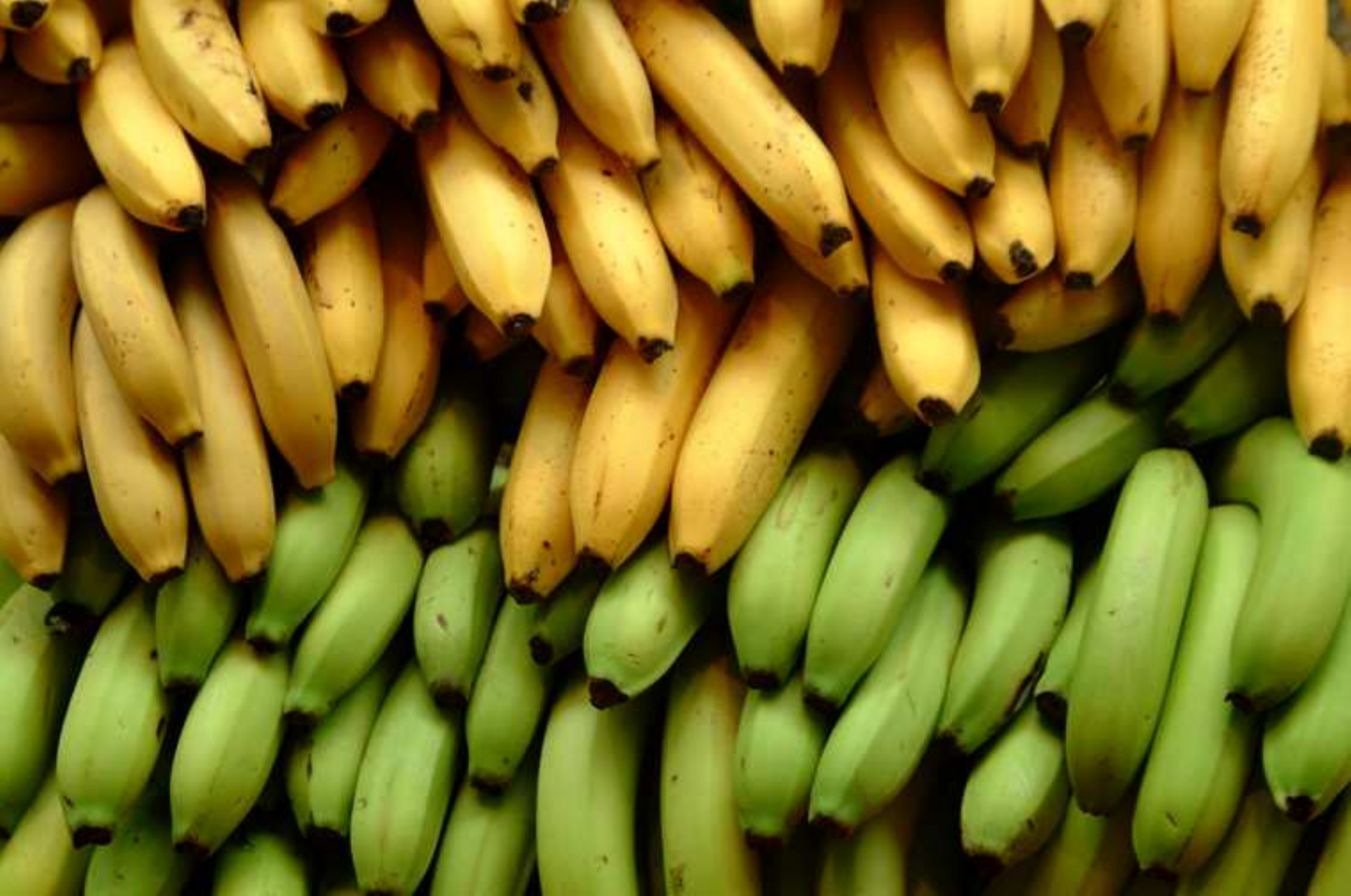,width=0.5in}
\psfig{figure=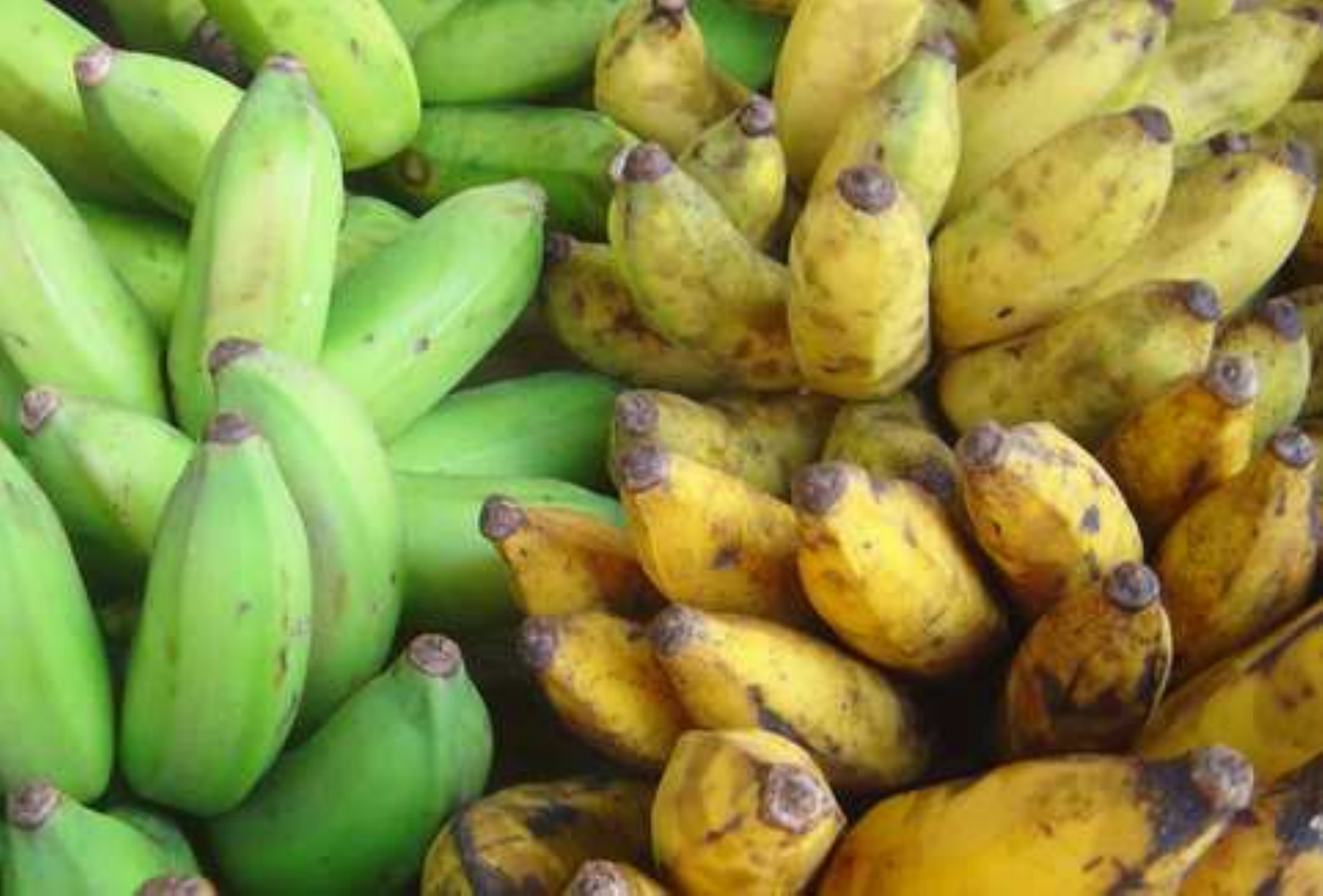,width=0.5in}
}
\centerline{\protect\tiny 148.3 \hspace{0.25in}120.3  \hspace{0.35in} 155.7\hspace{0.35in} 123.3 \hspace{0.35in} 122.7\hspace{0.3in} 35.1\hspace{0.35in} 38.6 \hspace{0.35in} 20.73 \hspace{0.35in}  0.3 \hspace{0.35in} 10.3 \hspace{0.35in} 4.6 \hspace{0.35in} 0.0
}
\caption{\protect\scriptsize
Recognition and evaluation of appearance with respect to the toasted bread (top two rows) and bananas. The top row has baked food while the middle row has fried foods. Below each image we show the distance.}\label{recognition}
\end{figure*}

Figure \ref{detection} shows a large set of example detections and segmentation
 of materials based on  single exemplars shown at the
left side of each row. The input image is shown above the detection image.
The pixels in Red/Green depict outlier pixels to the exemplar model. 
In the first row the exemplar  bok choy is used to detect a variety of vegetables that have different
range of green colors. An example of cooked greens (broccoli) as well as ones covered by a salad dressing are shown in addition to
raw vegetables. Similarly, examples for baked food are detected based on a toasted bread exemplar, fried foods are detected
based on an exemplar of fried Tofu, grilled foods are detected based on a grilled pork exemplar, bananas are detected based on
green/yellow exemplar and finally fires are detected based on an exemplar of flames.
Note that in the presence of white plates,  in grilled and fried food the white parts of the plate were 
detected since the color is close to the exit point of the polynomial. A similar event happens in very dark areas. 
Both can be reduced by either limiting the extrapolation of the representation or pre or post processing.

The examples in Figure \ref{detection} show that despite the use of a single exemplar the modeling approach is effective in capturing 
the essence of the variation in appearance. The combination of interpolation, extrapolation and enforcing a minimal length
of the model to be matched enhances the detection and reduces outliers.
The images are  taken under diverse illumination, indoor and outdoor scenes. While the representation currently does not
account for illumination change it is evident that a measure of robustness is already embedded in the representation.


\section {Recognition}

The color-subspace representation supports recognition and aesthetic evaluation of objects with respect to the exemplar.
Given a probe image we calculate how well its raw colors project on the polynomial curve via a tighter measure than 
 $L(S)$ in Equation \ref{sdistance}. Specifically, this measure uses an adaptive threshold $l_s$ (taking into out outlier ratio 
with respect to the length of $POLY$)  instead of a fixed scalar.
Figure \ref{recognition} shows examples of baked goods (top row), fried foods (middle row) and green/yellow bananas.
The baked and fried food images are matched against the exemplar of the toasted bread, while the banana images are matched against
the exemplar of green/yellow bananas (see exemplars in Figure \ref{detection}).  Distinguishing between baked and
fried food is a challenging objective that involves recovering subtle visual cues (e.g., specular reflections of leftover oil,
pattern of browning that indicates frying on a flat surface or deep frying, etc.). 

In Figure \ref{recognition} most scores of baked foods are high with the highest 
values corresponding to objects that have wide distributions of colors that match strongly the color spectrum 
of the exemplar image. The saltines (right most image) is the exception since the colors are more clustered and cover 
only small part of the curve. 
The middle row shows results for fried and grilled food
compared against the baking exemplar. Some objects easily pass for baked foods (the left most 7 objects).
The fried bread (sixth image form the left) is similar to the exemplar and clearly subtle cues are needed to make the 
classification. The scores of the five objects at the right are low and they are not classified as baked.
The bottom row shows the results of different yellow/green bananas. The highest scores are given to bananas that have similar
color characteristics to the exemplar shown in Figure \ref{detection}. The images on the right show different
color patterns and receive low scores.

\section{Summary}

This paper proposed that a single process tends to create gradual changes in color appearance
of objects and as a result a subspace of RGB can represent this information.
The paper did not address the impact of illumination variations, but illustrated using web images
that well-illuminated scenes of familiar objects tend to be close to subspace representations derived from exemplars.

The color representation via polynomials naturally interpolates and extrapolates from limited color samples.
The parametric representation allows recognizing colors that do not occur in the exemplar colors of the object.
Thus, providing a clear advantage with respect to other representations 
where unobserved data is not accounted for. 

The representation enables determining whether a set of pixels matches a significant part of the exemplar polynomial.
Therefore, the degree of correspondence between a set of spatially coherent pixels in the probe image and the
exemplar polynomial determines if a region complies with exemplar characteristics. For example, different toasts of bread
lead to different color distributions. However, as long as the distances between the probe pixels and the exemplar are small
and {\em sufficient} segment of the closest points on the polynomial is found than it is likely that image is that of toasted bread.